# High Dimensional Expectation-Maximization Algorithm: Statistical Optimization and Asymptotic Normality

Zhaoran Wang* Quanquan Gu* Yang Ning* Han Liu*

December 20th, 2014

**Abstract**

We provide a general theory of the expectation-maximization (EM) algorithm for inferring high dimensional latent variable models. In particular, we make two contributions: (i) For parameter estimation, we propose a novel high dimensional EM algorithm which naturally incorporates sparsity structure into parameter estimation. With an appropriate initialization, this algorithm converges at a geometric rate and attains an estimator with the (near-)optimal statistical rate of convergence. (ii) Based on the obtained estimator, we propose new inferential procedures for testing hypotheses and constructing confidence intervals for low dimensional components of high dimensional parameters. For a broad family of statistical models, our framework establishes the first computationally feasible approach for optimal estimation and asymptotic inference in high dimensions. Our theory is supported by thorough numerical results.

## 1 Introduction

The expectation-maximization (EM) algorithm (Dempster et al., 1977) is the most popular approach for calculating the maximum likelihood estimator of latent variable models. Nevertheless, due to the nonconcavity of the likelihood function of latent variable models, the EM algorithm generally only converges to a local maximum rather than the global one (Wu, 1983). On the other hand, existing statistical guarantees for latent variable models are only established for global optima (Bartholomew et al., 2011). Therefore, there exists a gap between computation and statistics.

Significant progress has been made toward closing the gap between the local maximum attained by the EM algorithm and the maximum likelihood estimator (Wu, 1983; Tseng, 2004; McLachlan and Krishnan, 2007; Chrétien and Hero, 2008; Balakrishnan et al., 2014). In particular, Wu (1983) first establishes general sufficient conditions for the convergence of the EM algorithm. Tseng (2004); Chrétien and Hero (2008) further improve this result by viewing the EM algorithm as a proximal point method applied to the Kullback-Leibler divergence. See McLachlan and Krishnan (2007) for a detailed survey. More recently, Balakrishnan et al. (2014) establish the first result that characterizes explicit statistical and computational rates of convergence for the EM algorithm. They prove that, given a suitable initialization, the EM algorithm converges at a geometric rate to a local maximum close to the maximum likelihood estimator. All these results are established in the low dimensional regime where the dimension $d$ is much smaller than the sample size $n$.

*Department of Operations Research and Financial Engineering, Princeton University, Princeton, NJ 08544, USA; Email: {zhaoran, qgu, yning, hanliu}@princeton.edu

In high dimensional regimes where the dimension $d$ is much larger than the sample size $n$, there exists no theoretical guarantee for the EM algorithm. In fact, when $d \gg n$, the maximum likelihood estimator is in general not well defined, unless the models are carefully regularized by sparsity-type assumptions. Furthermore, even if a regularized maximum likelihood estimator can be obtained in a computationally tractable manner, establishing the corresponding statistical properties, especially asymptotic normality, can still be challenging because of the existence of high dimensional nuisance parameters. To address such a challenge, we develop a general inferential theory of the EM algorithm for parameter estimation and uncertainty assessment of high dimensional latent variable models. In particular, we make two contributions in this paper:

- For high dimensional parameter estimation, we propose a novel high dimensional EM algorithm by attaching a truncation step to the expectation step (E-step) and maximization step (M-step). Such a truncation step effectively enforces the sparsity of the attained estimator and allows us to establish significantly improved statistical rate of convergence.
- Based upon the estimator attained by the high dimensional EM algorithm, we propose a family of decorrelated score and Wald statistics for testing hypotheses for low dimensional components of the high dimensional parameter. The decorrelated Wald statistic can be further used to construct optimal valid confidence intervals for low dimensional parameters of interest.

Under a unified analytic framework, we establish simultaneous statistical and computational guarantees for the proposed high dimensional EM algorithm and the respective uncertainty assessment procedures. Let $\boldsymbol{\beta}^* \in \mathbb{R}^d$ be the true parameter, $s^*$ be its sparsity level and $\{\boldsymbol{\beta}^{(t)}\}_{t=0}^T$ be the iterative solution sequence of the high dimensional EM algorithm with $T$ being the total number of iterations. In particular, we prove that:

- Given an appropriate initialization $\boldsymbol{\beta}^{\text{init}}$ with relative error upper bounded by a constant $\kappa \in (0,1)$, i.e., $\|\boldsymbol{\beta}^{\text{init}} - \boldsymbol{\beta}^*\|_2 / \|\boldsymbol{\beta}^*\|_2 \leq \kappa$, the iterative solution sequence $\{\boldsymbol{\beta}^{(t)}\}_{t=0}^T$ satisfies

$$\|\boldsymbol{\beta}^{(t)} - \boldsymbol{\beta}^*\|_2 \leq \underbrace{\Delta_1 \cdot \rho^{t/2}}_{\text{Optimization Error}} + \underbrace{\Delta_2 \cdot \overbrace{\sqrt{\frac{s^* \cdot \log d}{n}}}^{\text{Optimal Rate}}}_{\text{Statistical Error}} \tag{1.1}$$

  with high probability. Here $\rho \in (0,1)$, and $\Delta_1$, $\Delta_2$ are quantities that possibly depend on $\rho$, $\kappa$ and $\boldsymbol{\beta}^*$. As the optimization error term in (1.1) decreases to zero at a geometric rate with respect to $t$, the overall estimation error achieves the $\sqrt{s^* \cdot \log d / n}$ statistical rate of convergence (up to an extra factor of $\log n$), which is (near-)minimax-optimal. See Theorem 3.4 and the corresponding discussion for details.

- The proposed decorrelated score and Wald statistics are asymptotically normal. Moreover, their limiting variances and the size of the respective confidence interval are optimal in the sense that they attain the semiparametric information bound for the low dimensional components of interest in the presence of high dimensional nuisance parameters. See Theorems 4.6 and 4.7 for details.

Our framework allows two implementations of the M-step: the exact maximization versus approximate maximization. The former one calculates the maximizer exactly, while the latter one conducts an approximate maximization through a gradient ascent step. Our framework is quite general. We illustrate its effectiveness by applying it to three high dimensional latent variable models, including Gaussian mixture model, mixture of regression model and regression with missing covariates.

**Comparison with Related Work:** A closely related work is by Balakrishnan et al. (2014), which considers the low dimensional regime where $d$ is much smaller than $n$. Under certain initialization conditions, they prove that the EM algorithm converges at a geometric rate to some local optimum that attains the $\sqrt{d/n}$ statistical rate of convergence. They cover both maximization and gradient ascent implementations of the M-step, and establish the consequences for the three latent variable models considered in our paper under low dimensional settings. Our framework adopts their view of treating the EM algorithm as a perturbed version of gradient methods. However, to handle the challenge of high dimensionality, the key ingredient of our framework is the truncation step that enforces the sparsity structure along the solution path. Such a truncation operation poses significant challenges for both computational and statistical analysis. In detail, for computational analysis we need to carefully characterize the evolution of each intermediate solution's support and its effects on the evolution of the entire iterative solution sequence. For statistical analysis, we need to establish a fine-grained characterization of the entrywise statistical error, which is technically more challenging than just establishing the $\ell_2$-norm error employed by Balakrishnan et al. (2014). In high dimensional regimes, we need to establish the $\sqrt{s^* \cdot \log d/n}$ statistical rate of convergence, which is much sharper than their $\sqrt{d/n}$ rate when $d \gg n$. In addition to point estimation, we further construct confidence intervals and hypothesis tests for latent variable models in the high dimensional regime, which have not been established before.

High dimensionality poses significant challenges for assessing the uncertainty (e.g., constructing confidence intervals and testing hypotheses) of the constructed estimators. For example, Knight and Fu (2000) show that the limiting distribution of the Lasso estimator is not Gaussian even in the low dimensional regime. A variety of approaches have been proposed to correct the Lasso estimator to attain asymptotic normality, including the debiasing method (Javanmard and Montanari, 2014), the desparsification methods (Zhang and Zhang, 2014; van de Geer et al., 2014) as well as instrumental variable-based methods (Belloni et al., 2012, 2013, 2014). Meanwhile, Lockhart et al. (2014); Taylor et al. (2014); Lee et al. (2013) propose the post-selection procedures for exact inference. In addition, several authors propose methods based on data splitting (Wasserman and Roeder, 2009; Meinshausen et al., 2009), stability selection (Meinshausen and Bühlmann, 2010) and $\ell_2$-confidence sets (Nickl and van de Geer, 2013). However, these approaches mainly focus on generalized linear models rather than latent variable models. In addition, their results heavily rely on the fact that the estimator is a global optimum of a convex program. In comparison, our approach applies to a much broader family of statistical models with latent structures. For these latent variable models, it is computationally infeasible to obtain the global maximum of the penalized likelihood due to the nonconcavity of the likelihood function. Unlike existing approaches, our inferential theory is developed for the estimator attained by the proposed high dimensional EM algorithm, which is not necessarily a global optimum to any optimization formulation.

Another line of research for the estimation of latent variable models is the tensor method, which exploits the structures of third or higher order moments. See Anandkumar et al. (2014a,b,c) and the references therein. However, existing tensor methods primarily focus on the low dimensional regime where $d \ll n$. In addition, since the high order sample moments generally have a slow statistical rate of convergence, the estimators obtained by the tensor methods usually have a suboptimal statistical rate even for $d \ll n$. For example, Chaganty and Liang (2013) establish the $\sqrt{d^6/n}$ statistical rate of convergence for mixture of regression model, which is suboptimal compared with the $\sqrt{d/n}$ minimax lower bound. Similarly, in high dimensional settings, the statistical rates of convergence attained by tensor methods are significantly slower than the statistical rate obtained in this paper.

The three latent variable models considered in this paper have been well studied. Nevertheless, only a few works establish theoretical guarantees for the EM algorithm. In particular, for Gaussian mixture model, Dasgupta and Schulman (2000, 2007); Chaudhuri et al. (2009) establish parameter estimation guarantees for the EM algorithm and its extensions. For mixture of regression model, Yi et al. (2013) establish exact parameter recovery guarantees for the EM algorithm under a noiseless setting. For high dimensional mixture of regression model, Städler et al. (2010) analyze the gradient EM algorithm for the $\ell_1$-penalized log-likelihood. They establish support recovery guarantees for the attained local optimum but have no parameter estimation guarantees. In comparison with existing works, this paper establishes a general inferential framework for simultaneous parameter estimation and uncertainty assessment based on a novel high dimensional EM algorithm. Our analysis provides the first theoretical guarantee of parameter estimation and asymptotic inference in high dimensional regimes for the EM algorithm and its applications to a broad family of latent variable models.

**Notation:** Let $\mathbf{A} = [\mathbf{A}_{i,j}] \in \mathbb{R}^{d\times d}$ and $\mathbf{v} = (v_1, \ldots, v_d)^\top \in \mathbb{R}^d$. We define the $\ell_q$-norm $(q \geq 1)$ of $\mathbf{v}$ as $\|\mathbf{v}\|_q = \big(\sum_{j=1}^d |v_j|^q\big)^{1/q}$. Particularly, $\|\mathbf{v}\|_0$ denotes the number of nonzero entries of $\mathbf{v}$. For $q \geq 1$, we define $\|\mathbf{A}\|_q$ as the operator norm of $\mathbf{A}$. Specifically, $\|\mathbf{A}\|_2$ is the spectral norm. For a set $\mathcal{S}$, $|\mathcal{S}|$ denotes its cardinality. We denote the $d \times d$ identity matrix as $\mathbf{I}_d$. For index sets $\mathcal{I}, \mathcal{J} \subseteq \{1, \ldots, d\}$, we define $\mathbf{A}_{\mathcal{I},\mathcal{J}} \in \mathbb{R}^{d\times d}$ to be the matrix whose $(i, j)$-th entry is equal to $\mathbf{A}_{i,j}$ if $i \in \mathcal{I}$ and $j \in \mathcal{J}$, and zero otherwise. We define $\mathbf{v}_{\mathcal{I}}$ similarly. We denote $\otimes$ and $\odot$ to be the outer product and Hadamard product between vectors. The matrix $(p, q)$-norm, i.e., $\|\mathbf{A}\|_{p,q}$, is obtained by taking the $\ell_p$-norm of each row and then taking the $\ell_q$-norm of the obtained row norms. Let $\mathrm{supp}(\mathbf{v})$ be the support of $\mathbf{v}$, i.e., the index set of its nonzero entries. We use $C, C', \ldots$ to denote generic absolute constants. The values of these constants may vary from line to line. In addition, we denote $\|\cdot\|_{\psi_q}$ $(q \geq 1)$ to be the Orlicz norm of random variables. We will introduce more notations in §2.2.

The rest of the paper is organized as follows. In §2 we present the high dimensional EM algorithm and the corresponding procedures for inference, and then apply them to three latent variable models. In §3 and §4, we establish the main theoretical results for computation, parameter estimation and asymptotic inference, as well as their implications for specific latent variable models. In §5 we sketch the proof of the main results. In §6 we present the numerical results. In §7 we conclude the paper.

## 2 Methodology

We first introduce the high dimensional EM Algorithm. Then we present the respective procedures for asymptotic inference. Finally, we apply the proposed methods to three latent variable models.

### 2.1 High Dimensional EM Algorithm

Before we introduce the proposed high dimensional EM Algorithm (Algorithm 1), we briefly review the classical EM algorithm. Let $h_{\boldsymbol{\beta}}(\mathbf{y})$ be the probability density function of $\boldsymbol{Y} \in \mathcal{Y}$, where $\boldsymbol{\beta} \in \mathbb{R}^d$ is the model parameter. For latent variable models, we assume that $h_{\boldsymbol{\beta}}(\mathbf{y})$ is obtained by marginalizing over an unobserved latent variable $\boldsymbol{Z} \in \mathcal{Z}$, i.e.,

$$h_{\boldsymbol{\beta}}(\mathbf{y}) = \int_{\mathcal{Z}} f_{\boldsymbol{\beta}}(\mathbf{y}, \mathbf{z}) \, \mathrm{d}\mathbf{z}. \tag{2.1}$$

Given the $n$ observations $\mathbf{y}_1, \ldots, \mathbf{y}_n$ of $\boldsymbol{Y}$, the EM algorithm aims at maximizing the log-likelihood

$$\ell_n(\boldsymbol{\beta}) = \sum_{i=1}^n \log h_{\boldsymbol{\beta}}(\mathbf{y}_i). \tag{2.2}$$

Due to the unobserved latent variable $\boldsymbol{Z}$, it is difficult to directly evaluate $\ell_n(\boldsymbol{\beta})$. Instead, we turn to consider the difference between $\ell_n(\boldsymbol{\beta})$ and $\ell_n(\boldsymbol{\beta}')$. Let $k_{\boldsymbol{\beta}}(\mathbf{z} \mid \mathbf{y})$ be the density of $\boldsymbol{Z}$ conditioning on the observed variable $\boldsymbol{Y} = \mathbf{y}$, i.e.,

$$k_{\boldsymbol{\beta}}(\mathbf{z} \mid \mathbf{y}) = f_{\boldsymbol{\beta}}(\mathbf{y}, \mathbf{z}) / h_{\boldsymbol{\beta}}(\mathbf{y}). \tag{2.3}$$

According to (2.1) and (2.2), we have

$$\begin{aligned} \frac{1}{n} \cdot \big[\ell_n(\boldsymbol{\beta}) - \ell_n(\boldsymbol{\beta}')\big] &= \frac{1}{n} \sum_{i=1}^n \log\big[h_{\boldsymbol{\beta}}(\mathbf{y}_i)/h_{\boldsymbol{\beta}'}(\mathbf{y}_i)\big] = \frac{1}{n} \sum_{i=1}^n \log\left[\int_{\mathcal{Z}} \frac{f_{\boldsymbol{\beta}}(\mathbf{y}_i, \mathbf{z})}{h_{\boldsymbol{\beta}'}(\mathbf{y}_i)} \, \mathrm{d}\mathbf{z}\right] \\ &= \frac{1}{n} \sum_{i=1}^n \log\left[\int_{\mathcal{Z}} k_{\boldsymbol{\beta}'}(\mathbf{z} \mid \mathbf{y}_i) \cdot \frac{f_{\boldsymbol{\beta}}(\mathbf{y}_i, \mathbf{z})}{f_{\boldsymbol{\beta}'}(\mathbf{y}_i, \mathbf{z})} \, \mathrm{d}\mathbf{z}\right] \geq \frac{1}{n} \sum_{i=1}^n \int_{\mathcal{Z}} k_{\boldsymbol{\beta}'}(\mathbf{z} \mid \mathbf{y}_i) \cdot \log\left[\frac{f_{\boldsymbol{\beta}}(\mathbf{y}_i, \mathbf{z})}{f_{\boldsymbol{\beta}'}(\mathbf{y}_i, \mathbf{z})}\right] \mathrm{d}\mathbf{z}, \end{aligned} \tag{2.4}$$

where the third equality follows from (2.3) and the inequality is obtained from Jensen's inequality. On the right-hand side of (2.4) we have

$$\begin{aligned} &\frac{1}{n} \sum_{i=1}^n \int_{\mathcal{Z}} k_{\boldsymbol{\beta}'}(\mathbf{z} \mid \mathbf{y}_i) \cdot \log\left[\frac{f_{\boldsymbol{\beta}}(\mathbf{y}_i, \mathbf{z})}{f_{\boldsymbol{\beta}'}(\mathbf{y}_i, \mathbf{z})}\right] \mathrm{d}\mathbf{z} \\ &\quad = \underbrace{\frac{1}{n} \sum_{i=1}^n \int_{\mathcal{Z}} k_{\boldsymbol{\beta}'}(\mathbf{z} \mid \mathbf{y}_i) \cdot \log f_{\boldsymbol{\beta}}(\mathbf{y}_i, \mathbf{z}) \, \mathrm{d}\mathbf{z}}_{Q_n(\boldsymbol{\beta}; \boldsymbol{\beta}')} - \frac{1}{n} \sum_{i=1}^n \int_{\mathcal{Z}} k_{\boldsymbol{\beta}'}(\mathbf{z} \mid \mathbf{y}_i) \cdot \log f_{\boldsymbol{\beta}'}(\mathbf{y}_i, \mathbf{z}) \, \mathrm{d}\mathbf{z}. \end{aligned} \tag{2.5}$$

We define the first term on the right-hand side of (2.5) to be $Q_n(\boldsymbol{\beta}; \boldsymbol{\beta}')$. Correspondingly, we define its expectation to be $Q(\boldsymbol{\beta}; \boldsymbol{\beta}')$. Note the second term on the right-hand side of (2.5) doesn't depend

**Algorithm 1** High Dimensional EM Algorithm

1: **Parameter:** Sparsity Parameter $\widehat{s}$, Maximum Number of Iterations $T$
2: **Initialization:** $\widehat{\mathcal{S}}^{\text{init}} \leftarrow \text{supp}\big(\boldsymbol{\beta}^{\text{init}}, \widehat{s}\big)$, $\quad \boldsymbol{\beta}^{(0)} \leftarrow \text{trunc}\big(\boldsymbol{\beta}^{\text{init}}, \widehat{\mathcal{S}}^{\text{init}}\big)$
$\big\{$supp$(\cdot,\cdot)$ and trunc$(\cdot,\cdot)$ are defined in (2.6) and (2.7)$\big\}$
3: **For** $t = 0$ to $T-1$
4: **E-step:** Evaluate $Q_n\big(\boldsymbol{\beta}; \boldsymbol{\beta}^{(t)}\big)$
5: **M-step:** $\boldsymbol{\beta}^{(t+0.5)} \leftarrow M_n\big(\boldsymbol{\beta}^{(t)}\big)$ $\quad \big\{M_n(\cdot)$ is implemented as in Algorithm 2 or 3$\big\}$
6: **T-step:** $\widehat{\mathcal{S}}^{(t+0.5)} \leftarrow \text{supp}\big(\boldsymbol{\beta}^{(t+0.5)}, \widehat{s}\big)$, $\quad \boldsymbol{\beta}^{(t+1)} \leftarrow \text{trunc}\big(\boldsymbol{\beta}^{(t+0.5)}, \widehat{\mathcal{S}}^{(t+0.5)}\big)$
7: **End For**
8: **Output:** $\widehat{\boldsymbol{\beta}} \leftarrow \boldsymbol{\beta}^{(T)}$

on $\boldsymbol{\beta}$. Hence, given some fixed $\boldsymbol{\beta}'$, we can maximize the lower bound function $Q_n(\boldsymbol{\beta}; \boldsymbol{\beta}')$ over $\boldsymbol{\beta}$ to obtain a sufficiently large $\ell_n(\boldsymbol{\beta}) - \ell_n(\boldsymbol{\beta}')$. Based on such an observation, at the $t$-th iteration of the classical EM algorithm, we evaluate $Q_n\big(\boldsymbol{\beta}; \boldsymbol{\beta}^{(t)}\big)$ at the E-step and then perform $\max_{\boldsymbol{\beta}} Q_n\big(\boldsymbol{\beta}; \boldsymbol{\beta}^{(t)}\big)$ at the M-step. See McLachlan and Krishnan (2007) for more details.

The proposed high dimensional EM algorithm (Algorithm 1) is built upon the E-step and M-step (lines 4 and 5) of the classical EM algorithm. In addition to the exact maximization implementation of the M-step (Algorithm 2), we allow the gradient ascent implementation of the M-step (Algorithm 3), which performs an approximate maximization via a gradient ascent step. To handle the challenge of high dimensionality, in line 6 of Algorithm 1 we perform a truncation step (T-step) to enforce the sparsity structure. In detail, we define the supp$(\cdot,\cdot)$ function in line 6 as

$$\text{supp}(\boldsymbol{\beta}, s)\text{: The set of index } j\text{'s corresponding to the top } s \text{ largest } |\beta_j|\text{'s.} \tag{2.6}$$

Also, for an index set $\mathcal{S} \subseteq \{1, \ldots, d\}$, we define the trunc$(\cdot,\cdot)$ function in line 6 as

$$\big[\text{trunc}(\boldsymbol{\beta}, \mathcal{S})\big]_j = \begin{cases} \beta_j & j \in \mathcal{S}, \\ 0 & j \notin \mathcal{S}. \end{cases} \tag{2.7}$$

Note that $\boldsymbol{\beta}^{(t+0.5)}$ is the output of the M-step (line 5) at the $t$-th iteration of the high dimensional EM algorithm. To obtain $\boldsymbol{\beta}^{(t+1)}$, the T-step (line 6) preserves the entries of $\boldsymbol{\beta}^{(t+0.5)}$ with the top $\widehat{s}$ large magnitudes and sets the rest to zero. Here $\widehat{s}$ is a tuning parameter that controls the sparsity level (line 1). By iteratively performing the E-step, M-step and T-step, the high dimensional EM algorithm attains an $\widehat{s}$-sparse estimator $\widehat{\boldsymbol{\beta}} = \boldsymbol{\beta}^{(T)}$ (line 8). Here $T$ is the total number of iterations.

It is worth noting that, the truncation strategy employed here and its variants are widely adopted in the context of sparse linear regression and sparse principal component analysis. For example, see Yuan and Zhang (2013); Yuan et al. (2013) and the references therein. Nevertheless, we incorporate this truncation strategy into the EM algorithm for the first time. Also, our analysis is significantly different from existing works.

**Algorithm 2** Maximization Implementation of the M-step

1: **Input:** $\boldsymbol{\beta}^{(t)}$, $Q_n\big(\boldsymbol{\beta}; \boldsymbol{\beta}^{(t)}\big)$
2: **Output:** $M_n\big(\boldsymbol{\beta}^{(t)}\big) \leftarrow \text{argmax}_{\boldsymbol{\beta}}\, Q_n\big(\boldsymbol{\beta}; \boldsymbol{\beta}^{(t)}\big)$

**Algorithm 3** Gradient Ascent Implementation of the M-step

1: **Input:** $\boldsymbol{\beta}^{(t)}$, $Q_n\big(\boldsymbol{\beta};\boldsymbol{\beta}^{(t)}\big)$
2: **Parameter:** Stepsize $\eta > 0$
3: **Output:** $M_n\big(\boldsymbol{\beta}^{(t)}\big) \leftarrow \boldsymbol{\beta}^{(t)} + \eta \cdot \nabla Q_n\big(\boldsymbol{\beta}^{(t)};\boldsymbol{\beta}^{(t)}\big)$
$\big\{$The gradient is taken with respect to the first $\boldsymbol{\beta}^{(t)}$ in $Q_n\big(\boldsymbol{\beta}^{(t)};\boldsymbol{\beta}^{(t)}\big)\big\}$

### 2.2 Asymptotic Inference

In the sequel, we first introduce some additional notations. Then we present the proposed methods for asymptotic inference in high dimensions.

**Notation:** Let $\nabla_1 Q(\boldsymbol{\beta};\boldsymbol{\beta}')$ be the gradient with respect to $\boldsymbol{\beta}$ and $\nabla_2 Q(\boldsymbol{\beta};\boldsymbol{\beta}')$ be the gradient with respect to $\boldsymbol{\beta}'$. If there is no confusion, we simply denote $\nabla Q(\boldsymbol{\beta};\boldsymbol{\beta}') = \nabla_1 Q(\boldsymbol{\beta};\boldsymbol{\beta}')$ as in the previous sections. We define the higher order derivatives in the same manner, e.g., $\nabla^2_{1,2} Q(\boldsymbol{\beta};\boldsymbol{\beta}')$ is calculated by first taking derivative with respect to $\boldsymbol{\beta}$ and then with respect to $\boldsymbol{\beta}'$. For $\boldsymbol{\beta} = \big(\boldsymbol{\beta}_1^\top, \boldsymbol{\beta}_2^\top\big)^\top \in \mathbb{R}^d$ with $\boldsymbol{\beta}_1 \in \mathbb{R}^{d_1}$, $\boldsymbol{\beta}_2 \in \mathbb{R}^{d_2}$ and $d_1 + d_2 = d$, we use notations such as $\mathbf{v}_{\boldsymbol{\beta}_1} \in \mathbb{R}^{d_1}$ and $\mathbf{A}_{\boldsymbol{\beta}_1,\boldsymbol{\beta}_2} \in \mathbb{R}^{d_1 \times d_2}$ to denote the corresponding subvector of $\mathbf{v} \in \mathbb{R}^d$ and the submatrix of $\mathbf{A} \in \mathbb{R}^{d \times d}$.

We aim to conduct asymptotic inference for low dimensional components of the high dimensional parameter $\boldsymbol{\beta}^*$. Without loss of generality, we consider a single entry of $\boldsymbol{\beta}^*$. In particular, we assume $\boldsymbol{\beta}^* = \big[\alpha^*, (\boldsymbol{\gamma}^*)^\top\big]^\top$, where $\alpha^* \in \mathbb{R}$ is the entry of interest, while $\boldsymbol{\gamma}^* \in \mathbb{R}^{d-1}$ is treated as the nuisance parameter. In the following, we construct two hypothesis tests, namely, the decorrelated score and Wald tests. Based on the decorrelated Wald test, we further construct valid confidence interval for $\alpha^*$. It is worth noting that, our method and theory can be easily generalized to perform statistical inference for an arbitrary low dimensional subvector of $\boldsymbol{\beta}^*$.

**Decorrelated Score Test:** For score test, we are primarily interested in testing $H_0 : \alpha^* = 0$, since this null hypothesis characterizes the uncertainty in variable selection. Our method easily generalizes to $H_0 : \alpha^* = \alpha_0$ with $\alpha_0 \neq 0$. For notational simplicity, we define the following key quantity

$$T_n(\boldsymbol{\beta}) = \nabla^2_{1,1} Q_n(\boldsymbol{\beta};\boldsymbol{\beta}) + \nabla^2_{1,2} Q_n(\boldsymbol{\beta};\boldsymbol{\beta}) \in \mathbb{R}^{d \times d}. \tag{2.8}$$

Let $\boldsymbol{\beta} = \big(\alpha, \boldsymbol{\gamma}^\top\big)^\top$. We define the decorrelated score function $S_n(\cdot,\cdot) \in \mathbb{R}$ as

$$S_n(\boldsymbol{\beta}, \lambda) = \big[\nabla_1 Q_n(\boldsymbol{\beta};\boldsymbol{\beta})\big]_\alpha - w(\boldsymbol{\beta}, \lambda)^\top \cdot \big[\nabla_1 Q_n(\boldsymbol{\beta};\boldsymbol{\beta})\big]_{\boldsymbol{\gamma}}. \tag{2.9}$$

Here $w(\boldsymbol{\beta}, \lambda) \in \mathbb{R}^{d-1}$ is obtained using the following Dantzig selector (Candès and Tao, 2007)

$$w(\boldsymbol{\beta}, \lambda) = \operatorname*{argmin}_{\mathbf{w} \in \mathbb{R}^{d-1}} \|\mathbf{w}\|_1, \quad \text{subject to } \big\| \big[T_n(\boldsymbol{\beta})\big]_{\boldsymbol{\gamma},\alpha} - \big[T_n(\boldsymbol{\beta})\big]_{\boldsymbol{\gamma},\boldsymbol{\gamma}} \cdot \mathbf{w} \big\|_\infty \le \lambda, \tag{2.10}$$

where $\lambda > 0$ is a tuning parameter. Let $\widehat{\boldsymbol{\beta}} = \big(\widehat{\alpha}, \widehat{\boldsymbol{\gamma}}^\top\big)^\top$, where $\widehat{\boldsymbol{\beta}}$ is the estimator attained by the high dimensional EM algorithm (Algorithm 1). We define the decorrelated score statistic as

$$\sqrt{n} \cdot S_n\big(\widehat{\boldsymbol{\beta}}_0, \lambda\big) \Big/ \Big\{ -\big[T_n\big(\widehat{\boldsymbol{\beta}}_0\big)\big]_{\alpha|\boldsymbol{\gamma}} \Big\}^{1/2}, \tag{2.11}$$

where $\widehat{\boldsymbol{\beta}}_0 = \big(0, \widehat{\boldsymbol{\gamma}}^\top\big)^\top$, and $\big[T_n\big(\widehat{\boldsymbol{\beta}}_0\big)\big]_{\alpha|\boldsymbol{\gamma}} = \Big[1, -w\big(\widehat{\boldsymbol{\beta}}_0, \lambda\big)^\top\Big] \cdot T_n\big(\widehat{\boldsymbol{\beta}}_0\big) \cdot \Big[1, -w\big(\widehat{\boldsymbol{\beta}}_0, \lambda\big)^\top\Big]^\top.$

Here we use $\widehat{\boldsymbol{\beta}}_0$ instead of $\widehat{\boldsymbol{\beta}}$ since we are interested in the null hypothesis $H_0 : \alpha^* = 0$. We can also replace $\widehat{\boldsymbol{\beta}}_0$ with $\widehat{\boldsymbol{\beta}}$ and the theoretical results will remain the same. In §4 we will prove the proposed decorrelated score statistic in (2.11) is asymptotically $N(0,1)$. Consequently, the decorrelated score test with significance level $\delta \in (0,1)$ takes the form

$$\psi_S(\delta) = \mathbb{1}\left\{\sqrt{n} \cdot S_n\big(\widehat{\boldsymbol{\beta}}_0, \lambda\big) \Big/ \Big\{-\big[T_n\big(\widehat{\boldsymbol{\beta}}_0\big)\big]_{\alpha|\boldsymbol{\gamma}}\Big\}^{1/2} \notin \big[-\Phi^{-1}(1-\delta/2), \Phi^{-1}(1-\delta/2)\big]\right\},$$

where $\Phi^{-1}(\cdot)$ is the inverse function of the Gaussian cumulative distribution function. If $\psi_S(\delta) = 1$, we reject the null hypothesis $H_0 : \alpha^* = 0$. Correspondingly, the associated p-value takes the form

$$\text{p-value} = 2 \cdot \left[1 - \Phi\left(\left|\sqrt{n} \cdot S_n\big(\widehat{\boldsymbol{\beta}}_0, \lambda\big) \Big/ \Big\{-\big[T_n\big(\widehat{\boldsymbol{\beta}}_0\big)\big]_{\alpha|\boldsymbol{\gamma}}\Big\}^{1/2}\right|\right)\right].$$

The intuition for the decorrelated score statistic in (2.11) can be understood as follows. Since $\ell_n(\boldsymbol{\beta})$ is the log-likelihood, its score function is $\nabla \ell_n(\boldsymbol{\beta})$ and the Fisher information at $\boldsymbol{\beta}^*$ is $I(\boldsymbol{\beta}^*) = -\mathbb{E}_{\boldsymbol{\beta}^*}\big[\nabla^2 \ell_n(\boldsymbol{\beta}^*)\big]/n$, where $\mathbb{E}_{\boldsymbol{\beta}^*}(\cdot)$ means the expectation is taken under the model with parameter $\boldsymbol{\beta}^*$. The following lemma reveals the connection of $\nabla_1 Q_n(\cdot;\cdot)$ in (2.9) and $T_n(\cdot)$ in (2.11) with the score function and Fisher information.

**Lemma 2.1.** For the true parameter $\boldsymbol{\beta}^*$ and any $\boldsymbol{\beta} \in \mathbb{R}^d$, it holds that

$$\nabla_1 Q_n(\boldsymbol{\beta}; \boldsymbol{\beta}) = \nabla \ell_n(\boldsymbol{\beta})/n, \quad \text{and} \quad \mathbb{E}_{\boldsymbol{\beta}^*}\big[T_n(\boldsymbol{\beta}^*)\big] = -I(\boldsymbol{\beta}^*) = \mathbb{E}_{\boldsymbol{\beta}^*}\big[\nabla^2 \ell_n(\boldsymbol{\beta}^*)\big]/n. \tag{2.12}$$

*Proof.* See §C.1 for details. □

Recall that the log-likelihood $\ell_n(\boldsymbol{\beta})$ defined in (2.2) is difficult to evaluate due to the unobserved latent variable. Lemma 2.1 provides a feasible way to calculate or estimate the corresponding score function and Fisher information, since $Q_n(\cdot;\cdot)$ and $T_n(\cdot)$ have closed forms. The geometric intuition behind Lemma 2.1 can be understood as follows. By (2.4) and (2.5) we have

$$\ell_n(\boldsymbol{\beta}) \geq \ell_n(\boldsymbol{\beta}') + n \cdot Q_n(\boldsymbol{\beta}; \boldsymbol{\beta}') - \sum_{i=1}^{n} \int_{\mathcal{Z}} k_{\boldsymbol{\beta}'}(\mathbf{z} \mid \mathbf{y}_i) \cdot \log f_{\boldsymbol{\beta}'}(\mathbf{y}_i, \mathbf{z})\, \mathrm{d}\mathbf{z}. \tag{2.13}$$

By (2.12), both sides of (2.13) have the same gradient with respect to $\boldsymbol{\beta}$ at $\boldsymbol{\beta}' = \boldsymbol{\beta}$. Furthermore, by (2.5), (2.13) becomes an equality for $\boldsymbol{\beta}' = \boldsymbol{\beta}$. Therefore, the lower bound function on the right-hand side of (2.13) is tangent to $\ell_n(\boldsymbol{\beta})$ at $\boldsymbol{\beta}' = \boldsymbol{\beta}$. Meanwhile, according to (2.8), $T_n(\boldsymbol{\beta})$ defines a modified curvature for the right-hand side of (2.13), which is obtained by taking derivative with respect to $\boldsymbol{\beta}$, then setting $\boldsymbol{\beta}' = \boldsymbol{\beta}$ and taking the second order derivative with respect to $\boldsymbol{\beta}$. The second equation in (2.12) shows that the obtained curvature equals the curvature of $\ell_n(\boldsymbol{\beta})$ at $\boldsymbol{\beta} = \boldsymbol{\beta}^*$ in expectation (up to a renormalization factor of $n$). Therefore, $\nabla_1 Q_n(\boldsymbol{\beta}; \boldsymbol{\beta})$ gives the score function and $T_n(\boldsymbol{\beta}^*)$ gives a good estimate of the Fisher information $I(\boldsymbol{\beta}^*)$. Since $\boldsymbol{\beta}^*$ is unknown in practice, later we will use $T_n\big(\widehat{\boldsymbol{\beta}}\big)$ or $T_n\big(\widehat{\boldsymbol{\beta}}_0\big)$ to approximate $T_n(\boldsymbol{\beta}^*)$.

In the presence of the high dimensional nuisance parameter $\boldsymbol{\gamma}^* \in \mathbb{R}^{d-1}$, the classical score test is no longer applicable. In detail, the score test for $H_0 : \alpha^* = 0$ relies on the following Taylor expansion of the score function $\partial \ell_n(\cdot)/\partial \alpha$

$$\frac{1}{\sqrt{n}} \cdot \frac{\partial \ell_n(\overline{\boldsymbol{\beta}}_0)}{\partial \alpha} = \frac{1}{\sqrt{n}} \cdot \frac{\partial \ell_n(\boldsymbol{\beta}^*)}{\partial \alpha} + \frac{1}{\sqrt{n}} \cdot \frac{\partial^2 \ell_n(\boldsymbol{\beta}^*)}{\partial \alpha \partial \boldsymbol{\gamma}} \cdot (\overline{\boldsymbol{\gamma}} - \boldsymbol{\gamma}^*) + \overline{R}. \tag{2.14}$$

Here $\boldsymbol{\beta}^* = \big[0, (\boldsymbol{\gamma}^*)^\top\big]^\top$, $\overline{R}$ denotes the remainder term and $\overline{\boldsymbol{\beta}}_0 = \big(0, \overline{\boldsymbol{\gamma}}^\top\big)^\top$, where $\overline{\boldsymbol{\gamma}}$ is an estimator of the nuisance parameter $\boldsymbol{\gamma}^*$. The asymptotic normality of $1/\sqrt{n} \cdot \partial \ell_n\big(\overline{\boldsymbol{\beta}}_0\big)/\partial \alpha$ in (2.14) relies on the fact that $1/\sqrt{n} \cdot \partial \ell_n\big(\boldsymbol{\beta}_0^*\big)/\partial \alpha$ and $\sqrt{n} \cdot (\overline{\boldsymbol{\gamma}} - \boldsymbol{\gamma}^*)$ are jointly normal asymptotically and $\overline{R}$ is $o_{\mathbb{P}}(1)$. In low dimensional settings, such a necessary condition holds for $\overline{\boldsymbol{\gamma}}$ being the maximum likelihood estimator. However, in high dimensional settings, the maximum likelihood estimator can't guarantee that $\overline{R}$ is $o_{\mathbb{P}}(1)$, since $\|\overline{\boldsymbol{\gamma}} - \boldsymbol{\gamma}^*\|_2$ can be large due to the curse of dimensionality. Meanwhile, for $\overline{\boldsymbol{\gamma}}$ being sparsity-type estimators, in general the asymptotic normality of $\sqrt{n} \cdot (\overline{\boldsymbol{\gamma}} - \boldsymbol{\gamma}^*)$ doesn't hold. For example, let $\overline{\boldsymbol{\gamma}}$ be $\widehat{\boldsymbol{\gamma}}$, where $\widehat{\boldsymbol{\gamma}} \in \mathbb{R}^{d-1}$ is the subvector of $\widehat{\boldsymbol{\beta}}$, i.e., the estimator attained by the proposed high dimensional EM algorithm. Note that $\widehat{\boldsymbol{\gamma}}$ has many zero entries due to the truncation step. As $n \to \infty$, some entries of $\sqrt{n} \cdot (\widehat{\boldsymbol{\gamma}} - \boldsymbol{\gamma}^*)$ have limiting distributions with point mass at zero. Clearly, this limiting distribution is not Gaussian with nonzero variance. In fact, for a similar setting of high dimensional linear regression, Knight and Fu (2000) illustrate that for $\boldsymbol{\gamma}^\sharp$ being a subvector of the Lasso estimator and $\boldsymbol{\gamma}^*$ being the corresponding subvector of the true parameter, the limiting distribution of $\sqrt{n} \cdot (\boldsymbol{\gamma}^\sharp - \boldsymbol{\gamma}^*)$ is not Gaussian.

The decorrelated score function defined in (2.9) successfully addresses the above issues. In detail, according to (2.12) in Lemma 2.1 we have

$$\sqrt{n} \cdot S_n\big(\widehat{\boldsymbol{\beta}}_0, \lambda\big) = \frac{1}{\sqrt{n}} \cdot \frac{\partial \ell_n\big(\widehat{\boldsymbol{\beta}}_0\big)}{\partial \alpha} - \frac{1}{\sqrt{n}} \cdot w\big(\widehat{\boldsymbol{\beta}}_0, \lambda\big)^\top \cdot \frac{\partial \ell_n\big(\widehat{\boldsymbol{\beta}}_0\big)}{\partial \boldsymbol{\gamma}}. \tag{2.15}$$

Intuitively, if we replace $w\big(\widehat{\boldsymbol{\beta}}_0, \lambda\big)$ with $\mathbf{w} \in \mathbb{R}^{d-1}$ that satisfies

$$\mathbf{w}^\top \cdot \frac{\partial^2 \ell_n(\boldsymbol{\beta}^*)}{\partial^2 \boldsymbol{\gamma}} = \frac{\partial^2 \ell_n(\boldsymbol{\beta}^*)}{\partial \alpha \partial \boldsymbol{\gamma}}, \tag{2.16}$$

we have the following Taylor expansion of the decorrelated score function

$$\begin{aligned} \frac{1}{\sqrt{n}} \cdot \frac{\partial \ell_n\big(\widehat{\boldsymbol{\beta}}_0\big)}{\partial \alpha} - \frac{\mathbf{w}^\top}{\sqrt{n}} \cdot \frac{\partial \ell_n\big(\widehat{\boldsymbol{\beta}}_0\big)}{\partial \boldsymbol{\gamma}} = & \overbrace{\frac{1}{\sqrt{n}} \cdot \frac{\partial \ell_n(\boldsymbol{\beta}^*)}{\partial \alpha} - \frac{\mathbf{w}^\top}{\sqrt{n}} \cdot \frac{\partial \ell_n(\boldsymbol{\beta}^*)}{\partial \boldsymbol{\gamma}}}^{\text{(i)}} \\ & + \underbrace{\frac{1}{\sqrt{n}} \cdot \frac{\partial^2 \ell_n(\boldsymbol{\beta}^*)}{\partial \alpha \partial \boldsymbol{\gamma}} \cdot (\widehat{\boldsymbol{\gamma}} - \boldsymbol{\gamma}^*) - \frac{\mathbf{w}^\top}{\sqrt{n}} \cdot \frac{\partial^2 \ell_n(\boldsymbol{\beta}^*)}{\partial^2 \boldsymbol{\gamma}} \cdot (\widehat{\boldsymbol{\gamma}} - \boldsymbol{\gamma}^*)}_{\text{(i)}} + \widetilde{R}, \end{aligned} \tag{2.17}$$

where term (ii) is zero by (2.16). Therefore, we no longer require the asymptotic normality of $\widehat{\boldsymbol{\gamma}} - \boldsymbol{\gamma}^*$. Also, we will prove that the new remainder term $\widetilde{R}$ in (2.17) is $o_{\mathbb{P}}(1)$, since $\widehat{\boldsymbol{\gamma}}$ has a fast statistical rate of convergence. Now we only need to find the $\mathbf{w}$ that satisfies (2.16). Nevertheless, it is difficult to calculate the second order derivatives in (2.16), because it is hard to evaluate $\ell_n(\cdot)$. According to (2.12), we use the submatrices of $T_n(\cdot)$ to approximate the derivatives in (2.16). Since $\big[T_n(\boldsymbol{\beta})\big]_{\boldsymbol{\gamma},\boldsymbol{\gamma}}$ is not invertible in high dimensions, we use the Dantzig selector in (2.10) to approximately solve the linear system in (2.16). Based on this intuition, we can expect that $\sqrt{n}\cdot S_n\big(\widehat{\boldsymbol{\beta}}_0,\lambda\big)$ is asymptotically normal, since term (i) in (2.17) is a (rescaled) average of $n$ i.i.d. random variables for which we can apply the central limit theorem. Besides, we will prove that $-\big[T_n\big(\widehat{\boldsymbol{\beta}}_0\big)\big]_{\alpha|\boldsymbol{\gamma}}$ in (2.11) is a consistent estimator of $\sqrt{n}\cdot S_n\big(\widehat{\boldsymbol{\beta}}_0,\lambda\big)$'s asymptotic variance. Hence, we can expect that the decorrelated score statistic in (2.11) is asymptotically $N(0,1)$.

From a high-level perspective, we can view $w\big(\widehat{\boldsymbol{\beta}}_0,\lambda\big)^\top\cdot\partial\ell_n\big(\widehat{\boldsymbol{\beta}}_0\big)/\partial\boldsymbol{\gamma}$ in (2.15) as the projection of $\partial\ell_n\big(\widehat{\boldsymbol{\beta}}_0\big)/\partial\alpha$ onto the span of $\partial\ell_n\big(\widehat{\boldsymbol{\beta}}_0\big)/\partial\boldsymbol{\gamma}$, where $w\big(\widehat{\boldsymbol{\beta}}_0,\lambda\big)$ is the projection coefficient. Intuitively, such a projection guarantees that in (2.15), $S_n\big(\widehat{\boldsymbol{\beta}}_0,\lambda\big)$ is orthogonal (uncorrelated) with $\partial\ell_n\big(\widehat{\boldsymbol{\beta}}_0\big)/\partial\boldsymbol{\gamma}$, i.e., the score function with respect to the nuisance parameter $\boldsymbol{\gamma}$. In this way, the projection corrects the effects of the high dimensional nuisance parameter. According to this intuition of decorrelation, we name $S_n\big(\widehat{\boldsymbol{\beta}}_0,\lambda\big)$ as the decorrelated score function.

**Decorrelated Wald Test:** Based on the decorrelated score test, we propose the decorrelated Wald test. In detail, recall that $T_n(\cdot)$ is defined in (2.8). Let

$$\bar{\alpha}\big(\widehat{\boldsymbol{\beta}},\lambda\big)=\widehat{\alpha}-\Big\{\big[T_n\big(\widehat{\boldsymbol{\beta}}\big)\big]_{\alpha,\alpha}-w\big(\widehat{\boldsymbol{\beta}},\lambda\big)^\top\cdot\big[T_n\big(\widehat{\boldsymbol{\beta}}\big)\big]_{\boldsymbol{\gamma},\alpha}\Big\}^{-1}\cdot S_n\big(\widehat{\boldsymbol{\beta}},\lambda\big), \tag{2.18}$$

where $S_n(\cdot,\cdot)$ is the decorrelated score function in (2.9), $w(\cdot,\cdot)$ is defined in (2.10) and $\widehat{\boldsymbol{\beta}}=\big(\widehat{\alpha},\widehat{\boldsymbol{\gamma}}^\top\big)^\top$ is the estimator obtained by Algorithm 1. For testing the null hypothesis $H_0:\alpha^*=0$, we define the decorrelated Wald statistic as

$$\sqrt{n}\cdot\bar{\alpha}\big(\widehat{\boldsymbol{\beta}},\lambda\big)\cdot\Big\{-\big[T_n\big(\widehat{\boldsymbol{\beta}}\big)\big]_{\alpha|\boldsymbol{\gamma}}\Big\}^{1/2}, \tag{2.19}$$

where $\big[T_n\big(\widehat{\boldsymbol{\beta}}\big)\big]_{\alpha|\boldsymbol{\gamma}}$ is defined by replacing $\widehat{\boldsymbol{\beta}}_0$ with $\widehat{\boldsymbol{\beta}}$ in (2.11). In §4 we will prove that this statistic is asymptotically $N(0,1)$. Consequently, the decorrelated Wald test with significance level $\delta\in(0,1)$ takes the form

$$\psi_W(\delta)=\mathbb{1}\left\{\sqrt{n}\cdot\bar{\alpha}\big(\widehat{\boldsymbol{\beta}},\lambda\big)\cdot\Big\{-\big[T_n\big(\widehat{\boldsymbol{\beta}}\big)\big]_{\alpha|\boldsymbol{\gamma}}\Big\}^{1/2}\notin\big[-\Phi^{-1}(1-\delta/2),\Phi^{-1}(1-\delta/2)\big]\right\},$$

where $\Phi^{-1}(\cdot)$ is the inverse function of the Gaussian cumulative distribution function. If $\psi_W(\delta)=1$, we reject the null hypothesis $H_0:\alpha^*=0$. The associated p-value takes the form

$$\text{p-value}=2\cdot\left[1-\Phi\left(\left|\sqrt{n}\cdot\bar{\alpha}\big(\widehat{\boldsymbol{\beta}},\lambda\big)\cdot\Big\{-\big[T_n\big(\widehat{\boldsymbol{\beta}}\big)\big]_{\alpha|\boldsymbol{\gamma}}\Big\}^{1/2}\right|\right)\right].$$

In more general settings where $\alpha^*$ is possibly nonzero, in §4.1 we will prove that $\sqrt{n}\cdot\big[\bar{\alpha}\big(\widehat{\boldsymbol{\beta}},\lambda\big)-\alpha^*\big]\cdot\Big\{-\big[T_n\big(\widehat{\boldsymbol{\beta}}\big)\big]_{\alpha|\boldsymbol{\gamma}}\Big\}^{1/2}$ is asymptotically $N(0,1)$. Hence, we construct the two-sided confidence interval for $\alpha^*$ with confidence level $1-\delta$ as

$$\left[\bar{\alpha}\big(\widehat{\boldsymbol{\beta}},\lambda\big)-\frac{\Phi^{-1}(1-\delta/2)}{\sqrt{-n\cdot\big[T_n\big(\widehat{\boldsymbol{\beta}}\big)\big]_{\alpha|\boldsymbol{\gamma}}}},\ \ \bar{\alpha}\big(\widehat{\boldsymbol{\beta}},\lambda\big)+\frac{\Phi^{-1}(1-\delta/2)}{\sqrt{-n\cdot\big[T_n\big(\widehat{\boldsymbol{\beta}}\big)\big]_{\alpha|\boldsymbol{\gamma}}}}\right]. \tag{2.20}$$

The intuition for the decorrelated Wald statistic defined in (2.19) can be understood as follows. For notational simplicity, we define

$$\overline{S}_n(\boldsymbol{\beta},\widehat{\mathbf{w}}) = \bigl[\nabla_1 Q_n(\boldsymbol{\beta};\boldsymbol{\beta})\bigr]_{\alpha} - \widehat{\mathbf{w}}\cdot\bigl[\nabla_1 Q_n(\boldsymbol{\beta};\boldsymbol{\beta})\bigr]_{\boldsymbol{\gamma}}, \quad \text{where } \ \widehat{\mathbf{w}} = w\bigl(\widehat{\boldsymbol{\beta}},\lambda\bigr). \tag{2.21}$$

By the definitions in (2.9) and (2.21), we have $\overline{S}_n\bigl(\widehat{\boldsymbol{\beta}},\widehat{\mathbf{w}}\bigr) = S_n\bigl(\widehat{\boldsymbol{\beta}},\lambda\bigr)$. According to the same intuition for the asymptotic normality of $\sqrt{n}\cdot S_n\bigl(\widehat{\boldsymbol{\beta}}_0,\lambda\bigr)$ in the decorrelated score test, we can easily establish the asymptotic normality of $\sqrt{n}\cdot S_n\bigl(\widehat{\boldsymbol{\beta}},\lambda\bigr) = \sqrt{n}\cdot \overline{S}_n\bigl(\widehat{\boldsymbol{\beta}},\widehat{\mathbf{w}}\bigr)$. Based on the proof for the classical Wald test (van der Vaart, 2000), we can further establish the asymptotic normality of $\sqrt{n}\cdot\underline{\alpha}$, where $\underline{\alpha}$ is defined as the solution to

$$\overline{S}_n\bigl[\bigl(\alpha,\widehat{\boldsymbol{\gamma}}^\top\bigr)^\top,\widehat{\mathbf{w}}\bigr] = 0. \tag{2.22}$$

Nevertheless, it is difficult to calculate $\underline{\alpha}$. Instead, we consider the first order Taylor approximation

$$\overline{S}_n\bigl(\widehat{\boldsymbol{\beta}},\widehat{\mathbf{w}}\bigr) + (\alpha-\widehat{\alpha})\cdot\bigl[\nabla\overline{S}_n\bigl(\widehat{\boldsymbol{\beta}},\widehat{\mathbf{w}}\bigr)\bigr]_{\alpha} = 0, \quad \text{where } \ \widehat{\boldsymbol{\beta}} = \bigl(\widehat{\alpha},\widehat{\boldsymbol{\gamma}}^\top\bigr)^\top. \tag{2.23}$$

Here $\widehat{\mathbf{w}}$ is defined in (2.21) and the gradient is taken with respect to $\boldsymbol{\beta}$ in (2.21). According to (2.8) and (2.21), we have that in (2.23),

$$\begin{aligned}\bigl[\nabla\overline{S}_n\bigl(\widehat{\boldsymbol{\beta}},\widehat{\mathbf{w}}\bigr)\bigr]_{\alpha} &= \bigl[\nabla_{1,1}^2 Q_n\bigl(\widehat{\boldsymbol{\beta}};\widehat{\boldsymbol{\beta}}\bigr)+\nabla_{1,2}^2 Q_n\bigl(\widehat{\boldsymbol{\beta}};\widehat{\boldsymbol{\beta}}\bigr)\bigr]_{\alpha,\alpha} - \widehat{\mathbf{w}}^\top\cdot\bigl[\nabla_{1,1}^2 Q_n\bigl(\widehat{\boldsymbol{\beta}};\widehat{\boldsymbol{\beta}}\bigr)+\nabla_{1,2}^2 Q_n\bigl(\widehat{\boldsymbol{\beta}};\widehat{\boldsymbol{\beta}}\bigr)\bigr]_{\boldsymbol{\gamma},\alpha}\\ &= \bigl[T_n\bigl(\widehat{\boldsymbol{\beta}}\bigr)\bigr]_{\alpha,\alpha} - \widehat{\mathbf{w}}^\top\cdot\bigl[T_n\bigl(\widehat{\boldsymbol{\beta}}\bigr)\bigr]_{\boldsymbol{\gamma},\alpha}, \qquad \text{where } \ \widehat{\mathbf{w}} = w\bigl(\widehat{\boldsymbol{\beta}},\lambda\bigr).\end{aligned}$$

Hence, $\overline{\alpha}\bigl(\widehat{\boldsymbol{\beta}},\lambda\bigr)$ defined in (2.18) is the solution to (2.23). Alternatively, we can view $\overline{\alpha}\bigl(\widehat{\boldsymbol{\beta}},\lambda\bigr)$ as the output of one Newton-Raphson iteration applied to $\widehat{\alpha}$. Since (2.23) approximates (2.22), intuitively $\overline{\alpha}\bigl(\widehat{\boldsymbol{\beta}},\lambda\bigr)$ has similar statistical properties as $\underline{\alpha}$, i.e., the solution to (2.22). Therefore, we can expect that $\sqrt{n}\cdot\overline{\alpha}\bigl(\widehat{\boldsymbol{\beta}},\lambda\bigr)$ is asymptotically normal. Besides, we will prove that $-\bigl[T_n\bigl(\widehat{\boldsymbol{\beta}}_0\bigr)\bigr]_{\alpha|\boldsymbol{\gamma}}^{-1}$ is a consistent estimator of the asymptotic variance of $\sqrt{n}\cdot\overline{\alpha}\bigl(\widehat{\boldsymbol{\beta}},\lambda\bigr)$. Thus, the decorrelated Wald statistic in (2.19) is asymptotically $N(0,1)$. It is worth noting that, although the intuition of the decorrelated Wald statistic is the same as the one-step estimator proposed by Bickel (1975), here we don't require the $\sqrt{n}$-consistency of the initial estimator $\widehat{\alpha}$ to achieve the asymptotic normality of $\overline{\alpha}\bigl(\widehat{\boldsymbol{\beta}},\lambda\bigr)$.

## 2.3 Applications to Latent Variable Models

In the sequel, we introduce three latent variable models as examples. To apply the high dimensional EM algorithm in §2.1 and the methods for asymptotic inference in §2.2, we only need to specify the forms of $Q_n(\cdot;\cdot)$ defined in (2.5), $M_n(\cdot)$ in Algorithms 2 and 3, and $T_n(\cdot)$ in (2.8) for each model.

**Gaussian Mixture Model:** Let $\mathbf{y}_1,\ldots,\mathbf{y}_n$ be the $n$ i.i.d. realizations of $\boldsymbol{Y}\in\mathbb{R}^d$ and

$$\boldsymbol{Y} = Z\cdot\boldsymbol{\beta}^* + \boldsymbol{V}. \tag{2.24}$$

Here $Z$ is a Rademacher random variable, i.e., $\mathbb{P}(Z=+1)=\mathbb{P}(Z=-1)=1/2$, and $\boldsymbol{V}\sim N(\mathbf{0},\sigma^2\cdot\mathbf{I}_d)$ is independent of $Z$, where $\sigma$ is the standard deviation. We suppose $\sigma$ is known. In high dimensional settings, we assume that $\boldsymbol{\beta}^*\in\mathbb{R}^d$ is sparse. To avoid the degenerate case in which the two Gaussians in the mixture are identical, here we suppose that $\boldsymbol{\beta}^* \neq \mathbf{0}$. See §A.1 for the detailed forms of $Q_n(\cdot;\cdot)$, $M_n(\cdot)$ and $T_n(\cdot)$.

**Mixture of Regression Model:** We assume that $Y \in \mathbb{R}$ and $\boldsymbol{X} \in \mathbb{R}^d$ satisfy

$$Y = Z \cdot \boldsymbol{X}^\top \boldsymbol{\beta}^* + V, \tag{2.25}$$

where $\boldsymbol{X} \sim N(\mathbf{0}, \mathbf{I}_d)$, $V \sim N(0, \sigma^2)$ and $Z$ is a Rademacher random variable. Here $\boldsymbol{X}$, $V$ and $Z$ are independent. In the high dimensional regime, we assume $\boldsymbol{\beta}^* \in \mathbb{R}^d$ is sparse. To avoid the degenerate case, we suppose $\boldsymbol{\beta}^* \neq \mathbf{0}$. In addition, we assume that $\sigma$ is known. See §A.2 for the detailed forms of $Q_n(\cdot;\cdot)$, $M_n(\cdot)$ and $T_n(\cdot)$.

**Regression with Missing Covariates:** We assume that $Y \in \mathbb{R}$ and $\boldsymbol{X} \in \mathbb{R}^d$ satisfy

$$Y = \boldsymbol{X}^\top \boldsymbol{\beta}^* + V, \tag{2.26}$$

where $\boldsymbol{X} \sim N(\mathbf{0}, \mathbf{I}_d)$ and $V \sim N(0, \sigma^2)$ are independent. We assume $\boldsymbol{\beta}^*$ is sparse. Let $\mathbf{x}_1, \ldots, \mathbf{x}_n$ be the $n$ realizations of $\boldsymbol{X}$. We assume that each coordinate of $\mathbf{x}_i$ is missing (unobserved) independently with probability $p_{\mathrm{m}} \in [0, 1)$. We treat the missing covariates as the latent variable and suppose that $\sigma$ is known. See §A.3 for the detailed forms of $Q_n(\cdot;\cdot)$, $M_n(\cdot)$ and $T_n(\cdot)$.

## 3 Theory of Computation and Estimation

Before we lay out the main results, we introduce three technical conditions, which will significantly simplify our presentation. These conditions will be verified for the two implementations of the high dimensional EM algorithm and three latent variable models.

The first two conditions, proposed by Balakrishnan et al. (2014), characterize the properties of the population version lower bound function $Q(\cdot;\cdot)$, i.e., the expectation of $Q_n(\cdot;\cdot)$ defined in (2.5). We define the respective population version M-step as follows. For the maximization implementation of the M-step (Algorithm 2), we define

$$M(\boldsymbol{\beta}) = \operatorname*{argmax}_{\boldsymbol{\beta}'} Q(\boldsymbol{\beta}'; \boldsymbol{\beta}). \tag{3.1}$$

For the gradient ascent implementation of the M-step (Algorithm 3), we define

$$M(\boldsymbol{\beta}) = \boldsymbol{\beta} + \eta \cdot \nabla_1 Q(\boldsymbol{\beta}; \boldsymbol{\beta}), \tag{3.2}$$

where $\eta > 0$ is the stepsize in Algorithm 3. Hereafter, we employ $\mathcal{B}$ to denote the basin of attraction, i.e., the local region in which the proposed high dimensional EM algorithm enjoys desired statistical and computational guarantees.

**Condition 3.1.** We define two versions of this condition.

- Lipschitz-Gradient-1$(\gamma_1, \mathcal{B})$. For the true parameter $\boldsymbol{\beta}^*$ and any $\boldsymbol{\beta} \in \mathcal{B}$, we have

$$\big\| \nabla_1 Q\big[M(\boldsymbol{\beta}); \boldsymbol{\beta}^*\big] - \nabla_1 Q\big[M(\boldsymbol{\beta}); \boldsymbol{\beta}\big] \big\|_2 \leq \gamma_1 \cdot \|\boldsymbol{\beta} - \boldsymbol{\beta}^*\|_2, \tag{3.3}$$

  where $M(\cdot)$ is the population version M-step (maximization implementation) defined in (3.1).

- Lipschitz-Gradient-2($\gamma_2, \mathcal{B}$). For the true parameter $\boldsymbol{\beta}^*$ and any $\boldsymbol{\beta} \in \mathcal{B}$, we have

$$\big\|\nabla_1 Q(\boldsymbol{\beta}; \boldsymbol{\beta}^*) - \nabla_1 Q(\boldsymbol{\beta}; \boldsymbol{\beta})\big\|_2 \leq \gamma_2 \cdot \|\boldsymbol{\beta} - \boldsymbol{\beta}^*\|_2. \tag{3.4}$$

Condition 3.1 defines a variant of Lipschitz continuity for $\nabla_1 Q(\cdot;\cdot)$. Note that in (3.3) and (3.4), the gradient is taken with respect to the first variable of $Q(\cdot;\cdot)$. Meanwhile, the Lipschitz continuity is with respect to the second variable of $Q(\cdot;\cdot)$. Also, the Lipschitz property is defined only between the true parameter $\boldsymbol{\beta}^*$ and an arbitrary $\boldsymbol{\beta} \in \mathcal{B}$, rather than between two arbitrary $\boldsymbol{\beta}$'s. In the sequel, we will use (3.3) and (3.4) in the analysis of the two implementations of the M-step respectively.

**Condition 3.2** Concavity-Smoothness($\mu, \nu, \mathcal{B}$). For any $\boldsymbol{\beta}_1, \boldsymbol{\beta}_2 \in \mathcal{B}$, $Q(\cdot; \boldsymbol{\beta}^*)$ is $\mu$-smooth, i.e.,

$$Q(\boldsymbol{\beta}_1; \boldsymbol{\beta}^*) \geq Q(\boldsymbol{\beta}_2; \boldsymbol{\beta}^*) + (\boldsymbol{\beta}_1 - \boldsymbol{\beta}_2)^\top \cdot \nabla_1 Q(\boldsymbol{\beta}_2; \boldsymbol{\beta}^*) - \mu/2 \cdot \|\boldsymbol{\beta}_2 - \boldsymbol{\beta}_1\|_2^2, \tag{3.5}$$

and $\nu$-strongly concave, i.e.,

$$Q(\boldsymbol{\beta}_1; \boldsymbol{\beta}^*) \leq Q(\boldsymbol{\beta}_2; \boldsymbol{\beta}^*) + (\boldsymbol{\beta}_1 - \boldsymbol{\beta}_2)^\top \cdot \nabla_1 Q(\boldsymbol{\beta}_2; \boldsymbol{\beta}^*) - \nu/2 \cdot \|\boldsymbol{\beta}_2 - \boldsymbol{\beta}_1\|_2^2. \tag{3.6}$$

This condition indicates that, when the second variable of $Q(\cdot;\cdot)$ is fixed to be $\boldsymbol{\beta}^*$, the function is 'sandwiched' between two quadratic functions. Conditions 3.1 and 3.2 are essential to establishing the desired optimization results. In detail, Condition 3.2 ensures that standard convex optimization results for strongly convex and smooth objective functions, e.g., in Nesterov (2004), can be applied to $-Q(\cdot; \boldsymbol{\beta}^*)$. Since our analysis will not only involve $Q(\cdot; \boldsymbol{\beta}^*)$, but also $Q(\cdot; \boldsymbol{\beta})$ for all $\boldsymbol{\beta} \in \mathcal{B}$, we also need to quantify the difference between $Q(\cdot; \boldsymbol{\beta}^*)$ and $Q(\cdot; \boldsymbol{\beta})$ by Condition 3.1. It suggests that, this difference can be controlled in the sense that $\nabla_1 Q(\cdot; \boldsymbol{\beta})$ is Lipschitz with respect to $\boldsymbol{\beta}$. Consequently, for any $\boldsymbol{\beta} \in \mathcal{B}$, the behavior of $Q(\cdot; \boldsymbol{\beta})$ mimics that of $Q(\cdot; \boldsymbol{\beta}^*)$. Hence, standard convex optimization results can be adapted to analyze $-Q(\cdot; \boldsymbol{\beta})$ for any $\boldsymbol{\beta} \in \mathcal{B}$.

The third condition characterizes the statistical error between the sample version and population version M-steps, i.e., $M_n(\cdot)$ defined in Algorithms 2 and 3, and $M(\cdot)$ in (3.1) and (3.2). Recall that $\|\cdot\|_0$ denotes the total number of nonzero entries in a vector.

**Condition 3.3** Statistical-Error($\epsilon, \delta, s, n, \mathcal{B}$). For any fixed $\boldsymbol{\beta} \in \mathcal{B}$ with $\|\boldsymbol{\beta}\|_0 \leq s$, we have that

$$\big\|M(\boldsymbol{\beta}) - M_n(\boldsymbol{\beta})\big\|_\infty \leq \epsilon \tag{3.7}$$

holds with probability at least $1 - \delta$. Here $\epsilon > 0$ possibly depends on $\delta$, sparsity level $s$, sample size $n$, dimension $d$, as well as the basin of attraction $\mathcal{B}$.

In (3.7), the statistical error $\epsilon$ quantifies the $\ell_\infty$-norm of the difference between the population version and sample version M-steps. Particularly, we constrain the input $\boldsymbol{\beta}$ of $M(\cdot)$ and $M_n(\cdot)$ to be $s$-sparse. Such a condition is different from the one used by Balakrishnan et al. (2014). In detail, they quantify the statistical error with the $\ell_2$-norm and don't constrain the input of $M(\cdot)$ and $M_n(\cdot)$ to be sparse. Consequently, our subsequent statistical analysis differs from theirs. The reason we use the $\ell_\infty$-norm is that, it characterizes the more refined entrywise statistical error, which converges at a fast rate of $\sqrt{\log d/n}$ (possibly with extra factors depending on specific models). In comparison, the $\ell_2$-norm statistical error converges at a slow rate of $\sqrt{d/n}$, which doesn't decrease to zero as $n$ increases when $d \gg n$. Moreover, the fine-grained entrywise statistical error is crucial to quantifying the effects of the truncation step (line 6 of Algorithm 1) on the iterative solution sequence.

### 3.1 Main Results

Equipped with Conditions 3.1-3.3, we now lay out the computational and statistical results for the high dimensional EM algorithm. To simplify the technical analysis of the algorithm, we focus on its resampling version, which is illustrated in Algorithm 4.

**Algorithm 4** High Dimensional EM Algorithm with Resampling.

**Parameter:** Sparsity Parameter $\widehat{s}$, Maximum Number of Iterations $T$

**Initialization:** $\widehat{\mathcal{S}}^{\text{init}} \leftarrow \text{supp}\big(\boldsymbol{\beta}^{\text{init}}, \widehat{s}\big), \quad \boldsymbol{\beta}^{(0)} \leftarrow \text{trunc}\big(\boldsymbol{\beta}^{\text{init}}, \widehat{\mathcal{S}}^{\text{init}}\big)$,

$\big\{\text{supp}(\cdot,\cdot)$ and $\text{trunc}(\cdot,\cdot)$ are defined in (2.6) and (2.7)$\big\}$

Split the Dataset into $T$ Subsets of Size $n/T$

$\big\{$Without loss of generality, we assume $n/T$ is an integer$\big\}$

**For** $t = 0$ to $T-1$

  **E-step:** Evaluate $Q_{n/T}\big(\boldsymbol{\beta};\boldsymbol{\beta}^{(t)}\big)$ with the $t$-th Data Subset

  **M-step:** $\boldsymbol{\beta}^{(t+0.5)} \leftarrow M_{n/T}\big(\boldsymbol{\beta}^{(t)}\big)$

  $\big\{M_{n/T}(\cdot)$ is implemented as in Algorithm 2 or 3 with $Q_n(\cdot;\cdot)$ replaced by $Q_{n/T}(\cdot;\cdot)\big\}$

  **T-step:** $\widehat{\mathcal{S}}^{(t+0.5)} \leftarrow \text{supp}\big(\boldsymbol{\beta}^{(t+0.5)}, \widehat{s}\big), \quad \boldsymbol{\beta}^{(t+1)} \leftarrow \text{trunc}\big(\boldsymbol{\beta}^{(t+0.5)}, \widehat{\mathcal{S}}^{(t+0.5)}\big)$

**End For**

**Output:** $\widehat{\boldsymbol{\beta}} \leftarrow \boldsymbol{\beta}^{(T)}$

**Theorem 3.4.** For Algorithm 4, we define $\mathcal{B} = \big\{\boldsymbol{\beta} : \|\boldsymbol{\beta} - \boldsymbol{\beta}^*\|_2 \le R\big\}$, where $R = \kappa \cdot \|\boldsymbol{\beta}^*\|_2$ for some $\kappa \in (0,1)$. We assume that Condition $\mathsf{Concavity\text{-}Smoothness}(\mu,\nu,\mathcal{B})$ holds and $\big\|\boldsymbol{\beta}^{\text{init}} - \boldsymbol{\beta}^*\big\|_2 \le R/2$.

- For the maximization implementation of the M-step (Algorithm 2), we suppose that Condition $\mathsf{Lipschitz\text{-}Gradient\text{-}1}(\gamma_1,\mathcal{B})$ holds with $\gamma_1/\nu \in (0,1)$ and

$$\widehat{s} = \left\lceil C \cdot \max\left\{\frac{16}{(\nu/\gamma_1 - 1)^2}, \frac{4\cdot(1+\kappa)^2}{(1-\kappa)^2}\right\} \cdot s^* \right\rceil, \tag{3.8}$$

$$\big(\sqrt{\widehat{s}} + C'/\sqrt{1-\kappa}\cdot\sqrt{s^*}\big)\cdot\epsilon \le \min\left\{\big(1-\sqrt{\gamma_1/\nu}\big)^2 \cdot R,\ \frac{(1-\kappa)^2}{2\cdot(1+\kappa)} \cdot \|\boldsymbol{\beta}^*\|_2\right\}, \tag{3.9}$$

  where $C \ge 1$ and $C' > 0$ are absolute constants. Under Condition $\mathsf{Statistical\text{-}Error}(\epsilon, \delta/T, \widehat{s}, n/T, \mathcal{B})$ we have that, for $t = 1, \ldots, T$,

$$\big\|\boldsymbol{\beta}^{(t)} - \boldsymbol{\beta}^*\big\|_2 \le \underbrace{(\gamma_1/\nu)^{t/2}\cdot R}_{\text{Optimization Error}} + \underbrace{\frac{\big(\sqrt{\widehat{s}} + C'/\sqrt{1-\kappa}\cdot\sqrt{s^*}\big)\cdot\epsilon}{1-\sqrt{\gamma_1/\nu}}}_{\text{Statistical Error}} \tag{3.10}$$

  holds with probability at least $1-\delta$, where $C'$ is the same constant as in (3.9).

- For the gradient ascent implementation of the M-step (Algorithm 3), we suppose that Condition $\mathsf{Lipschitz\text{-}Gradient\text{-}2}(\gamma_2,\mathcal{B})$ holds with $1 - 2\cdot(\nu-\gamma_2)/(\nu+\mu) \in (0,1)$ and the stepsize in Algorithm

3 is set to $\eta = 2/(\nu+\mu)$. Meanwhile, we assume that

$$\widehat{s} = \left\lceil C \cdot \max\left\{ \frac{16}{\big\{1/\big[1-2\cdot(\nu-\gamma_2)/(\nu+\mu)\big]-1\big\}^2}, \frac{4\cdot(1+\kappa)^2}{(1-\kappa)^2} \right\} \cdot s^* \right\rceil, \tag{3.11}$$

$$\big(\sqrt{\widehat{s}} + C'/\sqrt{1-\kappa}\cdot\sqrt{s^*}\big)\cdot\epsilon \le \min\left\{ \left(1-\sqrt{1-2\cdot\frac{\nu-\gamma_2}{\nu+\mu}}\right)^2 \cdot R,\ \frac{(1-\kappa)^2}{2\cdot(1+\kappa)}\cdot\|\boldsymbol{\beta}^*\|_2 \right\}, \tag{3.12}$$

where $C \ge 1$ and $C' > 0$ are absolute constants. Under Condition Statistical-Error$(\epsilon, \delta/T, \widehat{s}, n/T, \mathcal{B})$ we have that, for $t = 1, \ldots, T$,

$$\big\|\boldsymbol{\beta}^{(t)} - \boldsymbol{\beta}^*\big\|_2 \le \underbrace{\left(1-2\cdot\frac{\nu-\gamma_2}{\nu+\mu}\right)^{t/2}\cdot R}_{\text{Optimization Error}} + \underbrace{\frac{\big(\sqrt{\widehat{s}}+C'/\sqrt{1-\kappa}\cdot\sqrt{s^*}\big)\cdot\epsilon}{1-\sqrt{1-2\cdot(\nu-\gamma_2)/(\nu+\mu)}}}_{\text{Statistical Error}} \tag{3.13}$$

holds with probability at least $1-\delta$, where $C'$ is the same constant as in (3.12).

*Proof.* See §5.1 for a detailed proof. □

The assumptions in (3.8) and (3.11) state that the sparsity parameter $\widehat{s}$ in Algorithm 4 is chosen to be sufficiently large and also of the same order as the true sparsity level $s^*$. These assumptions, which will be used by Lemma 5.1 in the proof of Theorem 3.4, ensure that the error incurred by the truncation step can be upper bounded. In addition, as will be shown for specific models in §3.2, the error term $\epsilon$ in Condition Statistical-Error$(\epsilon, \delta/T, \widehat{s}, n/T, \mathcal{B})$ decreases as sample size $n$ increases. By the assumptions in (3.8) and (3.11), $\big(\sqrt{\widehat{s}} + C'/\sqrt{1-\kappa}\cdot\sqrt{s^*}\big)$ is of the same order as $\sqrt{s^*}$. Therefore, the assumptions in (3.9) and (3.12) suggest the sample size $n$ is sufficiently large such that $\sqrt{s^*}\cdot\epsilon$ is sufficiently small. These assumptions guarantee that the entire iterative solution sequence remains within the basin of attraction $\mathcal{B}$ in the presence of statistical error. The assumptions in (3.8), (3.9), (3.11), (3.12) will be more explicit as we specify the values of $\gamma_1$, $\gamma_2$, $\nu$, $\mu$ and $\kappa$ for specific models.

Theorem 3.4 illustrates that, the upper bound of the overall estimation error can be decomposed into two terms. The first term is the upper bound of optimization error, which decreases to zero at a geometric rate of convergence, because we have $\gamma_1/\nu < 1$ in (3.10) and $1-2\cdot(\nu-\gamma_2)/(\nu+\mu) < 1$ in (3.13). Meanwhile, the second term is the upper bound of statistical error, which doesn't depend on $t$. Since $\big(\sqrt{\widehat{s}} + C'/\sqrt{1-\kappa}\cdot\sqrt{s^*}\big)$ is of the same order as $\sqrt{s^*}$, this term is proportional to $\sqrt{s^*}\cdot\epsilon$, where $\epsilon$ is the entrywise statistical error between $M(\cdot)$ and $M_n(\cdot)$. We will prove that, for a variety of models and the two implementations of the M-step, $\epsilon$ is roughly of the order $\sqrt{\log d/n}$. (There may be extra factors attached to $\epsilon$ depending on each specific model.) Hence, the statistical error term is roughly of the order $\sqrt{s^*\cdot\log d/n}$. Consequently, for a sufficiently large $t = T$ such that the optimization and statistical error terms in (3.10) or (3.13) are of the same order, the final estimator $\widehat{\boldsymbol{\beta}} = \boldsymbol{\beta}^{(T)}$ attains a (near-)optimal $\sqrt{s^*\cdot\log d/n}$ (possibly with extra factors) rate of convergence for a broad variety of high dimensional latent variable models.

## 3.2 Implications for Specific Models

To establish the corresponding results for specific models under the unified framework, we only need to establish Conditions 3.1-3.3 and determine the key quantities $R$, $\gamma_1$, $\gamma_2$, $\nu$, $\mu$, $\kappa$ and $\epsilon$. Recall that Conditions 3.1 and 3.2 and the models analyzed in our paper are identical to those in Balakrishnan et al. (2014). Meanwhile, note that Conditions 3.1 and 3.2 only involve the population version lower bound function $Q(\cdot;\cdot)$ and M-step $M(\cdot)$. Since Balakrishnan et al. (2014) prove that the quantities in Conditions 3.1 and 3.2 are independent of the dimension $d$ and sample size $n$, their corresponding results can be directly adapted. To establish the final results, it still remains to verify Condition 3.3 for each high dimensional latent variable model.

**Gaussian Mixture Model:** The following lemma, which is proved by Balakrishnan et al. (2014), verifies Conditions 3.1 and 3.2 for Gaussian mixture model. Recall that $\sigma$ is the standard deviation of each individual Gaussian distribution within the mixture.

**Lemma 3.5.** Suppose we have $\|\boldsymbol{\beta}^*\|_2/\sigma \geq r$, where $r > 0$ is a sufficiently large constant that denotes the minimum signal-to-noise ratio. There exists some absolute constant $C > 0$ such that Conditions Lipschitz-Gradient-1$(\gamma_1, \mathcal{B})$ and Concavity-Smoothness$(\mu, \nu, \mathcal{B})$ hold with

$$\gamma_1 = \exp\bigl(-C\cdot r^2\bigr), \quad \mu = \nu = 1, \quad \mathcal{B} = \bigl\{\boldsymbol{\beta} : \|\boldsymbol{\beta} - \boldsymbol{\beta}^*\|_2 \leq R\bigr\} \ \text{ with } \ R = \kappa\cdot\|\boldsymbol{\beta}^*\|_2, \ \kappa = 1/4. \tag{3.14}$$

*Proof.* See the proof of Corollary 1 in Balakrishnan et al. (2014) for details. □

Now we verify Condition 3.3 for the maximization implementation of the M-step (Algorithm 2).

**Lemma 3.6.** For the maximization implementation of the M-step (Algorithm 2), we have that for a sufficiently large $n$ and $\mathcal{B}$ specified in (3.14), Condition Statistical-Error$(\epsilon, \delta, s, n, \mathcal{B})$ holds with

$$\epsilon = C \cdot \bigl(\|\boldsymbol{\beta}^*\|_\infty + \sigma\bigr) \cdot \sqrt{\frac{\log d + \log(2/\delta)}{n}}. \tag{3.15}$$

*Proof.* See §B.4 for a detailed proof. □

The next theorem establishes the implication of Theorem 3.4 for Gaussian mixture model.

**Theorem 3.7.** We consider the maximization implementation of M-step (Algorithm 2). We assume $\|\boldsymbol{\beta}^*\|_2/\sigma \geq r$ holds with a sufficiently large $r > 0$, and $\mathcal{B}$ and $R$ are as defined in (3.14). We suppose the initialization $\boldsymbol{\beta}^{\text{init}}$ of Algorithm 4 satisfies $\bigl\|\boldsymbol{\beta}^{\text{init}} - \boldsymbol{\beta}^*\bigr\|_2 \leq R/2$. Let the sparsity parameter $\widehat{s}$ be

$$\widehat{s} = \Bigl\lceil C' \cdot \max\Bigl\{16 \cdot \bigl[\exp\bigl(C \cdot r^2\bigr) - 1\bigr]^{-2}, \ 100/9\Bigr\} \cdot s^*\Bigr\rceil \tag{3.16}$$

with $C$ specified in (3.14) and $C' \geq 1$. Let the total number of iterations $T$ in Algorithm 4 be

$$T = \left\lceil \frac{\log\bigl\{C' \cdot R/\bigl[\Delta^{\text{GMM}}(s^*) \cdot \sqrt{\log d/n}\bigr]\bigr\}}{C \cdot r^2/2} \right\rceil, \tag{3.17}$$
$$\text{where } \ \Delta^{\text{GMM}}(s^*) = \bigl(\sqrt{\widehat{s}} + C''\cdot\sqrt{s^*}\bigr) \cdot \bigl(\|\boldsymbol{\beta}^*\|_\infty + \sigma\bigr).$$

Meanwhile, we suppose the dimension $d$ is sufficiently large such that $T$ in (3.17) is upper bounded by $\sqrt{d}$, and the sample size $n$ is sufficiently large such that

$$C' \cdot \Delta^{\text{GMM}}(s^*) \cdot \sqrt{\frac{\log d \cdot T}{n}} \leq \min\Bigl\{\bigl[1 - \exp\bigl(-C \cdot r^2/2\bigr)\bigr]^2 \cdot R, \ 9/40 \cdot \|\boldsymbol{\beta}^*\|_2\Bigr\}. \tag{3.18}$$

We have that, with probability at least $1 - 2 \cdot d^{-1/2}$, the final estimator $\widehat{\boldsymbol{\beta}} = \boldsymbol{\beta}^{(T)}$ satisfies

$$\|\widehat{\boldsymbol{\beta}} - \boldsymbol{\beta}^*\|_2 \leq \frac{C' \cdot \Delta^{\mathrm{GMM}}(s^*)}{1 - \exp(-C \cdot r^2/2)} \cdot \sqrt{\frac{\log d \cdot T}{n}}. \tag{3.19}$$

*Proof.* First we plug the quantities in (3.14) and (3.15) into Theorem 3.4. Recall that Theorem 3.4 requires Condition Statistical-Error$(\epsilon, \delta/T, \widehat{s}, n/T, \mathcal{B})$. Thus we need to replace $\delta$ and $n$ with $\delta/T$ and $n/T$ in the definition of $\epsilon$ in (3.15). Then we set $\delta = 2 \cdot d^{-1/2}$. Since $T$ is specified in (3.17) and the dimension $d$ is sufficiently large such that $T \leq \sqrt{d}$, we have $\log\bigl[2/(\delta/T)\bigr] \leq \log d$ in the definition of $\epsilon$. By (3.16) and (3.18), we can then verify the assumptions in (3.8) and (3.9). Finally, by plugging in $T$ in (3.17) into (3.10) and taking $t = T$, we can verify that in (3.9) the optimization error term is smaller than the statistical error term up to a constant factor. Therefore, we obtain (3.19). □

To see the statistical rate of convergence with respect to $s^*$, $d$ and $n$, for the moment we assume that $R$, $r$, $\|\boldsymbol{\beta}^*\|_\infty$, $\|\boldsymbol{\beta}^*\|_2$ and $\sigma$ are absolute constants. From (3.16) and (3.17), we obtain $\widehat{s} = C \cdot s^*$ and therefore $\Delta^{\mathrm{GMM}}(s^*) = C' \cdot \sqrt{s^*}$, which implies $T = C''' \cdot \log\bigl[C'' \cdot \sqrt{n/(s^* \cdot \log d)}\bigr]$. Hence, according to (3.19) we have that, with high probaibility,

$$\|\widehat{\boldsymbol{\beta}} - \boldsymbol{\beta}^*\|_2 \leq C \cdot \sqrt{\frac{s^* \cdot \log d \cdot \log n}{n}}.$$

Because the minimax lower bound for estimating an $s^*$-sparse $d$-dimensional vector is $\sqrt{s^* \cdot \log d/n}$, the rate of convergence in (3.19) is optimal up to a factor of $\log n$. Such a logarithmic factor results from the resampling scheme in Algorithm 4, since we only utilize $n/T$ samples within each iteration. We expect that by directly analyzing Algorithm 1 we can eliminate such a logarithmic factor, which however incurs extra technical complexity for the analysis.

**Mixture of Regression Model:** The next lemma, proved by Balakrishnan et al. (2014), verifies Conditions 3.1 and 3.2 for mixture of regression model. Recall that $\boldsymbol{\beta}^*$ is the regression coefficient and $\sigma$ is the standard deviation of the Gaussian noise.

**Lemma 3.8.** Suppose $\|\boldsymbol{\beta}^*\|_2/\sigma \geq r$, where $r > 0$ is a sufficiently large constant that denotes the required minimum signal-to-noise ratio. Conditions Lipschitz-Gradient-1$(\gamma_1, \mathcal{B})$, Lipschitz-Gradient-2$(\gamma_2, \mathcal{B})$ and Concavity-Smoothness$(\mu, \nu, \mathcal{B})$ hold with

$$\begin{aligned} &\gamma_1 \in (0, 1/2), \quad \gamma_2 \in (0, 1/4), \quad \mu = \nu = 1, \\ &\mathcal{B} = \bigl\{\boldsymbol{\beta} : \|\boldsymbol{\beta} - \boldsymbol{\beta}^*\|_2 \leq R\bigr\} \ \text{ with } \ R = \kappa \cdot \|\boldsymbol{\beta}^*\|_2, \ \kappa = 1/32. \end{aligned} \tag{3.20}$$

*Proof.* See the proof of Corollary 3 in Balakrishnan et al. (2014) for details. □

The following lemma establishes Condition 3.3 for the two implementations of the M-step.

**Lemma 3.9.** For $\mathcal{B}$ specified in (3.20), we have the following results.

- For the maximization implementation of the M-step (line 5 of Algorithm 4), we have that Condition Statistical-Error$(\epsilon, \delta, s, n, \mathcal{B})$ holds with

$$\epsilon = C \cdot \Bigl[\max\bigl\{\|\boldsymbol{\beta}^*\|_2^2 + \sigma^2, \ 1\bigr\} + \|\boldsymbol{\beta}^*\|_2\Bigr] \cdot \sqrt{\frac{\log d + \log(4/\delta)}{n}} \tag{3.21}$$

  for sufficiently large sample size $n$ and absolute constant $C > 0$.

- For the gradient ascent implementation, Condition $\mathsf{Statistical\text{-}Error}(\epsilon, \delta, s, n, \mathcal{B})$ holds with

$$\epsilon = C \cdot \eta \cdot \max\bigl\{\|\boldsymbol{\beta}^*\|_2^2 + \sigma^2,\ 1,\ \sqrt{s} \cdot \|\boldsymbol{\beta}^*\|_2\bigr\} \cdot \sqrt{\frac{\log d + \log(4/\delta)}{n}} \tag{3.22}$$

  for sufficiently large sample size $n$ and $C > 0$, where $\eta$ denotes the stepsize in Algorithm 3.

*Proof.* See §B.5 for a detailed proof. □

The next theorem establishes the implication of Theorem 3.4 for mixture of regression model.

**Theorem 3.10.** Let $\|\boldsymbol{\beta}^*\|_2/\sigma \geq r$ with $r > 0$ sufficiently large. Assuming that $\mathcal{B}$ and $R$ are specified in (3.20) and the initialization $\boldsymbol{\beta}^{\text{init}}$ satisfies $\bigl\|\boldsymbol{\beta}^{\text{init}} - \boldsymbol{\beta}^*\bigr\|_2 \leq R/2$, we have the following results.

- For the maximization implementation of the M-step (Algorithm 2), let $\widehat{s}$ and $T$ be

$$\widehat{s} = \bigl\lceil C \cdot \max\{16,\ 132/31\} \cdot s^* \bigr\rceil, \quad T = \left\lceil \frac{\log\bigl\{C' \cdot R / \bigl[\Delta_1^{\text{MR}}(s^*) \cdot \sqrt{\log d / n}\bigr]\bigr\}}{\log \sqrt{2}} \right\rceil,$$

$$\text{where}\ \ \Delta_1^{\text{MR}}(s^*) = \bigl(\sqrt{\widehat{s}} + C'' \cdot \sqrt{s^*}\bigr) \cdot \Bigl[\max\bigl\{\|\boldsymbol{\beta}^*\|_2^2 + \sigma^2,\ 1\bigr\} + \|\boldsymbol{\beta}^*\|_2\Bigr], \quad \text{and} \quad C \geq 1.$$

  We suppose $d$ and $n$ are sufficiently large such that $T \leq \sqrt{d}$ and

$$C \cdot \Delta_1^{\text{MR}}(s^*) \cdot \sqrt{\frac{\log d \cdot T}{n}} \leq \min\Bigl\{\bigl(1 - 1/\sqrt{2}\bigr)^2 \cdot R,\ 3/8 \cdot \|\boldsymbol{\beta}^*\|_2\Bigr\}.$$

  Then with probability at least $1 - 4 \cdot d^{-1/2}$, the final estimator $\widehat{\boldsymbol{\beta}} = \boldsymbol{\beta}^{(T)}$ satisfies

$$\bigl\|\widehat{\boldsymbol{\beta}} - \boldsymbol{\beta}^*\bigr\|_2 \leq C' \cdot \Delta_1^{\text{MR}}(s^*) \cdot \sqrt{\frac{\log d \cdot T}{n}}. \tag{3.23}$$

- For the gradient ascent implementation of the M-step (Algorithm 3) with stepsize set to $\eta = 1$, let $\widehat{s}$ and $T$ be

$$\widehat{s} = \bigl\lceil C \cdot \max\{16/9,\ 132/31\} \cdot s^* \bigr\rceil, \quad T = \left\lceil \frac{\log\bigl\{C' \cdot R / \bigl[\Delta_2^{\text{MR}}(s^*) \cdot \sqrt{\log d / n}\bigr]\bigr\}}{\log 2} \right\rceil,$$

$$\text{where}\ \ \Delta_2^{\text{MR}}(s^*) = \bigl(\sqrt{\widehat{s}} + C'' \cdot \sqrt{s^*}\bigr) \cdot \max\Bigl\{\|\boldsymbol{\beta}^*\|_2^2 + \sigma^2,\ 1,\ \sqrt{\widehat{s}} \cdot \|\boldsymbol{\beta}^*\|_2\Bigr\}, \quad \text{and} \quad C \geq 1.$$

  We suppose $d$ and $n$ are sufficiently large such that $T \leq \sqrt{d}$ and

$$C \cdot \Delta_2^{\text{MR}}(s^*) \cdot \sqrt{\frac{\log d \cdot T}{n}} \leq \min\bigl\{R/4,\ 3/8 \cdot \|\boldsymbol{\beta}^*\|_2\bigr\}.$$

  Then with probability at least $1 - 4 \cdot d^{-1/2}$, the final estimator $\widehat{\boldsymbol{\beta}} = \boldsymbol{\beta}^{(T)}$ satisfies

$$\bigl\|\widehat{\boldsymbol{\beta}} - \boldsymbol{\beta}^*\bigr\|_2 \leq C' \cdot \Delta_2^{\text{MR}}(s^*) \cdot \sqrt{\frac{\log d \cdot T}{n}}. \tag{3.24}$$

*Proof.* The proof is similar to Theorem 3.7. $\square$

To understand the intuition of Theorem 3.10, we suppose that $\|\boldsymbol{\beta}^*\|_2$, $\sigma$, $R$ and $r$ are absolute constants, which implies $\widehat{s} = C \cdot s^*$ and $\Delta_1^{\mathrm{MR}}(s^*) = C' \cdot \sqrt{s^*}$, $\Delta_2^{\mathrm{MR}}(s^*) = C'' \cdot s^*$. Therefore, for the maximization and gradient ascent implementations of the M-step, we have $T = C' \cdot \log\big[n/(s^* \cdot \log d)\big]$ and $T = C'' \cdot \log\big\{n/\big[(s^*)^2 \cdot \log d\big]\big\}$ correspondingly. Hence, by (3.23) in Theorem 3.10 we have that, for the maximization implementation of the M-step,

$$\big\|\widehat{\boldsymbol{\beta}} - \boldsymbol{\beta}^*\big\|_2 \leq C \cdot \sqrt{\frac{s^* \cdot \log d \cdot \log n}{n}} \tag{3.25}$$

holds with high probability. Meanwhile, from (3.24) in Theorem 3.10 we have that, for the gradient ascent implementation of the M-step,

$$\big\|\widehat{\boldsymbol{\beta}} - \boldsymbol{\beta}^*\big\|_2 \leq C' \cdot s^* \cdot \sqrt{\frac{\log d \cdot \log n}{n}} \tag{3.26}$$

holds with high probability. The statistical rates in (3.25) and (3.26) attain the $\sqrt{s^* \cdot \log d / n}$ minimax lower bound up to factors of $\sqrt{\log n}$ and $\sqrt{s^* \cdot \log n}$ respectively and are therefore near-optimal. Note that the statistical rate of convergence attained by the gradient ascent implementation of the M-step is slower by a $\sqrt{s^*}$ factor than the rate of the maximization implementation. However, our discussion in §A.2 illustrates that, for mixture of regression model, the gradient ascent implementation doesn't involve estimating the inverse covariance of $\boldsymbol{X}$ in (2.25). Hence, the gradient ascent implementation is more computationally efficient, and is also applicable to the settings in which $\boldsymbol{X}$ has more general covariance structures.

**Regression with Missing Covariates:** Recall that we consider the linear regression model with covariates missing completely at random, which is defined in §2.3. The next lemma, which is proved by Balakrishnan et al. (2014), verifies Conditions 3.1 and 3.2 for this model. Recall that $p_{\mathrm{m}}$ denotes the probability that each covariate is missing and $\sigma$ is the standard deviation of the Gaussian noise.

**Lemma 3.11.** Suppose $\|\boldsymbol{\beta}^*\|_2/\sigma \leq r$, where $r > 0$ is a constant that denotes the required maximum signal-to-noise ratio. Assuming that we have

$$p_{\mathrm{m}} < 1/\big(1 + 2 \cdot b + 2 \cdot b^2\big), \quad \text{where} \;\; b = r^2 \cdot (1+\kappa)^2,$$

for some constant $\kappa \in (0,1)$, then we have that Conditions Lipschitz-Gradient-2$(\gamma_2, \mathcal{B})$ and Concavity-Smoothness$(\mu, \nu, \mathcal{B})$ hold with

$$\gamma_2 = \frac{b + p_{\mathrm{m}} \cdot \big(1 + 2 \cdot b + 2 \cdot b^2\big)}{1+b} < 1, \quad \mu = \nu = 1, \tag{3.27}$$
$$\mathcal{B} = \big\{\boldsymbol{\beta} : \|\boldsymbol{\beta} - \boldsymbol{\beta}^*\|_2 \leq R\big\} \;\; \text{with} \;\; R = \kappa \cdot \|\boldsymbol{\beta}^*\|_2.$$

*Proof.* See the proof of Corollary 6 in Balakrishnan et al. (2014) for details. $\square$

The next lemma proves Condition 3.3 for the gradient ascent implementation of the M-step.

**Lemma 3.12.** Suppose $\mathcal{B}$ is defined in (3.27) and $\|\boldsymbol{\beta}^*\|_2/\sigma \le r$. For the gradient ascent implementation of the M-step (Algorithm 3), Condition $\textsf{Statistical-Error}(\epsilon, \delta, s, n, \mathcal{B})$ holds with

$$\epsilon = C\cdot\eta\cdot\Big[\sqrt{s}\cdot\|\boldsymbol{\beta}^*\|_2\cdot(1+\kappa)\cdot(1+\kappa\cdot r)^2+\max\big\{(1+\kappa\cdot r)^2,\ \sigma^2+\|\boldsymbol{\beta}^*\|_2^2\big\}\Big]\cdot\sqrt{\frac{\log d+\log(12/\delta)}{n}} \tag{3.28}$$

for sufficiently large sample size $n$ and $C > 0$. Here $\eta$ denotes the stepsize in Algorithm 3.

*Proof.* See §B.6 for a detailed proof. □

Next we establish the implication of Theorem 3.4 for regression with missing covariates.

**Theorem 3.13.** We consider the gradient ascent implementation of M-step (Algorithm 3) in which $\eta = 1$. Under the assumptions of Lemma 3.11, we suppose the sparsity parameter $\widehat{s}$ takes the form of (3.11) and the initialization $\boldsymbol{\beta}^{\rm init}$ satisfies $\big\|\boldsymbol{\beta}^{\rm init}-\boldsymbol{\beta}^*\big\|_2 \le R/2$ with $\nu$, $\mu$, $\gamma_2$, $\kappa$ and $R$ specified in (3.27). For $r > 0$ specified in Lemma 3.11, let the total number of iterations $T$ in Algorithm 4 be

$$T = \Big\lceil \log\big\{C\cdot R/\big[\Delta^{\rm RMC}(s^*)\cdot\sqrt{\log d/n}\big]\big\}/\log\big(\sqrt{1/\gamma_2}\big)\Big\rceil, \tag{3.29}$$

where $\Delta^{\rm RMC}(s^*) = \left(\sqrt{\widehat{s}}+\dfrac{C'\cdot\sqrt{s^*}}{\sqrt{1-\kappa}}\right)\cdot\Big[\sqrt{\widehat{s}}\cdot\|\boldsymbol{\beta}^*\|_2\cdot(1+\kappa)\cdot(1+\kappa\cdot r)^2+\max\big\{(1+\kappa\cdot r)^2,\ \sigma^2+\|\boldsymbol{\beta}^*\|_2^2\big\}\Big]$.

We assume $d$ and $n$ are sufficiently large such that $T \le \sqrt{d}$ and

$$C\cdot\Delta^{\rm RMC}(s^*)\cdot\sqrt{\frac{\log d\cdot T}{n}} \le \min\Big\{\big(1-\sqrt{\gamma_2}\big)^2\cdot R,\ (1-\kappa)^2/\big[2\cdot(1+\kappa)\big]\cdot\|\boldsymbol{\beta}^*\|_2\Big\}.$$

We have that, with probability at least $1 - 12\cdot d^{-1/2}$, the final estimator $\widehat{\boldsymbol{\beta}} = \boldsymbol{\beta}^{(T)}$ satisfies

$$\big\|\widehat{\boldsymbol{\beta}}-\boldsymbol{\beta}^*\big\|_2 \le \frac{C'\cdot\Delta^{\rm RMC}(s^*)}{1-\sqrt{\gamma_2}}\cdot\sqrt{\frac{\log d\cdot T}{n}}. \tag{3.30}$$

*Proof.* The proof is similar to Theorem 3.7. □

Assuming $r$, $p_{\rm m}$ and $\kappa$ are absolute constants, we have $\Delta^{\rm RMC}(s^*) = C\cdot s^*$ in (3.29) since $\widehat{s} = C'\cdot s^*$. Hence, we obtain $T = C''\cdot\log\big\{n/\big[(s^*)^2\cdot\log d\big]\big\}$. By (3.30) we have that, with high probability,

$$\big\|\widehat{\boldsymbol{\beta}}-\boldsymbol{\beta}^*\big\|_2 \le C\cdot s^*\cdot\sqrt{\frac{\log d\cdot\log n}{n}},$$

which is near-optimal with respect to the $\sqrt{s^*\cdot\log d/n}$ minimax lower bound. It is worth noting that the assumption $\|\boldsymbol{\beta}^*\|_2/\sigma \le r$ in Lemma 3.12 requires the signal-to-noise ratio to be upper bounded rather than lower bounded, which differs from the assumptions for the previous two models. Such a counter-intuitive phenomenon is explained by Loh and Wainwright (2012).

## 4 Theory of Inference

To simplify the presentation of the unified framework, we first lay out several technical conditions, which will later be verified for each model. Some of the notations used in this section are introduced in §2.2. The first condition characterizes the statistical rate of convergence of the estimator attained by the high dimensional EM algorithm (Algorithm 4).

**Condition 4.1** Parameter-Estimation$\big(\zeta^{\mathrm{EM}}\big)$. It holds that

$$\big\|\widehat{\boldsymbol{\beta}} - \boldsymbol{\beta}^*\big\|_1 = O_{\mathbb{P}}\big(\zeta^{\mathrm{EM}}\big),$$

where $\zeta^{\mathrm{EM}}$ scales with $s^*$, $d$ and $n$.

Since both $\widehat{\boldsymbol{\beta}}$ and $\boldsymbol{\beta}^*$ are sparse, we can verify this condition for each model based on the $\ell_2$-norm recovery results in Theorems 3.7, 3.10 and 3.13. The second condition quantifies the statistical error between the gradients of $Q_n(\boldsymbol{\beta}^*;\boldsymbol{\beta}^*)$ and $Q(\boldsymbol{\beta}^*;\boldsymbol{\beta}^*)$.

**Condition 4.2** Gradient-Statistical-Error$\big(\zeta^{\mathrm{G}}\big)$. For the true parameter $\boldsymbol{\beta}^*$, it holds that

$$\big\|\nabla_1 Q_n(\boldsymbol{\beta}^*;\boldsymbol{\beta}^*) - \nabla_1 Q(\boldsymbol{\beta}^*;\boldsymbol{\beta}^*)\big\|_\infty = O_{\mathbb{P}}\big(\zeta^{\mathrm{G}}\big),$$

where $\zeta^{\mathrm{G}}$ scales with $s^*$, $d$ and $n$.

Note that for the gradient ascent implementation of the M-step (Algorithm 3) and its population version defined in (3.2), we have

$$\big\|M_n(\boldsymbol{\beta}^*) - M(\boldsymbol{\beta}^*)\big\|_\infty = \eta\cdot\big\|\nabla_1 Q_n(\boldsymbol{\beta}^*;\boldsymbol{\beta}^*) - \nabla_1 Q(\boldsymbol{\beta}^*;\boldsymbol{\beta}^*)\big\|_\infty.$$

Thus, we can verify Condition 4.2 for each model following the proof of Lemmas 3.6, 3.9 and 3.12. Recall that $T_n(\cdot)$ is defined in (2.8). The following condition quantifies the difference between $T_n(\boldsymbol{\beta}^*)$ and its expectation. Recall that $\|\cdot\|_{\infty,\infty}$ denotes the maximum magnitude of all entries in a matrix.

**Condition 4.3** $T_n(\cdot)$-Concentration$\big(\zeta^{\mathrm{T}}\big)$. For the true parameter $\boldsymbol{\beta}^*$, it holds that

$$\big\|T_n(\boldsymbol{\beta}^*) - \mathbb{E}_{\boldsymbol{\beta}^*}\big[T_n(\boldsymbol{\beta}^*)\big]\big\|_{\infty,\infty} = O_{\mathbb{P}}\big(\zeta^{\mathrm{T}}\big),$$

where $\zeta^{\mathrm{T}}$ scales with $d$ and $n$.

By Lemma 2.1 we have $\mathbb{E}_{\boldsymbol{\beta}^*}\big[T_n(\boldsymbol{\beta}^*)\big] = -I(\boldsymbol{\beta}^*)$. Hence, Condition 4.3 characterizes the statistical rate of convergence of $T_n(\boldsymbol{\beta}^*)$ for estimating the Fisher information. Since $\boldsymbol{\beta}^*$ is unknown in practice, we use $T_n\big(\widehat{\boldsymbol{\beta}}\big)$ or $T_n\big(\widehat{\boldsymbol{\beta}}_0\big)$ to approximate $T_n(\boldsymbol{\beta}^*)$, where $\widehat{\boldsymbol{\beta}}_0$ is defined in (2.11). The next condition quantifies the accuracy of this approximation.

**Condition 4.4** $T_n(\cdot)$-Lipschitz$\big(\zeta^{\mathrm{L}}\big)$. For the true parameter $\boldsymbol{\beta}^*$ and any $\boldsymbol{\beta}$, we have

$$\big\|T_n(\boldsymbol{\beta}) - T_n(\boldsymbol{\beta}^*)\big\|_{\infty,\infty} = O_{\mathbb{P}}\big(\zeta^{\mathrm{L}}\big)\cdot\|\boldsymbol{\beta} - \boldsymbol{\beta}^*\|_1,$$

where $\zeta^{\mathrm{L}}$ scales with $d$ and $n$.

Condition 4.4 characterizes the Lipschitz continuity of $T_n(\cdot)$. We consider $\boldsymbol{\beta}=\widehat{\boldsymbol{\beta}}$. Since Condition 4.1 ensures $\|\widehat{\boldsymbol{\beta}}-\boldsymbol{\beta}^*\|_1$ converges to zero at the rate of $\zeta^{\rm EM}$, Condition 4.4 implies $T_n(\widehat{\boldsymbol{\beta}})$ converges to $T_n(\boldsymbol{\beta}^*)$ entrywise at the rate of $\zeta^{\rm EM}\cdot\zeta^{\rm L}$. In other words, $T_n(\widehat{\boldsymbol{\beta}})$ is a good approximation of $T_n(\boldsymbol{\beta}^*)$.

In the following, we lay out an assumption on several population quantities and the sample size $n$. Recall that $\boldsymbol{\beta}^*=[\alpha^*,(\boldsymbol{\gamma}^*)^\top]^\top$, where $\alpha^*\in\mathbb{R}$ is the entry of interest, while $\boldsymbol{\gamma}^*\in\mathbb{R}^{d-1}$ is the nuisance parameter. By the notations introduced in §2.2, $[I(\boldsymbol{\beta}^*)]_{\boldsymbol{\gamma},\boldsymbol{\gamma}}\in\mathbb{R}^{(d-1)\times(d-1)}$ and $[I(\boldsymbol{\beta}^*)]_{\boldsymbol{\gamma},\alpha}\in\mathbb{R}^{(d-1)\times 1}$ denote the submatrices of the Fisher information matrix $I(\boldsymbol{\beta}^*)\in\mathbb{R}^{d\times d}$. We define $\mathbf{w}^*$, $s^*_{\mathbf{w}}$ and $\mathcal{S}^*_{\mathbf{w}}$ as

$$\mathbf{w}^*=[I(\boldsymbol{\beta}^*)]^{-1}_{\boldsymbol{\gamma},\boldsymbol{\gamma}}\cdot[I(\boldsymbol{\beta}^*)]_{\boldsymbol{\gamma},\alpha}\in\mathbb{R}^{d-1},\quad s^*_{\mathbf{w}}=\|\mathbf{w}^*\|_0,\quad \text{and}\quad \mathcal{S}^*_{\mathbf{w}}=\mathrm{supp}(\mathbf{w}^*). \tag{4.1}$$

We define $\lambda_1[I(\boldsymbol{\beta}^*)]$ and $\lambda_d[I(\boldsymbol{\beta}^*)]$ as the largest and smallest eigenvalues of $I(\boldsymbol{\beta}^*)$, and

$$[I(\boldsymbol{\beta}^*)]_{\alpha|\boldsymbol{\gamma}}=[I(\boldsymbol{\beta}^*)]_{\alpha,\alpha}-[I(\boldsymbol{\beta}^*)]^\top_{\boldsymbol{\gamma},\alpha}\cdot[I(\boldsymbol{\beta}^*)]^{-1}_{\boldsymbol{\gamma},\boldsymbol{\gamma}}\cdot[I(\boldsymbol{\beta}^*)]_{\boldsymbol{\gamma},\alpha}\in\mathbb{R}. \tag{4.2}$$

According to (4.1) and (4.2), we can easily verify that

$$[I(\boldsymbol{\beta}^*)]_{\alpha|\boldsymbol{\gamma}}=[1,-(\mathbf{w}^*)^\top]\cdot I(\boldsymbol{\beta}^*)\cdot[1,-(\mathbf{w}^*)^\top]^\top. \tag{4.3}$$

The following assumption ensures that $\lambda_d[I(\boldsymbol{\beta}^*)]>0$. Hence, $[I(\boldsymbol{\beta}^*)]_{\boldsymbol{\gamma},\boldsymbol{\gamma}}$ in (4.1) is invertible. Also, according to (4.3) and the fact that $\lambda_d[I(\boldsymbol{\beta}^*)]>0$, we have $[I(\boldsymbol{\beta}^*)]_{\alpha|\boldsymbol{\gamma}}>0$.

**Assumption 4.5.** We impose the following assumptions.

- For positive absolute constants $\rho_{\max}$ and $\rho_{\min}$, we assume

$$\rho_{\max}\geq\lambda_1[I(\boldsymbol{\beta}^*)]\geq\lambda_d[I(\boldsymbol{\beta}^*)]\geq\rho_{\min}>0,\quad [I(\boldsymbol{\beta}^*)]_{\alpha|\boldsymbol{\gamma}}=O(1),\quad [I(\boldsymbol{\beta}^*)]^{-1}_{\alpha|\boldsymbol{\gamma}}=O(1). \tag{4.4}$$

- The tuning parameter $\lambda$ of the Dantzig selector in (2.10) is set to

$$\lambda=C\cdot(\zeta^{\rm T}+\zeta^{\rm L}\cdot\zeta^{\rm EM})\cdot(1+\|\mathbf{w}^*\|_1), \tag{4.5}$$

  where $C\geq 1$ is a sufficiently large absolute constant. We suppose the sample size $n$ is sufficiently large such that

$$\max\{\|\mathbf{w}^*\|_1,\ 1\}\cdot s^*_{\mathbf{w}}\cdot\lambda=o(1),\quad \zeta^{\rm EM}=o(1),\quad s^*_{\mathbf{w}}\cdot\lambda\cdot\zeta^{\rm G}=o(1/\sqrt{n}), \tag{4.6}$$
$$\lambda\cdot\zeta^{\rm EM}=o(1/\sqrt{n}),\quad \max\{1,\ \|\mathbf{w}^*\|_1\}\cdot\zeta^{\rm L}\cdot(\zeta^{\rm EM})^2=o(1/\sqrt{n}).$$

The assumption on $\lambda_d[I(\boldsymbol{\beta}^*)]$ guarantees that the Fisher information matrix is positive definite. The other assumptions in (4.4) guarantee the existence of the asymptotic variance of $\sqrt{n}\cdot S_n(\widehat{\boldsymbol{\beta}}_0,\lambda)$ in the score statistic defined in (2.11) and that of $\sqrt{n}\cdot\overline{\alpha}(\widehat{\boldsymbol{\beta}},\lambda)$ in the Wald statistic defined in (2.19). Similar assumptions are standard in existing asymptotic inference results. For example, for mixture of regression model, Khalili and Chen (2007) impose variants of these assumptions.

For specific models, we will show that $\zeta^{\rm EM}$, $\zeta^{\rm G}$, $\zeta^{\rm T}$ and $\lambda$ all decrease with $n$, while $\zeta^{\rm L}$ increases with $n$ at a slow rate. Therefore, the assumptions in (4.6) ensure that the sample size $n$ is sufficiently large. We will make these assumptions more explicit after we specify $\zeta^{\rm EM}$, $\zeta^{\rm G}$, $\zeta^{\rm T}$ and $\zeta^{\rm L}$ for each model. Note the assumptions in (4.6) imply that $s_{\mathbf{w}}^* = \|\mathbf{w}^*\|_0$ needs to be small. For instance, for $\lambda$ specified in (4.5), $\max\{\|\mathbf{w}^*\|_1, 1\} \cdot s_{\mathbf{w}}^* \cdot \lambda = o(1)$ in (4.6) implies $s_{\mathbf{w}}^* \cdot \zeta^{\mathrm{T}} = o(1)$. In the following, we will prove that $\zeta^{\mathrm{T}}$ is of the order $\sqrt{\log d/n}$. Hence, we require that $s_{\mathbf{w}}^* = o\big(\sqrt{n/\log d}\big) \ll d-1$, i.e., $\mathbf{w}^* \in \mathbb{R}^{d-1}$ is sparse. We can understand this sparsity assumption as follows.

- According to the definition of $\mathbf{w}^*$ in (4.1), we have $\big[I(\boldsymbol{\beta}^*)\big]_{\boldsymbol{\gamma},\boldsymbol{\gamma}} \cdot \mathbf{w}^* = \big[I(\boldsymbol{\beta}^*)\big]_{\boldsymbol{\gamma},\alpha}$. Therefore, such a sparsity assumption suggests that $\big[I(\boldsymbol{\beta}^*)\big]_{\boldsymbol{\gamma},\alpha}$ lies within the span of a few columns of $\big[I(\boldsymbol{\beta}^*)\big]_{\boldsymbol{\gamma},\boldsymbol{\gamma}}$.
- By block matrix inversion formula we have $\big\{\big[I(\boldsymbol{\beta}^*)\big]^{-1}\big\}_{\boldsymbol{\gamma},\alpha} = \delta \cdot \mathbf{w}^*$, where $\delta \in \mathbb{R}$. Hence, it can also be understood as a sparsity assumption on the $(d-1)\times 1$ submatrix of $\big[I(\boldsymbol{\beta}^*)\big]^{-1}$.

Such a sparsity assumption on $\mathbf{w}^*$ is necessary, because otherwise it is difficult to accurately estimate $\mathbf{w}^*$ in high dimensional regimes. In the context of high dimensional generalized linear models, Zhang and Zhang (2014); van de Geer et al. (2014) impose similar sparsity assumptions.

## 4.1 Main Results

Equipped with Conditions 4.1-4.4 and Assumption 4.5, we are now ready to establish our theoretical results to justify the inferential methods proposed in §2.2. We first cover the decorrelated score test and then the decorrelated Wald test. Finally, we establish the optimality of our proposed methods.

**Decorrelated Score Test:** The next theorem establishes the asymptotic normality of the decorrelated score statistic defined in (2.11).

**Theorem 4.6.** We consider $\boldsymbol{\beta}^* = \big[\alpha^*, (\boldsymbol{\gamma}^*)^\top\big]^\top$ with $\alpha^* = 0$. Under Assumption 4.5 and Conditions 4.1-4.4, we have that for $n \to \infty$,

$$\sqrt{n} \cdot S_n\big(\widehat{\boldsymbol{\beta}}_0, \lambda\big) \Big/ \Big\{-\big[T_n\big(\widehat{\boldsymbol{\beta}}_0\big)\big]_{\alpha|\boldsymbol{\gamma}}\Big\}^{1/2} \xrightarrow{D} N(0,1), \tag{4.7}$$

where $\widehat{\boldsymbol{\beta}}_0$ and $\big[T_n\big(\widehat{\boldsymbol{\beta}}_0\big)\big]_{\alpha|\boldsymbol{\gamma}} \in \mathbb{R}$ are defined in (2.11).

*Proof.* See §5.2 for a detailed proof. □

**Decorrelated Wald Test:** The next theorem provides the asymptotic normality of $\sqrt{n} \cdot \big[\overline{\alpha}\big(\widehat{\boldsymbol{\beta}}, \lambda\big) - \alpha^*\big] \cdot \big\{-\big[T_n\big(\widehat{\boldsymbol{\beta}}\big)\big]_{\alpha|\boldsymbol{\gamma}}\big\}^{1/2}$, which implies the confidence interval for $\alpha^*$ in (2.20) is valid. In particular, for testing the null hypothesis $H_0\colon \alpha^* = 0$, the next theorem implies the asymptotic normality of the decorrelated Wald statistic defined in (2.19).

**Theorem 4.7.** Under Assumption 4.5 and Conditions 4.1-4.4, we have that for $n \to \infty$,

$$\sqrt{n} \cdot \big[\overline{\alpha}\big(\widehat{\boldsymbol{\beta}}, \lambda\big) - \alpha^*\big] \cdot \Big\{-\big[T_n\big(\widehat{\boldsymbol{\beta}}\big)\big]_{\alpha|\boldsymbol{\gamma}}\Big\}^{1/2} \xrightarrow{D} N(0,1), \tag{4.8}$$

where $\big[T_n\big(\widehat{\boldsymbol{\beta}}\big)\big]_{\alpha|\boldsymbol{\gamma}} \in \mathbb{R}$ is defined by replacing $\widehat{\boldsymbol{\beta}}_0$ with $\widehat{\boldsymbol{\beta}}$ in (2.11).

*Proof.* See §5.2 for a detailed proof. □

**Optimality:** In Lemma 5.3 in the proof of Theorem 4.6 we show that, the limiting variance of the decorrelated score function $\sqrt{n} \cdot S_n\big(\widehat{\boldsymbol{\beta}}_0, \lambda\big)$ is $\big[I(\boldsymbol{\beta}^*)\big]_{\alpha|\boldsymbol{\gamma}}$, which is defined in (4.2). In fact, van der Vaart (2000) proves that for inferring $\alpha^*$ in the presence of the nuisance parameter $\boldsymbol{\gamma}^*$, $\big[I(\boldsymbol{\beta}^*)\big]_{\alpha|\boldsymbol{\gamma}}$ is named the semiparametric efficient information, i.e., the minimum limiting variance of the (rescaled) score function. Our proposed decorrelated score function achieves such an information lower bound and is hence optimal. Because the decorrelated Wald test and the respective confidence interval are built on the decorrelated score function, the limiting variance of $\sqrt{n} \cdot \big[\overline{\alpha}\big(\widehat{\boldsymbol{\beta}}, \lambda\big) - \alpha^*\big]$ and the size of the confidence interval are also optimal in terms of the semiparametric information lower bound.

## 4.2 Implications for Specific Models

To establish the high dimensional inference results for each model, we only need to verify Conditions 4.1-4.4 and determine the key quantities $\zeta^{\mathrm{EM}}$, $\zeta^{\mathrm{G}}$, $\zeta^{\mathrm{T}}$ and $\zeta^{\mathrm{L}}$. In the following, we focus on Gaussian mixture and mixture of regression models.

**Gaussian Mixture Model:** The following lemma verifies Conditions 4.1 and 4.2.

**Lemma 4.8.** We have that Conditions 4.1 and 4.2 hold with

$$\zeta^{\mathrm{EM}} = \frac{\sqrt{\widehat{s}} \cdot \Delta^{\mathrm{GMM}}(s^*)}{1 - \exp\big(-C \cdot r^2/2\big)} \cdot \sqrt{\frac{\log d \cdot T}{n}}, \quad \text{and} \quad \zeta^{\mathrm{G}} = \big(\|\boldsymbol{\beta}^*\|_\infty + \sigma\big) \cdot \sqrt{\frac{\log d}{n}},$$

where $\widehat{s}$, $\Delta^{\mathrm{GMM}}(s^*)$, $r$ and $T$ are as defined in Theorem 3.7.

*Proof.* See §C.6 for a detailed proof. □

By our discussion that follows Theorem 3.7, we have that $\widehat{s}$ and $\Delta^{\mathrm{GMM}}(s^*)$ are of the same order as $s^*$ and $\sqrt{s^*}$ respectively, and $T$ is roughly of the order $\sqrt{\log n}$. Therefore, $\zeta^{\mathrm{EM}}$ is roughly of the order $s^* \cdot \sqrt{\log d/n \cdot \log n}$. The following lemma verifies Condition 4.3 for Gaussian mixture model.

**Lemma 4.9.** We have that Condition 4.3 holds with

$$\zeta^{\mathrm{T}} = \big(\|\boldsymbol{\beta}^*\|_\infty^2 + \sigma^2\big)/\sigma^2 \cdot \sqrt{\frac{\log d}{n}}.$$

*Proof.* See §C.7 for a detailed proof. □

The following lemma establishes Condition 4.4 for Gaussian mixture model.

**Lemma 4.10.** We have that Condition 4.4 holds with

$$\zeta^{\mathrm{L}} = \big(\|\boldsymbol{\beta}^*\|_\infty^2 + \sigma^2\big)^{3/2}/\sigma^4 \cdot \big(\log d + \log n\big)^{3/2}.$$

*Proof.* See §C.8 for a detailed proof. □

Equipped with Lemmas 4.8-4.10, we establish the inference results for Gaussian mixture model.

**Theorem 4.11.** Under Assumption 4.5, we have that for $n \to \infty$, (4.7) and (4.8) hold for Gaussian mixture model. Also, the proposed confidence interval for $\alpha^*$ in (2.20) is valid.

In fact, for Gaussian mixture model we can make (4.6) in Assumption 4.5 more transparent by plugging in $\zeta^{\mathrm{EM}}$, $\zeta^{\mathrm{G}}$, $\zeta^{\mathrm{T}}$ and $\zeta^{\mathrm{L}}$ specified above. Particularly, for simplicity we assume all quantities except $s^*_{\mathbf{w}}$, $s^*$, $d$ and $n$ are absolute constants. Then we can verify that (4.6) holds if

$$\max\{s^*_{\mathbf{w}},\ s^*\}^2 \cdot (s^*)^2 \cdot (\log d)^5 = o\big[n/(\log n)^2\big]. \tag{4.9}$$

According to the discussion following Theorem 3.7, we require $s^* \cdot \log d = o(n/\log n)$ for the estimator $\widehat{\boldsymbol{\beta}}$ to be consistent. In comparison, (4.9) illustrates that high dimensional inference requires a higher sample complexity than parameter estimation. In the context of high dimensional generalized linear models, Zhang and Zhang (2014); van de Geer et al. (2014) also observe the same phenomenon.

**Mixture of Regression Model:** The following lemma verifies Conditions 4.1 and 4.2. Recall that $\widehat{s}$, $T$, $\Delta_1^{\mathrm{MR}}(s^*)$ and $\Delta_2^{\mathrm{MR}}(s^*)$ are defined in Theorem 3.10, while $\sigma$ denotes the standard deviation of the Gaussian noise in mixture of regression model.

**Lemma 4.12.** We have that Conditions 4.1 and 4.2 hold with

$$\zeta^{\mathrm{EM}} = \sqrt{\widehat{s}} \cdot \Delta^{\mathrm{MR}}(s^*) \cdot \sqrt{\frac{\log d \cdot T}{n}}, \quad \text{and} \quad \zeta^{\mathrm{G}} = \max\Big\{\|\boldsymbol{\beta}^*\|_2^2 + \sigma^2,\ 1,\ \sqrt{s^*} \cdot \|\boldsymbol{\beta}^*\|_2\Big\} \cdot \sqrt{\frac{\log d}{n}},$$

where we have $\Delta^{\mathrm{MR}}(s^*) = \Delta_1^{\mathrm{MR}}(s^*)$ for the maximization implementation of the M-step (Algorithm 2), and $\Delta^{\mathrm{MR}}(s^*) = \Delta_2^{\mathrm{MR}}(s^*)$ for the gradient ascent implementation of the M-step (Algorithm 3).

*Proof.* See §C.9 for a detailed proof. □

By our discussion that follows Theorem 3.10, we have that $\widehat{s}$ is of the same order as $s^*$. For the maximization implementation of the M-step (Algorithm 2), we have that $\Delta^{\mathrm{MR}}(s^*) = \Delta_1^{\mathrm{MR}}(s^*)$ is of the same order as $\sqrt{s^*}$. Meanwhile, for the gradient ascent implementation in Algorithm 3, we have that $\Delta^{\mathrm{MR}}(s^*) = \Delta_2^{\mathrm{MR}}(s^*)$ is of the same order as $s^*$. Hence, $\zeta^{\mathrm{EM}}$ is of the order $s^* \cdot \sqrt{\log d/n \cdot \log n}$ or $(s^*)^{3/2} \cdot \sqrt{\log d/n \cdot \log n}$ correspondingly, since $T$ is roughly of the order $\sqrt{\log n}$. The next lemma establishes Condition 4.3 for mixture of regression model.

**Lemma 4.13.** We have that Condition 4.3 holds with

$$\zeta^{\mathrm{T}} = \big(\log n + \log d\big) \cdot \big[\big(\log n + \log d\big) \cdot \|\boldsymbol{\beta}^*\|_1^2 + \sigma^2\big]/\sigma^2 \cdot \sqrt{\frac{\log d}{n}}.$$

*Proof.* See §C.10 for a detailed proof. □

The following lemma establishes Condition 4.4 for mixture of regression model.

**Lemma 4.14.** We have that Condition 4.4 holds with

$$\zeta^{\mathrm{L}} = \big(\|\boldsymbol{\beta}^*\|_1 + \sigma\big)^3 \cdot \big(\log n + \log d\big)^3/\sigma^4.$$

*Proof.* See §C.11 for a detailed proof. □

Equipped with Lemmas 4.12-4.14, we are now ready to establish the high dimensional inference results for mixture of regression model.

**Theorem 4.15.** For mixture of regression model, under Assumption 4.5, both (4.7) and (4.8) hold as $n \to \infty$. Also, the proposed confidence interval for $\alpha^*$ in (2.20) is valid.

Similar to the discussion that follows Theorem 4.11, we can make (4.6) in Assumption 4.5 more explicit by plugging in $\zeta^{\rm EM}$, $\zeta^{\rm G}$, $\zeta^{\rm T}$ and $\zeta^{\rm L}$ specified in Lemmas 4.12-4.14. Assuming all quantities except $s_{\mathbf{w}}^*$, $s^*$, $d$ and $n$ are absolute constants, we have that (4.6) holds if

$$\max\bigl\{s_{\mathbf{w}}^*,\ s^*\bigr\}^2 \cdot (s^*)^4 \cdot (\log d)^8 = o\bigl[n/(\log n)^2\bigr].$$

In contrast, for high dimensional estimation, we only require $(s^*)^2 \cdot \log d = o(n/\log n)$ to ensure the consistency of $\widehat{\boldsymbol{\beta}}$ by our discussion following Theorem 3.10.

# 5 Proof of Main Results

We lay out a proof sketch of the main theory. First we prove the results in Theorem 3.4 for parameter estimation and computation. Then we establish the results in Theorems 4.6 and 4.7 for inference.

## 5.1 Proof of Results for Computation and Estimation

**Proof of Theorem 3.4:** First we introduce some notations. Recall that the $\mathrm{trunc}(\cdot,\cdot)$ function is defined in (2.7). We define $\overline{\boldsymbol{\beta}}^{(t+0.5)}, \overline{\boldsymbol{\beta}}^{(t+1)} \in \mathbb{R}^d$ as

$$\overline{\boldsymbol{\beta}}^{(t+0.5)} = M\bigl(\boldsymbol{\beta}^{(t)}\bigr), \qquad \overline{\boldsymbol{\beta}}^{(t+1)} = \mathrm{trunc}\bigl(\overline{\boldsymbol{\beta}}^{(t+0.5)}, \widehat{\mathcal{S}}^{(t+0.5)}\bigr). \tag{5.1}$$

As defined in (3.1) or (3.2), $M(\cdot)$ is the population version M-step with the maximization or gradient ascent implementation. Here $\widehat{\mathcal{S}}^{(t+0.5)}$ denotes the set of index $j$'s with the top $\widehat{s}$ largest $\bigl|\beta_j^{(t+0.5)}\bigr|$'s. It is worth noting $\widehat{\mathcal{S}}^{(t+0.5)}$ is calculated based on $\boldsymbol{\beta}^{(t+0.5)}$ in the truncation step (line 6 of Algorithm 4), rather than based on $\overline{\boldsymbol{\beta}}^{(t+0.5)}$ defined in (5.1).

Our goal is to characterize the relationship between $\bigl\|\boldsymbol{\beta}^{(t+1)} - \boldsymbol{\beta}^*\bigr\|_2$ and $\bigl\|\boldsymbol{\beta}^{(t)} - \boldsymbol{\beta}^*\bigr\|_2$. According to the definition of the truncation step (line 6 of Algorithm 4) and triangle inequality, we have

$$\begin{aligned}
\bigl\|\boldsymbol{\beta}^{(t+1)} - \boldsymbol{\beta}^*\bigr\|_2 &= \Bigl\|\mathrm{trunc}\bigl(\boldsymbol{\beta}^{(t+0.5)}, \widehat{\mathcal{S}}^{(t+0.5)}\bigr) - \boldsymbol{\beta}^*\Bigr\|_2 \\
&\le \Bigl\|\mathrm{trunc}\bigl(\boldsymbol{\beta}^{(t+0.5)}, \widehat{\mathcal{S}}^{(t+0.5)}\bigr) - \mathrm{trunc}\bigl(\overline{\boldsymbol{\beta}}^{(t+0.5)}, \widehat{\mathcal{S}}^{(t+0.5)}\bigr)\Bigr\|_2 + \Bigl\|\mathrm{trunc}\bigl(\overline{\boldsymbol{\beta}}^{(t+0.5)}, \widehat{\mathcal{S}}^{(t+0.5)}\bigr) - \boldsymbol{\beta}^*\Bigr\|_2 \\
&= \underbrace{\Bigl\|\mathrm{trunc}\bigl(\boldsymbol{\beta}^{(t+0.5)}, \widehat{\mathcal{S}}^{(t+0.5)}\bigr) - \mathrm{trunc}\bigl(\overline{\boldsymbol{\beta}}^{(t+0.5)}, \widehat{\mathcal{S}}^{(t+0.5)}\bigr)\Bigr\|_2}_{\text{(i)}} + \underbrace{\bigl\|\overline{\boldsymbol{\beta}}^{(t+1)} - \boldsymbol{\beta}^*\bigr\|_2}_{\text{(ii)}},
\end{aligned} \tag{5.2}$$

where the last equality is obtained from (5.1). According to the definition of the $\mathrm{trunc}(\cdot,\cdot)$ function in (2.7), the two terms within term (i) both have support $\widehat{\mathcal{S}}^{(t+0.5)}$ with cardinality $\widehat{s}$. Thus, we have

$$\begin{aligned}
\Bigl\|\mathrm{trunc}\bigl(\boldsymbol{\beta}^{(t+0.5)}, \widehat{\mathcal{S}}^{(t+0.5)}\bigr) - \mathrm{trunc}\bigl(\overline{\boldsymbol{\beta}}^{(t+0.5)}, \widehat{\mathcal{S}}^{(t+0.5)}\bigr)\Bigr\|_2 &= \Bigl\|\bigl(\boldsymbol{\beta}^{(t+0.5)} - \overline{\boldsymbol{\beta}}^{(t+0.5)}\bigr)_{\widehat{\mathcal{S}}^{(t+0.5)}}\Bigr\|_2 \\
&\le \sqrt{\widehat{s}} \cdot \Bigl\|\bigl(\boldsymbol{\beta}^{(t+0.5)} - \overline{\boldsymbol{\beta}}^{(t+0.5)}\bigr)_{\widehat{\mathcal{S}}^{(t+0.5)}}\Bigr\|_\infty \\
&\le \sqrt{\widehat{s}} \cdot \bigl\|\boldsymbol{\beta}^{(t+0.5)} - \overline{\boldsymbol{\beta}}^{(t+0.5)}\bigr\|_\infty.
\end{aligned} \tag{5.3}$$

Since we have $\boldsymbol{\beta}^{(t+0.5)} = M_n(\boldsymbol{\beta}^{(t)})$ and $\overline{\boldsymbol{\beta}}^{(t+0.5)} = M(\boldsymbol{\beta}^{(t)})$, we can further establish an upper bound for the right-hand side by invoking Condition 3.3.

Our subsequent proof will establish an upper bound for term (ii) in (5.2) in two steps. We first characterize the relationship between $\|\overline{\boldsymbol{\beta}}^{(t+1)} - \boldsymbol{\beta}^*\|_2$ and $\|\overline{\boldsymbol{\beta}}^{(t+0.5)} - \boldsymbol{\beta}^*\|_2$ and then the relationship between $\|\overline{\boldsymbol{\beta}}^{(t+0.5)} - \boldsymbol{\beta}^*\|_2$ and $\|\boldsymbol{\beta}^{(t)} - \boldsymbol{\beta}^*\|_2$. The next lemma accomplishes the first step. Recall that $\widehat{s}$ is the sparsity parameter in Algorithm 4, while $s^*$ is the sparsity level of the true parameter $\boldsymbol{\beta}^*$.

**Lemma 5.1.** Suppose that we have

$$\|\overline{\boldsymbol{\beta}}^{(t+0.5)} - \boldsymbol{\beta}^*\|_2 \leq \kappa \cdot \|\boldsymbol{\beta}^*\|_2 \tag{5.4}$$

for some $\kappa \in (0,1)$. Assuming that we have

$$\widehat{s} \geq \frac{4 \cdot (1+\kappa)^2}{(1-\kappa)^2} \cdot s^*, \quad \text{and} \quad \sqrt{\widehat{s}} \cdot \|\boldsymbol{\beta}^{(t+0.5)} - \overline{\boldsymbol{\beta}}^{(t+0.5)}\|_\infty \leq \frac{(1-\kappa)^2}{2 \cdot (1+\kappa)} \cdot \|\boldsymbol{\beta}^*\|_2, \tag{5.5}$$

then it holds that

$$\|\overline{\boldsymbol{\beta}}^{(t+1)} - \boldsymbol{\beta}^*\|_2 \leq \frac{C \cdot \sqrt{s^*}}{\sqrt{1-\kappa}} \cdot \|\boldsymbol{\beta}^{(t+0.5)} - \overline{\boldsymbol{\beta}}^{(t+0.5)}\|_\infty + \big(1 + 4 \cdot \sqrt{s^*/\widehat{s}}\big)^{1/2} \cdot \|\overline{\boldsymbol{\beta}}^{(t+0.5)} - \boldsymbol{\beta}^*\|_2. \tag{5.6}$$

*Proof.* The proof is based on fine-grained analysis of the relationship between $\widehat{\mathcal{S}}^{(t+0.5)}$ and the true support $\mathcal{S}^*$. In particular, we focus on three index sets, namely, $\mathcal{I}_1 = \mathcal{S}^* \setminus \widehat{\mathcal{S}}^{(t+0.5)}$, $\mathcal{I}_2 = \mathcal{S}^* \cap \widehat{\mathcal{S}}^{(t+0.5)}$ and $\mathcal{I}_3 = \widehat{\mathcal{S}}^{(t+0.5)} \setminus \mathcal{S}^*$. Among them, $\mathcal{I}_2$ characterizes the similarity between $\widehat{\mathcal{S}}^{(t+0.5)}$ and $\mathcal{S}^*$, while $\mathcal{I}_1$ and $\mathcal{I}_3$ characterize their difference. The key proof strategy is to represent the three distances in (5.6) with the $\ell_2$-norms of the restrictions of $\overline{\boldsymbol{\beta}}^{(t+0.5)}$ and $\boldsymbol{\beta}^*$ on the three index sets. In particular, we focus on $\|\overline{\boldsymbol{\beta}}^{(t+0.5)}_{\mathcal{I}_1}\|_2$ and $\|\boldsymbol{\beta}^*_{\mathcal{I}_1}\|_2$. In order to quantify these $\ell_2$-norms, we establish a fine-grained characterization for the absolute values of $\overline{\boldsymbol{\beta}}^{(t+0.5)}$'s entries that are selected and eliminated within the truncation operation $\overline{\boldsymbol{\beta}}^{(t+1)} \leftarrow \mathrm{trunc}(\overline{\boldsymbol{\beta}}^{(t+0.5)}, \widehat{\mathcal{S}}^{(t+0.5)})$. See §B.1 for a detailed proof. $\square$

Lemma 5.1 is central to the proof of Theorem 3.4. In detail, the assumption in (5.4) guarantees $\overline{\boldsymbol{\beta}}^{(t+0.5)}$ is within the basin of attraction. In (5.5), the first assumption ensures the sparsity parameter $\widehat{s}$ in Algorithm 4 is set to be sufficiently large, while second ensures the statistical error is sufficiently small. These assumptions will be verified in the proof of Theorem 3.4. The intuition behind (5.6) is as follows. Let $\overline{\mathcal{S}}^{(t+0.5)} = \mathrm{supp}(\overline{\boldsymbol{\beta}}^{(t+0.5)}, \widehat{s})$, where $\mathrm{supp}(\cdot,\cdot)$ is defined in (2.6). By triangle inequality, the left-hand side of (5.6) satisfies

$$\|\overline{\boldsymbol{\beta}}^{(t+1)} - \boldsymbol{\beta}^*\|_2 \leq \underbrace{\Big\|\overline{\boldsymbol{\beta}}^{(t+1)} - \mathrm{trunc}\big(\overline{\boldsymbol{\beta}}^{(t+0.5)}, \overline{\mathcal{S}}^{(t+0.5)}\big)\Big\|_2}_{\text{(i)}} + \underbrace{\Big\|\mathrm{trunc}\big(\overline{\boldsymbol{\beta}}^{(t+0.5)}, \overline{\mathcal{S}}^{(t+0.5)}\big) - \boldsymbol{\beta}^*\Big\|_2}_{\text{(ii)}}. \tag{5.7}$$

Intuitively, the two terms on right-hand side of (5.6) reflect terms (i) and (ii) in (5.7) correspondingly. In detail, for term (i) in (5.7), recall that according to (5.1) and line 6 of Algorithm 4 we have

$$\overline{\boldsymbol{\beta}}^{(t+1)} = \mathrm{trunc}\big(\overline{\boldsymbol{\beta}}^{(t+0.5)}, \widehat{\mathcal{S}}^{(t+0.5)}\big), \quad \text{where} \;\; \widehat{\mathcal{S}}^{(t+0.5)} = \mathrm{supp}\big(\boldsymbol{\beta}^{(t+0.5)}, \widehat{s}\big).$$

As the sample size $n$ is sufficiently large, $\overline{\boldsymbol{\beta}}^{(t+0.5)}$ and $\boldsymbol{\beta}^{(t+0.5)}$ are close, since they are attained by the population version and sample version M-steps correspondingly. Hence, $\overline{\mathcal{S}}^{(t+0.5)} = \text{supp}\big(\overline{\boldsymbol{\beta}}^{(t+0.5)}, \widehat{s}\big)$ and $\widehat{\mathcal{S}}^{(t+0.5)} = \text{supp}\big(\boldsymbol{\beta}^{(t+0.5)}, \widehat{s}\big)$ should be similar. Thus, in term (i), $\boldsymbol{\beta}^{(t+1)} = \text{trunc}\big(\boldsymbol{\beta}^{(t+0.5)}, \widehat{\mathcal{S}}^{(t+0.5)}\big)$ should be close to $\text{trunc}\big(\overline{\boldsymbol{\beta}}^{(t+0.5)}, \overline{\mathcal{S}}^{(t+0.5)}\big)$ up to some statistical error, which is reflected by the first term on the right-hand side of (5.6).

Also, we turn to quantify the relationship between $\big\|\overline{\boldsymbol{\beta}}^{(t+0.5)} - \boldsymbol{\beta}^*\big\|_2$ in (5.6) and term (ii) in (5.7). The truncation in term (ii) preserves the top $\widehat{s}$ coordinates of $\overline{\boldsymbol{\beta}}^{(t+0.5)}$ with the largest magnitudes while setting others to zero. Intuitively speaking, the truncation incurs additional error to $\overline{\boldsymbol{\beta}}^{(t+0.5)}$'s distance to $\boldsymbol{\beta}^*$. Meanwhile, note that when $\overline{\boldsymbol{\beta}}^{(t+0.5)}$ is close to $\boldsymbol{\beta}^*$, $\overline{\mathcal{S}}^{(t+0.5)}$ is similar to $\mathcal{S}^*$. Therefore, the incurred error can be controlled, because the truncation doesn't eliminate many relevant entries. In particular, as shown in the second term on the right-hand side of (5.6), such incurred error decays as $\widehat{s}$ increases, since in this case $\widehat{\mathcal{S}}^{(t+0.5)}$ includes more entries. According to the discussion for term (i) in (5.7), $\overline{\mathcal{S}}^{(t+0.5)}$ is similar to $\widehat{\mathcal{S}}^{(t+0.5)}$, which implies that $\overline{\mathcal{S}}^{(t+0.5)}$ should also cover more entries. Thus, fewer relevant entries are wrongly eliminated by the truncation and hence the incurred error is smaller. The extreme case is that, when $\widehat{s} \to \infty$, term (ii) in (5.7) becomes $\big\|\overline{\boldsymbol{\beta}}^{(t+0.5)} - \boldsymbol{\beta}^*\big\|_2$, which indicates that no additional error is incurred by the truncation. Correspondingly, the second term on the right-hand side of (5.6) also becomes $\big\|\overline{\boldsymbol{\beta}}^{(t+0.5)} - \boldsymbol{\beta}^*\big\|_2$.

Next, we turn to characterize the relationship between $\big\|\overline{\boldsymbol{\beta}}^{(t+0.5)} - \boldsymbol{\beta}^*\big\|_2$ and $\big\|\boldsymbol{\beta}^{(t)} - \boldsymbol{\beta}^*\big\|_2$. Recall $\overline{\boldsymbol{\beta}}^{(t+0.5)} = M\big(\boldsymbol{\beta}^{(t)}\big)$ is defined in (5.1). The next lemma, which is adapted from Theorems 1 and 3 in Balakrishnan et al. (2014), characterizes the contraction property of the population version M-step defined in (3.1) or (3.2).

**Lemma 5.2.** Under the assumptions of Theorem 3.4, the following results hold for $\boldsymbol{\beta}^{(t)} \in \mathcal{B}$.

- For the maximization implementation of the M-step (Algorithm 2), we have

$$\big\|\overline{\boldsymbol{\beta}}^{(t+0.5)} - \boldsymbol{\beta}^*\big\|_2 \le (\gamma_1/\nu) \cdot \big\|\boldsymbol{\beta}^{(t)} - \boldsymbol{\beta}^*\big\|_2. \tag{5.8}$$

- For the gradient ascent implementation of the M-step (Algorithm 3), we have

$$\big\|\overline{\boldsymbol{\beta}}^{(t+0.5)} - \boldsymbol{\beta}^*\big\|_2 \le \left(1 - 2 \cdot \frac{\nu - \gamma_2}{\nu + \mu}\right) \cdot \big\|\boldsymbol{\beta}^{(t)} - \boldsymbol{\beta}^*\big\|_2. \tag{5.9}$$

Here $\gamma_1$, $\gamma_2$, $\mu$ and $\nu$ are defined in Conditions 3.1 and 3.2.

*Proof.* The proof strategy is to first characterize the M-step using $Q(\cdot;\boldsymbol{\beta}^*)$. According to Condition Concavity-Smoothness$(\mu, \nu, \mathcal{B})$, $-Q(\cdot;\boldsymbol{\beta}^*)$ is $\nu$-strongly convex and $\mu$-smooth, and thus enjoys desired optimization guarantees. Then Condition Lipschitz-Gradient-1$(\gamma_1, \mathcal{B})$ or Lipschitz-Gradient-2$(\gamma_2, \mathcal{B})$ is invoked to characterize the difference between $Q(\cdot;\boldsymbol{\beta}^*)$ and $Q\big(\cdot;\boldsymbol{\beta}^{(t)}\big)$. We provide the proof in §B.2 for the sake of completeness. □

Equipped with Lemmas 5.1 and 5.2, we are now ready to prove Theorem 3.4.

*Proof.* To unify the subsequent proof for the maximization and gradient implementations of the M-step, we employ $\rho \in (0,1)$ to denote $\gamma_1/\nu$ in (5.8) or $1-2\cdot(\nu-\gamma_2)/(\nu+\mu)$ in (5.9). By the definitions of $\overline{\boldsymbol{\beta}}^{(t+0.5)}$ and $\boldsymbol{\beta}^{(t+0.5)}$ in (5.1) and Algorithm 4, Condition $\mathsf{Statistical\text{-}Error}(\epsilon, \delta/T, \widehat{s}, n/T, \mathcal{B})$ implies

$$\big\|\boldsymbol{\beta}^{(t+0.5)} - \overline{\boldsymbol{\beta}}^{(t+0.5)}\big\|_\infty = \Big\|M_{n/T}\big(\boldsymbol{\beta}^{(t)}\big) - M\big(\boldsymbol{\beta}^{(t)}\big)\Big\|_\infty \le \epsilon$$

holds with probability at least $1-\delta/T$. Then by taking union bound we have that, the event

$$\mathcal{E} = \Big\{\big\|\boldsymbol{\beta}^{(t+0.5)} - \overline{\boldsymbol{\beta}}^{(t+0.5)}\big\|_\infty \le \epsilon, \quad \text{for all } t \in \{0, \ldots, T-1\}\Big\} \tag{5.10}$$

occurs with probability at least $1-\delta$. Conditioning on $\mathcal{E}$, in the following we prove that

$$\big\|\boldsymbol{\beta}^{(t)} - \boldsymbol{\beta}^*\big\|_2 \le \frac{\big(\sqrt{\widehat{s}} + C/\sqrt{1-\kappa}\cdot\sqrt{s^*}\big)\cdot\epsilon}{1-\sqrt{\rho}} + \rho^{t/2}\cdot\big\|\boldsymbol{\beta}^{(0)} - \boldsymbol{\beta}^*\big\|_2, \quad \text{for all } t \in \{1, \ldots, T\} \tag{5.11}$$

by mathematical induction.

Before we lay out the proof, we first prove $\boldsymbol{\beta}^{(0)} \in \mathcal{B}$. Recall $\boldsymbol{\beta}^{\mathrm{init}}$ is the initialization of Algorithm 4 and $R$ is the radius of the basin of attraction $\mathcal{B}$. By the assumption of Theorem 3.4, we have

$$\big\|\boldsymbol{\beta}^{\mathrm{init}} - \boldsymbol{\beta}^*\big\|_2 \le R/2. \tag{5.12}$$

Therefore, (5.12) implies $\big\|\boldsymbol{\beta}^{\mathrm{init}} - \boldsymbol{\beta}^*\big\|_2 < \kappa\cdot\|\boldsymbol{\beta}^*\|_2$ since $R = \kappa\cdot\|\boldsymbol{\beta}^*\|_2$. Invoking the auxiliary result in Lemma B.1, we obtain

$$\big\|\boldsymbol{\beta}^{(0)} - \boldsymbol{\beta}^*\big\|_2 \le \big(1 + 4\cdot\sqrt{s^*/\widehat{s}}\big)^{1/2}\cdot\big\|\boldsymbol{\beta}^{\mathrm{init}} - \boldsymbol{\beta}^*\big\|_2 \le \big(1 + 4\cdot\sqrt{1/4}\big)^{1/2}\cdot R/2 < R. \tag{5.13}$$

Here the second inequality is from (5.12) as well as the assumption in (3.8) or (3.11), which implies $s^*/\widehat{s} \le (1-\kappa)^2/\big[4\cdot(1+\kappa)^2\big] \le 1/4$. Thus, (5.13) implies $\boldsymbol{\beta}^{(0)} \in \mathcal{B}$. In the sequel, we prove that (5.11) holds for $t = 1$. By invoking Lemma 5.2 and setting $t = 0$ in (5.8) or (5.9), we obtain

$$\big\|\overline{\boldsymbol{\beta}}^{(0.5)} - \boldsymbol{\beta}^*\big\|_2 \le \rho\cdot\big\|\boldsymbol{\beta}^{(0)} - \boldsymbol{\beta}^*\big\|_2 \le \rho\cdot R < R = \kappa\cdot\|\boldsymbol{\beta}^*\|_2,$$

where the second inequality is from (5.13). Hence, the assumption in (5.4) of Lemma 5.1 holds for $\overline{\boldsymbol{\beta}}^{(0.5)}$. Furthermore, by the assumptions in (3.8), (3.9), (3.11) and (3.12) of Theorem 3.4, we can also verify that the assumptions in (5.5) of Lemma 5.1 hold conditioning on the event $\mathcal{E}$ defined in (5.10). By invoking Lemma 5.1 we have that (5.6) holds for $t = 0$. Further plugging $\big\|\boldsymbol{\beta}^{(t+0.5)} - \overline{\boldsymbol{\beta}}^{(t+0.5)}\big\|_\infty \le \epsilon$ in (5.10) into (5.6) with $t = 0$, we obtain

$$\big\|\overline{\boldsymbol{\beta}}^{(1)} - \boldsymbol{\beta}^*\big\|_2 \le \frac{C\cdot\sqrt{s^*}}{\sqrt{1-\kappa}}\cdot\epsilon + \big(1 + 4\cdot\sqrt{s^*/\widehat{s}}\big)^{1/2}\cdot\big\|\overline{\boldsymbol{\beta}}^{(0.5)} - \boldsymbol{\beta}^*\big\|_2. \tag{5.14}$$

Setting $t = 0$ in (5.8) or (5.9) of Lemma 5.2 and then plugging (5.8) or (5.9) into (5.14), we obtain

$$\big\|\overline{\boldsymbol{\beta}}^{(1)} - \boldsymbol{\beta}^*\big\|_2 \le \frac{C\cdot\sqrt{s^*}}{\sqrt{1-\kappa}}\cdot\epsilon + \big(1 + 4\cdot\sqrt{s^*/\widehat{s}}\big)^{1/2}\cdot\rho\cdot\big\|\boldsymbol{\beta}^{(0)} - \boldsymbol{\beta}^*\big\|_2. \tag{5.15}$$

For $t=0$, plugging (5.3) into term (i) in (5.2), and (5.15) into term (ii) in (5.2), and then applying $\big\|\boldsymbol{\beta}^{(t+0.5)}-\overline{\boldsymbol{\beta}}^{(t+0.5)}\big\|_\infty\le\epsilon$ with $t=0$ in (5.10), we obtain

$$\begin{aligned}\big\|\boldsymbol{\beta}^{(1)}-\boldsymbol{\beta}^*\big\|_2&\le\sqrt{\widehat{s}}\cdot\big\|\boldsymbol{\beta}^{(0.5)}-\overline{\boldsymbol{\beta}}^{(0.5)}\big\|_\infty+\frac{C\cdot\sqrt{s^*}}{\sqrt{1-\kappa}}\cdot\epsilon+\big(1+4\cdot\sqrt{s^*/\widehat{s}}\big)^{1/2}\cdot\rho\cdot\big\|\boldsymbol{\beta}^{(0)}-\boldsymbol{\beta}^*\big\|_2\\&\le\big(\sqrt{\widehat{s}}+C/\sqrt{1-\kappa}\cdot\sqrt{s^*}\big)\cdot\epsilon+\big(1+4\cdot\sqrt{s^*/\widehat{s}}\big)^{1/2}\cdot\rho\cdot\big\|\boldsymbol{\beta}^{(0)}-\boldsymbol{\beta}^*\big\|_2.\end{aligned}\tag{5.16}$$

By our assumption that $\widehat{s}\ge16\cdot(1/\rho-1)^{-2}\cdot s^*$ in (3.8) or (3.11), we have $\big(1+4\cdot\sqrt{s^*/\widehat{s}}\big)^{1/2}\le1/\sqrt{\rho}$ in (5.16). Hence, from (5.16) we obtain

$$\big\|\boldsymbol{\beta}^{(1)}-\boldsymbol{\beta}^*\big\|_2\le\big(\sqrt{\widehat{s}}+C/\sqrt{1-\kappa}\cdot\sqrt{s^*}\big)\cdot\epsilon+\sqrt{\rho}\cdot\big\|\boldsymbol{\beta}^{(0)}-\boldsymbol{\beta}^*\big\|_2,\tag{5.17}$$

which implies that (5.11) holds for $t=1$, since we have $1-\sqrt{\rho}<1$ in (5.11).

Suppose we have that (5.11) holds for some $t\ge1$. By (5.11) we have

$$\begin{aligned}\big\|\boldsymbol{\beta}^{(t)}-\boldsymbol{\beta}^*\big\|_2&\le\frac{\big(\sqrt{\widehat{s}}+C/\sqrt{1-\kappa}\cdot\sqrt{s^*}\big)\cdot\epsilon}{1-\sqrt{\rho}}+\rho^{t/2}\cdot\big\|\boldsymbol{\beta}^{(0)}-\boldsymbol{\beta}^*\big\|_2\\&\le\big(1-\sqrt{\rho}\big)\cdot R+\sqrt{\rho}\cdot R=R,\end{aligned}\tag{5.18}$$

where the second inequality is from (5.13) and our assumption $\big(\sqrt{\widehat{s}}+C/\sqrt{1-\kappa}\cdot\sqrt{s^*}\big)\cdot\epsilon\le\big(1-\sqrt{\rho}\big)^2\cdot R$ in (3.9) or (3.12). Therefore, by (5.18) we have $\boldsymbol{\beta}^{(t)}\in\mathcal{B}$. Then (5.8) or (5.9) in Lemma 5.2 implies

$$\big\|\overline{\boldsymbol{\beta}}^{(t+0.5)}-\boldsymbol{\beta}^*\big\|_2\le\rho\cdot\big\|\boldsymbol{\beta}^{(t)}-\boldsymbol{\beta}^*\big\|_2\le\rho\cdot R<R=\kappa\cdot\|\boldsymbol{\beta}^*\|_2,$$

where the third inequality is from $\rho\in(0,1)$. Following the same proof for (5.17), we obtain

$$\begin{aligned}\big\|\boldsymbol{\beta}^{(t+1)}-\boldsymbol{\beta}^*\big\|_2&\le\big(\sqrt{\widehat{s}}+C/\sqrt{1-\kappa}\cdot\sqrt{s^*}\big)\cdot\epsilon+\sqrt{\rho}\cdot\big\|\boldsymbol{\beta}^{(t)}-\boldsymbol{\beta}^*\big\|_2\\&\le\left(1+\frac{\sqrt{\rho}}{1-\sqrt{\rho}}\right)\cdot\big(\sqrt{\widehat{s}}+C/\sqrt{1-\kappa}\cdot\sqrt{s^*}\big)\cdot\epsilon+\sqrt{\rho}\cdot\rho^{t/2}\cdot\big\|\boldsymbol{\beta}^{(0)}-\boldsymbol{\beta}^*\big\|_2\\&=\frac{\big(\sqrt{\widehat{s}}+C/\sqrt{1-\kappa}\cdot\sqrt{s^*}\big)\cdot\epsilon}{1-\sqrt{\rho}}+\rho^{(t+1)/2}\cdot\big\|\boldsymbol{\beta}^{(0)}-\boldsymbol{\beta}^*\big\|_2.\end{aligned}$$

Here the second inequality is obtained by plugging in (5.11) for $t$. Hence we have that (5.11) holds for $t+1$. By induction, we conclude that (5.11) holds conditioning on the event $\mathcal{E}$ defined in (5.10), which occurs with probability at least $1-\delta$. By plugging the specific definitions of $\rho$ into (5.11), and applying $\big\|\boldsymbol{\beta}^{(0)}-\boldsymbol{\beta}^*\big\|_2\le R$ in (5.13) to the right-hand side of (5.11), we obtain (3.10) and (3.13). □

## 5.2 Proof of Results for Inference

**Proof of Theorem 4.6:** We establish the asymptotic normality of the decorrelated score statistic defined in (2.11) in two steps. We first prove the asymptotic normality of $\sqrt{n}\cdot S_n\big(\widehat{\boldsymbol{\beta}}_0,\lambda\big)$, where $\widehat{\boldsymbol{\beta}}_0$ is defined in (2.11) and $S_n(\cdot,\cdot)$ is defined in (2.9). Then we prove that $-\big[T_n\big(\widehat{\boldsymbol{\beta}}_0\big)\big]_{\alpha|\boldsymbol{\gamma}}$ defined in (2.11) is a consistent estimator of $\sqrt{n}\cdot S_n\big(\widehat{\boldsymbol{\beta}}_0,\lambda\big)$'s asymptotic variance. The next lemma accomplishes the first step. Recall $I(\boldsymbol{\beta}^*)=-\mathbb{E}_{\boldsymbol{\beta}^*}\big[\nabla^2\ell_n(\boldsymbol{\beta}^*)\big]/n$ is the Fisher information for $\ell_n(\boldsymbol{\beta}^*)$ defined in (2.2).

**Lemma 5.3.** Under the assumptions of Theorem 4.6, we have that for $n \to \infty$,

$$\sqrt{n} \cdot S_n\big(\widehat{\boldsymbol{\beta}}_0, \lambda\big) \xrightarrow{D} N\big(0, \big[I(\boldsymbol{\beta}^*)\big]_{\alpha|\boldsymbol{\gamma}}\big),$$

where $\big[I(\boldsymbol{\beta}^*)\big]_{\alpha|\boldsymbol{\gamma}}$ is defined in (4.2).

*Proof.* Our proof consists of two steps. Note that by the definition in (2.9) we have

$$\sqrt{n} \cdot S_n\big(\widehat{\boldsymbol{\beta}}_0, \lambda\big) = \sqrt{n} \cdot \big[\nabla_1 Q_n\big(\widehat{\boldsymbol{\beta}}_0; \widehat{\boldsymbol{\beta}}_0\big)\big]_\alpha - \sqrt{n} \cdot w\big(\widehat{\boldsymbol{\beta}}_0, \lambda\big)^\top \cdot \big[\nabla_1 Q_n\big(\widehat{\boldsymbol{\beta}}_0; \widehat{\boldsymbol{\beta}}_0\big)\big]_{\boldsymbol{\gamma}}. \tag{5.19}$$

Recall that $\mathbf{w}^* = \big[I(\boldsymbol{\beta}^*)\big]_{\boldsymbol{\gamma},\boldsymbol{\gamma}}^{-1} \cdot \big[I(\boldsymbol{\beta}^*)\big]_{\boldsymbol{\gamma},\alpha}$ is defined in (4.1). At the first step, we prove

$$\sqrt{n} \cdot S_n\big(\widehat{\boldsymbol{\beta}}_0, \lambda\big) = \sqrt{n} \cdot \big[\nabla_1 Q_n(\boldsymbol{\beta}^*; \boldsymbol{\beta}^*)\big]_\alpha - \sqrt{n} \cdot (\mathbf{w}^*)^\top \cdot \big[\nabla_1 Q_n(\boldsymbol{\beta}^*; \boldsymbol{\beta}^*)\big]_{\boldsymbol{\gamma}} + o_{\mathbb{P}}(1). \tag{5.20}$$

In other words, replacing $\widehat{\boldsymbol{\beta}}_0$ and $w\big(\widehat{\boldsymbol{\beta}}_0, \lambda\big)$ in (5.19) with the corresponding population quantities $\boldsymbol{\beta}^*$ and $\mathbf{w}^*$ only introduces an $o_{\mathbb{P}}(1)$ error term. Meanwhile, by Lemma 2.1 we have $\nabla_1 Q_n(\boldsymbol{\beta}^*; \boldsymbol{\beta}^*) = \nabla \ell_n(\boldsymbol{\beta}^*)/n$. Recall that $\ell_n(\cdot)$ is the log-likelihood defined in (2.2), which implies that in (5.20)

$$\sqrt{n} \cdot \big[\nabla_1 Q_n(\boldsymbol{\beta}^*; \boldsymbol{\beta}^*)\big]_\alpha - \sqrt{n} \cdot (\mathbf{w}^*)^\top \cdot \big[\nabla_1 Q_n(\boldsymbol{\beta}^*; \boldsymbol{\beta}^*)\big]_{\boldsymbol{\gamma}} = \sqrt{n} \cdot \big[1, -(\mathbf{w}^*)^\top\big] \cdot \nabla \ell_n(\boldsymbol{\beta}^*)/n$$

is a (rescaled) average of $n$ i.i.d. random variables. At the second step, we calculate the mean and variance of each term within this average and invoke the central limit theorem. Finally we combine these two steps by invoking Slutsky's theorem. See §C.3 for a detailed proof. □

The next lemma establishes the consistency of $-\big[T_n\big(\widehat{\boldsymbol{\beta}}_0\big)\big]_{\alpha|\boldsymbol{\gamma}}$ for estimating $\big[I(\boldsymbol{\beta}^*)\big]_{\alpha|\boldsymbol{\gamma}}$. Recall that $\big[T_n\big(\widehat{\boldsymbol{\beta}}_0\big)\big]_{\alpha|\boldsymbol{\gamma}} \in \mathbb{R}$ and $\big[I(\boldsymbol{\beta}^*)\big]_{\alpha|\boldsymbol{\gamma}} \in \mathbb{R}$ are defined in (2.11) and (4.2) respectively.

**Lemma 5.4.** Under the assumptions of Theorem 4.6, we have

$$\big[T_n\big(\widehat{\boldsymbol{\beta}}_0\big)\big]_{\alpha|\boldsymbol{\gamma}} + \big[I(\boldsymbol{\beta}^*)\big]_{\alpha|\boldsymbol{\gamma}} = o_{\mathbb{P}}(1). \tag{5.21}$$

*Proof.* For notational simplicity, we abbreviate $w\big(\widehat{\boldsymbol{\beta}}_0, \lambda\big)$ in the definition of $\big[T_n\big(\widehat{\boldsymbol{\beta}}_0\big)\big]_{\alpha|\boldsymbol{\gamma}}$ as $\widehat{\mathbf{w}}_0$. By (2.11) and (4.3), we have

$$\big[T_n\big(\widehat{\boldsymbol{\beta}}_0\big)\big]_{\alpha|\boldsymbol{\gamma}} = \big(1, -\widehat{\mathbf{w}}_0^\top\big) \cdot T_n\big(\widehat{\boldsymbol{\beta}}_0\big) \cdot \big(1, -\widehat{\mathbf{w}}_0^\top\big)^\top, \quad \big[I(\boldsymbol{\beta}^*)\big]_{\alpha|\boldsymbol{\gamma}} = \big[1, -(\mathbf{w}^*)^\top\big] \cdot I(\boldsymbol{\beta}^*) \cdot \big[1, -(\mathbf{w}^*)^\top\big]^\top.$$

First, we establish the relationship between $\widehat{\mathbf{w}}_0$ and $\mathbf{w}^*$ by analyzing the Dantzig selector in (2.10). Meanwhile, by Lemma 2.1 we have $\mathbb{E}_{\boldsymbol{\beta}^*}\big[T_n(\boldsymbol{\beta}^*)\big] = -I(\boldsymbol{\beta}^*)$. Then by triangle inequality we have

$$\Big|T_n\big(\widehat{\boldsymbol{\beta}}_0\big) + I(\boldsymbol{\beta}^*)\Big| \leq \underbrace{\Big|T_n\big(\widehat{\boldsymbol{\beta}}_0\big) - T_n(\boldsymbol{\beta}^*)\Big|}_{\text{(i)}} + \underbrace{\Big|T_n(\boldsymbol{\beta}^*) - \mathbb{E}_{\boldsymbol{\beta}^*}\big[T_n(\boldsymbol{\beta}^*)\big]\Big|}_{\text{(ii)}}.$$

We prove term (i) is $o_{\mathbb{P}}(1)$ by quantifying the Lipschitz continuity of $T_n(\cdot)$ using Condition 4.4. We then prove term (ii) is $o_{\mathbb{P}}(1)$ by concentration analysis. Together with the result on the relationship between $\widehat{\mathbf{w}}_0$ and $\mathbf{w}^*$ we establish (5.21). See §C.4 for a detailed proof. □

Combining Lemmas 5.3 and 5.4 using Slutsky's theorem, we obtain Theorem 4.6.

**Proof of Theorem 4.7:** Similar to the proof of Theorem 4.6, we first prove $\sqrt{n}\cdot\big[\overline{\alpha}\big(\widehat{\boldsymbol{\beta}},\lambda\big)-\alpha^*\big]$ is asymptotically normal, where $\overline{\alpha}\big(\widehat{\boldsymbol{\beta}},\lambda\big)$ is defined in (2.18), while $\widehat{\boldsymbol{\beta}}$ is the estimator attained by the high dimensional EM algorithm (Algorithm 4). By Lemma 5.4, we then show that $-\big\{\big[T_n\big(\widehat{\boldsymbol{\beta}}\big)\big]_{\alpha|\boldsymbol{\gamma}}\big\}^{-1}$ is a consistent estimator of $\sqrt{n}\cdot\big[\overline{\alpha}\big(\widehat{\boldsymbol{\beta}},\lambda\big)-\alpha^*\big]$'s asymptotic variance. Here recall that $\big[T_n\big(\widehat{\boldsymbol{\beta}}\big)\big]_{\alpha|\boldsymbol{\gamma}}$ is defined by replacing $\widehat{\boldsymbol{\beta}}_0$ with $\widehat{\boldsymbol{\beta}}$ in (2.11). The following lemma accomplishes the first step.

**Lemma 5.5.** Under the assumptions of Theorem 4.7, we have that for $n\to\infty$,

$$\sqrt{n}\cdot\big[\overline{\alpha}\big(\widehat{\boldsymbol{\beta}},\lambda\big)-\alpha^*\big]\xrightarrow{D}N\big(0,\big[I(\boldsymbol{\beta}^*)\big]_{\alpha|\boldsymbol{\gamma}}^{-1}\big),\tag{5.22}$$

where $\big[I(\boldsymbol{\beta}^*)\big]_{\alpha|\boldsymbol{\gamma}}>0$ is defined in (4.2).

*Proof.* Our proof consists of two steps. Similar to the proof of Lemma 5.3, we first prove

$$\sqrt{n}\cdot\big[\overline{\alpha}\big(\widehat{\boldsymbol{\beta}},\lambda\big)-\alpha^*\big]=-\sqrt{n}\cdot\big[I(\boldsymbol{\beta}^*)\big]_{\alpha|\boldsymbol{\gamma}}^{-1}\cdot\Big\{\big[\nabla_1Q_n(\boldsymbol{\beta}^*;\boldsymbol{\beta}^*)\big]_{\alpha}-(\mathbf{w}^*)^\top\cdot\big[\nabla_1Q_n(\boldsymbol{\beta}^*;\boldsymbol{\beta}^*)\big]_{\boldsymbol{\gamma}}\Big\}+o_{\mathbb{P}}(1).$$

Then we prove that, as $n\to\infty$,

$$\sqrt{n}\cdot\big[I(\boldsymbol{\beta}^*)\big]_{\alpha|\boldsymbol{\gamma}}^{-1}\cdot\Big\{\big[\nabla_1Q_n(\boldsymbol{\beta}^*;\boldsymbol{\beta}^*)\big]_{\alpha}-(\mathbf{w}^*)^\top\cdot\big[\nabla_1Q_n(\boldsymbol{\beta}^*;\boldsymbol{\beta}^*)\big]_{\boldsymbol{\gamma}}\Big\}\xrightarrow{D}N\big(0,\big[I(\boldsymbol{\beta}^*)\big]_{\alpha|\boldsymbol{\gamma}}^{-1}\big).$$

Combining the two steps by Slutsky's theorem, we obtain (5.22). See §C.5 for a detailed proof. $\square$

Now we accomplish the second step for proving Theorem 4.7. Note that by replacing $\widehat{\boldsymbol{\beta}}_0$ with $\widehat{\boldsymbol{\beta}}$ in the proof of Lemma 5.4, we have

$$\big[T_n\big(\widehat{\boldsymbol{\beta}}\big)\big]_{\alpha|\boldsymbol{\gamma}}+\big[I(\boldsymbol{\beta}^*)\big]_{\alpha|\boldsymbol{\gamma}}=o_{\mathbb{P}}(1).\tag{5.23}$$

Meanwhile, by the definitions of $\mathbf{w}^*$ and $\big[I(\boldsymbol{\beta}^*)\big]_{\alpha|\boldsymbol{\gamma}}$ in (4.1) and (4.2), we have

$$\begin{aligned}\big[I(\boldsymbol{\beta}^*)\big]_{\alpha|\boldsymbol{\gamma}}&=\big[I(\boldsymbol{\beta}^*)\big]_{\alpha,\alpha}-\big[I(\boldsymbol{\beta}^*)\big]_{\boldsymbol{\gamma},\alpha}^\top\cdot\big[I(\boldsymbol{\beta}^*)\big]_{\boldsymbol{\gamma},\boldsymbol{\gamma}}^{-1}\cdot\big[I(\boldsymbol{\beta}^*)\big]_{\boldsymbol{\gamma},\alpha}\\&=\big[I(\boldsymbol{\beta}^*)\big]_{\alpha,\alpha}-2\cdot(\mathbf{w}^*)^\top\cdot\big[I(\boldsymbol{\beta}^*)\big]_{\boldsymbol{\gamma},\alpha}+(\mathbf{w}^*)^\top\cdot\big[I(\boldsymbol{\beta}^*)\big]_{\boldsymbol{\gamma},\boldsymbol{\gamma}}\cdot\mathbf{w}^*=\big[1,(\mathbf{w}^*)^\top\big]\cdot I(\boldsymbol{\beta}^*)\cdot\big[1,(\mathbf{w}^*)^\top\big]^\top.\end{aligned}\tag{5.24}$$

Note that (4.4) in Assumption 4.5 implies that $I(\boldsymbol{\beta}^*)$ is positive definite, which yields $\big[I(\boldsymbol{\beta}^*)\big]_{\alpha|\boldsymbol{\gamma}}>0$ by (5.24). Also, (4.4) gives $\big[I(\boldsymbol{\beta}^*)\big]_{\alpha|\boldsymbol{\gamma}}=O(1)$ and $\big[I(\boldsymbol{\beta}^*)\big]_{\alpha|\boldsymbol{\gamma}}^{-1}=O(1)$. Hence, by (5.23) we have

$$-\big[T_n\big(\widehat{\boldsymbol{\beta}}\big)\big]_{\alpha|\boldsymbol{\gamma}}^{-1}=\big[I(\boldsymbol{\beta}^*)\big]_{\alpha|\boldsymbol{\gamma}}^{-1}+o_{\mathbb{P}}(1).\tag{5.25}$$

Combining (5.22) in Lemma 5.5 and (5.25) by invoking Slutsky's theorem, we obtain Theorem 4.7.

# 6 Numerical Results

In this section, we lay out numerical results illustrating the computational and statistical efficiency of the methods proposed in §2. In §6.1 and §6.2, we present the results for parameter estimation and asymptotic inference respectively. Throughout §6, we focus on the high dimensional EM algorithm without resampling, which is illustrated in Algorithm 1.

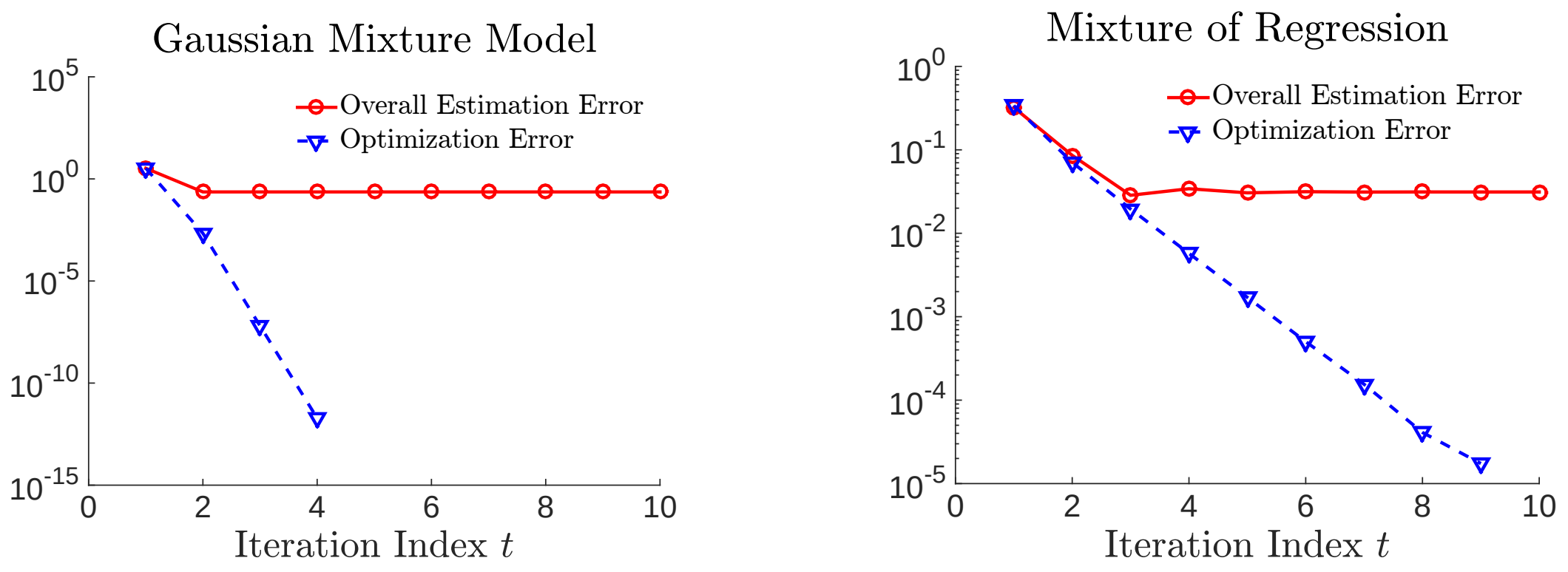

(a) (b)

Figure 1: Evolution of the optimization error $\|\boldsymbol{\beta}^{(t)} - \widehat{\boldsymbol{\beta}}\|_2$ and overall estimation error $\|\boldsymbol{\beta}^{(t)} - \boldsymbol{\beta}^*\|_2$ (in logarithmic scale). (a) The high dimensional EM algorithm for Gaussian mixture model (with the maximization implementation of the M-step). (b) The high dimensional EM algorithm for mixture of regression model (with the gradient ascent implementation of the M-step). Here note that in (a) the optimization error for $t = 5, \ldots, 10$ is truncated due to arithmetic underflow.

### 6.1 Computation and Estimation

We empirically study two latent variable models: Gaussian mixture model and mixture of regression model. The first synthetic dataset is generated according to the Gaussian mixture model defined in (2.24). The second one is generated according to the mixture of regression model defined in (2.25). We set $d = 256$ and $s^* = 5$. In particular, the first 5 entries of $\boldsymbol{\beta}^*$ are set to $\{4, 4, 4, 6, 6\}$, while the rest 251 entries are set to zero. For Gaussian mixture model, the standard deviation of $\boldsymbol{V}$ in (2.24) is set to $\sigma = 1$. For mixture of regression model, the standard deviation of the Gaussian noise $V$ in (2.25) is $\sigma = 0.1$. For both datasets, we generate $n = 100$ data points.

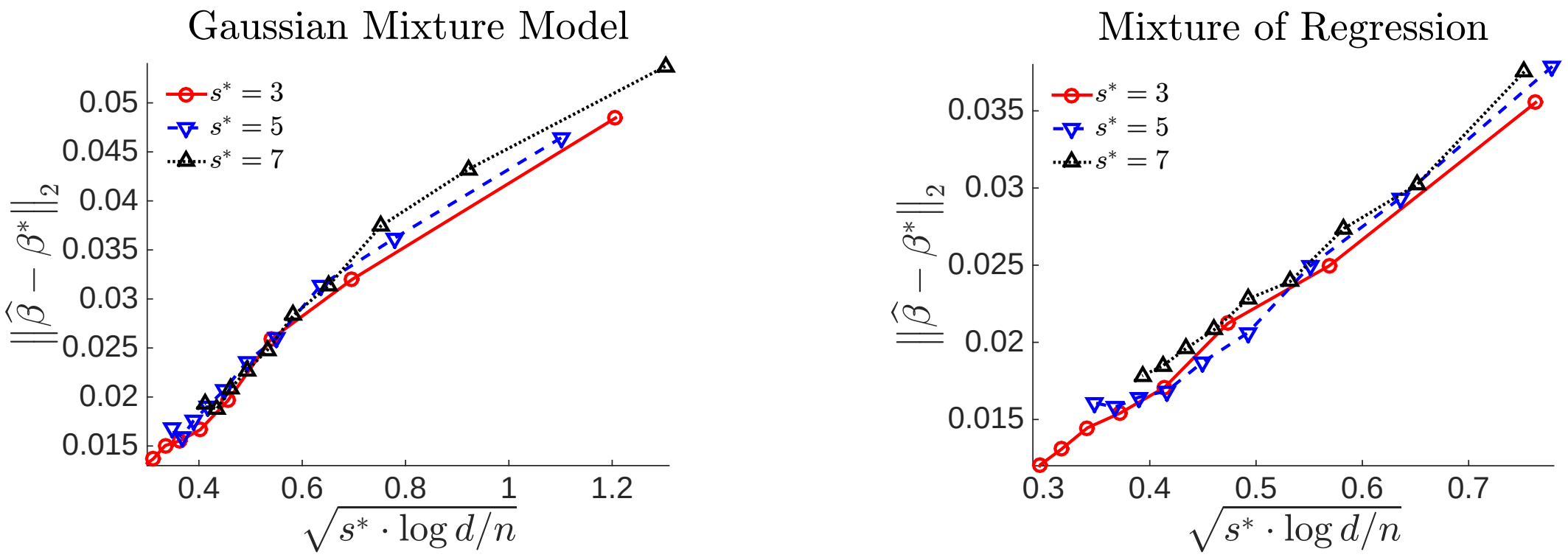

(a) (b)

Figure 2: Statistical error $\|\widehat{\boldsymbol{\beta}} - \boldsymbol{\beta}^*\|_2$ plotted against $\sqrt{s^* \cdot \log d / n}$ with fixed $d = 128$ and varying $s^*$ and $n$. (a) The high dimensional EM algorithm for Gaussian mixture model (with the maximization implementation of the M-step). (b) The high dimensional EM algorithm for mixture of regression model (with the gradient ascent implementation of the M-step).

We apply the proposed high dimensional EM algorithm to both datasets. In particular, we apply the maximization implementation of the M-step (Algorithm 2) to Gaussian mixture model, and the gradient ascent implementation of the M-step (Algorithm 3) to mixture of regression model. We set the stepsize in Algorithm 3 to be $\eta = 1$. In Figure 1, we illustrate the evolution of the optimization error $\|\boldsymbol{\beta}^{(t)} - \widehat{\boldsymbol{\beta}}\|_2$ and overall estimation error $\|\boldsymbol{\beta}^{(t)} - \boldsymbol{\beta}^*\|_2$ with respect to the iteration index $t$. Here $\widehat{\boldsymbol{\beta}}$ is the final estimator and $\boldsymbol{\beta}^*$ is the true parameter.

Figure 1 illustrates the geometric decay of the optimization error, as predicted by Theorem 3.4. As the optimization error decreases to zero, $\boldsymbol{\beta}^{(t)}$ converges to the final estimator $\widehat{\boldsymbol{\beta}}$ and the overall estimation error $\|\boldsymbol{\beta}^{(t)} - \boldsymbol{\beta}^*\|_2$ converges to the statistical error $\|\widehat{\boldsymbol{\beta}} - \boldsymbol{\beta}^*\|_2$. In Figure 2, we illustrate the statistical rate of convergence of $\|\widehat{\boldsymbol{\beta}} - \boldsymbol{\beta}^*\|_2$. In particular, we plot $\|\widehat{\boldsymbol{\beta}} - \boldsymbol{\beta}^*\|_2$ against $\sqrt{s^* \cdot \log d / n}$ with varying $s^*$ and $n$. Figure 2 illustrates that the statistical error exhibits a linear dependency on $\sqrt{s^* \cdot \log d / n}$. Hence, the proposed high dimensional EM algorithm empirically achieves an estimator with the optimal $\sqrt{s^* \cdot \log d / n}$ statistical rate of convergence.

### 6.2 Asymptotic Inference

To examine the validity of the proposed hypothesis testing procedures, we consider the same setting as in §6.1. Recall that we have $\boldsymbol{\beta}^* = [4, 4, 4, 6, 6, 0, \ldots, 0]^\top \in \mathbb{R}^{256}$. We consider the null hypothesis $H_0 : \beta_{10}^* = 0$. We construct the decorrelated score and Wald statistics based on the estimator obtained in the previous section, and fix the significance level to $\delta = 0.05$. We repeat the experiment for 500 times and calculate the rejection rate as the type I error. In Table 1, we report the type I errors of the decorrelated score and Wald tests. In detail, Table 1 illustrates that the type I errors achieved by the proposed hypothesis testing procedures are close to the significance level, which validates our proposed hypothesis testing procedures.

Table 1: Type I errors of the decorrelated Score and Wald tests

| | Gaussian Mixture Model | Mixture of Regression |
|---|---|---|
| Decorrelated Score Test | 0.052 | 0.050 |
| Decorrelated Wald Test | 0.049 | 0.049 |

## 7 Conclusion

We propose a novel high dimensional EM algorithm which naturally incorporates sparsity structure. Our theory shows that, with a suitable initialization, the proposed algorithm converges at a geometric rate and achieves an estimator with the (near-)optimal statistical rate of convergence. Beyond point estimation, we further propose the decorrelated score and Wald statistics for testing hypotheses and constructing confidence intervals for low dimensional components of high dimensional parameters. We apply the proposed algorithmic framework to a broad family of high dimensional latent variable models. For these models, our framework establishes the first computationally feasible approach for optimal parameter estimation and asymptotic inference under high dimensional settings. Thorough numerical simulations are provided to back up our theory.

## Acknowledgement

The authors are grateful for the support of NSF CAREER Award DMS1454377, NSF IIS1408910, NSF IIS1332109, NIH R01MH102339, NIH R01GM083084, and NIH R01HG06841.

# A Applications to Latent Variable Models

We provide the specific forms of $Q_n(\cdot;\cdot)$, $M_n(\cdot)$ and $T_n(\cdot)$ for the models defined in §2.3. Recall that $Q_n(\cdot;\cdot)$ is defined in (2.5), $M_n(\cdot)$ is defined in Algorithms 2 and 3, and $T_n(\cdot)$ is defined in (2.8).

## A.1 Gaussian Mixture Model

Let $\mathbf{y}_1, \ldots, \mathbf{y}_n$ be the $n$ realizations of $\boldsymbol{Y}$. For the E-step (line 4 of Algorithm 1), we have

$$Q_n(\boldsymbol{\beta}'; \boldsymbol{\beta}) = -\frac{1}{2n} \sum_{i=1}^{n} \omega_{\boldsymbol{\beta}}(\mathbf{y}_i) \cdot \|\mathbf{y}_i - \boldsymbol{\beta}'\|_2^2 + \big[1 - \omega_{\boldsymbol{\beta}}(\mathbf{y}_i)\big] \cdot \|\mathbf{y}_i + \boldsymbol{\beta}'\|_2^2, \tag{A.1}$$

$$\text{where} \quad \omega_{\boldsymbol{\beta}}(\mathbf{y}) = \frac{1}{1 + \exp\big(-\langle \boldsymbol{\beta}, \mathbf{y} \rangle / \sigma^2\big)}.$$

The maximization implementation (Algorithm 2) of the M-step takes the form

$$M_n(\boldsymbol{\beta}) = \frac{2}{n} \sum_{i=1}^{n} \omega_{\boldsymbol{\beta}}(\mathbf{y}_i) \cdot \mathbf{y}_i - \frac{1}{n} \sum_{i=1}^{n} \mathbf{y}_i. \tag{A.2}$$

Meanwhile, for the gradient ascent implementation (Algorithm 3) of the M-step, we have

$$M_n(\boldsymbol{\beta}) = \boldsymbol{\beta} + \eta \cdot \nabla_1 Q_n(\boldsymbol{\beta}; \boldsymbol{\beta}), \quad \text{where} \quad \nabla_1 Q_n(\boldsymbol{\beta}; \boldsymbol{\beta}) = \frac{1}{n} \sum_{i=1}^{n} \big[2 \cdot \omega_{\boldsymbol{\beta}}(\mathbf{y}_i) - 1\big] \cdot \mathbf{y}_i - \boldsymbol{\beta}.$$

Here $\eta > 0$ is the stepsize. For asymptotic inference, $T_n(\cdot)$ in (2.8) takes the form

$$T_n(\boldsymbol{\beta}) = \frac{1}{n} \sum_{i=1}^{n} \nu_{\boldsymbol{\beta}}(\mathbf{y}_i) \cdot \mathbf{y}_i \cdot \mathbf{y}_i^\top - \mathbf{I}_d, \quad \text{where} \quad \nu_{\boldsymbol{\beta}}(\mathbf{y}) = \frac{4/\sigma^2}{\big[1 + \exp\big(-2 \cdot \langle \boldsymbol{\beta}, \mathbf{y} \rangle / \sigma^2\big)\big] \cdot \big[1 + \exp\big(2 \cdot \langle \boldsymbol{\beta}, \mathbf{y} \rangle / \sigma^2\big)\big]}.$$

## A.2 Mixture of Regression Model

Let $y_1, \ldots, y_n$ and $\mathbf{x}_1, \ldots, \mathbf{x}_n$ be the $n$ realizations of $Y$ and $\boldsymbol{X}$. For the E-step (line 4 of Algorithm 1), we have

$$Q_n(\boldsymbol{\beta}'; \boldsymbol{\beta}) = -\frac{1}{2n} \sum_{i=1}^{n} \omega_{\boldsymbol{\beta}}(\mathbf{x}_i, y_i) \cdot \big(y_i - \langle \mathbf{x}_i, \boldsymbol{\beta}' \rangle\big)^2 + \big[1 - \omega_{\boldsymbol{\beta}}(\mathbf{x}_i, y_i)\big] \cdot \big(y_i + \langle \mathbf{x}_i, \boldsymbol{\beta}' \rangle\big)^2, \tag{A.3}$$

$$\text{where} \quad \omega_{\boldsymbol{\beta}}(\mathbf{x}, y) = \frac{1}{1 + \exp\big(-y \cdot \langle \boldsymbol{\beta}, \mathbf{x} \rangle / \sigma^2\big)}.$$

For the maximization implementation (Algorithm 2) of the M-step (line 5 of Algorithm 1), we have that $M_n(\boldsymbol{\beta}) = \operatorname{argmax}_{\boldsymbol{\beta}'} Q_n(\boldsymbol{\beta}'; \boldsymbol{\beta})$ satisfies

$$\widehat{\boldsymbol{\Sigma}} \cdot M_n(\boldsymbol{\beta}) = \frac{1}{n} \sum_{i=1}^{n} \big[2 \cdot \omega_{\boldsymbol{\beta}}(\mathbf{x}_i, y_i) - 1\big] \cdot y_i \cdot \mathbf{x}_i, \quad \text{where} \quad \widehat{\boldsymbol{\Sigma}} = \frac{1}{n} \sum_{i=1}^{n} \mathbf{x}_i \cdot \mathbf{x}_i^\top. \tag{A.4}$$

However, in high dimensional regimes, the sample covariance matrix $\widehat{\mathbf{\Sigma}}$ is not invertible. To estimate the inverse covariance matrix of $\boldsymbol{X}$, we use the CLIME estimator proposed by Cai et al. (2011), i.e.,

$$\widehat{\mathbf{\Theta}} = \mathop{\mathrm{argmin}}_{\mathbf{\Theta}\in\mathbb{R}^{d\times d}} \|\mathbf{\Theta}\|_{1,1}, \quad \text{subject to} \quad \big\|\widehat{\mathbf{\Sigma}}\cdot\mathbf{\Theta} - \mathbf{I}_d\big\|_{\infty,\infty} \le \lambda^{\text{CLIME}}, \tag{A.5}$$

where $\|\cdot\|_{1,1}$ and $\|\cdot\|_{\infty,\infty}$ are the sum and maximum of the absolute values of all entries respectively, and $\lambda^{\text{CLIME}} > 0$ is a tuning parameter. Based on (A.4), we modify the maximization implementation of the M-step to be

$$M_n(\boldsymbol{\beta}) = \widehat{\mathbf{\Theta}} \cdot \frac{1}{n}\sum_{i=1}^{n}\big[2\cdot\omega_{\boldsymbol{\beta}}(\mathbf{x}_i, y_i) - 1\big]\cdot y_i \cdot \mathbf{x}_i. \tag{A.6}$$

For the gradient ascent implementation (Algorithm 3) of the M-step, we have

$$M_n(\boldsymbol{\beta}) = \boldsymbol{\beta} + \eta\cdot\nabla_1 Q_n(\boldsymbol{\beta};\boldsymbol{\beta}), \tag{A.7}$$

$$\text{where} \quad \nabla_1 Q_n(\boldsymbol{\beta},\boldsymbol{\beta}) = \frac{1}{n}\sum_{i=1}^{n}\big[2\cdot\omega_{\boldsymbol{\beta}}(\mathbf{x}_i, y_i)\cdot y_i\cdot\mathbf{x}_i - \mathbf{x}_i\cdot\mathbf{x}_i^\top\cdot\boldsymbol{\beta}\big].$$

Here $\eta > 0$ is the stepsize. For asymptotic inference, $T_n(\cdot)$ in (2.8) takes the form

$$T_n(\boldsymbol{\beta}) = \frac{1}{n}\sum_{i=1}^{n}\nu_{\boldsymbol{\beta}}(\mathbf{x}_i, y_i)\cdot\mathbf{x}_i\cdot\mathbf{x}_i^\top\cdot y_i^2 - \frac{1}{n}\sum_{i=1}^{n}\mathbf{x}_i\cdot\mathbf{x}_i^\top,$$

$$\text{where} \quad \nu_{\boldsymbol{\beta}}(\mathbf{x}, y) = \frac{4/\sigma^2}{\big[1+\exp\big(-2\cdot y\cdot\langle\boldsymbol{\beta},\mathbf{x}\rangle/\sigma^2\big)\big]\cdot\big[1+\exp\big(2\cdot y\cdot\langle\boldsymbol{\beta},\mathbf{x}\rangle/\sigma^2\big)\big]}.$$

It is worth noting that, for the maximization implementation of the M-step, the CLIME estimator in (A.5) requires that $\mathbf{\Sigma}^{-1}$ is sparse, where $\mathbf{\Sigma}$ is the population covariance of $\boldsymbol{X}$. Since we assume $\boldsymbol{X}\sim N(\mathbf{0},\mathbf{I}_d)$, this requirement is satisfied. Nevertheless, for more general settings where $\mathbf{\Sigma}$ doesn't possess such a structure, the gradient ascent implementation of the M-step is a better choice, since it doesn't require inverse covariance estimation and is also more efficient in computation.

## A.3 Regression with Missing Covariates

For notational simplicity, we define $\mathbf{z}_i\in\mathbb{R}^d$ $(i=1,\ldots,n)$ as the vector with $z_{i,j}=1$ if $x_{i,j}$ is observed and $z_{i,j}=0$ if $x_{i,j}$ is missing. Let $\mathbf{x}_i^{\text{obs}}$ be the subvector corresponding to the observed component of $\mathbf{x}_i$. For the E-step (line 4 of Algorithm 1), we have

$$Q_n(\boldsymbol{\beta}';\boldsymbol{\beta}) = \frac{1}{n}\sum_{i=1}^{n} y_i\cdot(\boldsymbol{\beta}')^\top\cdot m_{\boldsymbol{\beta}}\big(\mathbf{x}_i^{\text{obs}}, y_i\big) - \frac{1}{2}\cdot(\boldsymbol{\beta}')^\top\cdot K_{\boldsymbol{\beta}}\big(\mathbf{x}_i^{\text{obs}}, y_i\big)\cdot\boldsymbol{\beta}'. \tag{A.8}$$

Here $m_{\boldsymbol{\beta}}(\cdot,\cdot)\in\mathbb{R}^d$ and $K_{\boldsymbol{\beta}}(\cdot,\cdot)\in\mathbb{R}^{d\times d}$ are defined as

$$m_{\boldsymbol{\beta}}\big(\mathbf{x}_i^{\text{obs}}, y_i\big) = \mathbf{z}_i\odot\mathbf{x}_i + \frac{y_i - \langle\boldsymbol{\beta},\mathbf{z}_i\odot\mathbf{x}_i\rangle}{\sigma^2 + \big\|(\mathbf{1}-\mathbf{z}_i)\odot\boldsymbol{\beta}\big\|_2^2}\cdot(\mathbf{1}-\mathbf{z}_i)\odot\boldsymbol{\beta}, \tag{A.9}$$

$$\begin{aligned} K_{\boldsymbol{\beta}}\big(\mathbf{x}_i^{\text{obs}}, y_i\big) = {} & \mathrm{diag}(\mathbf{1}-\mathbf{z}_i) + m_{\boldsymbol{\beta}}\big(\mathbf{x}_i^{\text{obs}}, y_i\big)\cdot\big[m_{\boldsymbol{\beta}}\big(\mathbf{x}_i^{\text{obs}}, y_i\big)\big]^\top \\ & - \big[(\mathbf{1}-\mathbf{z}_i)\odot m_{\boldsymbol{\beta}}\big(\mathbf{x}_i^{\text{obs}}, y_i\big)\big]\cdot\big[(\mathbf{1}-\mathbf{z}_i)\odot m_{\boldsymbol{\beta}}\big(\mathbf{x}_i^{\text{obs}}, y_i\big)\big]^\top, \end{aligned} \tag{A.10}$$

where $\odot$ denotes the Hadamard product and $\mathrm{diag}(\mathbf{1}-\mathbf{z}_i)$ is defined as the $d\times d$ diagonal matrix with the entries of $\mathbf{1}-\mathbf{z}_i \in \mathbb{R}^d$ on its diagonal. Note that maximizing $Q_n(\boldsymbol{\beta}';\boldsymbol{\beta})$ over $\boldsymbol{\beta}'$ requires inverting $K_{\boldsymbol{\beta}}(\mathbf{x}_i^{\mathrm{obs}}, y_i)$, which may be not invertible in high dimensions. Thus, we consider the gradient ascent implementation (Algorithm 3) of the M-step (line 5 of Algorithm 1), in which we have

$$
\begin{aligned}
& M_n(\boldsymbol{\beta}) = \boldsymbol{\beta} + \eta \cdot \nabla_1 Q_n(\boldsymbol{\beta};\boldsymbol{\beta}), \qquad\qquad (A.11)\\
& \text{where} \ \ \nabla_1 Q_n(\boldsymbol{\beta};\boldsymbol{\beta}) = \frac{1}{n}\sum_{i=1}^n y_i \cdot m_{\boldsymbol{\beta}}(\mathbf{x}_i^{\mathrm{obs}}, y_i) - K_{\boldsymbol{\beta}}(\mathbf{x}_i^{\mathrm{obs}}, y_i)\cdot \boldsymbol{\beta}.
\end{aligned}
$$

Here $\eta > 0$ is the stepsize. For asymptotic inference, we can similarly calculate $T_n(\cdot)$ according to its definition in (2.8). However, here we omit the detailed form of $T_n(\cdot)$ since it is overly complicated.

# B Proof of Results for Computation and Estimation

We provide the detailed proof of the main results in §3 for computation and parameter estimation. We first lay out the proof for the general framework, and then the proof for specific models.

## B.1 Proof of Lemma 5.1

*Proof.* Recall $\overline{\boldsymbol{\beta}}^{(t+0.5)}$ and $\overline{\boldsymbol{\beta}}^{(t+1)}$ are defined in (5.1). Note that in (5.4) of Lemma 5.1 we assume

$$\big\|\overline{\boldsymbol{\beta}}^{(t+0.5)} - \boldsymbol{\beta}^*\big\|_2 \le \kappa \cdot \|\boldsymbol{\beta}^*\|_2, \qquad (B.1)$$

which implies

$$(1-\kappa)\cdot\|\boldsymbol{\beta}^*\|_2 \le \big\|\overline{\boldsymbol{\beta}}^{(t+0.5)}\big\|_2 \le (1+\kappa)\cdot\|\boldsymbol{\beta}^*\|_2. \qquad (B.2)$$

For notational simplicity, we define

$$\bar{\boldsymbol{\theta}} = \overline{\boldsymbol{\beta}}^{(t+0.5)}/\big\|\overline{\boldsymbol{\beta}}^{(t+0.5)}\big\|_2, \quad \boldsymbol{\theta} = \boldsymbol{\beta}^{(t+0.5)}/\big\|\overline{\boldsymbol{\beta}}^{(t+0.5)}\big\|_2, \quad \text{and} \quad \boldsymbol{\theta}^* = \boldsymbol{\beta}^*/\|\boldsymbol{\beta}^*\|_2. \qquad (B.3)$$

Note that $\bar{\boldsymbol{\theta}}$ and $\boldsymbol{\theta}^*$ are unit vectors, while $\boldsymbol{\theta}$ is not, since it is obtained by normalizing $\boldsymbol{\beta}^{(t+0.5)}$ with $\big\|\overline{\boldsymbol{\beta}}^{(t+0.5)}\big\|_2$. Recall that the $\mathrm{supp}(\cdot,\cdot)$ function is defined in (2.6). Hence we have

$$\mathrm{supp}(\boldsymbol{\theta}^*) = \mathrm{supp}(\boldsymbol{\beta}^*) = \mathcal{S}^*, \quad \text{and} \quad \mathrm{supp}(\boldsymbol{\theta},\widehat{s}) = \mathrm{supp}\big(\boldsymbol{\beta}^{(t+0.5)},\widehat{s}\big) = \widehat{\mathcal{S}}^{(t+0.5)}, \qquad (B.4)$$

where the last equality follows from line 6 of Algorithm 4. To ease the notation, we define

$$\mathcal{I}_1 = \mathcal{S}^*\setminus\widehat{\mathcal{S}}^{(t+0.5)}, \quad \mathcal{I}_2 = \mathcal{S}^*\cap\widehat{\mathcal{S}}^{(t+0.5)}, \quad \text{and} \quad \mathcal{I}_3 = \widehat{\mathcal{S}}^{(t+0.5)}\setminus\mathcal{S}^*. \qquad (B.5)$$

Let $s_1 = |\mathcal{I}_1|$, $s_2 = |\mathcal{I}_2|$ and $s_3 = |\mathcal{I}_3|$ correspondingly. Also, we define $\Delta = \langle\bar{\boldsymbol{\theta}},\boldsymbol{\theta}^*\rangle$. Note that

$$\Delta = \big\langle\bar{\boldsymbol{\theta}},\boldsymbol{\theta}^*\big\rangle = \sum_{j\in\mathcal{S}^*}\bar{\theta}_j\cdot\theta_j^* = \sum_{j\in\mathcal{I}_1}\bar{\theta}_j\cdot\theta_j^* + \sum_{j\in\mathcal{I}_2}\bar{\theta}_j\cdot\theta_j^* \le \big\|\bar{\boldsymbol{\theta}}_{\mathcal{I}_1}\big\|_2\cdot\big\|\boldsymbol{\theta}^*_{\mathcal{I}_1}\big\|_2 + \big\|\bar{\boldsymbol{\theta}}_{\mathcal{I}_2}\big\|_2\cdot\big\|\boldsymbol{\theta}^*_{\mathcal{I}_2}\big\|_2. \qquad (B.6)$$

Here the first equality is from $\text{supp}(\boldsymbol{\theta}^*) = \mathcal{S}^*$, the second equality is from (B.5) and the last inequality is from Cauchy-Schwarz inequality. Furthermore, from (B.6) we have

$$\begin{aligned}\Delta^2 \le \Big(\|\bar{\boldsymbol{\theta}}_{\mathcal{I}_1}\|_2 \cdot \|\boldsymbol{\theta}^*_{\mathcal{I}_1}\|_2 + \|\bar{\boldsymbol{\theta}}_{\mathcal{I}_2}\|_2 \cdot \|\boldsymbol{\theta}^*_{\mathcal{I}_2}\|_2\Big)^2 &\le \|\bar{\boldsymbol{\theta}}_{\mathcal{I}_1}\|_2^2 \cdot \Big(\|\boldsymbol{\theta}^*_{\mathcal{I}_1}\|_2^2 + \|\boldsymbol{\theta}^*_{\mathcal{I}_2}\|_2^2\Big) + \|\bar{\boldsymbol{\theta}}_{\mathcal{I}_2}\|_2^2 \cdot \Big(\|\boldsymbol{\theta}^*_{\mathcal{I}_1}\|_2^2 + \|\boldsymbol{\theta}^*_{\mathcal{I}_2}\|_2^2\Big) \\ &= \|\bar{\boldsymbol{\theta}}_{\mathcal{I}_1}\|_2^2 + \|\bar{\boldsymbol{\theta}}_{\mathcal{I}_2}\|_2^2 \\ &\le 1 - \|\bar{\boldsymbol{\theta}}_{\mathcal{I}_3}\|_2^2. \end{aligned} \tag{B.7}$$

To obtain the second inequality, we expand the square and apply $2ab \le a^2 + b^2$. In the equality and the last inequality of (B.7), we use the fact that $\boldsymbol{\theta}^*$ and $\bar{\boldsymbol{\theta}}$ are both unit vectors.

By (2.6) and (B.4), $\widehat{\mathcal{S}}^{(t+0.5)}$ contains the index $j$'s with the top $\widehat{s}$ largest $|\beta_j^{(t+0.5)}|$'s. Therefore, we have

$$\frac{\|\boldsymbol{\beta}^{(t+0.5)}_{\mathcal{I}_3}\|_2^2}{s_3} = \frac{\sum_{j\in\mathcal{I}_3}\big(\beta_j^{(t+0.5)}\big)^2}{s_3} \ge \frac{\sum_{j\in\mathcal{I}_1}\big(\beta_j^{(t+0.5)}\big)^2}{s_1} = \frac{\|\boldsymbol{\beta}^{(t+0.5)}_{\mathcal{I}_1}\|_2^2}{s_1}, \tag{B.8}$$

because from (B.5) we have $\mathcal{I}_3 \subseteq \widehat{\mathcal{S}}^{(t+0.5)}$ and $\mathcal{I}_1 \cap \widehat{\mathcal{S}}^{(t+0.5)} = \varnothing$. Taking square roots of both sides of (B.8) and then dividing them by $\|\bar{\boldsymbol{\beta}}^{(t+0.5)}\|_2$ (which is nonzero according to (B.2)), by the definition of $\boldsymbol{\theta}$ in (B.3) we obtain

$$\frac{\|\boldsymbol{\theta}_{\mathcal{I}_3}\|_2}{\sqrt{s_3}} \ge \frac{\|\boldsymbol{\theta}_{\mathcal{I}_1}\|_2}{\sqrt{s_1}}. \tag{B.9}$$

Equipped with (B.9), we now quantify the relationship between $\|\bar{\boldsymbol{\theta}}_{\mathcal{I}_3}\|_2$ and $\|\bar{\boldsymbol{\theta}}_{\mathcal{I}_1}\|_2$. For notational simplicity, let

$$\widetilde{\epsilon} = 2 \cdot \|\bar{\boldsymbol{\theta}} - \boldsymbol{\theta}\|_\infty = 2 \cdot \|\bar{\boldsymbol{\beta}}^{(t+0.5)} - \boldsymbol{\beta}^{(t+0.5)}\|_\infty / \|\bar{\boldsymbol{\beta}}^{(t+0.5)}\|_2. \tag{B.10}$$

Note that we have

$$\max\left\{\frac{\|\boldsymbol{\theta}_{\mathcal{I}_3} - \bar{\boldsymbol{\theta}}_{\mathcal{I}_3}\|_2}{\sqrt{s_3}}, \frac{\|\boldsymbol{\theta}_{\mathcal{I}_1} - \bar{\boldsymbol{\theta}}_{\mathcal{I}_1}\|_2}{\sqrt{s_1}}\right\} \le \max\Big\{\|\boldsymbol{\theta}_{\mathcal{I}_3} - \bar{\boldsymbol{\theta}}_{\mathcal{I}_3}\|_\infty, \|\boldsymbol{\theta}_{\mathcal{I}_1} - \bar{\boldsymbol{\theta}}_{\mathcal{I}_1}\|_\infty\Big\} \le \|\bar{\boldsymbol{\theta}} - \boldsymbol{\theta}\|_\infty = \widetilde{\epsilon}/2,$$

which implies

$$\begin{aligned}\frac{\|\bar{\boldsymbol{\theta}}_{\mathcal{I}_3}\|_2}{\sqrt{s_3}} \ge \frac{\|\boldsymbol{\theta}_{\mathcal{I}_3}\|_2}{\sqrt{s_3}} - \frac{\|\boldsymbol{\theta}_{\mathcal{I}_3} - \bar{\boldsymbol{\theta}}_{\mathcal{I}_3}\|_2}{\sqrt{s_3}} &\ge \frac{\|\boldsymbol{\theta}_{\mathcal{I}_1}\|_2}{\sqrt{s_1}} - \frac{\|\boldsymbol{\theta}_{\mathcal{I}_3} - \bar{\boldsymbol{\theta}}_{\mathcal{I}_3}\|_2}{\sqrt{s_3}} \\ &\ge \frac{\|\bar{\boldsymbol{\theta}}_{\mathcal{I}_1}\|_2}{\sqrt{s_1}} - \frac{\|\bar{\boldsymbol{\theta}}_{\mathcal{I}_1} - \boldsymbol{\theta}_{\mathcal{I}_1}\|_2}{\sqrt{s_1}} - \frac{\|\boldsymbol{\theta}_{\mathcal{I}_3} - \bar{\boldsymbol{\theta}}_{\mathcal{I}_3}\|_2}{\sqrt{s_3}} \ge \frac{\|\bar{\boldsymbol{\theta}}_{\mathcal{I}_1}\|_2}{\sqrt{s_1}} - \widetilde{\epsilon}, \end{aligned} \tag{B.11}$$

where second inequality is obtained from (B.9), while the first and third are from triangle inequality. Plugging (B.11) into (B.7), we obtain

$$\Delta^2 \le 1 - \|\bar{\boldsymbol{\theta}}_{\mathcal{I}_3}\|_2^2 \le 1 - \Big(\sqrt{s_3/s_1} \cdot \|\bar{\boldsymbol{\theta}}_{\mathcal{I}_1}\|_2 - \sqrt{s_3} \cdot \widetilde{\epsilon}\Big)^2.$$

Since by definition we have $\Delta = \langle \overline{\boldsymbol{\theta}}, \boldsymbol{\theta}^* \rangle \in [-1, 1]$, solving for $\|\overline{\boldsymbol{\theta}}_{\mathcal{I}_1}\|_2$ in the above inequality yields

$$\|\overline{\boldsymbol{\theta}}_{\mathcal{I}_1}\|_2 \le \sqrt{s_1/s_3} \cdot \sqrt{1-\Delta^2} + \sqrt{s_1} \cdot \widetilde{\epsilon} \le \sqrt{s^*/\widehat{s}} \cdot \sqrt{1-\Delta^2} + \sqrt{s^*} \cdot \widetilde{\epsilon}. \tag{B.12}$$

Here we employ the fact that $s_1 \le s^*$ and $s_1/s_3 \le (s_1+s_2)/(s_3+s_2) = s^*/\widehat{s}$, which follows from (B.5) and our assumption in (5.5) that $s^*/\widehat{s} \le (1-\kappa)^2/\big[4\cdot(1+\kappa)^2\big] < 1$.

In the following, we prove that the right-hand side of (B.12) is upper bounded by $\Delta$, i.e.,

$$\sqrt{s^*/\widehat{s}} \cdot \sqrt{1-\Delta^2} + \sqrt{s^*} \cdot \widetilde{\epsilon} \le \Delta. \tag{B.13}$$

We can verify that a sufficient condition for (B.13) to hold is that

$$\begin{aligned}\Delta &\ge \frac{\sqrt{s^*} \cdot \widetilde{\epsilon} + \big[s^* \cdot \widetilde{\epsilon}^2 - \big(s^*/\widehat{s} + 1\big) \cdot \big(s^* \cdot \widetilde{\epsilon}^2 - s^*/\widehat{s}\big)\big]^{1/2}}{s^*/\widehat{s} + 1} \\ &= \frac{\sqrt{s^*} \cdot \widetilde{\epsilon} + \big[-\big(s^* \cdot \widetilde{\epsilon}\big)^2/\widehat{s} + \big(s^*/\widehat{s} + 1\big) \cdot \big(s^*/\widehat{s}\big)\big]^{1/2}}{s^*/\widehat{s} + 1},\end{aligned} \tag{B.14}$$

which is obtained by solving for $\Delta$ in (B.13). When we are solving for $\Delta$ in (B.13), we use the fact that $\sqrt{s^*} \cdot \widetilde{\epsilon} \le \Delta$, which holds because

$$\sqrt{s^*} \cdot \widetilde{\epsilon} \le \sqrt{\widehat{s}} \cdot \widetilde{\epsilon} = 2 \cdot \frac{\sqrt{\widehat{s}} \cdot \big\|\overline{\boldsymbol{\beta}}^{(t+0.5)} - \boldsymbol{\beta}^{(t+0.5)}\big\|_\infty}{\big\|\overline{\boldsymbol{\beta}}^{(t+0.5)}\big\|_2} \le \frac{1-\kappa}{1+\kappa} \le \Delta. \tag{B.15}$$

The first inequality is from our assumption in (5.5) that $s^*/\widehat{s} \le (1-\kappa)^2/\big[4\cdot(1+\kappa)^2\big] < 1$. The equality is from the definition of $\widetilde{\epsilon}$ in (B.10). The second inequality follows from our assumption in (5.5) that

$$\sqrt{\widehat{s}} \cdot \big\|\boldsymbol{\beta}^{(t+0.5)} - \overline{\boldsymbol{\beta}}^{(t+0.5)}\big\|_\infty \le \frac{(1-\kappa)^2}{2 \cdot (1+\kappa)} \cdot \|\boldsymbol{\beta}^*\|_2$$

and the first inequality in (B.2). To prove the last inequality in (B.15), we note that (B.1) implies

$$\big\|\overline{\boldsymbol{\beta}}^{(t+0.5)}\big\|_2^2 + \|\boldsymbol{\beta}^*\|_2^2 - 2 \cdot \big\langle \overline{\boldsymbol{\beta}}^{(t+0.5)}, \boldsymbol{\beta}^* \big\rangle = \big\|\overline{\boldsymbol{\beta}}^{(t+0.5)} - \boldsymbol{\beta}^*\big\|_2^2 \le \kappa^2 \cdot \|\boldsymbol{\beta}^*\|_2^2.$$

This together with (B.3) implies

$$\begin{aligned}\Delta = \langle \overline{\boldsymbol{\theta}}, \boldsymbol{\theta}^* \rangle = \frac{\big\langle \overline{\boldsymbol{\beta}}^{(t+0.5)}, \boldsymbol{\beta}^* \big\rangle}{\big\|\overline{\boldsymbol{\beta}}^{(t+0.5)}\big\|_2 \cdot \|\boldsymbol{\beta}^*\|_2} &\ge \frac{\big\|\overline{\boldsymbol{\beta}}^{(t+0.5)}\big\|_2^2 + \|\boldsymbol{\beta}^*\|_2^2 - \kappa^2 \cdot \|\boldsymbol{\beta}^*\|_2^2}{2 \cdot \big\|\overline{\boldsymbol{\beta}}^{(t+0.5)}\big\|_2 \cdot \|\boldsymbol{\beta}^*\|_2} \\ &\ge \frac{(1-\kappa)^2 + 1 - \kappa^2}{2 \cdot (1+\kappa)} = \frac{1-\kappa}{1+\kappa},\end{aligned} \tag{B.16}$$

where in the second inequality we use both sides of (B.2). In summary, we have that (B.15) holds. Now we verify that (B.14) holds. By (B.15) we have

$$\sqrt{\widehat{s}} \cdot \widetilde{\epsilon} \le \frac{1-\kappa}{1+\kappa} < 1 < \sqrt{(s^* + \widehat{s})/\widehat{s}},$$

which implies $\widetilde{\epsilon} \le \sqrt{s^*+\widehat{s}}/\widehat{s}$. For the right-hand side of (B.14) we have

$$\frac{\sqrt{s^*}\cdot\widetilde{\epsilon}+\big[-\big(s^*\cdot\widetilde{\epsilon}\big)^2/\widehat{s}+\big(s^*/\widehat{s}+1\big)\cdot\big(s^*/\widehat{s}\big)\big]^{1/2}}{s^*/\widehat{s}+1} \le \frac{\sqrt{s^*}\cdot\widetilde{\epsilon}+\big[\big(s^*/\widehat{s}+1\big)\cdot\big(s^*/\widehat{s}\big)\big]^{1/2}}{s^*/\widehat{s}+1} \le 2\cdot\sqrt{s^*/(s^*+\widehat{s})}, \tag{B.17}$$

where the last inequality is obtained by plugging in $\widetilde{\epsilon} \le \sqrt{s^*+\widehat{s}}/\widehat{s}$. Meanwhile, note that we have

$$2\cdot\sqrt{s^*/(s^*+\widehat{s})} \le 2\cdot\sqrt{1/\big[1+4\cdot(1+\kappa)^2/(1-\kappa)^2\big]} \le (1-\kappa)/(1+\kappa) \le \Delta, \tag{B.18}$$

where the first inequality is from our assumption in (5.5) that $s^*/\widehat{s} \le (1-\kappa)^2/\big[4\cdot(1+\kappa)^2\big]$, while the last inequality is from (B.16). Combining (B.17) and (B.18), we then obtain (B.14). By (B.14) we further establish (B.13), i.e., the right-hand side of (B.12) is upper bounded by $\Delta$, which implies

$$\big\|\overline{\boldsymbol{\theta}}_{\mathcal{I}_1}\big\|_2 \le \Delta. \tag{B.19}$$

Furthermore, according to (B.6) we have

$$\Delta \le \big\|\overline{\boldsymbol{\theta}}_{\mathcal{I}_1}\big\|_2\cdot\big\|\boldsymbol{\theta}^*_{\mathcal{I}_1}\big\|_2+\big\|\overline{\boldsymbol{\theta}}_{\mathcal{I}_2}\big\|_2\cdot\big\|\boldsymbol{\theta}^*_{\mathcal{I}_2}\big\|_2 \le \big\|\overline{\boldsymbol{\theta}}_{\mathcal{I}_1}\big\|_2\cdot\big\|\boldsymbol{\theta}^*_{\mathcal{I}_1}\big\|_2+\sqrt{1-\big\|\overline{\boldsymbol{\theta}}_{\mathcal{I}_1}\big\|_2^2}\cdot\sqrt{1-\big\|\boldsymbol{\theta}^*_{\mathcal{I}_1}\big\|_2^2}, \tag{B.20}$$

where in the last inequality we use the fact $\boldsymbol{\theta}^*$ and $\overline{\boldsymbol{\theta}}$ are unit vectors. Now we solve for $\big\|\boldsymbol{\theta}^*_{\mathcal{I}_1}\big\|_2$ in (B.20). According to (B.19) and the fact that $\big\|\boldsymbol{\theta}^*_{\mathcal{I}_1}\big\|_2 \le \|\boldsymbol{\theta}^*\|_2 = 1$, on the right-hand side of (B.20) we have $\big\|\overline{\boldsymbol{\theta}}_{\mathcal{I}_1}\big\|_2\cdot\big\|\boldsymbol{\theta}^*_{\mathcal{I}_1}\big\|_2 \le \big\|\overline{\boldsymbol{\theta}}_{\mathcal{I}_1}\big\|_2 \le \Delta$. Thus, we have

$$\Big(\Delta-\big\|\overline{\boldsymbol{\theta}}_{\mathcal{I}_1}\big\|_2\cdot\big\|\boldsymbol{\theta}^*_{\mathcal{I}_1}\big\|_2\Big)^2 \le \Big(1-\big\|\overline{\boldsymbol{\theta}}_{\mathcal{I}_1}\big\|_2^2\Big)\cdot\Big(1-\big\|\boldsymbol{\theta}^*_{\mathcal{I}_1}\big\|_2^2\Big).$$

Further by solving for $\big\|\boldsymbol{\theta}^*_{\mathcal{I}_1}\big\|_2$ in the above inequality, we obtain

$$\begin{aligned}\big\|\boldsymbol{\theta}^*_{\mathcal{I}_1}\big\|_2 \le \big\|\overline{\boldsymbol{\theta}}_{\mathcal{I}_1}\big\|_2\cdot\Delta+\sqrt{1-\big\|\overline{\boldsymbol{\theta}}_{\mathcal{I}_1}\big\|_2^2}\cdot\sqrt{1-\Delta^2} &\le \big\|\overline{\boldsymbol{\theta}}_{\mathcal{I}_1}\big\|_2+\sqrt{1-\Delta^2}\\ &\le \big(1+\sqrt{s^*/\widehat{s}}\big)\cdot\sqrt{1-\Delta^2}+\sqrt{s^*}\cdot\widetilde{\epsilon},\end{aligned} \tag{B.21}$$

where in the second inequality we use the fact that $\Delta\le 1$, which follows from its definition, while in the last inequality we plug in (B.12). Then combining (B.12) and (B.21), we obtain

$$\big\|\overline{\boldsymbol{\theta}}_{\mathcal{I}_1}\big\|_2\cdot\big\|\boldsymbol{\theta}^*_{\mathcal{I}_1}\big\|_2 \le \Big[\sqrt{s^*/\widehat{s}}\cdot\sqrt{1-\Delta^2}+\sqrt{s^*}\cdot\widetilde{\epsilon}\Big]\cdot\Big[\big(1+\sqrt{s^*/\widehat{s}}\big)\cdot\sqrt{1-\Delta^2}+\sqrt{s^*}\cdot\widetilde{\epsilon}\Big]. \tag{B.22}$$

Note that by (5.1) and the definition of $\overline{\boldsymbol{\theta}}$ in (B.3), we have

$$\overline{\boldsymbol{\beta}}^{(t+1)} = \mathrm{trunc}\big(\overline{\boldsymbol{\beta}}^{(t+0.5)},\widehat{\mathcal{S}}^{(t+0.5)}\big) = \mathrm{trunc}\big(\overline{\boldsymbol{\theta}},\widehat{\mathcal{S}}^{(t+0.5)}\big)\cdot\big\|\overline{\boldsymbol{\beta}}^{(t+0.5)}\big\|_2.$$

Therefore, we have

$$\begin{aligned}\Big\langle\overline{\boldsymbol{\beta}}^{(t+1)}/\big\|\overline{\boldsymbol{\beta}}^{(t+0.5)}\big\|_2,\boldsymbol{\beta}^*/\|\boldsymbol{\beta}^*\|_2\Big\rangle = \Big\langle\mathrm{trunc}\big(\overline{\boldsymbol{\theta}},\widehat{\mathcal{S}}^{(t+0.5)}\big),\boldsymbol{\theta}^*\Big\rangle = \big\langle\overline{\boldsymbol{\theta}}_{\mathcal{I}_2},\boldsymbol{\theta}^*_{\mathcal{I}_2}\big\rangle &= \big\langle\overline{\boldsymbol{\theta}},\boldsymbol{\theta}^*\big\rangle-\big\langle\overline{\boldsymbol{\theta}}_{\mathcal{I}_1},\boldsymbol{\theta}^*_{\mathcal{I}_1}\big\rangle\\ &\ge \big\langle\overline{\boldsymbol{\theta}},\boldsymbol{\theta}^*\big\rangle-\big\|\overline{\boldsymbol{\theta}}_{\mathcal{I}_1}\big\|_2\cdot\big\|\boldsymbol{\theta}^*_{\mathcal{I}_1}\big\|_2,\end{aligned}$$

where the second and third equalities follow from (B.5). Let $\overline{\chi}=\|\overline{\boldsymbol{\beta}}^{(t+0.5)}\|_2\cdot\|\boldsymbol{\beta}^*\|_2$. Plugging (B.22) into the right-hand side of the above inequality and then multiplying $\overline{\chi}$ on both sides, we obtain

$$
\begin{aligned}
&\langle\overline{\boldsymbol{\beta}}^{(t+1)},\boldsymbol{\beta}^*\rangle \qquad\qquad \text{(B.23)}\\
&\geq\langle\overline{\boldsymbol{\beta}}^{(t+0.5)},\boldsymbol{\beta}^*\rangle-\Big[\sqrt{s^*/\widehat{s}}\cdot\sqrt{\overline{\chi}\cdot(1-\Delta^2)}+\sqrt{s^*}\cdot\sqrt{\overline{\chi}}\cdot\widetilde{\epsilon}\Big]\cdot\Big[\big(1+\sqrt{s^*/\widehat{s}}\big)\cdot\sqrt{\overline{\chi}\cdot(1-\Delta^2)}+\sqrt{s^*}\cdot\sqrt{\overline{\chi}}\cdot\widetilde{\epsilon}\Big]\\
&=\langle\overline{\boldsymbol{\beta}}^{(t+0.5)},\boldsymbol{\beta}^*\rangle-\big(\sqrt{s^*/\widehat{s}}+s^*/\widehat{s}\big)\cdot\overline{\chi}\cdot(1-\Delta^2)-\big(1+2\cdot\sqrt{s^*/\widehat{s}}\big)\cdot\underbrace{\sqrt{\overline{\chi}\cdot(1-\Delta^2)}}_{\text{(i)}}\cdot\sqrt{s^*}\cdot\underbrace{\sqrt{\overline{\chi}}\cdot\widetilde{\epsilon}}_{\text{(ii)}}-\big(\sqrt{s^*}\cdot\sqrt{\overline{\chi}}\cdot\widetilde{\epsilon}\big)^2.
\end{aligned}
$$

For term (i) in (B.23), note that $\sqrt{1-\Delta^2}\leq\sqrt{2\cdot(1-\Delta)}$. By (B.3) and the definition that $\Delta=\langle\overline{\boldsymbol{\theta}},\boldsymbol{\theta}^*\rangle$, for term (i) we have

$$
\begin{aligned}
\sqrt{\overline{\chi}\cdot(1-\Delta^2)}&\leq\sqrt{2\cdot\overline{\chi}\cdot(1-\Delta)}\leq\sqrt{2\cdot\|\overline{\boldsymbol{\beta}}^{(t+0.5)}\|_2\cdot\|\boldsymbol{\beta}^*\|_2-2\cdot\langle\overline{\boldsymbol{\beta}}^{(t+0.5)},\boldsymbol{\beta}^*\rangle} \qquad \text{(B.24)}\\
&\leq\sqrt{\|\overline{\boldsymbol{\beta}}^{(t+0.5)}\|_2^2+\|\boldsymbol{\beta}^*\|_2^2-2\cdot\langle\overline{\boldsymbol{\beta}}^{(t+0.5)},\boldsymbol{\beta}^*\rangle}=\|\overline{\boldsymbol{\beta}}^{(t+0.5)}-\boldsymbol{\beta}^*\|_2.
\end{aligned}
$$

For term (ii) in (B.23), by the definition of $\widetilde{\epsilon}$ in (B.10) we have

$$
\begin{aligned}
\sqrt{\overline{\chi}}\cdot\widetilde{\epsilon}&=\sqrt{\|\overline{\boldsymbol{\beta}}^{(t+0.5)}\|_2\cdot\|\boldsymbol{\beta}^*\|_2}\cdot2\cdot\|\overline{\boldsymbol{\beta}}^{(t+0.5)}-\boldsymbol{\beta}^{(t+0.5)}\|_\infty/\|\overline{\boldsymbol{\beta}}^{(t+0.5)}\|_2\\
&=2\cdot\|\overline{\boldsymbol{\beta}}^{(t+0.5)}-\boldsymbol{\beta}^{(t+0.5)}\|_\infty\cdot\sqrt{\|\boldsymbol{\beta}^*\|_2/\|\overline{\boldsymbol{\beta}}^{(t+0.5)}\|_2}\leq\frac{2}{\sqrt{1-\kappa}}\cdot\|\overline{\boldsymbol{\beta}}^{(t+0.5)}-\boldsymbol{\beta}^{(t+0.5)}\|_\infty, \qquad \text{(B.25)}
\end{aligned}
$$

where the last inequality is obtained from (B.2). Plugging (B.24) and (B.25) into (B.23), we obtain

$$
\begin{aligned}
\langle\overline{\boldsymbol{\beta}}^{(t+1)},\boldsymbol{\beta}^*\rangle\geq\langle\overline{\boldsymbol{\beta}}^{(t+0.5)},\boldsymbol{\beta}^*\rangle&-\big(\sqrt{s^*/\widehat{s}}+s^*/\widehat{s}\big)\cdot\|\overline{\boldsymbol{\beta}}^{(t+0.5)}-\boldsymbol{\beta}^*\|_2^2 \qquad \text{(B.26)}\\
&-\big(1+2\cdot\sqrt{s^*/\widehat{s}}\big)\cdot\|\overline{\boldsymbol{\beta}}^{(t+0.5)}-\boldsymbol{\beta}^*\|_2\cdot\frac{2\cdot\sqrt{s^*}}{\sqrt{1-\kappa}}\cdot\|\overline{\boldsymbol{\beta}}^{(t+0.5)}-\boldsymbol{\beta}^{(t+0.5)}\|_\infty\\
&-\frac{4\cdot s^*}{1-\kappa}\cdot\|\overline{\boldsymbol{\beta}}^{(t+0.5)}-\boldsymbol{\beta}^{(t+0.5)}\|_\infty^2.
\end{aligned}
$$

Meanwhile, according to (5.1) we have that $\overline{\boldsymbol{\beta}}^{(t+1)}$ is obtained by truncating $\overline{\boldsymbol{\beta}}^{(t+0.5)}$, which implies

$$
\|\overline{\boldsymbol{\beta}}^{(t+1)}\|_2^2+\|\boldsymbol{\beta}^*\|_2^2\leq\|\overline{\boldsymbol{\beta}}^{(t+0.5)}\|_2^2+\|\boldsymbol{\beta}^*\|_2^2. \qquad \text{(B.27)}
$$

Subtracting two times both sides of (B.26) from (B.27), we obtain

$$
\begin{aligned}
\|\overline{\boldsymbol{\beta}}^{(t+1)}-\boldsymbol{\beta}^*\|_2^2\leq\big(1+2\cdot\sqrt{s^*/\widehat{s}}+2\cdot s^*/\widehat{s}\big)&\cdot\|\overline{\boldsymbol{\beta}}^{(t+0.5)}-\boldsymbol{\beta}^*\|_2^2\\
&+\big(1+2\cdot\sqrt{s^*/\widehat{s}}\big)\cdot\frac{4\cdot\sqrt{s^*}}{\sqrt{1-\kappa}}\cdot\|\overline{\boldsymbol{\beta}}^{(t+0.5)}-\boldsymbol{\beta}^{(t+0.5)}\|_\infty\cdot\|\overline{\boldsymbol{\beta}}^{(t+0.5)}-\boldsymbol{\beta}^*\|_2\\
&+\frac{8\cdot s^*}{1-\kappa}\cdot\|\overline{\boldsymbol{\beta}}^{(t+0.5)}-\boldsymbol{\beta}^{(t+0.5)}\|_\infty^2.
\end{aligned}
$$

We can easily verify that the above inequality implies

$$
\begin{aligned}
\|\overline{\boldsymbol{\beta}}^{(t+1)}-\boldsymbol{\beta}^*\|_2^2\leq\big(1+2\cdot\sqrt{s^*/\widehat{s}}+2\cdot s^*/\widehat{s}\big)\cdot\Big[\|\overline{\boldsymbol{\beta}}^{(t+0.5)}-\boldsymbol{\beta}^*\|_2+\frac{2\cdot\sqrt{s^*}}{\sqrt{1-\kappa}}\cdot\|\overline{\boldsymbol{\beta}}^{(t+0.5)}-\boldsymbol{\beta}^{(t+0.5)}\|_\infty\Big]^2\\
+\frac{8\cdot s^*}{1-\kappa}\cdot\|\overline{\boldsymbol{\beta}}^{(t+0.5)}-\boldsymbol{\beta}^{(t+0.5)}\|_\infty^2.
\end{aligned}
$$

Taking square roots of both sides and utilizing the fact that $\sqrt{a^2+b^2} \le a+b$ $(a,b>0)$, we obtain

$$
\begin{aligned}
\big\|\overline{\boldsymbol\beta}^{(t+1)} - \boldsymbol\beta^*\big\|_2 \le \big(1 + 4\cdot\sqrt{s^*/\widehat{s}}\big)^{1/2} \cdot \big\|\overline{\boldsymbol\beta}^{(t+0.5)} - \boldsymbol\beta^*\big\|_2 \\
+ \frac{C\cdot\sqrt{s^*}}{\sqrt{1-\kappa}} \cdot \big\|\overline{\boldsymbol\beta}^{(t+0.5)} - \boldsymbol\beta^{(t+0.5)}\big\|_\infty,
\end{aligned}
\tag{B.28}
$$

where $C > 0$ is an absolute constant. Here we utilize the fact that $s^*/\widehat{s} \le \sqrt{s^*/\widehat{s}}$ and

$$
1 + 2\cdot\sqrt{s^*/\widehat{s}} + 2\cdot s^*/\widehat{s} \le 5,
$$

both of which follow from our assumption that $s^*/\widehat{s} \le (1-\kappa)^2/\big[4\cdot(1+\kappa)^2\big] < 1$ in (5.5). By (B.28) we conclude the proof of Lemma 5.1. □

## B.2 Proof of Lemma 5.2

In the following, we prove (5.8) and (5.9) for the maximization and gradient ascent implementation of the M-step correspondingly.

**Proof of** (5.8)**:** To prove (5.8) for the maximization implementation of the M-step (Algorithm 2), note that by the self-consistency property (McLachlan and Krishnan, 2007) we have

$$
\boldsymbol\beta^* = \operatorname*{argmax}_{\boldsymbol\beta} Q(\boldsymbol\beta;\boldsymbol\beta^*). \tag{B.29}
$$

Hence, $\boldsymbol\beta^*$ satisfies the following first-order optimality condition

$$
\big\langle \boldsymbol\beta - \boldsymbol\beta^*, \nabla_1 Q(\boldsymbol\beta^*;\boldsymbol\beta^*)\big\rangle \le 0, \quad \text{for all } \boldsymbol\beta,
$$

where $\nabla_1 Q(\cdot,\cdot)$ denotes the gradient taken with respect to the first variable. In particular, it implies

$$
\big\langle \overline{\boldsymbol\beta}^{(t+0.5)} - \boldsymbol\beta^*, \nabla_1 Q(\boldsymbol\beta^*;\boldsymbol\beta^*)\big\rangle \le 0. \tag{B.30}
$$

Meanwhile, by (5.1) and the definition of $M(\cdot)$ in (3.1), we have

$$
\overline{\boldsymbol\beta}^{(t+0.5)} = M\big(\boldsymbol\beta^{(t)}\big) = \operatorname*{argmax}_{\boldsymbol\beta} Q\big(\boldsymbol\beta;\boldsymbol\beta^{(t)}\big).
$$

Hence we have the following first-order optimality condition

$$
\big\langle \boldsymbol\beta - \overline{\boldsymbol\beta}^{(t+0.5)}, \nabla_1 Q\big(\overline{\boldsymbol\beta}^{(t+0.5)};\boldsymbol\beta^{(t)}\big)\big\rangle \le 0, \quad \text{for all } \boldsymbol\beta,
$$

which implies

$$
\big\langle \boldsymbol\beta^* - \overline{\boldsymbol\beta}^{(t+0.5)}, \nabla_1 Q\big(\overline{\boldsymbol\beta}^{(t+0.5)};\boldsymbol\beta^{(t)}\big)\big\rangle \le 0. \tag{B.31}
$$

Combining (B.30) and (B.31), we then obtain

$$
\big\langle \boldsymbol\beta^* - \overline{\boldsymbol\beta}^{(t+0.5)}, -\nabla_1 Q(\boldsymbol\beta^*;\boldsymbol\beta^*)\big\rangle \le \big\langle \boldsymbol\beta^* - \overline{\boldsymbol\beta}^{(t+0.5)}, -\nabla_1 Q\big(\overline{\boldsymbol\beta}^{(t+0.5)};\boldsymbol\beta^{(t)}\big)\big\rangle,
$$

which implies

$$\begin{aligned}&\big\langle \boldsymbol{\beta}^* - \overline{\boldsymbol{\beta}}^{(t+0.5)}, \nabla_1 Q\big(\overline{\boldsymbol{\beta}}^{(t+0.5)};\boldsymbol{\beta}^*\big) - \nabla_1 Q(\boldsymbol{\beta}^*;\boldsymbol{\beta}^*)\big\rangle\\ &\quad\le \big\langle \boldsymbol{\beta}^* - \overline{\boldsymbol{\beta}}^{(t+0.5)}, \nabla_1 Q\big(\overline{\boldsymbol{\beta}}^{(t+0.5)};\boldsymbol{\beta}^*\big) - \nabla_1 Q\big(\overline{\boldsymbol{\beta}}^{(t+0.5)};\boldsymbol{\beta}^{(t)}\big)\big\rangle. \qquad \text{(B.32)}\end{aligned}$$

In the following, we establish upper and lower bounds for both sides of (B.32) correspondingly. By applying Condition Lipschitz-Gradient-1$(\gamma_1,\mathcal{B})$, for the right-hand side of (B.32) we have

$$\begin{aligned}&\big\langle \boldsymbol{\beta}^* - \overline{\boldsymbol{\beta}}^{(t+0.5)}, \nabla_1 Q\big(\overline{\boldsymbol{\beta}}^{(t+0.5)};\boldsymbol{\beta}^*\big) - \nabla_1 Q\big(\overline{\boldsymbol{\beta}}^{(t+0.5)};\boldsymbol{\beta}^{(t)}\big)\big\rangle\\ &\quad\le \big\|\boldsymbol{\beta}^* - \overline{\boldsymbol{\beta}}^{(t+0.5)}\big\|_2 \cdot \Big\|\nabla_1 Q\big(\overline{\boldsymbol{\beta}}^{(t+0.5)};\boldsymbol{\beta}^*\big) - \nabla_1 Q\big(\overline{\boldsymbol{\beta}}^{(t+0.5)};\boldsymbol{\beta}^{(t)}\big)\Big\|_2\\ &\quad\le \gamma_1 \cdot \big\|\boldsymbol{\beta}^* - \overline{\boldsymbol{\beta}}^{(t+0.5)}\big\|_2 \cdot \big\|\boldsymbol{\beta}^* - \boldsymbol{\beta}^{(t)}\big\|_2, \qquad \text{(B.33)}\end{aligned}$$

where the last inequality is from (3.3). Meanwhile, for the left-hand side of (B.32), we have

$$Q\big(\overline{\boldsymbol{\beta}}^{(t+0.5)};\boldsymbol{\beta}^*\big) \le Q(\boldsymbol{\beta}^*;\boldsymbol{\beta}^*) + \big\langle \nabla_1 Q(\boldsymbol{\beta}^*;\boldsymbol{\beta}^*), \overline{\boldsymbol{\beta}}^{(t+0.5)} - \boldsymbol{\beta}^*\big\rangle - \nu/2 \cdot \big\|\overline{\boldsymbol{\beta}}^{(t+0.5)} - \boldsymbol{\beta}^*\big\|_2^2, \qquad \text{(B.34)}$$

$$Q(\boldsymbol{\beta}^*;\boldsymbol{\beta}^*) \le Q\big(\overline{\boldsymbol{\beta}}^{(t+0.5)};\boldsymbol{\beta}^*\big) + \big\langle \nabla_1 Q\big(\overline{\boldsymbol{\beta}}^{(t+0.5)};\boldsymbol{\beta}^*\big), \boldsymbol{\beta}^* - \overline{\boldsymbol{\beta}}^{(t+0.5)}\big\rangle - \nu/2 \cdot \big\|\overline{\boldsymbol{\beta}}^{(t+0.5)} - \boldsymbol{\beta}^*\big\|_2^2 \qquad \text{(B.35)}$$

by (3.6) in Condition Concavity-Smoothness$(\mu,\nu,\mathcal{B})$. By adding (B.34) and (B.35), we obtain

$$\nu \cdot \big\|\overline{\boldsymbol{\beta}}^{(t+0.5)} - \boldsymbol{\beta}^*\big\|_2^2 \le \big\langle \boldsymbol{\beta}^* - \overline{\boldsymbol{\beta}}^{(t+0.5)}, \nabla_1 Q\big(\overline{\boldsymbol{\beta}}^{(t+0.5)};\boldsymbol{\beta}^*\big) - \nabla_1 Q(\boldsymbol{\beta}^*;\boldsymbol{\beta}^*)\big\rangle. \qquad \text{(B.36)}$$

Plugging (B.33) and (B.36) into (B.32), we obtain

$$\nu \cdot \big\|\overline{\boldsymbol{\beta}}^{(t+0.5)} - \boldsymbol{\beta}^*\big\|_2^2 \le \gamma_1 \cdot \big\|\boldsymbol{\beta}^* - \overline{\boldsymbol{\beta}}^{(t+0.5)}\big\|_2 \cdot \big\|\boldsymbol{\beta}^* - \boldsymbol{\beta}^{(t)}\big\|_2,$$

which implies (5.8) in Lemma 5.2.

**Proof of** (5.9)**:** We turn to prove (5.9). The self-consistency property in (B.29) implies that $\boldsymbol{\beta}^*$ is the maximizer of $Q(\cdot;\boldsymbol{\beta}^*)$. Furthermore, (3.5) and (3.6) in Condition Concavity-Smoothness$(\mu,\nu,\mathcal{B})$ ensure that $-Q(\cdot;\boldsymbol{\beta}^*)$ is $\mu$-smooth and $\nu$-strongly convex. By invoking standard optimization results for minimizing strongly convex and smooth objective functions, e.g., in Nesterov (2004), for stepsize $\eta = 2/(\nu+\mu)$, we have

$$\Big\|\boldsymbol{\beta}^{(t)} + \eta \cdot \nabla_1 Q\big(\boldsymbol{\beta}^{(t)};\boldsymbol{\beta}^*\big) - \boldsymbol{\beta}^*\Big\|_2 \le \left(\frac{\mu-\nu}{\mu+\nu}\right) \cdot \big\|\boldsymbol{\beta}^{(t)} - \boldsymbol{\beta}^*\big\|_2, \qquad \text{(B.37)}$$

i.e., the gradient ascent step decreases the distance to $\boldsymbol{\beta}^*$ by a multiplicative factor. Hence, for the gradient ascent implementation of the M-step, i.e., $M(\cdot)$ defined in (3.2), we have

$$\begin{aligned}\big\|\overline{\boldsymbol{\beta}}^{(t+0.5)} - \boldsymbol{\beta}^*\big\|_2 &= \Big\|M\big(\boldsymbol{\beta}^{(t)}\big) - \boldsymbol{\beta}^*\Big\|_2\\ &= \Big\|\boldsymbol{\beta}^{(t)} + \eta \cdot \nabla_1 Q\big(\boldsymbol{\beta}^{(t)};\boldsymbol{\beta}^{(t)}\big) - \boldsymbol{\beta}^*\Big\|_2\\ &\le \Big\|\boldsymbol{\beta}^{(t)} + \eta \cdot \nabla_1 Q\big(\boldsymbol{\beta}^{(t)};\boldsymbol{\beta}^*\big) - \boldsymbol{\beta}^*\Big\|_2 + \eta \cdot \Big\|\nabla_1 Q\big(\boldsymbol{\beta}^{(t)};\boldsymbol{\beta}^*\big) - \nabla_1 Q\big(\boldsymbol{\beta}^{(t)};\boldsymbol{\beta}^{(t)}\big)\Big\|_2\\ &\le \left(\frac{\mu-\nu}{\mu+\nu}\right) \cdot \big\|\boldsymbol{\beta}^{(t)} - \boldsymbol{\beta}^*\big\|_2 + \eta \cdot \gamma_2 \cdot \big\|\boldsymbol{\beta}^{(t)} - \boldsymbol{\beta}^*\big\|_2, \qquad \text{(B.38)}\end{aligned}$$

where the last inequality is from (B.37) and (3.4) in Condition Lipschitz-Gradient-2$(\gamma_2, \mathcal{B})$. Plugging $\eta = 2/(\nu + \mu)$ into (B.38), we obtain

$$\big\|\overline{\boldsymbol{\beta}}^{(t+0.5)} - \boldsymbol{\beta}^*\big\|_2 \le \left(\frac{\mu - \nu + 2\cdot\gamma_2}{\mu + \nu}\right) \cdot \big\|\boldsymbol{\beta}^{(t)} - \boldsymbol{\beta}^*\big\|_2,$$

which implies (5.9). Thus, we conclude the proof of Lemma 5.2.

## B.3 Auxiliary Lemma for Proving Theorem 3.4

The following lemma characterizes the initialization step in line 2 of Algorithm 4.

**Lemma B.1.** Suppose that we have $\big\|\boldsymbol{\beta}^{\rm init} - \boldsymbol{\beta}^*\big\|_2 \le \kappa \cdot \|\boldsymbol{\beta}^*\|_2$ for some $\kappa \in (0,1)$. Assuming that $\widehat{s} \ge 4\cdot(1+\kappa)^2/(1-\kappa)^2 \cdot s^*$, we have $\big\|\boldsymbol{\beta}^{(0)} - \boldsymbol{\beta}^*\big\|_2 \le \big(1 + 4\cdot\sqrt{s^*/\widehat{s}}\big)^{1/2} \cdot \big\|\boldsymbol{\beta}^{\rm init} - \boldsymbol{\beta}^*\big\|_2$.

*Proof.* Following the same proof of Lemma 5.1 with both $\overline{\boldsymbol{\beta}}^{(t+0.5)}$ and $\boldsymbol{\beta}^{(t+0.5)}$ replaced with $\boldsymbol{\beta}^{\rm init}$, $\overline{\boldsymbol{\beta}}^{(t+1)}$ replaced with $\boldsymbol{\beta}^{(0)}$ and $\widehat{\mathcal{S}}^{(t+0.5)}$ replaced with $\widehat{\mathcal{S}}^{\rm init}$, we reach the conclusion. □

## B.4 Proof of Lemma 3.6

*Proof.* Recall that $Q(\cdot;\cdot)$ is the expectation of $Q_n(\cdot;\cdot)$. According to (A.1) and (3.1), we have

$$M(\boldsymbol{\beta}) = \mathbb{E}\big[2\cdot\omega_{\boldsymbol{\beta}}(\boldsymbol{Y})\cdot\boldsymbol{Y} - \boldsymbol{Y}\big]$$

with $\omega_{\boldsymbol{\beta}}(\cdot)$ being the weight function defined in (A.1), which together with (A.2) implies

$$M_n(\boldsymbol{\beta}) - M(\boldsymbol{\beta}) = \frac{1}{n}\sum_{i=1}^n \big[2\cdot\omega_{\boldsymbol{\beta}}(\mathbf{y}_i) - 1\big]\cdot\mathbf{y}_i - \mathbb{E}\Big\{\big[2\cdot\omega_{\boldsymbol{\beta}}(\boldsymbol{Y}) - 1\big]\cdot\boldsymbol{Y}\Big\}. \tag{B.39}$$

Recall $\mathbf{y}_i$ is the $i$-th realization of $\boldsymbol{Y}$, which follows the mixture distribution. For any $u>0$, we have

$$\begin{aligned}\mathbb{E}\Big\{\exp\Big[u\cdot\big\|M_n(\boldsymbol{\beta}) - M(\boldsymbol{\beta})\big\|_\infty\Big]\Big\} &= \mathbb{E}\Big\{\max_{j\in\{1,\ldots,d\}} \exp\Big[u\cdot\big|\big[M_n(\boldsymbol{\beta}) - M(\boldsymbol{\beta})\big]_j\big|\Big]\Big\} \\ &\le \sum_{j=1}^d \mathbb{E}\Big\{\exp\Big[u\cdot\big|\big[M_n(\boldsymbol{\beta}) - M(\boldsymbol{\beta})\big]_j\big|\Big]\Big\}. \end{aligned} \tag{B.40}$$

Based on (B.39), we apply the symmetrization result in Lemma D.4 to the right-hand side of (B.40). Then we have

$$\mathbb{E}\Big\{\exp\Big[u\cdot\big\|M_n(\boldsymbol{\beta}) - M(\boldsymbol{\beta})\big\|_\infty\Big]\Big\} \le \sum_{j=1}^d \mathbb{E}\Big\{\exp\Big[u\cdot\Big|\frac{1}{n}\sum_{i=1}^n \xi_i\cdot\big[2\cdot\omega_{\boldsymbol{\beta}}(\mathbf{y}_i) - 1\big]\cdot y_{i,j}\Big|\Big]\Big\}, \tag{B.41}$$

where $\xi_1,\ldots,\xi_n$ are i.i.d. Rademacher random variables that are independent of $\mathbf{y}_1,\ldots,\mathbf{y}_n$. Then we invoke the contraction result in Lemma D.5 by setting

$$f(y_{i,j}) = y_{i,j}, \quad \mathcal{F} = \{f\}, \quad \psi_i(v) = \big[2\cdot\omega_{\boldsymbol{\beta}}(\mathbf{y}_i) - 1\big]\cdot v, \quad \text{and} \quad \phi(v) = \exp(u\cdot v),$$

where $u$ is the variable of the moment generating function in (B.40). From the definition of $\omega_{\boldsymbol{\beta}}(\cdot)$ in (A.1) we have $\big|2\cdot\omega_{\boldsymbol{\beta}}(\mathbf{y}_i)-1\big|\leq 1$, which implies

$$\big|\psi_i(v)-\psi_i(v')\big|\leq\Big|\big[2\cdot\omega_{\boldsymbol{\beta}}(\mathbf{y}_i)-1\big]\cdot(v-v')\Big|\leq|v-v'|,\quad\text{for all } v,v'\in\mathbb{R}.$$

Therefore, by Lemma D.5 we obtain

$$\mathbb{E}\bigg\{\exp\bigg[u\cdot\bigg|\frac{1}{n}\sum_{i=1}^{n}\xi_i\cdot\big[2\cdot\omega_{\boldsymbol{\beta}}(\mathbf{y}_i)-1\big]\cdot y_{i,j}\bigg|\bigg]\bigg\}\leq\mathbb{E}\bigg\{\exp\bigg[2\cdot u\cdot\bigg|\frac{1}{n}\sum_{i=1}^{n}\xi_i\cdot y_{i,j}\bigg|\bigg]\bigg\}\tag{B.42}$$

for the right-hand side of (B.41), where $j\in\{1,\ldots,d\}$. Here note that in Gaussian mixture model we have $y_{i,j}=z_i\cdot\beta_j^*+v_{i,j}$, where $z_i$ is a Rademacher random variable and $v_{i,j}\sim N(0,\sigma^2)$. Therefore, according to Example 5.8 in Vershynin (2010) we have $\|z_i\cdot\beta_j^*\|_{\psi_2}\leq|\beta_j^*|$ and $\|v_{i,j}\|_{\psi_2}\leq C\cdot\sigma$. Hence by Lemma D.1 we have

$$\|y_{i,j}\|_{\psi_2}=\big\|z_i\cdot\beta_j^*+v_{i,j}\big\|_{\psi_2}\leq C\cdot\sqrt{|\beta_j^*|^2+C'\cdot\sigma^2}\leq C\cdot\sqrt{\|\boldsymbol{\beta}^*\|_\infty^2+C'\cdot\sigma^2}.$$

Since $|\xi_i\cdot y_{i,j}|=|y_{i,j}|$, $\xi_i\cdot y_{i,j}$ and $y_{i,j}$ have the same $\psi_2$-norm. Because $\xi_i$ is a Rademacher random variable independent of $y_{i,j}$, we have $\mathbb{E}(\xi_i\cdot y_{i,j})=0$. By Lemma 5.5 in Vershynin (2010), we obtain

$$\mathbb{E}\big[\exp(u'\cdot\xi_i\cdot y_{i,j})\big]\leq\exp\Big[(u')^2\cdot C\cdot\big(\|\boldsymbol{\beta}^*\|_\infty^2+C'\cdot\sigma^2\big)\Big],\quad\text{for all } u'\in\mathbb{R}.\tag{B.43}$$

Hence, for the right-hand side of (B.42) we have

$$\begin{aligned}\mathbb{E}\bigg\{\exp\bigg[2\cdot u\cdot\bigg|\frac{1}{n}\sum_{i=1}^{n}\xi_i\cdot y_{i,j}\bigg|\bigg]\bigg\}&\leq\mathbb{E}\bigg(\max\bigg\{\exp\bigg[2\cdot u\cdot\frac{1}{n}\sum_{i=1}^{n}\xi_i\cdot y_{i,j}\bigg],\ \exp\bigg[-2\cdot u\cdot\frac{1}{n}\sum_{i=1}^{n}\xi_i\cdot y_{i,j}\bigg]\bigg\}\bigg)\\&\leq\mathbb{E}\bigg\{\exp\bigg[2\cdot u\cdot\frac{1}{n}\sum_{i=1}^{n}\xi_i\cdot y_{i,j}\bigg]\bigg\}+\mathbb{E}\bigg\{\exp\bigg[-2\cdot u\cdot\frac{1}{n}\sum_{i=1}^{n}\xi_i\cdot y_{i,j}\bigg]\bigg\}\\&\leq 2\cdot\exp\Big[C\cdot u^2\cdot\big(\|\boldsymbol{\beta}^*\|_\infty^2+C'\cdot\sigma^2\big)/n\Big].\end{aligned}\tag{B.44}$$

Here the last inequality is obtained by plugging (B.43) with $u'=2\cdot u/n$ and $u'=-2\cdot u/n$ respectively into the two terms. Plugging (B.44) into (B.42) and then into (B.41), we obtain

$$\mathbb{E}\left\{\exp\Big[u\cdot\big\|M_n(\boldsymbol{\beta})-M(\boldsymbol{\beta})\big\|_\infty\Big]\right\}\leq 2\cdot d\cdot\exp\Big[C\cdot u^2\cdot\big(\|\boldsymbol{\beta}^*\|_\infty^2+C'\cdot\sigma^2\big)/n\Big].$$

By Chernoff bound we have that, for all $u>0$ and $v>0$,

$$\begin{aligned}\mathbb{P}\Big[\big\|M_n(\boldsymbol{\beta})-M(\boldsymbol{\beta})\big\|_\infty>v\Big]&\leq\mathbb{E}\left\{\exp\Big[u\cdot\big\|M_n(\boldsymbol{\beta})-M(\boldsymbol{\beta})\big\|_\infty\Big]\right\}/\exp(u\cdot v)\\&\leq 2\cdot\exp\Big[C\cdot u^2\cdot\big(\|\boldsymbol{\beta}^*\|_\infty^2+C'\cdot\sigma^2\big)/n-u\cdot v+\log d\Big].\end{aligned}$$

Minimizing the right-hand side over $u$ we obtain

$$\mathbb{P}\Big[\big\|M_n(\boldsymbol{\beta})-M(\boldsymbol{\beta})\big\|_\infty>v\Big]\leq 2\cdot\exp\Big\{-n\cdot v^2\Big/\Big[4\cdot C\cdot\big(\|\boldsymbol{\beta}^*\|_\infty^2+C'\cdot\sigma^2\big)\Big]+\log d\Big\}.$$

Setting the right-hand side to be $\delta$, we have that

$$v=C\cdot\big(\|\boldsymbol{\beta}^*\|_\infty^2+C'\cdot\sigma^2\big)^{1/2}\cdot\sqrt{\frac{\log d+\log(2/\delta)}{n}}\leq C''\cdot\big(\|\boldsymbol{\beta}^*\|_\infty+\sigma\big)\cdot\sqrt{\frac{\log d+\log(2/\delta)}{n}}$$

holds for some absolute constants $C$, $C'$ and $C''$, which completes the proof of Lemma 3.6. □

### B.5 Proof of Lemma 3.9

In the sequel, we first establish the result for the maximization implementation of the M-step and then for the gradient ascent implementation.

**Maximization Implementation:** For the maximization implementation we need to estimate the inverse covariance matrix $\boldsymbol{\Theta}^* = \boldsymbol{\Sigma}^{-1}$ with the CLIME estimator $\widehat{\boldsymbol{\Theta}}$ defined in (A.5). The following lemma from Cai et al. (2011) quantifies the statistical rate of convergence of $\widehat{\boldsymbol{\Theta}}$. Recall that $\|\cdot\|_{1,\infty}$ is defined as the maximum of the row $\ell_1$-norms of a matrix.

**Lemma B.2.** For $\boldsymbol{\Sigma} = \mathbf{I}_d$ and $\lambda^{\mathrm{CLIME}} = C \cdot \sqrt{\log d/n}$ in (A.5), we have that

$$\big\|\widehat{\boldsymbol{\Theta}} - \boldsymbol{\Theta}^*\big\|_{1,\infty} \leq C' \cdot \sqrt{\frac{\log d + \log(1/\delta)}{n}}$$

holds with probability at least $1-\delta$, where $C$ and $C'$ are positive absolute constants.

*Proof.* See the proof of Theorem 6 in Cai et al. (2011) for details. □

Now we are ready to prove (3.21) of Lemma 3.9.

*Proof.* Recall that $Q(\cdot;\cdot)$ is the expectation of $Q_n(\cdot;\cdot)$. According to (A.3) and (3.1), we have

$$M(\boldsymbol{\beta}) = \mathbb{E}\Big\{\big[2\cdot\omega_{\boldsymbol{\beta}}(\boldsymbol{X}, Y) - 1\big]\cdot Y \cdot \boldsymbol{X}\Big\} \tag{B.45}$$

with $\omega_{\boldsymbol{\beta}}(\cdot,\cdot)$ being the weight function defined in (A.3), which together with (A.6) implies

$$M_n(\boldsymbol{\beta}) - M(\boldsymbol{\beta}) = \widehat{\boldsymbol{\Theta}} \cdot \frac{1}{n}\sum_{i=1}^n \big[2\cdot\omega_{\boldsymbol{\beta}}(\mathbf{x}_i, y_i) - 1\big]\cdot y_i \cdot \mathbf{x}_i - \mathbb{E}\Big\{\big[2\cdot\omega_{\boldsymbol{\beta}}(\boldsymbol{X}, Y) - 1\big]\cdot Y \cdot \boldsymbol{X}\Big\}.$$

Here $\widehat{\boldsymbol{\Theta}}$ is the CLIME estimator defined in (A.5). For notational simplicity, we denote

$$\overline{\omega}_i = 2\cdot\omega_{\boldsymbol{\beta}}(\mathbf{x}_i, y_i) - 1, \quad \text{and} \quad \overline{\omega} = 2\cdot\omega_{\boldsymbol{\beta}}(\boldsymbol{X}, Y) - 1. \tag{B.46}$$

It is worth noting that both $\overline{\omega}_i$ and $\overline{\omega}$ depend on $\boldsymbol{\beta}$. Note that we have

$$\begin{aligned}\big\|M_n(\boldsymbol{\beta}) - M(\boldsymbol{\beta})\big\|_\infty &\leq \bigg\|\widehat{\boldsymbol{\Theta}}\cdot\bigg[\frac{1}{n}\sum_{i=1}^n \overline{\omega}_i\cdot y_i\cdot\mathbf{x}_i - \mathbb{E}(\overline{\omega}\cdot Y\cdot\boldsymbol{X})\bigg]\bigg\|_\infty + \Big\|\big(\widehat{\boldsymbol{\Theta}} - \mathbf{I}_d\big)\cdot\mathbb{E}(\overline{\omega}\cdot Y\cdot\boldsymbol{X})\Big\|_\infty \\ &\leq \underbrace{\big\|\widehat{\boldsymbol{\Theta}}\big\|_{1,\infty}}_{\text{(i)}}\cdot\underbrace{\bigg\|\frac{1}{n}\sum_{i=1}^n \overline{\omega}_i\cdot y_i\cdot\mathbf{x}_i - \mathbb{E}(\overline{\omega}\cdot Y\cdot\boldsymbol{X})\bigg\|_\infty}_{\text{(ii)}} + \underbrace{\big\|\widehat{\boldsymbol{\Theta}} - \mathbf{I}_d\big\|_{1,\infty}}_{\text{(iii)}}\cdot\underbrace{\big\|\mathbb{E}(\overline{\omega}\cdot Y\cdot\boldsymbol{X})\big\|_\infty}_{\text{(iv)}}.\end{aligned} \tag{B.47}$$

**Analysis of Term (i):** For term (i) in (B.47), recall that by our model assumption we have $\boldsymbol{\Sigma} = \mathbf{I}_d$, which implies $\boldsymbol{\Theta}^* = \boldsymbol{\Sigma}^{-1} = \mathbf{I}_d$. By Lemma B.2, for a sufficiently large sample size $n$, we have that

$$\big\|\widehat{\boldsymbol{\Theta}}\big\|_{1,\infty} \leq \big\|\widehat{\boldsymbol{\Theta}} - \mathbf{I}_d\big\|_{1,\infty} + \|\mathbf{I}_d\|_{1,\infty} \leq 1/2 + 1 = 3/2 \tag{B.48}$$

holds with probability at least $1-\delta/4$.

**Analysis of Term (ii):** For term (ii) in (B.47), we have that for $u>0$,

$$\begin{aligned}
\mathbb{E}\biggl\{\exp\biggl[u\cdot\biggl\|\frac{1}{n}\sum_{i=1}^{n}\overline{\omega}_i\cdot y_i\cdot\mathbf{x}_i-\mathbb{E}(\overline{\omega}\cdot Y\cdot\boldsymbol{X})\biggr\|_{\infty}\biggr]\biggr\} &= \mathbb{E}\biggl\{\max_{j\in\{1,\ldots,d\}}\exp\biggl[u\cdot\biggl|\frac{1}{n}\sum_{i=1}^{n}\overline{\omega}_i\cdot y_i\cdot x_{i,j}-\mathbb{E}(\overline{\omega}\cdot Y\cdot X_j)\biggr|\biggr]\biggr\}\\
&\le \sum_{j=1}^{d}\mathbb{E}\biggl\{\exp\biggl[u\cdot\biggl|\frac{1}{n}\sum_{i=1}^{n}\overline{\omega}_i\cdot y_i\cdot x_{i,j}-\mathbb{E}(\overline{\omega}\cdot Y\cdot X_j)\biggr|\biggr]\biggr\}\\
&\le \sum_{j=1}^{d}\mathbb{E}\biggl\{\exp\biggl[u\cdot\biggl|\frac{1}{n}\sum_{i=1}^{n}\xi_i\cdot\overline{\omega}_i\cdot y_i\cdot x_{i,j}\biggr|\biggr]\biggr\}, \qquad \text{(B.49)}
\end{aligned}$$

where $\xi_1,\ldots,\xi_n$ are i.i.d. Rademacher random variables. The last inequality follows from Lemma D.4. Furthermore, for the right-hand side of (B.49), we invoke the contraction result in Lemma D.5 by setting

$$f(y_i\cdot x_{i,j})=y_i\cdot x_{i,j},\quad \mathcal{F}=\{f\},\quad \psi_i(v)=\overline{\omega}_i\cdot v,\quad \text{and}\quad \phi(v)=\exp(u\cdot v),$$

where $u$ is the variable of the moment generating function in (B.49). From the definitions in (A.3) and (B.46) we have $|\overline{\omega}_i|=\bigl|2\cdot\omega_{\boldsymbol{\beta}}(\mathbf{x}_i,y_i)-1\bigr|\le 1$, which implies

$$\bigl|\psi_i(v)-\psi_i(v')\bigr|\le\Bigl|\bigl[2\cdot\omega_{\boldsymbol{\beta}}(\mathbf{x}_i,y_i)-1\bigr]\cdot(v-v')\Bigr|\le|v-v'|,\quad \text{for all } v,v'\in\mathbb{R}.$$

By Lemma D.5, we obtain

$$\mathbb{E}\biggl\{\exp\biggl[u\cdot\biggl|\frac{1}{n}\sum_{i=1}^{n}\xi_i\cdot\overline{\omega}_i\cdot y_i\cdot x_{i,j}\biggr|\biggr]\biggr\}\le\mathbb{E}\biggl\{\exp\biggl[2\cdot u\cdot\biggl|\frac{1}{n}\sum_{i=1}^{n}\xi_i\cdot y_i\cdot x_{i,j}\biggr|\biggr]\biggr\} \qquad \text{(B.50)}$$

for $j\in\{1,\ldots,d\}$ on the right-hand side of (B.49). Recall that in mixture of regression model we have $y_i=z_i\cdot\langle\boldsymbol{\beta}^*,\mathbf{x}_i\rangle+v_i$, where $z_i$ is a Rademacher random variable, $v_i\sim N(0,\sigma^2)$, and $\mathbf{x}_i\sim N(\mathbf{0},\mathbf{I}_d)$. Then by Example 5.8 in Vershynin (2010) we have $\|z_i\cdot\langle\boldsymbol{\beta}^*,\mathbf{x}_i\rangle\|_{\psi_2}=\|\langle\boldsymbol{\beta}^*,\mathbf{x}_i\rangle\|_{\psi_2}\le C\cdot\|\boldsymbol{\beta}^*\|_2$ and $\|v_{i,j}\|_{\psi_2}\le C'\cdot\sigma$. By Lemma D.1 we further have

$$\|y_i\|_{\psi_2}=\bigl\|z_i\cdot\langle\boldsymbol{\beta}^*,\mathbf{x}_i\rangle+v_i\bigr\|_{\psi_2}\le\sqrt{C\cdot\|\boldsymbol{\beta}^*\|_2^2+C'\cdot\sigma^2}.$$

Note that we have $\|x_{i,j}\|_{\psi_2}\le C''$ since $x_{i,j}\sim N(0,1)$. Therefore, by Lemma D.2 we have

$$\|\xi_i\cdot y_i\cdot x_{i,j}\|_{\psi_1}=\|y_i\cdot x_{i,j}\|_{\psi_1}\le\max\bigl\{C\cdot\|\boldsymbol{\beta}^*\|_2^2+C'\cdot\sigma^2,\ C''\bigr\}\le C'''\cdot\max\bigl\{\|\boldsymbol{\beta}^*\|_2^2+\sigma^2,\ 1\bigr\}.$$

Since $\xi_i$ is a Rademacher random variable independent of $y_i\cdot x_{i,j}$, we have $\mathbb{E}(\xi_i\cdot y_i\cdot x_{i,j})=0$. Hence, by Lemma 5.15 in Vershynin (2010), we obtain

$$\mathbb{E}\bigl[\exp(u'\cdot\xi_i\cdot y_i\cdot x_{i,j})\bigr]\le\exp\Bigl[C\cdot(u')^2\cdot\max\bigl\{\|\boldsymbol{\beta}^*\|_2^2+\sigma^2,\ 1\bigr\}^2\Bigr] \qquad \text{(B.51)}$$

for all $|u'| \le C'/\max\{\|\boldsymbol{\beta}^*\|_2^2 + \sigma^2, 1\}$. Hence we have

$$\begin{aligned}
\mathbb{E}\biggl\{\exp\biggl[2\cdot u\cdot\biggl|\frac{1}{n}\sum_{i=1}^n \xi_i\cdot y_i\cdot x_{i,j}\biggr|\biggr]\biggr\} &\le \mathbb{E}\biggl(\max\biggl\{\exp\biggl[2\cdot u\cdot\frac{1}{n}\sum_{i=1}^n \xi_i\cdot y_i\cdot x_{i,j}\biggr],\ \exp\biggl[-2\cdot u\cdot\frac{1}{n}\sum_{i=1}^n \xi_i\cdot y_i\cdot x_{i,j}\biggr]\biggr\}\biggr)\\
&\le \mathbb{E}\biggl\{\exp\biggl[2\cdot u\cdot\frac{1}{n}\sum_{i=1}^n \xi_i\cdot y_i\cdot x_{i,j}\biggr]\biggr\} + \mathbb{E}\biggl\{\exp\biggl[-2\cdot u\cdot\frac{1}{n}\sum_{i=1}^n \xi_i\cdot y_i\cdot x_{i,j}\biggr]\biggr\}\\
&\le 2\cdot\exp\Bigl[C\cdot u^2\cdot\max\bigl\{\|\boldsymbol{\beta}^*\|_2^2+\sigma^2,\ 1\bigr\}^2/n\Bigr]. \qquad \text{(B.52)}
\end{aligned}$$

The last inequality is obtained by plugging (B.51) with $u' = 2\cdot u/n$ and $u' = -2\cdot u/n$ correspondingly into the two terms. Here $|u| \le C'\cdot n/\max\{\|\boldsymbol{\beta}^*\|_2^2 + \sigma^2, 1\}$. Plugging (B.52) into (B.50) and further into (B.49), we obtain

$$\mathbb{E}\biggl\{\exp\biggl[u\cdot\biggl\|\frac{1}{n}\sum_{i=1}^n \overline{\omega}_i\cdot y_i\cdot\mathbf{x}_i - \mathbb{E}(\overline{\omega}\cdot Y\cdot\boldsymbol{X})\biggr\|_\infty\biggr]\biggr\} \le 2\cdot d\cdot\exp\Bigl[C\cdot u^2\cdot\max\bigl\{\|\boldsymbol{\beta}^*\|_2^2+\sigma^2,\ 1\bigr\}^2/n\Bigr].$$

By Chernoff bound we have that, for all $v > 0$ and $|u| \le C'\cdot n/\max\{\|\boldsymbol{\beta}^*\|_2^2 + \sigma^2, 1\}$,

$$\begin{aligned}
&\mathbb{P}\biggl[\biggl\|\frac{1}{n}\sum_{i=1}^n \overline{\omega}_i\cdot y_i\cdot\mathbf{x}_i - \mathbb{E}(\overline{\omega}\cdot Y\cdot\boldsymbol{X})\biggr\|_\infty > v\biggr]\\
&\quad\le \mathbb{E}\biggl\{\exp\biggl[u\cdot\biggl\|\frac{1}{n}\sum_{i=1}^n \overline{\omega}_i\cdot y_i\cdot\mathbf{x}_i - \mathbb{E}(\overline{\omega}\cdot Y\cdot\boldsymbol{X})\biggr\|_\infty\biggr]\biggr\}\Big/\exp(u\cdot v)\\
&\quad\le 2\cdot\exp\Bigl[C\cdot u^2\cdot\max\bigl\{\|\boldsymbol{\beta}^*\|_2^2+\sigma^2,\ 1\bigr\}^2/n - u\cdot v + \log d\Bigr].
\end{aligned}$$

Minimizing over $u$ on its right-hand side we have that, for $0 < v \le C''\cdot\max\{\|\boldsymbol{\beta}^*\|_2^2 + \sigma^2, 1\}$,

$$\begin{aligned}
&\mathbb{P}\biggl[\biggl\|\frac{1}{n}\sum_{i=1}^n \overline{\omega}_i\cdot y_i\cdot\mathbf{x}_i - \mathbb{E}(\overline{\omega}\cdot Y\cdot\boldsymbol{X})\biggr\|_\infty > v\biggr]\\
&\quad\le 2\cdot\exp\Bigl\{-n\cdot v^2\Big/\Bigl[4\cdot C\cdot\max\bigl\{\|\boldsymbol{\beta}^*\|_2^2+\sigma^2,\ 1\bigr\}^2\Bigr] + \log d\Bigr\}.
\end{aligned}$$

Setting the right-hand side of the above inequality to be $\delta/2$, we have that

$$\biggl\|\frac{1}{n}\sum_{i=1}^n \overline{\omega}_i\cdot y_i\cdot\mathbf{x}_i - \mathbb{E}(\overline{\omega}\cdot Y\cdot\boldsymbol{X})\biggr\|_\infty \le v = C\cdot\max\bigl\{\|\boldsymbol{\beta}^*\|_2^2+\sigma^2,\ 1\bigr\}\cdot\sqrt{\frac{\log d + \log(4/\delta)}{n}} \qquad \text{(B.53)}$$

holds with probability at least $1 - \delta/2$ for a sufficiently large $n$.

**Analysis of Term (iii):** For term (iii) in (B.47), by Lemma B.2 we have

$$\bigl\|\widehat{\boldsymbol{\Theta}} - \mathbf{I}_d\bigr\|_{1,\infty} \le C\cdot\sqrt{\frac{\log d + \log(4/\delta)}{n}} \qquad \text{(B.54)}$$

with probability at least $1 - \delta/4$ for a sufficiently large $n$.

**Analysis of Term (iv):** For term (iv) in (B.47), recall that by (B.45) and (B.46) we have

$$M(\boldsymbol{\beta}) = \mathbb{E}\Big\{\big[2\cdot\omega_{\boldsymbol{\beta}}(\boldsymbol{X},Y)-1\big]\cdot Y\cdot\boldsymbol{X}\Big\} = \mathbb{E}(\overline{\omega}\cdot Y\cdot\boldsymbol{X}),$$

which implies

$$\begin{aligned}\big\|\mathbb{E}(\overline{\omega}\cdot Y\cdot\boldsymbol{X})\big\|_\infty = \big\|M(\boldsymbol{\beta})\big\|_\infty &\le \big\|M(\boldsymbol{\beta})-\boldsymbol{\beta}^*\big\|_2 + \|\boldsymbol{\beta}^*\|_2\\ &\le \|\boldsymbol{\beta}-\boldsymbol{\beta}^*\|_2 + \|\boldsymbol{\beta}^*\|_2 \le (1+1/32)\cdot\|\boldsymbol{\beta}^*\|_2, \end{aligned} \tag{B.55}$$

where the first inequality follows from triangle inequality and $\|\cdot\|_\infty \le \|\cdot\|_2$, the second inequality is from the proof of (5.8) in Lemma 5.2 with $\overline{\boldsymbol{\beta}}^{(t+0.5)}$ replaced with $\boldsymbol{\beta}$ and the fact that $\gamma_1/\nu < 1$ in (5.8), and the third inequality holds since in Condition $\mathsf{Statistical\text{-}Error}(\epsilon,\delta,s,n,\mathcal{B})$ we suppose that $\boldsymbol{\beta}\in\mathcal{B}$, and for mixture of regression model $\mathcal{B}$ is specified in (3.20).

Plugging (B.48), (B.53), (B.54) and (B.55) into (B.47), by union bound we have that

$$\begin{aligned}\big\|M_n(\boldsymbol{\beta})-M(\boldsymbol{\beta})\big\|_\infty &\le C\cdot\max\big\{\|\boldsymbol{\beta}^*\|_2^2+\sigma^2,\ 1\big\}\cdot\sqrt{\frac{\log d+\log(4/\delta)}{n}} + C'\cdot\|\boldsymbol{\beta}^*\|_2\cdot\sqrt{\frac{\log d+\log(4/\delta)}{n}}\\ &\le C''\cdot\Big[\max\big\{\|\boldsymbol{\beta}^*\|_2^2+\sigma^2,\ 1\big\}+\|\boldsymbol{\beta}^*\|_2\Big]\cdot\sqrt{\frac{\log d+\log(4/\delta)}{n}}\end{aligned}$$

holds with probability at least $1-\delta$. Therefore, we conclude the proof of (3.21) in Lemma 3.9. □

**Gradient Ascent Implementation:** In the following, we prove (3.22) in Lemma 3.9.

*Proof.* Recall that $Q(\cdot;\cdot)$ is the expectation of $Q_n(\cdot;\cdot)$. According to (A.3) and (3.2), we have

$$M(\boldsymbol{\beta}) = \boldsymbol{\beta} + \eta\cdot\mathbb{E}\big[2\cdot\omega_{\boldsymbol{\beta}}(\boldsymbol{X},Y)\cdot Y\cdot\boldsymbol{X}-\boldsymbol{\beta}\big]$$

with $\omega_{\boldsymbol{\beta}}(\cdot,\cdot)$ being the weight function defined in (A.3), which together with (A.7) implies

$$\begin{aligned}&\big\|M_n(\boldsymbol{\beta})-M(\boldsymbol{\beta})\big\|_\infty \\ &\quad= \bigg\|\eta\cdot\frac{1}{n}\sum_{i=1}^n\big[2\cdot\omega_{\boldsymbol{\beta}}(\mathbf{x}_i,y_i)\cdot y_i\cdot\mathbf{x}_i - \mathbf{x}_i\cdot\mathbf{x}_i^\top\cdot\boldsymbol{\beta}\big] - \eta\cdot\mathbb{E}\big[2\cdot\omega_{\boldsymbol{\beta}}(\boldsymbol{X},Y)\cdot Y\cdot\boldsymbol{X}-\boldsymbol{\beta}\big]\bigg\|_\infty\\ &\quad\le \eta\cdot\underbrace{\bigg\|\frac{1}{n}\sum_{i=1}^n\big[2\cdot\omega_{\boldsymbol{\beta}}(\mathbf{x}_i,y_i)\cdot y_i\cdot\mathbf{x}_i\big] - \mathbb{E}\big[2\cdot\omega_{\boldsymbol{\beta}}(\boldsymbol{X},Y)\cdot Y\cdot\boldsymbol{X}\big]\bigg\|_\infty}_{\text{(i)}} + \eta\cdot\underbrace{\bigg\|\frac{1}{n}\sum_{i=1}^n\mathbf{x}_i\cdot\mathbf{x}_i^\top\cdot\boldsymbol{\beta}-\boldsymbol{\beta}\bigg\|_\infty}_{\text{(ii)}}.\end{aligned} \tag{B.56}$$

Here $\eta>0$ denotes the stepsize in Algorithm 3.

**Analysis of Term (i):** For term (i) in (B.56), we redefine $\overline{\omega}_i$ and $\overline{\omega}$ in (B.46) as

$$\overline{\omega}_i = 2\cdot\omega_{\boldsymbol{\beta}}(\mathbf{x}_i,y_i), \quad \text{and} \quad \overline{\omega} = 2\cdot\omega_{\boldsymbol{\beta}}(\boldsymbol{X},Y). \tag{B.57}$$

Note that $|\overline{\omega}_i| = \big|2 \cdot \omega_{\boldsymbol{\beta}}(\mathbf{x}_i, y_i)\big| \leq 2$. Following the same way we establish the upper bound of term (ii) in (B.47), under the new definitions of $\overline{\omega}_i$ and $\overline{\omega}$ in (B.57) we have that

$$\left\|\frac{1}{n}\sum_{i=1}^{n}\big[2\cdot\omega_{\boldsymbol{\beta}}(\mathbf{x}_i, y_i)\cdot y_i\cdot\mathbf{x}_i\big] - \mathbb{E}\big[2\cdot\omega_{\boldsymbol{\beta}}(\boldsymbol{X}, Y)\cdot Y\cdot\boldsymbol{X}\big]\right\|_\infty$$
$$\leq C\cdot\max\big\{\|\boldsymbol{\beta}^*\|_2^2+\sigma^2,\ 1\big\}\cdot\sqrt{\frac{\log d+\log(4/\delta)}{n}}$$

holds with probability at least $1-\delta/2$.

**Analysis of Term (ii):** For term (ii) in (B.56), we have

$$\left\|\frac{1}{n}\sum_{i=1}^{n}\mathbf{x}_i\cdot\mathbf{x}_i^\top\cdot\boldsymbol{\beta}-\boldsymbol{\beta}\right\|_\infty \leq \underbrace{\left\|\frac{1}{n}\sum_{i=1}^{n}\mathbf{x}_i\cdot\mathbf{x}_i^\top-\mathbf{I}_d\right\|_{\infty,\infty}}_{\text{(ii).a}}\cdot\underbrace{\|\boldsymbol{\beta}\|_1}_{\text{(ii).b}}.$$

For term (ii).a, recall by our model assumption we have $\mathbb{E}\big(\boldsymbol{X}\cdot\boldsymbol{X}^\top\big)=\mathbf{I}_d$ and $\mathbf{x}_i$'s are the independent realizations of $\boldsymbol{X}$. Hence we have

$$\left\|\frac{1}{n}\sum_{i=1}^{n}\mathbf{x}_i\cdot\mathbf{x}_i^\top-\mathbf{I}_d\right\|_{\infty,\infty}\leq\max_{j\in\{1,\ldots,d\}}\max_{k\in\{1,\ldots,d\}}\left|\frac{1}{n}\sum_{i=1}^{n}x_{i,j}\cdot x_{i,k}-\mathbb{E}(X_j\cdot X_k)\right|.$$

Since $X_j, X_k \sim N(0,1)$, according to Example 5.8 in Vershynin (2010) we have $\|X_j\|_{\psi_2} = \|X_k\|_{\psi_2} \leq C$. By Lemma D.2, $X_j\cdot X_k$ is a sub-exponential random variable with $\|X_j\cdot X_k\|_{\psi_1}\leq C'$. Moreover, we have $\big\|X_j\cdot X_k-\mathbb{E}(X_j\cdot X_k)\big\|_{\psi_1}\leq C''$ by Lemma D.3. Then by Bernstein's inequality (Proposition 5.16 in Vershynin (2010)) and union bound, we have

$$\mathbb{P}\left[\left\|\frac{1}{n}\sum_{i=1}^{n}\mathbf{x}_i\cdot\mathbf{x}_i^\top-\mathbf{I}_d\right\|_{\infty,\infty}>v\right]\leq 2\cdot d^2\cdot\exp\big(-C\cdot n\cdot v^2\big)$$

for $0<v\leq C'$ and a sufficiently large sample size $n$. Setting its right-hand side to be $\delta/2$, we have

$$\left\|\frac{1}{n}\sum_{i=1}^{n}\mathbf{x}_i\cdot\mathbf{x}_i^\top-\mathbf{I}_d\right\|_{\infty,\infty}\leq C''\cdot\sqrt{\frac{2\cdot\log d+\log(4/\delta)}{n}}$$

holds with probability at least $1-\delta/2$. For term (ii).b we have $\|\boldsymbol{\beta}\|_1\leq\sqrt{s}\cdot\|\boldsymbol{\beta}\|_2$, since in Condition $\mathsf{Statistical\text{-}Error}(\epsilon,\delta,s,n,\mathcal{B})$ we assume $\|\boldsymbol{\beta}\|_0\leq s$. Furthermore, we have $\|\boldsymbol{\beta}\|_2\leq\|\boldsymbol{\beta}^*\|_2+\|\boldsymbol{\beta}^*-\boldsymbol{\beta}\|_2\leq(1+1/32)\cdot\|\boldsymbol{\beta}^*\|_2$, because in Condition $\mathsf{Statistical\text{-}Error}(\epsilon,\delta,s,n,\mathcal{B})$ we assume that $\boldsymbol{\beta}\in\mathcal{B}$, and for mixture of regression model $\mathcal{B}$ is specified in (3.20).

Plugging the above results into the right-hand side of (B.56), by union bound we have that

$$\big\|M_n(\boldsymbol{\beta})-M(\boldsymbol{\beta})\big\|_\infty\leq\eta\cdot C\cdot\max\big\{\|\boldsymbol{\beta}^*\|_2^2+\sigma^2,\ 1\big\}\cdot\sqrt{\frac{\log d+\log(4/\delta)}{n}}$$
$$+\eta\cdot C'\cdot\sqrt{\frac{2\cdot\log d+\log(4/\delta)}{n}}\cdot\sqrt{s}\cdot\|\boldsymbol{\beta}^*\|_2$$
$$\leq\eta\cdot C''\cdot\max\big\{\|\boldsymbol{\beta}^*\|_2^2+\sigma^2,\ 1,\ \sqrt{s}\cdot\|\boldsymbol{\beta}^*\|_2\big\}\cdot\sqrt{\frac{\log d+\log(4/\delta)}{n}}$$

holds with probability at least $1-\delta$. Therefore, we conclude the proof of (3.22) in Lemma 3.9. $\square$

## B.6 Proof of Lemma 3.12

*Proof.* Recall that $Q(\cdot;\cdot)$ is the expectation of $Q_n(\cdot;\cdot)$. Let $\boldsymbol{X}^{\text{obs}}$ be the subvector corresponding to the observed component of $\boldsymbol{X}$ in (2.26). By (A.8) and (3.2), we have

$$M(\boldsymbol{\beta}) = \boldsymbol{\beta} + \eta \cdot \mathbb{E}\Big[K_{\boldsymbol{\beta}}\big(\boldsymbol{X}^{\text{obs}}, Y\big) \cdot \boldsymbol{\beta} + m_{\boldsymbol{\beta}}\big(\boldsymbol{X}^{\text{obs}}, Y\big) \cdot Y\Big]$$

with $\eta > 0$ being the stepsize in Algorithm 3, which together with (A.11) implies

$$\begin{aligned}
\big\|M_n(\boldsymbol{\beta}) - M(\boldsymbol{\beta})\big\|_\infty &= \eta \cdot \Bigg\| \bigg\{ \frac{1}{n}\sum_{i=1}^n K_{\boldsymbol{\beta}}\big(\mathbf{x}_i^{\text{obs}}, y_i\big) - \mathbb{E}\big[K_{\boldsymbol{\beta}}\big(\boldsymbol{X}^{\text{obs}}, Y\big)\big] \bigg\} \cdot \boldsymbol{\beta} \\
&\qquad + \frac{1}{n}\sum_{i=1}^n m_{\boldsymbol{\beta}}\big(\mathbf{x}_i^{\text{obs}}, y_i\big) \cdot y_i - \mathbb{E}\big[m_{\boldsymbol{\beta}}\big(\boldsymbol{X}^{\text{obs}}, Y\big) \cdot Y\big] \Bigg\|_\infty \\
&\le \eta \cdot \underbrace{\bigg\| \frac{1}{n}\sum_{i=1}^n K_{\boldsymbol{\beta}}\big(\mathbf{x}_i^{\text{obs}}, y_i\big) - \mathbb{E}\big[K_{\boldsymbol{\beta}}\big(\boldsymbol{X}^{\text{obs}}, Y\big)\big] \bigg\|_{\infty,\infty}}_{\text{(i)}} \cdot \underbrace{\|\boldsymbol{\beta}\|_1}_{\text{(ii)}} \qquad \text{(B.58)} \\
&\qquad + \eta \cdot \underbrace{\bigg\| \frac{1}{n}\sum_{i=1}^n m_{\boldsymbol{\beta}}\big(\mathbf{x}_i^{\text{obs}}, y_i\big) \cdot y_i - \mathbb{E}\big[m_{\boldsymbol{\beta}}\big(\boldsymbol{X}^{\text{obs}}, Y\big) \cdot Y\big] \bigg\|_\infty}_{\text{(iii)}}.
\end{aligned}$$

Here $K_{\boldsymbol{\beta}}(\cdot,\cdot) \in \mathbb{R}^{d\times d}$ and $m_{\boldsymbol{\beta}}(\cdot,\cdot) \in \mathbb{R}^d$ are defined in (A.9) and (A.10). To ease the notation, let

$$K_{\boldsymbol{\beta}}\big(\mathbf{x}_i^{\text{obs}}, y_i\big) = \overline{\mathbf{K}}^i, \quad K_{\boldsymbol{\beta}}\big(\boldsymbol{X}^{\text{obs}}, Y\big) = \overline{\mathbf{K}}, \quad m_{\boldsymbol{\beta}}\big(\mathbf{x}_i^{\text{obs}}, y_i\big) = \overline{\mathbf{m}}^i, \quad m_{\boldsymbol{\beta}}\big(\boldsymbol{X}^{\text{obs}}, Y\big) = \overline{\mathbf{m}}. \qquad \text{(B.59)}$$

Let the entries of $\boldsymbol{Z} \in \mathbb{R}^d$ be i.i.d. Bernoulli random variables, each of which is zero with probability $p_{\text{m}}$, and $\mathbf{z}_1, \ldots, \mathbf{z}_n$ be the $n$ independent realizations of $\boldsymbol{Z}$. If $z_{i,j} = 1$, then $x_{i,j}$ is observed, otherwise $x_{i,j}$ is missing (unobserved).

**Analysis of Term (i):** For term (i) in (B.58), recall that we have

$$\overline{\mathbf{K}} = \text{diag}(\mathbf{1} - \boldsymbol{Z}) + \overline{\mathbf{m}} \cdot \overline{\mathbf{m}}^\top - \big[(\mathbf{1} - \boldsymbol{Z}) \odot \overline{\mathbf{m}}\big] \cdot \big[(\mathbf{1} - \boldsymbol{Z}) \odot \overline{\mathbf{m}}\big]^\top,$$

where $\text{diag}(\mathbf{1} - \boldsymbol{Z})$ denotes the $d \times d$ diagonal matrix with the entries of $\mathbf{1} - \boldsymbol{Z} \in \mathbb{R}^d$ on its diagonal, and $\odot$ denotes the Hadamard product. Therefore, by union bound we have

$$\begin{aligned}
&\mathbb{P}\Bigg[\bigg\|\frac{1}{n}\sum_{i=1}^n \overline{\mathbf{K}}^i - \mathbb{E}\,\overline{\mathbf{K}}\bigg\|_{\infty,\infty} > v\Bigg] \qquad \text{(B.60)} \\
&\quad \le \sum_{j=1}^d \sum_{k=1}^d \Bigg\{ \mathbb{P}\Bigg[\bigg|\frac{1}{n}\sum_{i=1}^n \big[\text{diag}(\mathbf{1}-\mathbf{z}_i)\big]_{j,k} - \mathbb{E}\big[\text{diag}(\mathbf{1}-\boldsymbol{Z})\big]_{j,k}\bigg| > v/3\Bigg] \\
&\qquad + \mathbb{P}\Bigg[\bigg|\frac{1}{n}\sum_{i=1}^n \overline{m}_j^i \cdot \overline{m}_k^i - \mathbb{E}(\overline{m}_j \cdot \overline{m}_k)\bigg| > v/3\Bigg]\Bigg\} \\
&\qquad + \mathbb{P}\Bigg[\bigg|\frac{1}{n}\sum_{i=1}^n (1-z_{i,j}) \cdot \overline{m}_j^i \cdot (1-z_{i,k}) \cdot \overline{m}_k^i - \mathbb{E}\big[(1-Z_j) \cdot \overline{m}_j \cdot (1-Z_k) \cdot \overline{m}_k\big]\bigg| > v/3\Bigg]\Bigg\}.
\end{aligned}$$

According to Lemma B.3 we have, for all $j \in \{1, \ldots, d\}$, $\overline{m}_j$ is a zero-mean sub-Gaussian random variable with $\|\overline{m}_j\|_{\psi_2} \leq C \cdot (1 + \kappa \cdot r)$. Then by Lemmas D.2 and D.3 we have

$$\big\|\overline{m}_j \cdot \overline{m}_k - \mathbb{E}(\overline{m}_j \cdot \overline{m}_k)\big\|_{\psi_1} \leq 2 \cdot \|\overline{m}_j \cdot \overline{m}_k\|_{\psi_1} \leq C' \cdot \max\big\{\|\overline{m}_j\|_{\psi_2}^2,\ \|\overline{m}_k\|_{\psi_2}^2\big\} \leq C'' \cdot (1 + \kappa \cdot r)^2.$$

Meanwhile, since $|1 - Z_j| \leq 1$, it holds that $\big\|(1 - Z_j) \cdot \overline{m}_j\big\|_{\psi_2} \leq \|\overline{m}_j\|_{\psi_2} \leq C \cdot (1 + \kappa \cdot r)$. Similarly, by Lemmas D.2 and D.3 we have

$$\big\|(1 - Z_j) \cdot \overline{m}_j \cdot (1 - Z_k) \cdot \overline{m}_k - \mathbb{E}\big[(1 - Z_j) \cdot \overline{m}_j \cdot (1 - Z_k) \cdot \overline{m}_k\big]\big\|_{\psi_1} \leq C'' \cdot (1 + \kappa \cdot r)^2.$$

In addition, for the first term on the right-hand side of (B.60) we have

$$\Big\|\big[\mathrm{diag}(\mathbf{1} - \boldsymbol{Z})\big]_{j,k} - \mathbb{E}\big[\mathrm{diag}(\mathbf{1} - \boldsymbol{Z})\big]_{j,k}\Big\|_{\psi_2} \leq 2 \cdot \Big\|\big[\mathrm{diag}(\mathbf{1} - \boldsymbol{Z})\big]_{j,k}\Big\|_{\psi_2} \leq 2,$$

where the first inequality is from Lemma D.3 and the second inequality follows from Example 5.8 in Vershynin (2010) since $\big[\mathrm{diag}(\mathbf{1} - \boldsymbol{Z})\big]_{j,k} \in [0, 1]$. Applying Hoeffding's inequality to the first term on the right-hand side of (B.60) and Bernstein's inequality to the second and third terms, we obtain

$$\mathbb{P}\bigg[\bigg\|\frac{1}{n}\sum_{i=1}^n \overline{\mathbf{K}}^i - \mathbb{E}\,\overline{\mathbf{K}}\bigg\|_{\infty,\infty} > v\bigg] \leq d^2 \cdot 2 \cdot \exp\big(-C \cdot n \cdot v^2\big) + d^2 \cdot 4 \cdot \exp\big[-C' \cdot n \cdot v^2/(1 + \kappa \cdot r)^4\big].$$

Setting the two terms on the right-hand side of the above inequality to be $\delta/6$ and $\delta/3$ respectively, we have that

$$\begin{aligned}\bigg\|\frac{1}{n}\sum_{i=1}^n \overline{\mathbf{K}}^i - \mathbb{E}\,\overline{\mathbf{K}}\bigg\|_{\infty,\infty} &\leq v = C'' \cdot \max\big\{(1 + \kappa \cdot r)^2, ; 1\big\} \cdot \sqrt{\frac{\log d + \log(12/\delta)}{n}} \\ &= C'' \cdot (1 + \kappa \cdot r)^2 \cdot \sqrt{\frac{\log d + \log(12/\delta)}{n}}\end{aligned}$$

holds with probability at least $1 - \delta/2$, for sufficiently large constant $C''$ and sample size $n$.

**Analysis of Term (ii):** For term (ii) in (B.58) we have

$$\|\boldsymbol{\beta}\|_1 \leq \sqrt{s} \cdot \|\boldsymbol{\beta}\|_2 \leq \sqrt{s} \cdot \|\boldsymbol{\beta}^*\|_2 + \sqrt{s} \cdot \|\boldsymbol{\beta}^* - \boldsymbol{\beta}\|_2 \leq \sqrt{s} \cdot (1 + \kappa) \cdot \|\boldsymbol{\beta}^*\|_2,$$

where the first inequality holds because in Condition Statistical-Error$(\epsilon, \delta, s, n, \mathcal{B})$ we assume $\|\boldsymbol{\beta}\|_0 \leq s$, while the last inequality holds since in Condition Statistical-Error$(\epsilon, \delta, s, n, \mathcal{B})$ we assume that $\boldsymbol{\beta} \in \mathcal{B}$, and for regression with missing covariates $\mathcal{B}$ is specified in (3.27).

**Analysis of Term (iii):** For term (iii) in (B.58), by union bound we have

$$\begin{aligned}\mathbb{P}\bigg[\bigg\|\frac{1}{n}\sum_{i=1}^n \overline{\mathbf{m}}^i \cdot y_i - \mathbb{E}(\overline{\mathbf{m}} \cdot Y)\bigg\|_{\infty} > v\bigg] &\leq \sum_{j=1}^d \mathbb{P}\bigg[\bigg|\frac{1}{n}\sum_{i=1}^n \overline{m}_j^i \cdot y_i - \mathbb{E}(\overline{m}_j \cdot Y)\bigg| > v\bigg] \\ &\leq d \cdot 2 \cdot \exp\Big[C \cdot n \cdot v^2\Big/ \max\Big\{(1 + \kappa \cdot r)^4,\ \big(\sigma^2 + \|\boldsymbol{\beta}^*\|_2^2\big)^2\Big\}\Big]\end{aligned} \tag{B.61}$$

for a sufficiently large $n$. Here the second inequality is from Bernstein's inequality, since we have

$$\big\|\overline{m}_j \cdot Y - \mathbb{E}(\overline{m}_j \cdot Y)\big\|_{\psi_1} \leq C' \cdot \max\big\{\|\overline{m}_j\|_{\psi_2}^2,\ \|Y\|_{\psi_2}^2\big\} \leq C'' \cdot \max\big\{(1 + \kappa \cdot r)^2,\ \sigma^2 + \|\boldsymbol{\beta}^*\|_2^2\big\}.$$

Here the first inequality follows from Lemmas D.2 and D.3 and the second inequality is from Lemma B.3 and the fact that $Y = \langle \boldsymbol{X}, \boldsymbol{\beta}^* \rangle + V \sim N\big(0, \|\boldsymbol{\beta}^*\|_2^2 + \sigma^2\big)$, because $\boldsymbol{X} \sim N(0, \mathbf{I}_d)$ and $V \sim N(0, \sigma^2)$ are independent. Setting the right-hand side of (B.61) to be $\delta/2$, we have that

$$\bigg\| \frac{1}{n} \sum_{i=1}^{n} \overline{\mathbf{m}}^i \cdot y_i - \mathbb{E}(\overline{\mathbf{m}} \cdot Y) \bigg\|_\infty \le v = C \cdot \max\big\{(1 + \kappa \cdot r),\ \sigma^2 + \|\boldsymbol{\beta}^*\|_2^2\big\} \cdot \sqrt{\frac{\log d + \log(4/\delta)}{n}}$$

holds with probability at least $1 - \delta/2$ for sufficiently large constant $C$ and sample size $n$.

Finally, plugging in the upper bounds for terms (i)-(iii) in (B.58) we have that

$$\begin{aligned} &\big\| M_n(\boldsymbol{\beta}) - M(\boldsymbol{\beta}) \big\|_\infty \\ &\le C \cdot \eta \cdot \Big[\sqrt{s} \cdot \|\boldsymbol{\beta}^*\|_2 \cdot (1+\kappa) \cdot (1 + \kappa \cdot r)^2 + \max\big\{(1 + \kappa \cdot r)^2,\ \sigma^2 + \|\boldsymbol{\beta}^*\|_2^2\big\}\Big] \cdot \sqrt{\frac{\log d + \log(12/\delta)}{n}} \end{aligned}$$

holds with probability at least $1 - \delta$. Therefore, we conclude the proof of Lemma 3.12. □

The following auxiliary lemma used in the proof of Lemma 3.12 characterizes the sub-Gaussian property of $\overline{\mathbf{m}}$ defined in (B.59).

**Lemma B.3.** Under the assumptions of Lemma 3.12, the random vector $\overline{\mathbf{m}} \in \mathbb{R}^d$ defined in (B.59) is sub-Gaussian with mean zero and $\|\overline{m}_j\|_{\psi_2} \le C \cdot (1 + \kappa \cdot r)$ for all $j \in \{1, \ldots, d\}$, where $C > 0$ is a sufficiently large absolute constant.

*Proof.* In the proof of Lemma 3.12, $\boldsymbol{Z}$'s entries are i.i.d. Bernoulli random variables, which satisfy $\mathbb{P}(Z_j = 0) = p_{\mathrm{m}}$ for all $j \in \{1, \ldots, d\}$. Meanwhile, from the definitions in (A.9) and (B.59), we have

$$\overline{\mathbf{m}} = \boldsymbol{Z} \odot \boldsymbol{X} + \frac{Y - \langle \boldsymbol{\beta}, \boldsymbol{Z} \odot \boldsymbol{X} \rangle}{\sigma^2 + \big\|(\mathbf{1} - \boldsymbol{Z}) \odot \boldsymbol{\beta}\big\|_2^2} \cdot (\mathbf{1} - \boldsymbol{Z}) \odot \boldsymbol{\beta}.$$

Since we have $Y = \langle \boldsymbol{\beta}^*, \boldsymbol{X} \rangle + V$ and $-\langle \boldsymbol{\beta}, \boldsymbol{Z} \odot \boldsymbol{X} \rangle = -\langle \boldsymbol{\beta}, \boldsymbol{X} \rangle + \big\langle \boldsymbol{\beta}, (1 - \boldsymbol{Z}) \odot \boldsymbol{X} \big\rangle$, it holds that

$$\overline{m}_j = \underbrace{Z_j \cdot X_j}_{\text{(i)}} + \underbrace{\frac{V \cdot (1 - Z_j) \cdot \beta_j}{\sigma^2 + \big\|(\mathbf{1} - \boldsymbol{Z}) \odot \boldsymbol{\beta}\big\|_2^2}}_{\text{(ii)}} + \underbrace{\frac{\langle \boldsymbol{\beta}^* - \boldsymbol{\beta}, \boldsymbol{X} \rangle \cdot (1 - Z_j) \cdot \beta_j}{\sigma^2 + \big\|(\mathbf{1} - \boldsymbol{Z}) \odot \boldsymbol{\beta}\big\|_2^2}}_{\text{(iii)}} + \underbrace{\frac{\big\langle \boldsymbol{\beta}, (1 - \boldsymbol{Z}) \odot \boldsymbol{X} \big\rangle \cdot (1 - Z_j) \cdot \beta_j}{\sigma^2 + \big\|(\mathbf{1} - \boldsymbol{Z}) \odot \boldsymbol{\beta}\big\|_2^2}}_{\text{(iv)}}. \tag{B.62}$$

Since $\boldsymbol{X} \sim N(\mathbf{0}, \mathbf{I}_d)$ is independent of $\boldsymbol{Z}$, we can verify that $\mathbb{E}(\overline{m}_j) = 0$. In the following we provide upper bounds for the $\psi_2$-norms of terms (i)-(iv) in (B.62).

**Analysis of Terms (i) and (ii):** For term (i), since $X_j \sim N(0, 1)$, we have $\|Z_j \cdot X_j\|_{\psi_2} \le \|X_j\|_{\psi_2} \le C$, where the last inequality follows from Example 5.8 in Vershynin (2010). Meanwhile, for term (ii)

in (B.62), we have that for any $u > 0$,

$$\mathbb{P}\left[\left|\frac{V\cdot(1-Z_j)\cdot\beta_j}{\sigma^2+\big\|(\mathbf{1}-\boldsymbol{Z})\odot\boldsymbol{\beta}\big\|_2^2}\right| > u\right] \leq \mathbb{P}\left[\frac{|V|\cdot\big\|(1-\boldsymbol{Z})\odot\boldsymbol{\beta}\big\|_2}{\sigma^2+\big\|(\mathbf{1}-\boldsymbol{Z})\odot\boldsymbol{\beta}\big\|_2^2} > u\right]$$

$$= \mathbb{E}_{\boldsymbol{Z}}\left\{\mathbb{P}\left[\frac{|V|\cdot\big\|(1-\boldsymbol{Z})\odot\boldsymbol{\beta}\big\|_2}{\sigma^2+\big\|(\mathbf{1}-\boldsymbol{Z})\odot\boldsymbol{\beta}\big\|_2^2} > u\,\middle|\,\boldsymbol{Z}\right]\right\}$$

$$\leq \mathbb{E}_{\boldsymbol{Z}}\left\{2\cdot\exp\left(-C\cdot u^2\cdot\left[\sigma^2+\big\|(\mathbf{1}-\boldsymbol{Z})\odot\boldsymbol{\beta}\big\|_2^2\right]^2\Big/\left[\sigma^2\cdot\big\|(\mathbf{1}-\boldsymbol{Z})\odot\boldsymbol{\beta}\big\|_2^2\right]\right)\right\}$$

$$\leq \mathbb{E}_{\boldsymbol{Z}}\big\{2\cdot\exp\big(-C'\cdot u^2\big)\big\} = 2\cdot\exp\big(-C'\cdot u^2\big), \tag{B.63}$$

where the second last inequality is from the fact that $\|V\|_{\psi_2} \leq C''\cdot\sigma$, while the last inequality holds because we have

$$\left[\sigma^2+\big\|(\mathbf{1}-\boldsymbol{Z})\odot\boldsymbol{\beta}\big\|_2^2\right]^2 \geq \sigma^2\cdot\big\|(\mathbf{1}-\boldsymbol{Z})\odot\boldsymbol{\beta}\big\|_2^2. \tag{B.64}$$

According to Lemma 5.5 in Vershynin (2010), by (B.63) we conclude that the $\psi_2$-norm of term (ii) in (B.62) is upper bounded by some absolute constant $C > 0$.

**Analysis of Term (iii):** For term (iii), by the same conditioning argument in (B.63), we have

$$\mathbb{P}\left[\left|\frac{\langle\boldsymbol{\beta}^*-\boldsymbol{\beta},\boldsymbol{X}\rangle\cdot(1-Z_j)\cdot\beta_j}{\sigma^2+\big\|(\mathbf{1}-\boldsymbol{Z})\odot\boldsymbol{\beta}\big\|_2^2}\right| > u\right] \tag{B.65}$$

$$\leq \mathbb{E}_{\boldsymbol{Z}}\left\{2\cdot\exp\left(-C\cdot u^2\cdot\left[\sigma^2+\big\|(\mathbf{1}-\boldsymbol{Z})\odot\boldsymbol{\beta}\big\|_2^2\right]^2\Big/\left[\|\boldsymbol{\beta}^*-\boldsymbol{\beta}\|_2^2\cdot\big\|(\mathbf{1}-\boldsymbol{Z})\odot\boldsymbol{\beta}\big\|_2^2\right]\right)\right\},$$

where we utilize the fact that $\big\|\langle\boldsymbol{\beta}^*-\boldsymbol{\beta},\boldsymbol{X}\rangle\big\|_{\psi_2} \leq C'\cdot\|\boldsymbol{\beta}^*-\boldsymbol{\beta}\|_2$, since $\langle\boldsymbol{\beta}^*-\boldsymbol{\beta},\boldsymbol{X}\rangle\sim N\big(0,\|\boldsymbol{\beta}^*-\boldsymbol{\beta}\|_2^2\big)$. Note that in Condition Statistical-Error$(\epsilon,\delta,s,n,\mathcal{B})$ we assume $\boldsymbol{\beta}\in\mathcal{B}$, and for regression with missing covariates $\mathcal{B}$ is specified in (3.27). Hence we have

$$\|\boldsymbol{\beta}^*-\boldsymbol{\beta}\|_2^2 \leq \kappa^2\cdot\|\boldsymbol{\beta}^*\|_2^2 \leq \kappa^2\cdot r^2\cdot\sigma^2,$$

where the first inequality follows from (3.27), and the second inequality follows from our assumption that $\|\boldsymbol{\beta}^*\|_2/\sigma \leq r$ on the maximum signal-to-noise ratio. By invoking (B.64), from (B.65) we have

$$\mathbb{P}\left[\left|\frac{\langle\boldsymbol{\beta}^*-\boldsymbol{\beta},\boldsymbol{X}\rangle\cdot(1-Z_j)\cdot\beta_j}{\sigma^2+\big\|(\mathbf{1}-\boldsymbol{Z})\odot\boldsymbol{\beta}\big\|_2^2}\right| > u\right] \leq 2\cdot\exp\left[-C'\cdot u^2/\big(\kappa^2\cdot r^2\big)\right],$$

which further implies that the $\psi_2$-norm of term (iii) in (B.62) is upper bounded by $C''\cdot\kappa\cdot r$.

**Analysis of Term (iv):** For term (iv), note that $\big\langle\boldsymbol{\beta},(1-\boldsymbol{Z})\odot\boldsymbol{X}\big\rangle=\big\langle(1-\boldsymbol{Z})\odot\boldsymbol{\beta},\boldsymbol{X}\big\rangle$. By invoking the same conditioning argument in (B.63), we have

$$\mathbb{P}\left[\left|\frac{\big\langle\boldsymbol{\beta},(1-\boldsymbol{Z})\odot\boldsymbol{X}\big\rangle\cdot(1-Z_j)\cdot\beta_j}{\sigma^2+\big\|(\mathbf{1}-\boldsymbol{Z})\odot\boldsymbol{\beta}\big\|_2^2}\right| > u\right] \tag{B.66}$$

$$\leq \mathbb{E}_{\boldsymbol{Z}}\left\{2\cdot\exp\left(-C\cdot u^2\cdot\left[\sigma^2+\big\|(\mathbf{1}-\boldsymbol{Z})\odot\boldsymbol{\beta}\big\|_2^2\right]^2\Big/\big\|(\mathbf{1}-\boldsymbol{Z})\odot\boldsymbol{\beta}\big\|_2^4\right)\right\}$$

$$\leq \mathbb{E}_{\boldsymbol{Z}}\left[2\cdot\exp\big(-C\cdot u^2\big)\right] = 2\cdot\exp\big(-C\cdot u^2\big),$$

where we utilize the fact that $\big\|\langle(1-\boldsymbol{Z})\odot\boldsymbol{\beta},\boldsymbol{X}\rangle\big\|_{\psi_2}\le C'\cdot\big\|(1-\boldsymbol{Z})\odot\boldsymbol{\beta}\big\|_2$ when conditioning on $\boldsymbol{Z}$. By (B.66) we have that the $\psi_2$-norm of term (iv) in (B.62) is upper bounded by $C''>0$.

Combining the above upper bounds for the $\psi_2$-norms of terms (i)-(iv) in (B.62), by Lemma D.1 we have that

$$\|\overline{m}_j\|_{\psi_2}\le\sqrt{C+C'\cdot\kappa^2\cdot r^2}\le C''\cdot(1+\kappa\cdot r)$$

holds for a sufficiently large absolute constant $C''>0$. Therefore, we establish Lemma B.3. □

# C Proof of Results for Inference

In the following, we provide the detailed proof of the theoretical results for asymptotic inference in §4. We first present the proof of the general results, and then the proof for specific models.

## C.1 Proof of Lemma 2.1

*Proof.* In the sequel we establish the two equations in (2.12) respectively.

**Proof of the First Equation:** According to the definition of the lower bound function $Q_n(\cdot;\cdot)$ in (2.5), we have

$$Q_n(\boldsymbol{\beta}';\boldsymbol{\beta})=\frac{1}{n}\sum_{i=1}^n\int_{\mathcal{Z}}k_{\boldsymbol{\beta}}(\mathbf{z}\mid\mathbf{y}_i)\cdot\log f_{\boldsymbol{\beta}'}(\mathbf{y}_i,\mathbf{z})\,\mathrm{d}\mathbf{z}. \tag{C.1}$$

Here $k_{\boldsymbol{\beta}}(\mathbf{z}\mid\mathbf{y}_i)$ is the density of the latent variable $\boldsymbol{Z}$ conditioning on the observed variable $\boldsymbol{Y}=\mathbf{y}_i$ under the model with parameter $\boldsymbol{\beta}$. Hence we obtain

$$\nabla_1 Q_n(\boldsymbol{\beta};\boldsymbol{\beta})=\frac{1}{n}\sum_{i=1}^n\int_{\mathcal{Z}}k_{\boldsymbol{\beta}}(\mathbf{z}\mid\mathbf{y}_i)\cdot\frac{\partial f_{\boldsymbol{\beta}}(\mathbf{y}_i,\mathbf{z})/\partial\boldsymbol{\beta}}{f_{\boldsymbol{\beta}}(\mathbf{y}_i,\mathbf{z})}\,\mathrm{d}\mathbf{z}=\frac{1}{n}\sum_{i=1}^n\int_{\mathcal{Z}}\frac{\partial f_{\boldsymbol{\beta}}(\mathbf{y}_i,\mathbf{z})/\partial\boldsymbol{\beta}}{h_{\boldsymbol{\beta}}(\mathbf{y}_i)}\,\mathrm{d}\mathbf{z}, \tag{C.2}$$

where $h_{\boldsymbol{\beta}}(\mathbf{y}_i)$ is the marginal density function of $\boldsymbol{Y}$ evaluated at $\mathbf{y}_i$, and the second equality follows from the fact that

$$k_{\boldsymbol{\beta}}(\mathbf{z}\mid\mathbf{y}_i)=f_{\boldsymbol{\beta}}(\mathbf{y}_i,\mathbf{z})/h_{\boldsymbol{\beta}}(\mathbf{y}_i), \tag{C.3}$$

since $k_{\boldsymbol{\beta}}(\mathbf{z}\mid\mathbf{y}_i)$ is the conditional density. According to the definition in (2.2), we have

$$\nabla\ell_n(\boldsymbol{\beta})=\sum_{i=1}^n\frac{\partial\log h_{\boldsymbol{\beta}}(\mathbf{y}_i)}{\partial\boldsymbol{\beta}}=\sum_{i=1}^n\frac{\partial h_{\boldsymbol{\beta}}(\mathbf{y}_i)/\partial\boldsymbol{\beta}}{h_{\boldsymbol{\beta}}(\mathbf{y}_i)}=\sum_{i=1}^n\int_{\mathcal{Z}}\frac{\partial f_{\boldsymbol{\beta}}(\mathbf{y}_i,\mathbf{z})/\partial\boldsymbol{\beta}}{h_{\boldsymbol{\beta}}(\mathbf{y}_i)}\,\mathrm{d}\mathbf{z}, \tag{C.4}$$

where the last equality is from (2.1). Comparing (C.2) and (C.4), we obtain $\nabla_1 Q_n(\boldsymbol{\beta};\boldsymbol{\beta})=\nabla\ell_n(\boldsymbol{\beta})/n$.

**Proof of the Second Equation:** For the second equation in (2.12), by (C.1) and (C.3) we have

$$Q_n(\boldsymbol{\beta}';\boldsymbol{\beta})=\frac{1}{n}\sum_{i=1}^n\int_{\mathcal{Z}}\frac{f_{\boldsymbol{\beta}}(\mathbf{y}_i,\mathbf{z})}{h_{\boldsymbol{\beta}}(\mathbf{y}_i)}\cdot\log f_{\boldsymbol{\beta}'}(\mathbf{y}_i,\mathbf{z})\,\mathrm{d}\mathbf{z}.$$

By calculation we obtain

$$\nabla^2_{1,2} Q_n(\boldsymbol{\beta};\boldsymbol{\beta}) = \frac{1}{n}\sum_{i=1}^n \int_{\mathcal{Z}} \frac{\partial f_{\boldsymbol{\beta}}(\mathbf{y}_i,\mathbf{z})/\partial\boldsymbol{\beta}}{f_{\boldsymbol{\beta}}(\mathbf{y}_i,\mathbf{z})} \otimes \left\{ \frac{\partial f_{\boldsymbol{\beta}}(\mathbf{y}_i,\mathbf{z})/\partial\boldsymbol{\beta}\cdot h_{\boldsymbol{\beta}}(\mathbf{y}_i)}{\big[h_{\boldsymbol{\beta}}(\mathbf{y}_i)\big]^2} - \frac{f_{\boldsymbol{\beta}}(\mathbf{y}_i,\mathbf{z})\cdot \partial h_{\boldsymbol{\beta}}(\mathbf{y}_i)/\partial\boldsymbol{\beta}}{\big[h_{\boldsymbol{\beta}}(\mathbf{y}_i)\big]^2}\right\} \mathrm{d}\mathbf{z}. \tag{C.5}$$

Here $\otimes$ denotes the vector outer product. Note that in (C.5) we have

$$\begin{aligned}
\int_{\mathcal{Z}} \frac{\partial f_{\boldsymbol{\beta}}(\mathbf{y}_i,\mathbf{z})/\partial\boldsymbol{\beta}}{f_{\boldsymbol{\beta}}(\mathbf{y}_i,\mathbf{z})} \otimes \frac{\partial f_{\boldsymbol{\beta}}(\mathbf{y}_i,\mathbf{z})/\partial\boldsymbol{\beta}}{h_{\boldsymbol{\beta}}(\mathbf{y}_i)} \mathrm{d}\mathbf{z} &= \int_{\mathcal{Z}} \left[\frac{\partial f_{\boldsymbol{\beta}}(\mathbf{y}_i,\mathbf{z})/\partial\boldsymbol{\beta}}{f_{\boldsymbol{\beta}}(\mathbf{y}_i,\mathbf{z})}\right]^{\otimes 2} \cdot \frac{f_{\boldsymbol{\beta}}(\mathbf{y}_i,\mathbf{z})}{h_{\boldsymbol{\beta}}(\mathbf{y}_i)} \mathrm{d}\mathbf{z} \\
&= \int_{\mathcal{Z}} \left[\frac{\partial f_{\boldsymbol{\beta}}(\mathbf{y}_i,\mathbf{z})/\partial\boldsymbol{\beta}}{f_{\boldsymbol{\beta}}(\mathbf{y}_i,\mathbf{z})}\right]^{\otimes 2} \cdot k_{\boldsymbol{\beta}}(\mathbf{z}\mid \mathbf{y}_i)\, \mathrm{d}\mathbf{z} \\
&= \mathbb{E}_{\boldsymbol{\beta}}\Big[\widetilde{S}_{\boldsymbol{\beta}}(\boldsymbol{Y},\boldsymbol{Z})^{\otimes 2} \;\Big|\; \boldsymbol{Y}=\mathbf{y}_i\Big],
\end{aligned} \tag{C.6}$$

where $\mathbf{v}^{\otimes 2}$ denotes $\mathbf{v}\otimes\mathbf{v}$. Here $\widetilde{S}_{\boldsymbol{\beta}}(\cdot,\cdot)$ is defined as

$$\widetilde{S}_{\boldsymbol{\beta}}(\mathbf{y},\mathbf{z}) = \frac{\partial \log f_{\boldsymbol{\beta}}(\mathbf{y},\mathbf{z})}{\partial\boldsymbol{\beta}} = \frac{\partial f_{\boldsymbol{\beta}}(\mathbf{y},\mathbf{z})/\partial\boldsymbol{\beta}}{f_{\boldsymbol{\beta}}(\mathbf{y},\mathbf{z})} \in \mathbb{R}^d, \tag{C.7}$$

i.e., the score function for the complete likelihood, which involves both the observed variable $\boldsymbol{Y}$ and the latent variable $\boldsymbol{Z}$. Meanwhile, in (C.5) we have

$$\begin{aligned}
\int_{\mathcal{Z}} \frac{\partial f_{\boldsymbol{\beta}}(\mathbf{y}_i,\mathbf{z})/\partial\boldsymbol{\beta}}{f_{\boldsymbol{\beta}}(\mathbf{y}_i,\mathbf{z})} \otimes \frac{f_{\boldsymbol{\beta}}(\mathbf{y}_i,\mathbf{z})\cdot\partial h_{\boldsymbol{\beta}}(\mathbf{y}_i)/\partial\boldsymbol{\beta}}{\big[h_{\boldsymbol{\beta}}(\mathbf{y}_i)\big]^2} \mathrm{d}\mathbf{z} &= \left[\int_{\mathcal{Z}} \frac{\partial f_{\boldsymbol{\beta}}(\mathbf{y}_i,\mathbf{z})/\partial\boldsymbol{\beta}}{h_{\boldsymbol{\beta}}(\mathbf{y}_i)} \mathrm{d}\mathbf{z}\right] \otimes \frac{\partial h_{\boldsymbol{\beta}}(\mathbf{y}_i)/\partial\boldsymbol{\beta}}{h_{\boldsymbol{\beta}}(\mathbf{y}_i)} \\
&= \left[\int_{\mathcal{Z}} \frac{\partial f_{\boldsymbol{\beta}}(\mathbf{y}_i,\mathbf{z})/\partial\boldsymbol{\beta}}{h_{\boldsymbol{\beta}}(\mathbf{y}_i)} \mathrm{d}\mathbf{z}\right]^{\otimes 2},
\end{aligned} \tag{C.8}$$

where in the last equality we utilize the fact that

$$\int_{\mathcal{Z}} f_{\boldsymbol{\beta}}(\mathbf{y}_i,\mathbf{z})\,\mathrm{d}\mathbf{z} = h_{\boldsymbol{\beta}}(\mathbf{y}_i), \tag{C.9}$$

because $h_{\boldsymbol{\beta}}(\cdot)$ is the marginal density function of $\boldsymbol{Y}$. By (C.3) and (C.7), for the right-hand side of (C.8) we have

$$\mathbb{E}_{\boldsymbol{\beta}}\Big[\widetilde{S}_{\boldsymbol{\beta}}(\boldsymbol{Y},\boldsymbol{Z}) \;\Big|\; \boldsymbol{Y}=\mathbf{y}_i\Big] = \int_{\mathcal{Z}} \frac{\partial f_{\boldsymbol{\beta}}(\mathbf{y}_i,\mathbf{z})/\partial\boldsymbol{\beta}}{f_{\boldsymbol{\beta}}(\mathbf{y}_i,\mathbf{z})}\cdot k_{\boldsymbol{\beta}}(\mathbf{z}\mid\mathbf{y}_i)\,\mathrm{d}\mathbf{z} = \int_{\mathcal{Z}} \frac{\partial f_{\boldsymbol{\beta}}(\mathbf{y}_i,\mathbf{z})/\partial\boldsymbol{\beta}}{h_{\boldsymbol{\beta}}(\mathbf{y}_i)}\mathrm{d}\mathbf{z}. \tag{C.10}$$

Plugging (C.10) into (C.8) and then plugging (C.6) and (C.8) into (C.5) we obtain

$$\nabla^2_{1,2} Q_n(\boldsymbol{\beta};\boldsymbol{\beta}) = \frac{1}{n}\sum_{i=1}^n \left(\mathbb{E}_{\boldsymbol{\beta}}\Big[\widetilde{S}_{\boldsymbol{\beta}}(\boldsymbol{Y},\boldsymbol{Z})^{\otimes 2} \;\Big|\; \boldsymbol{Y}=\mathbf{y}_i\Big] - \Big\{\mathbb{E}_{\boldsymbol{\beta}}\Big[\widetilde{S}_{\boldsymbol{\beta}}(\boldsymbol{Y},\boldsymbol{Z}) \;\Big|\; \boldsymbol{Y}=\mathbf{y}_i\Big]\Big\}^{\otimes 2}\right).$$

Setting $\boldsymbol{\beta}=\boldsymbol{\beta}^*$ in the above equality, we obtain

$$\mathbb{E}_{\boldsymbol{\beta}^*}\big[\nabla^2_{1,2} Q_n(\boldsymbol{\beta}^*;\boldsymbol{\beta}^*)\big] = \mathbb{E}_{\boldsymbol{\beta}^*}\Big\{\mathrm{Cov}_{\boldsymbol{\beta}^*}\Big[\widetilde{S}_{\boldsymbol{\beta}^*}(\boldsymbol{Y},\boldsymbol{Z}) \;\Big|\; \boldsymbol{Y}\Big]\Big\}. \tag{C.11}$$

Meanwhile, for $\boldsymbol{\beta} = \boldsymbol{\beta}^*$, by the property of Fisher information we have

$$I(\boldsymbol{\beta}^*) = \mathrm{Cov}_{\boldsymbol{\beta}^*}\big[\nabla \ell_n(\boldsymbol{\beta}^*)\big] = \mathrm{Cov}_{\boldsymbol{\beta}^*}\bigg[\frac{\partial \log h_{\boldsymbol{\beta}^*}(\boldsymbol{Y})}{\partial \boldsymbol{\beta}}\bigg] = \mathrm{Cov}_{\boldsymbol{\beta}^*}\Big\{\mathbb{E}_{\boldsymbol{\beta}^*}\Big[\widetilde{S}_{\boldsymbol{\beta}^*}(\boldsymbol{Y}, \boldsymbol{Z}) \mid \boldsymbol{Y}\Big]\Big\}. \tag{C.12}$$

Here the last equality is obtained by taking $\boldsymbol{\beta} = \boldsymbol{\beta}^*$ in

$$\begin{aligned}\frac{\partial \log h_{\boldsymbol{\beta}}(\boldsymbol{Y})}{\partial \boldsymbol{\beta}} = \frac{\partial h_{\boldsymbol{\beta}}(\boldsymbol{Y})}{\partial \boldsymbol{\beta}} \cdot \frac{1}{h_{\boldsymbol{\beta}}(\boldsymbol{Y})} = \int_{\mathcal{Z}} \frac{\partial f_{\boldsymbol{\beta}}(\boldsymbol{Y}, \mathbf{z})/\partial \boldsymbol{\beta}}{h_{\boldsymbol{\beta}}(\boldsymbol{Y})}\,\mathrm{d}\mathbf{z} &= \int_{\mathcal{Z}} \frac{\partial f_{\boldsymbol{\beta}}(\boldsymbol{Y}, \mathbf{z})/\partial \boldsymbol{\beta}}{f_{\boldsymbol{\beta}}(\boldsymbol{Y}, \mathbf{z})} \cdot k_{\boldsymbol{\beta}}(\mathbf{z} \mid \boldsymbol{Y})\,\mathrm{d}\mathbf{z} \\ &= \int_{\mathcal{Z}} \widetilde{S}_{\boldsymbol{\beta}}(\boldsymbol{Y}, \mathbf{z}) \cdot k_{\boldsymbol{\beta}}(\mathbf{z} \mid \boldsymbol{Y})\,\mathrm{d}\mathbf{z} \\ &= \mathbb{E}_{\boldsymbol{\beta}}\Big[\widetilde{S}_{\boldsymbol{\beta}}(\boldsymbol{Y}, \boldsymbol{Z}) \mid \boldsymbol{Y}\Big],\end{aligned}$$

where the second equality follows from (C.9), the third equality follows from (C.3), while the second last equality follows from (C.7). Combining (C.11) and (C.12), by the law of total variance we have

$$\begin{aligned} I(\boldsymbol{\beta}^*) + \mathbb{E}_{\boldsymbol{\beta}^*}\big[\nabla^2_{1,2} Q_n(\boldsymbol{\beta}^*; \boldsymbol{\beta}^*)\big] &= \mathrm{Cov}_{\boldsymbol{\beta}^*}\Big\{\mathbb{E}_{\boldsymbol{\beta}^*}\Big[\widetilde{S}_{\boldsymbol{\beta}^*}(\boldsymbol{Y}, \boldsymbol{Z}) \mid \boldsymbol{Y}\Big]\Big\} + \mathbb{E}_{\boldsymbol{\beta}^*}\Big\{\mathrm{Cov}_{\boldsymbol{\beta}^*}\Big[\widetilde{S}_{\boldsymbol{\beta}^*}(\boldsymbol{Y}, \boldsymbol{Z}) \mid \boldsymbol{Y}\Big]\Big\} \\ &= \mathrm{Cov}_{\boldsymbol{\beta}^*}\Big[\widetilde{S}_{\boldsymbol{\beta}^*}(\boldsymbol{Y}, \boldsymbol{Z})\Big]. \end{aligned} \tag{C.13}$$

In the following, we prove

$$\mathbb{E}_{\boldsymbol{\beta}^*}\big[\nabla^2_{1,1} Q_n(\boldsymbol{\beta}^*; \boldsymbol{\beta}^*)\big] = -\,\mathrm{Cov}_{\boldsymbol{\beta}^*}\Big[\widetilde{S}_{\boldsymbol{\beta}^*}(\boldsymbol{Y}, \boldsymbol{Z})\Big]. \tag{C.14}$$

According to (C.1) we have

$$\nabla^2_{1,1} Q_n(\boldsymbol{\beta}; \boldsymbol{\beta}) = \frac{1}{n}\sum_{i=1}^{n} \int_{\mathcal{Z}} k_{\boldsymbol{\beta}}(\mathbf{z} \mid \mathbf{y}_i) \cdot \frac{\partial^2 \log f_{\boldsymbol{\beta}}(\mathbf{y}_i, \mathbf{z})}{\partial^2 \boldsymbol{\beta}}\,\mathrm{d}\mathbf{z} = \frac{1}{n}\sum_{i=1}^{n} \mathbb{E}_{\boldsymbol{\beta}}\bigg[\frac{\partial^2 \log f_{\boldsymbol{\beta}}(\boldsymbol{Y}, \boldsymbol{Z})}{\partial^2 \boldsymbol{\beta}} \;\bigg|\; \boldsymbol{Y} = \mathbf{y}_i\bigg]. \tag{C.15}$$

Let $\widetilde{\ell}(\boldsymbol{\beta}) = \log f_{\boldsymbol{\beta}}(\boldsymbol{Y}, \boldsymbol{Z})$ be the complete log-likelihood, which involves both the observed variable $\boldsymbol{Y}$ and the latent variable $\boldsymbol{Z}$, and $\widetilde{I}(\boldsymbol{\beta})$ be the corresponding Fisher information. By setting $\boldsymbol{\beta} = \boldsymbol{\beta}^*$ in (C.15) and taking expectation, we obtain

$$\mathbb{E}_{\boldsymbol{\beta}^*}\big[\nabla^2_{1,1} Q_n(\boldsymbol{\beta}^*; \boldsymbol{\beta}^*)\big] = \mathbb{E}_{\boldsymbol{\beta}^*}\bigg\{\mathbb{E}_{\boldsymbol{\beta}^*}\bigg[\frac{\partial^2 \log f_{\boldsymbol{\beta}^*}(\boldsymbol{Y}, \boldsymbol{Z})}{\partial^2 \boldsymbol{\beta}} \;\bigg|\; \boldsymbol{Y}\bigg]\bigg\} = \mathbb{E}_{\boldsymbol{\beta}^*}\big[\nabla^2 \widetilde{\ell}(\boldsymbol{\beta}^*)\big] = -\widetilde{I}(\boldsymbol{\beta}^*). \tag{C.16}$$

Since $\widetilde{S}_{\boldsymbol{\beta}}(\boldsymbol{Y}, \boldsymbol{Z})$ defined in (C.7) is the score function for the complete log-likelihood $\widetilde{\ell}(\boldsymbol{\beta})$, according to the relationship between the score function and Fisher information, we have

$$\widetilde{I}(\boldsymbol{\beta}^*) = \mathrm{Cov}_{\boldsymbol{\beta}^*}\Big[\widetilde{S}_{\boldsymbol{\beta}^*}(\boldsymbol{Y}, \boldsymbol{Z})\Big],$$

which together with (C.16) implies (C.14). By further plugging (C.14) into (C.13), we obtain

$$\mathbb{E}_{\boldsymbol{\beta}^*}\big[\nabla^2_{1,1} Q_n(\boldsymbol{\beta}^*; \boldsymbol{\beta}^*) + \nabla^2_{1,2} Q_n(\boldsymbol{\beta}^*; \boldsymbol{\beta}^*)\big] = -I(\boldsymbol{\beta}^*),$$

which establishes the first equality of the second equation in (2.12). In addition, the second equality of the second equation in (2.12) follows from the property of Fisher information. Thus, we conclude the proof of Lemma 2.1. □

## C.2 Auxiliary Lemmas for Proving Theorems 4.6 and 4.7

In this section, we lay out several lemmas on the Dantzig selector defined in (2.10). The first lemma, which is from Bickel et al. (2009), characterizes the cone condition for the Dantzig selector.

**Lemma C.1.** Any feasible solution $\mathbf{w}$ in (2.10) satisfies

$$\left\| \left[ w(\boldsymbol{\beta}, \lambda) - \mathbf{w} \right]_{\overline{\mathcal{S}_{\mathbf{w}}}} \right\|_1 \leq \left\| \left[ w(\boldsymbol{\beta}, \lambda) - \mathbf{w} \right]_{\mathcal{S}_{\mathbf{w}}} \right\|_1,$$

where $w(\boldsymbol{\beta}, \lambda)$ is the minimizer of (2.10), $\mathcal{S}_{\mathbf{w}}$ is the support of $\mathbf{w}$ and $\overline{\mathcal{S}_{\mathbf{w}}}$ is its complement.

*Proof.* See Lemma B.3 in Bickel et al. (2009) for a detailed proof. □

In the sequel, we focus on analyzing $w(\widehat{\boldsymbol{\beta}}, \lambda)$. The results for $w(\widehat{\boldsymbol{\beta}}_0, \lambda)$ can be obtained similarly. The next lemma characterizes the restricted eigenvalue of $T_n(\widehat{\boldsymbol{\beta}})$, which is defined as

$$\widehat{\rho}_{\min} = \inf_{\mathbf{v} \in \mathcal{C}} \frac{\mathbf{v}^\top \cdot \left[ -T_n(\widehat{\boldsymbol{\beta}}) \right] \cdot \mathbf{v}}{\|\mathbf{v}\|_2^2}, \quad \text{where } \mathcal{C} = \Big\{ \mathbf{v} : \|\mathbf{v}_{\overline{\mathcal{S}^*_{\mathbf{w}}}}\|_1 \leq \|\mathbf{v}_{\mathcal{S}^*_{\mathbf{w}}}\|_1, \quad \mathbf{v} \neq \mathbf{0} \Big\}. \tag{C.17}$$

Here $\mathcal{S}^*_{\mathbf{w}}$ is the support of $\mathbf{w}^*$ defined in (4.1).

**Lemma C.2.** Under Assumption 4.5 and Conditions 4.1, 4.3 and 4.4, for a sufficiently large sample size $n$, we have $\widehat{\rho}_{\min} \geq \rho_{\min}/2 > 0$ with high probability, where $\rho_{\min}$ is specified in (4.4).

*Proof.* By triangle inequality we have

$$\begin{aligned} \widehat{\rho}_{\min} \geq \inf_{\mathbf{v} \in \mathcal{C}} \frac{\mathbf{v}^\top \cdot \left[ -T_n(\widehat{\boldsymbol{\beta}}) \right] \cdot \mathbf{v}}{\|\mathbf{v}\|_2^2} &\geq \inf_{\mathbf{v} \in \mathcal{C}} \frac{\mathbf{v}^\top \cdot I(\boldsymbol{\beta}^*) \cdot \mathbf{v} - \left| \mathbf{v}^\top \cdot \left[ I(\boldsymbol{\beta}^*) + T_n(\widehat{\boldsymbol{\beta}}) \right] \cdot \mathbf{v} \right|}{\|\mathbf{v}\|_2^2} \\ &\geq \underbrace{\inf_{\mathbf{v} \in \mathcal{C}} \frac{\mathbf{v}^\top \cdot I(\boldsymbol{\beta}^*) \cdot \mathbf{v}}{\|\mathbf{v}\|_2^2}}_{\text{(i)}} - \underbrace{\sup_{\mathbf{v} \in \mathcal{C}} \frac{\left| \mathbf{v}^\top \cdot \left[ I(\boldsymbol{\beta}^*) + T_n(\widehat{\boldsymbol{\beta}}) \right] \cdot \mathbf{v} \right|}{\|\mathbf{v}\|_2^2}}_{\text{(ii)}}, \end{aligned} \tag{C.18}$$

where $\mathcal{C}$ is defined in (C.17).

**Analysis of Term (i):** For term (i) in (C.18), by (4.4) in Assumption 4.5 we have

$$\inf_{\mathbf{v} \in \mathcal{C}} \frac{\mathbf{v}^\top \cdot I(\boldsymbol{\beta}^*) \cdot \mathbf{v}}{\|\mathbf{v}\|_2^2} \geq \inf_{\mathbf{v} \neq \mathbf{0}} \frac{\mathbf{v}^\top \cdot I(\boldsymbol{\beta}^*) \cdot \mathbf{v}}{\|\mathbf{v}\|_2^2} = \lambda_d \left[ I(\boldsymbol{\beta}^*) \right] \geq \rho_{\min}. \tag{C.19}$$

**Analysis of Term (ii):** For term (ii) in (C.18) we have

$$\sup_{\mathbf{v} \in \mathcal{C}} \frac{\left| \mathbf{v}^\top \cdot \left[ I(\boldsymbol{\beta}^*) + T_n(\widehat{\boldsymbol{\beta}}) \right] \cdot \mathbf{v} \right|}{\|\mathbf{v}\|_2^2} \leq \sup_{\mathbf{v} \in \mathcal{C}} \frac{\|\mathbf{v}\|_1^2 \cdot \left\| I(\boldsymbol{\beta}^*) + T_n(\widehat{\boldsymbol{\beta}}) \right\|_{\infty,\infty}}{\|\mathbf{v}\|_2^2}. \tag{C.20}$$

By the definition of $\mathcal{C}$ in (C.17), for any $\mathbf{v} \in \mathcal{C}$ we have

$$\|\mathbf{v}\|_1 = \left\|\mathbf{v}_{\mathcal{S}^*_{\mathbf{w}}}\right\|_1 + \left\|\mathbf{v}_{\overline{\mathcal{S}^*_{\mathbf{w}}}}\right\|_1 \leq 2 \cdot \left\|\mathbf{v}_{\mathcal{S}^*_{\mathbf{w}}}\right\|_1 \leq 2 \cdot \sqrt{s^*_{\mathbf{w}}} \cdot \left\|\mathbf{v}_{\mathcal{S}^*_{\mathbf{w}}}\right\|_2 \leq 2 \cdot \sqrt{s^*_{\mathbf{w}}} \cdot \|\mathbf{v}\|_2.$$

Therefore, the right-hand side of (C.20) is upper bounded by

$$4 \cdot s^*_{\mathbf{w}} \cdot \Big\| I(\boldsymbol{\beta}^*) + T_n\big(\widehat{\boldsymbol{\beta}}\big) \Big\|_{\infty,\infty} \leq \underbrace{4 \cdot s^*_{\mathbf{w}} \cdot \big\| I(\boldsymbol{\beta}^*) + T_n(\boldsymbol{\beta}^*) \big\|_{\infty,\infty}}_{\text{(ii).a}} + \underbrace{4 \cdot s^*_{\mathbf{w}} \cdot \Big\| T\big(\widehat{\boldsymbol{\beta}}\big) - T_n(\boldsymbol{\beta}^*) \Big\|_{\infty,\infty}}_{\text{(ii).b}}.$$

For term (ii).a, by Lemma 2.1 and Condition 4.3 we have

$$4 \cdot s^*_{\mathbf{w}} \cdot \big\| I(\boldsymbol{\beta}^*) + T_n(\boldsymbol{\beta}^*) \big\|_{\infty,\infty} = 4 \cdot s^*_{\mathbf{w}} \cdot \big\| T_n(\boldsymbol{\beta}^*) - \mathbb{E}_{\boldsymbol{\beta}^*}\left[T_n(\boldsymbol{\beta}^*)\right] \big\|_{\infty,\infty} = O_{\mathbb{P}}\big(s^*_{\mathbf{w}} \cdot \zeta^{\mathrm{T}}\big) = o_{\mathbb{P}}(1),$$

where the last equality is from (4.6) in Assumption 4.5, since for $\lambda$ specified in (4.5) we have

$$s^*_{\mathbf{w}} \cdot \zeta^{\mathrm{T}} \leq s^*_{\mathbf{w}} \cdot \lambda \leq \max\big\{\|\mathbf{w}^*\|_1,\ 1\big\} \cdot s^*_{\mathbf{w}} \cdot \lambda = o(1).$$

For term (ii).b, by Conditions 4.1 and 4.4 we have

$$4 \cdot s^*_{\mathbf{w}} \cdot \Big\| T\big(\widehat{\boldsymbol{\beta}}\big) - T_n(\boldsymbol{\beta}^*) \Big\|_{\infty,\infty} = 4 \cdot s^*_{\mathbf{w}} \cdot O_{\mathbb{P}}\big(\zeta^{\mathrm{L}}\big) \cdot \big\| \widehat{\boldsymbol{\beta}} - \boldsymbol{\beta}^* \big\|_1 = O_{\mathbb{P}}\big(s^*_{\mathbf{w}} \cdot \zeta^{\mathrm{L}} \cdot \zeta^{\mathrm{EM}}\big) = o_{\mathbb{P}}(1),$$

where the last equality is also from (4.6) in Assumption 4.5, since for $\lambda$ specified in (4.5) we have

$$s^*_{\mathbf{w}} \cdot \zeta^{\mathrm{L}} \cdot \zeta^{\mathrm{EM}} \leq s^*_{\mathbf{w}} \cdot \lambda \leq \max\big\{\|\mathbf{w}^*\|_1,\ 1\big\} \cdot s^*_{\mathbf{w}} \cdot \lambda = o(1).$$

Hence, term (ii) in (C.18) is $o_{\mathbb{P}}(1)$. Since $\rho_{\min}$ is an absolute constant, for a sufficiently large $n$ we have that term (ii) is upper bounded by $\rho_{\min}/2$ with high probability. Further by plugging this and (C.19) into (C.18), we conclude that $\widehat{\rho}_{\min} \geq \rho_{\min}/2$ holds with high probability. □

The next lemma quantifies the statistical accuracy of $w\big(\widehat{\boldsymbol{\beta}}, \lambda\big)$, where $w(\cdot,\cdot)$ is defined in (2.10).

**Lemma C.3.** Under Assumption 4.5 and Conditions 4.1-4.4, for $\lambda$ specified in (4.5) we have that

$$\max\Big\{\Big\| w\big(\widehat{\boldsymbol{\beta}}, \lambda\big) - \mathbf{w}^* \Big\|_1,\ \Big\| w\big(\widehat{\boldsymbol{\beta}}_0, \lambda\big) - \mathbf{w}^* \Big\|_1\Big\} \leq \frac{16 \cdot s^*_{\mathbf{w}} \cdot \lambda}{\rho_{\min}}$$

holds with high probability. Here $\rho_{\min}$ is specified in (4.4), while $\mathbf{w}^*$ and $s^*_{\mathbf{w}}$ are defined (4.1).

*Proof.* For $\lambda$ specified in (4.5), we verify that $\mathbf{w}^*$ is a feasible solution in (2.10) with high probability. For notational simplicity, we define the following event

$$\mathcal{E} = \Big\{\Big\| \big[T_n\big(\widehat{\boldsymbol{\beta}}\big)\big]_{\gamma,\alpha} - \big[T_n\big(\widehat{\boldsymbol{\beta}}\big)\big]_{\gamma,\gamma} \cdot \mathbf{w}^* \Big\|_\infty \leq \lambda\Big\}. \tag{C.21}$$

By the definition of $\mathbf{w}^*$ in (4.1), we have $\big[I(\boldsymbol{\beta}^*)\big]_{\gamma,\alpha} - \big[I(\boldsymbol{\beta}^*)\big]_{\gamma,\gamma} \cdot \mathbf{w}^* = 0$. Hence, we have

$$\begin{aligned} \Big\| \big[T_n\big(\widehat{\boldsymbol{\beta}}\big)\big]_{\gamma,\alpha} - \big[T_n\big(\widehat{\boldsymbol{\beta}}\big)\big]_{\gamma,\gamma} \cdot \mathbf{w}^* \Big\|_\infty &= \Big\| \big[T_n\big(\widehat{\boldsymbol{\beta}}\big) + I(\boldsymbol{\beta}^*)\big]_{\gamma,\alpha} - \big[T_n\big(\widehat{\boldsymbol{\beta}}\big) + I(\boldsymbol{\beta}^*)\big]_{\gamma,\gamma} \cdot \mathbf{w}^* \Big\|_\infty \\ &\leq \Big\| T_n\big(\widehat{\boldsymbol{\beta}}\big) + I(\boldsymbol{\beta}^*) \Big\|_{\infty,\infty} + \Big\| T_n\big(\widehat{\boldsymbol{\beta}}\big) + I(\boldsymbol{\beta}^*) \Big\|_{\infty,\infty} \cdot \|\mathbf{w}^*\|_1, \end{aligned} \tag{C.22}$$

where the last inequality is from triangle inequality and Hölder's inequality. Note that we have

$$\Big\| T_n\big(\widehat{\boldsymbol{\beta}}\big) + I(\boldsymbol{\beta}^*) \Big\|_{\infty,\infty} \leq \big\| T_n(\boldsymbol{\beta}^*) + I(\boldsymbol{\beta}^*) \big\|_{\infty,\infty} + \Big\| T\big(\widehat{\boldsymbol{\beta}}\big) - T_n(\boldsymbol{\beta}^*) \Big\|_{\infty,\infty}. \tag{C.23}$$

On the right-hand side, by Lemma 2.1 and Condition 4.3 we have

$$\big\|T_n(\boldsymbol{\beta}^*) + I(\boldsymbol{\beta}^*)\big\|_{\infty,\infty} = \big\|T_n(\boldsymbol{\beta}^*) - \mathbb{E}_{\boldsymbol{\beta}^*}\big[T_n(\boldsymbol{\beta}^*)\big]\big\|_{\infty,\infty} = O_{\mathbb{P}}\big(\zeta^{\mathrm{T}}\big),$$

while by Conditions 4.1 and 4.4 we have

$$\Big\|T\big(\widehat{\boldsymbol{\beta}}\big) - T_n(\boldsymbol{\beta}^*)\Big\|_{\infty,\infty} = O_{\mathbb{P}}\big(\zeta^{\mathrm{L}}\big)\cdot\big\|\widehat{\boldsymbol{\beta}} - \boldsymbol{\beta}^*\big\|_1 = O_{\mathbb{P}}\big(\zeta^{\mathrm{L}}\cdot\zeta^{\mathrm{EM}}\big).$$

Plugging the above equations into (C.23) and further plugging (C.23) into (C.22), by (4.5) we have

$$\Big\|\big[T_n\big(\widehat{\boldsymbol{\beta}}\big)\big]_{\boldsymbol{\gamma},\alpha} - \big[T_n\big(\widehat{\boldsymbol{\beta}}\big)\big]_{\boldsymbol{\gamma},\boldsymbol{\gamma}}\cdot\mathbf{w}^*\Big\|_\infty \le C\cdot\big(\zeta^{\mathrm{T}} + \zeta^{\mathrm{L}}\cdot\zeta^{\mathrm{EM}}\big)\cdot\big(1 + \|\mathbf{w}^*\|_1\big) = \lambda$$

holds with high probability for a sufficiently large absolute constant $C \ge 1$. In other words, $\mathcal{E}$ occurs with high probability. The subsequent proof will be conditioning on $\mathcal{E}$ and the following event

$$\mathcal{E}' = \big\{\widehat{\rho}_{\min} \ge \rho_{\min}/2 > 0\big\}, \tag{C.24}$$

which also occurs with high probability according to Lemma C.2. Here $\widehat{\rho}_{\min}$ is defined in (C.17).

For notational simplicity, we denote $w\big(\widehat{\boldsymbol{\beta}}, \lambda\big) = \widehat{\mathbf{w}}$. By triangle inequality we have

$$\begin{aligned}\Big\|\big[T_n\big(\widehat{\boldsymbol{\beta}}\big)\big]_{\boldsymbol{\gamma},\boldsymbol{\gamma}}\cdot(\widehat{\mathbf{w}} - \mathbf{w}^*)\Big\|_\infty &\le \Big\|\big[T_n\big(\widehat{\boldsymbol{\beta}}\big)\big]_{\boldsymbol{\gamma},\alpha} - \big[T_n\big(\widehat{\boldsymbol{\beta}}\big)\big]_{\boldsymbol{\gamma},\boldsymbol{\gamma}}\cdot\mathbf{w}^*\Big\|_\infty + \Big\|\big[T_n\big(\widehat{\boldsymbol{\beta}}\big)\big]_{\boldsymbol{\gamma},\boldsymbol{\gamma}}\cdot\widehat{\mathbf{w}} - \big[T_n\big(\widehat{\boldsymbol{\beta}}\big)\big]_{\boldsymbol{\gamma},\alpha}\Big\|_\infty \\ &\le 2\cdot\lambda,\end{aligned} \tag{C.25}$$

where the last inequality follows from (2.10) and (C.21). Moreover, by (C.17) and (C.24) we have

$$(\widehat{\mathbf{w}} - \mathbf{w}^*)^\top\cdot\big[-T_n\big(\widehat{\boldsymbol{\beta}}\big)\big]_{\boldsymbol{\gamma},\boldsymbol{\gamma}}\cdot(\widehat{\mathbf{w}} - \mathbf{w}^*) \ge \widehat{\rho}_{\min}\cdot\|\widehat{\mathbf{w}} - \mathbf{w}^*\|_2^2 \ge \rho_{\min}/2\cdot\|\widehat{\mathbf{w}} - \mathbf{w}^*\|_2^2. \tag{C.26}$$

Meanwhile, by Lemma C.1 we have

$$\|\widehat{\mathbf{w}} - \mathbf{w}^*\|_1 = \big\|(\widehat{\mathbf{w}} - \mathbf{w}^*)_{\mathcal{S}^*_{\mathbf{w}}}\big\|_1 + \big\|(\widehat{\mathbf{w}} - \mathbf{w}^*)_{\overline{\mathcal{S}^*_{\mathbf{w}}}}\big\|_1 \le 2\cdot\big\|(\widehat{\mathbf{w}} - \mathbf{w}^*)_{\mathcal{S}^*_{\mathbf{w}}}\big\|_1 \le 2\cdot\sqrt{s^*_{\mathbf{w}}}\cdot\|\widehat{\mathbf{w}} - \mathbf{w}^*\|_2.$$

Plugging the above inequality into (C.26), we obtain

$$(\widehat{\mathbf{w}} - \mathbf{w}^*)^\top\cdot\big[-T_n\big(\widehat{\boldsymbol{\beta}}\big)\big]_{\boldsymbol{\gamma},\boldsymbol{\gamma}}\cdot(\widehat{\mathbf{w}} - \mathbf{w}^*) \ge \rho_{\min}/(8\cdot s^*_{\mathbf{w}})\cdot\|\widehat{\mathbf{w}} - \mathbf{w}^*\|_1^2. \tag{C.27}$$

Note that by (C.25), the left-hand side of (C.27) is upper bounded by

$$\|\widehat{\mathbf{w}} - \mathbf{w}^*\|_1\cdot\Big\|\big[T_n\big(\widehat{\boldsymbol{\beta}}\big)\big]_{\boldsymbol{\gamma},\boldsymbol{\gamma}}\cdot(\widehat{\mathbf{w}} - \mathbf{w}^*)\Big\|_\infty \le \|\widehat{\mathbf{w}} - \mathbf{w}^*\|_1\cdot 2\cdot\lambda. \tag{C.28}$$

By (C.27) and (C.28), we then obtain $\|\widehat{\mathbf{w}} - \mathbf{w}^*\|_1 \le 16\cdot s^*_{\mathbf{w}}\cdot\lambda/\rho_{\min}$ conditioning on $\mathcal{E}$ and $\mathcal{E}'$, both of which hold with high probability. Note that the proof for $w\big(\widehat{\boldsymbol{\beta}}_0, \lambda\big)$ follows similarly. Therefore, we conclude the proof of Lemma C.3. □

## C.3 Proof of Lemma 5.3

*Proof.* Our proof strategy is as follows. First we prove that

$$\sqrt{n}\cdot S_n\big(\widehat{\boldsymbol{\beta}}_0,\lambda\big)=\sqrt{n}\cdot\big[\nabla_1 Q_n(\boldsymbol{\beta}^*;\boldsymbol{\beta}^*)\big]_{\alpha}-\sqrt{n}\cdot(\mathbf{w}^*)^\top\cdot\big[\nabla_1 Q_n(\boldsymbol{\beta}^*;\boldsymbol{\beta}^*)\big]_{\boldsymbol{\gamma}}+o_{\mathbb{P}}(1), \tag{C.29}$$

where $\boldsymbol{\beta}^*$ is the true parameter and $\mathbf{w}^*$ is defined in (4.1). We then prove

$$\sqrt{n}\cdot\big[\nabla_1 Q_n(\boldsymbol{\beta}^*;\boldsymbol{\beta}^*)\big]_{\alpha}-\sqrt{n}\cdot(\mathbf{w}^*)^\top\cdot\big[\nabla_1 Q_n(\boldsymbol{\beta}^*;\boldsymbol{\beta}^*)\big]_{\boldsymbol{\gamma}}\xrightarrow{D}N\big(0,\big[I(\boldsymbol{\beta}^*)\big]_{\alpha|\boldsymbol{\gamma}}\big), \tag{C.30}$$

where $\big[I(\boldsymbol{\beta}^*)\big]_{\alpha|\boldsymbol{\gamma}}$ is defined in (4.2). Throughout the proof, we abbreviate $w\big(\widehat{\boldsymbol{\beta}}_0,\lambda\big)$ as $\widehat{\mathbf{w}}_0$. Also, it is worth noting that our analysis is under the null hypothesis where $\boldsymbol{\beta}^*=\big[\alpha^*,(\boldsymbol{\gamma}^*)^\top\big]^\top$ with $\alpha^*=0$.

**Proof of (C.29):** For (C.29), by the definition of the decorrelated score function in (2.9) we have

$$S_n\big(\widehat{\boldsymbol{\beta}}_0,\lambda\big)=\big[\nabla_1 Q_n\big(\widehat{\boldsymbol{\beta}}_0;\widehat{\boldsymbol{\beta}}_0\big)\big]_{\alpha}-\widehat{\mathbf{w}}_0^\top\cdot\big[\nabla_1 Q_n\big(\widehat{\boldsymbol{\beta}}_0;\widehat{\boldsymbol{\beta}}_0\big)\big]_{\boldsymbol{\gamma}}.$$

By the mean-value theorem, we obtain

$$\begin{aligned}S_n\big(\widehat{\boldsymbol{\beta}}_0,\lambda\big)=&\overbrace{\big[\nabla_1 Q_n(\boldsymbol{\beta}^*;\boldsymbol{\beta}^*)\big]_{\alpha}-\widehat{\mathbf{w}}_0^\top\cdot\big[\nabla_1 Q_n(\boldsymbol{\beta}^*;\boldsymbol{\beta}^*)\big]_{\boldsymbol{\gamma}}}^{\text{(i)}}\\&+\underbrace{\big[T_n(\boldsymbol{\beta}^\sharp)\big]_{\boldsymbol{\gamma},\alpha}^\top\cdot\big(\widehat{\boldsymbol{\beta}}_0-\boldsymbol{\beta}^*\big)-\widehat{\mathbf{w}}_0^\top\cdot\big[T_n(\boldsymbol{\beta}^\sharp)\big]_{\boldsymbol{\gamma},\boldsymbol{\gamma}}\cdot\big(\widehat{\boldsymbol{\beta}}_0-\boldsymbol{\beta}^*\big)}_{\text{(ii)}},\end{aligned} \tag{C.31}$$

where we have $T_n(\boldsymbol{\beta})=\nabla^2_{1,1}Q_n(\boldsymbol{\beta};\boldsymbol{\beta})+\nabla^2_{1,2}Q_n(\boldsymbol{\beta};\boldsymbol{\beta})$ as defined in (2.8), and $\boldsymbol{\beta}^\sharp$ is an intermediate value between $\boldsymbol{\beta}^*$ and $\widehat{\boldsymbol{\beta}}_0$.

**Analysis of Term (i):** For term (i) in (C.31), we have

$$\begin{aligned}&\big[\nabla_1 Q_n(\boldsymbol{\beta}^*;\boldsymbol{\beta}^*)\big]_{\alpha}-\widehat{\mathbf{w}}_0^\top\cdot\big[\nabla_1 Q_n(\boldsymbol{\beta}^*;\boldsymbol{\beta}^*)\big]_{\boldsymbol{\gamma}}\\&\quad=\big[\nabla_1 Q_n(\boldsymbol{\beta}^*;\boldsymbol{\beta}^*)\big]_{\alpha}-(\mathbf{w}^*)^\top\cdot\big[\nabla_1 Q_n(\boldsymbol{\beta}^*;\boldsymbol{\beta}^*)\big]_{\boldsymbol{\gamma}}+(\mathbf{w}^*-\widehat{\mathbf{w}}_0)^\top\cdot\big[\nabla_1 Q_n(\boldsymbol{\beta}^*;\boldsymbol{\beta}^*)\big]_{\boldsymbol{\gamma}}.\end{aligned} \tag{C.32}$$

For the right-hand side of (C.32), we have

$$(\mathbf{w}^*-\widehat{\mathbf{w}}_0)^\top\cdot\big[\nabla_1 Q_n(\boldsymbol{\beta}^*;\boldsymbol{\beta}^*)\big]_{\boldsymbol{\gamma}}\le\|\mathbf{w}^*-\widehat{\mathbf{w}}_0\|_1\cdot\big\|\big[\nabla_1 Q_n(\boldsymbol{\beta}^*;\boldsymbol{\beta}^*)\big]_{\boldsymbol{\gamma}}\big\|_\infty. \tag{C.33}$$

By Lemma C.3, we have $\|\mathbf{w}^*-\widehat{\mathbf{w}}_0\|_1=O_{\mathbb{P}}(s^*_{\mathbf{w}}\cdot\lambda)$, where $\lambda$ is specified in (4.5). Meanwhile, we have

$$\begin{aligned}\big\|\big[\nabla_1 Q_n(\boldsymbol{\beta}^*;\boldsymbol{\beta}^*)\big]_{\boldsymbol{\gamma}}\big\|_\infty\le\big\|\nabla_1 Q_n(\boldsymbol{\beta}^*;\boldsymbol{\beta}^*)\big\|_\infty&=\big\|\nabla_1 Q_n(\boldsymbol{\beta}^*;\boldsymbol{\beta}^*)-\nabla_1 Q(\boldsymbol{\beta}^*;\boldsymbol{\beta}^*)\big\|_\infty\\&=O_{\mathbb{P}}\big(\zeta^{\mathrm{G}}\big),\end{aligned}$$

where the first equality follows from the self-consistency property (McLachlan and Krishnan, 2007) that $\boldsymbol{\beta}^*=\mathrm{argmax}_{\boldsymbol{\beta}}\,Q(\boldsymbol{\beta};\boldsymbol{\beta}^*)$, which gives $\nabla_1 Q(\boldsymbol{\beta}^*;\boldsymbol{\beta}^*)=\mathbf{0}$. Here the last equality is from Condition 4.2. Therefore, (C.33) implies

$$(\mathbf{w}^*-\widehat{\mathbf{w}}_0)^\top\cdot\big[\nabla_1 Q_n(\boldsymbol{\beta}^*;\boldsymbol{\beta}^*)\big]_{\boldsymbol{\gamma}}=O_{\mathbb{P}}\big(s^*_{\mathbf{w}}\cdot\lambda\cdot\zeta^{\mathrm{G}}\big)=o_{\mathbb{P}}(1/\sqrt{n}),$$

where the second equality is from $s^*_{\mathbf{w}} \cdot \lambda \cdot \zeta^{\mathrm{G}} = o(1/\sqrt{n})$ in (4.6) of Assumption 4.5. Thus, by (C.32) we conclude that term (i) in (C.31) equals

$$\big[\nabla_1 Q_n(\boldsymbol{\beta}^*;\boldsymbol{\beta}^*)\big]_\alpha - (\mathbf{w}^*)^\top \cdot \big[\nabla_1 Q_n(\boldsymbol{\beta}^*;\boldsymbol{\beta}^*)\big]_{\boldsymbol{\gamma}} + o_{\mathbb{P}}(1/\sqrt{n}).$$

**Analysis of Term (ii):** By triangle inequality, term (ii) in (C.31) is upper bounded by

$$\begin{aligned}
&\underbrace{\Big|\big[T_n(\widehat{\boldsymbol{\beta}}_0)\big]_{\boldsymbol{\gamma},\alpha}^\top \cdot \big(\widehat{\boldsymbol{\beta}}_0 - \boldsymbol{\beta}^*\big) - \widehat{\mathbf{w}}_0^\top \cdot \big[T_n(\widehat{\boldsymbol{\beta}}_0)\big]_{\boldsymbol{\gamma},\boldsymbol{\gamma}} \cdot \big(\widehat{\boldsymbol{\beta}}_0 - \boldsymbol{\beta}^*\big)\Big|}_{\text{(ii).a}} \qquad \text{(C.34)}\\
&+ \underbrace{\Big|\big[T_n(\boldsymbol{\beta}^\sharp)\big]_{\boldsymbol{\gamma},\alpha}^\top \cdot \big(\widehat{\boldsymbol{\beta}}_0 - \boldsymbol{\beta}^*\big) - \big[T_n(\widehat{\boldsymbol{\beta}}_0)\big]_{\boldsymbol{\gamma},\alpha}^\top \cdot \big(\widehat{\boldsymbol{\beta}}_0 - \boldsymbol{\beta}^*\big)\Big|}_{\text{(ii).b}}\\
&+ \underbrace{\Big|\widehat{\mathbf{w}}_0^\top \cdot \big[T_n(\widehat{\boldsymbol{\beta}}_0)\big]_{\boldsymbol{\gamma},\boldsymbol{\gamma}} \cdot \big(\widehat{\boldsymbol{\beta}}_0 - \boldsymbol{\beta}^*\big) - \widehat{\mathbf{w}}_0^\top \cdot \big[T_n(\boldsymbol{\beta}^\sharp)\big]_{\boldsymbol{\gamma},\boldsymbol{\gamma}} \cdot \big(\widehat{\boldsymbol{\beta}}_0 - \boldsymbol{\beta}^*\big)\Big|}_{\text{(ii).c}}.
\end{aligned}$$

By Hölder's inequality, term (ii).a in (C.34) is upper bounded by

$$\begin{aligned}
\big\|\widehat{\boldsymbol{\beta}}_0 - \boldsymbol{\beta}^*\big\|_1 \cdot \Big\|\big[T_n(\widehat{\boldsymbol{\beta}}_0)\big]_{\boldsymbol{\gamma},\alpha} - \widehat{\mathbf{w}}_0^\top \cdot \big[T_n(\widehat{\boldsymbol{\beta}}_0)\big]_{\boldsymbol{\gamma},\boldsymbol{\gamma}}\Big\|_\infty &= \big\|\widehat{\boldsymbol{\beta}}_0 - \boldsymbol{\beta}^*\big\|_1 \cdot \lambda\\
&\le O_{\mathbb{P}}\big(\zeta^{\mathrm{EM}}\big) \cdot \lambda = o_{\mathbb{P}}(1/\sqrt{n}), \qquad \text{(C.35)}
\end{aligned}$$

where the first inequality holds because $\widehat{\mathbf{w}}_0 = w\big(\widehat{\boldsymbol{\beta}}_0, \lambda\big)$ is a feasible solution in (2.10). Meanwhile, Condition 4.1 gives $\big\|\widehat{\boldsymbol{\beta}} - \boldsymbol{\beta}^*\big\|_1 = O_{\mathbb{P}}\big(\zeta^{\mathrm{EM}}\big)$. Also note that by definition we have $\big(\widehat{\boldsymbol{\beta}}_0\big)_\alpha = (\boldsymbol{\beta}^*)_\alpha = 0$, which implies $\big\|\widehat{\boldsymbol{\beta}}_0 - \boldsymbol{\beta}^*\big\|_1 \le \big\|\widehat{\boldsymbol{\beta}} - \boldsymbol{\beta}^*\big\|_1$. Hence, we have

$$\big\|\widehat{\boldsymbol{\beta}}_0 - \boldsymbol{\beta}^*\big\|_1 = O_{\mathbb{P}}\big(\zeta^{\mathrm{EM}}\big), \qquad \text{(C.36)}$$

which implies the first equality in (C.35). The last equality in (C.35) follows from $\zeta^{\mathrm{EM}} \cdot \lambda = o(1/\sqrt{n})$ in (4.6) of Assumption 4.5. Note that term (ii).b in (C.34) is upper bounded by

$$\Big\|\big[T_n(\boldsymbol{\beta}^\sharp)\big]_{\boldsymbol{\gamma},\alpha} - \big[T_n(\widehat{\boldsymbol{\beta}}_0)\big]_{\boldsymbol{\gamma},\alpha}\Big\|_\infty \cdot \big\|\widehat{\boldsymbol{\beta}}_0 - \boldsymbol{\beta}^*\big\|_1 \le \Big\|T_n(\boldsymbol{\beta}^\sharp) - T_n(\widehat{\boldsymbol{\beta}}_0)\Big\|_{\infty,\infty} \cdot \big\|\widehat{\boldsymbol{\beta}}_0 - \boldsymbol{\beta}^*\big\|_1. \qquad \text{(C.37)}$$

For the first term on the right-hand side of (C.37), by triangle inequality we have

$$\Big\|T_n(\boldsymbol{\beta}^\sharp) - T_n(\widehat{\boldsymbol{\beta}}_0)\Big\|_{\infty,\infty} \le \Big\|T_n(\boldsymbol{\beta}^\sharp) - T_n(\boldsymbol{\beta}^*)\Big\|_{\infty,\infty} + \Big\|T_n(\widehat{\boldsymbol{\beta}}_0) - T_n(\boldsymbol{\beta}^*)\Big\|_{\infty,\infty}.$$

By Condition 4.4, we have

$$\Big\|T_n(\widehat{\boldsymbol{\beta}}_0) - T_n(\boldsymbol{\beta}^*)\Big\|_{\infty,\infty} = O_{\mathbb{P}}\big(\zeta^{\mathrm{L}}\big) \cdot \big\|\widehat{\boldsymbol{\beta}}_0 - \boldsymbol{\beta}^*\big\|_1, \qquad \text{(C.38)}$$

and

$$\Big\|T_n(\boldsymbol{\beta}^\sharp) - T_n(\boldsymbol{\beta}^*)\Big\|_{\infty,\infty} = O_{\mathbb{P}}\big(\zeta^{\mathrm{L}}\big) \cdot \big\|\boldsymbol{\beta}^\sharp - \boldsymbol{\beta}^*\big\|_1 \le O_{\mathbb{P}}\big(\zeta^{\mathrm{L}}\big) \cdot \big\|\widehat{\boldsymbol{\beta}}_0 - \boldsymbol{\beta}^*\big\|_1, \qquad \text{(C.39)}$$

where the last inequality in (C.39) holds because $\boldsymbol{\beta}^\sharp$ is defined as an intermediate value between $\boldsymbol{\beta}^*$ and $\widehat{\boldsymbol{\beta}}_0$. Further by plugging (C.36) into (C.38), (C.39) as well as the second term on the right-hand side of (C.37), we have that term (ii).b in (C.34) is $O_{\mathbb{P}}\big[\zeta^{\mathrm{L}}\cdot(\zeta^{\mathrm{EM}})^2\big]$. Moreover, by our assumption in (4.6) of Assumption 4.5 we have

$$\zeta^{\mathrm{L}}\cdot\big(\zeta^{\mathrm{EM}}\big)^2\leq\max\big\{1,\ \|\mathbf{w}^*\|_1\big\}\cdot\zeta^{\mathrm{L}}\cdot\big(\zeta^{\mathrm{EM}}\big)^2=o(1/\sqrt{n}).$$

Thus, we conclude that term (ii).b is $o_{\mathbb{P}}(1/\sqrt{n})$. Similarly, term (ii).c in (C.34) is upper bounded by

$$\|\widehat{\mathbf{w}}_0\|_1\cdot\Big\|T_n(\boldsymbol{\beta}^\sharp)-T_n(\widehat{\boldsymbol{\beta}}_0)\Big\|_{\infty,\infty}\cdot\big\|\widehat{\boldsymbol{\beta}}_0-\boldsymbol{\beta}^*\big\|_1. \tag{C.40}$$

By triangle inequality and Lemma C.3, the first term in (C.40) is upper bounded by

$$\|\mathbf{w}^*\|_1+\|\widehat{\mathbf{w}}_0-\mathbf{w}^*\|_1=\|\mathbf{w}^*\|_1+O_{\mathbb{P}}(s^*_{\mathbf{w}}\cdot\lambda).$$

Meanwhile, for the second and third terms in (C.40), by the same analysis for term (ii).b in (C.34) we have

$$\Big\|T_n(\boldsymbol{\beta}^\sharp)-T_n(\widehat{\boldsymbol{\beta}}_0)\Big\|_{\infty,\infty}\cdot\big\|\widehat{\boldsymbol{\beta}}_0-\boldsymbol{\beta}^*\big\|_1=O_{\mathbb{P}}\big[\zeta^{\mathrm{L}}\cdot\big(\zeta^{\mathrm{EM}}\big)^2\big].$$

By (4.6) in Assumption 4.5, since $s^*_{\mathbf{w}}\cdot\lambda=o(1)$, we have

$$\begin{aligned}\big(\|\mathbf{w}^*\|_1+s^*_{\mathbf{w}}\cdot\lambda\big)\cdot\zeta^{\mathrm{L}}\cdot\big(\zeta^{\mathrm{EM}}\big)^2&\leq\big[\max\big\{1,\ \|\mathbf{w}^*\|_1\big\}+o(1)\big]\cdot\zeta^{\mathrm{L}}\cdot\big(\zeta^{\mathrm{EM}}\big)^2\\&=o(1/\sqrt{n}).\end{aligned}$$

Therefore, term (ii).c in (C.34) is $o_{\mathbb{P}}(1/\sqrt{n})$. Hence, by (C.34) we conclude that term (ii) in (C.31) is $o_{\mathbb{P}}(1/\sqrt{n})$. Combining the analysis for terms (i) and (ii) in (C.31), we then obtain (C.29). In the sequel, we turn to prove the second part on asymptotic normality.

**Proof of (C.30):** Note that by Lemma 2.1, we have

$$\begin{aligned}\sqrt{n}\cdot\big[\nabla_1 Q_n(\boldsymbol{\beta}^*;\boldsymbol{\beta}^*)\big]_\alpha-\sqrt{n}\cdot(\mathbf{w}^*)^\top\cdot\big[\nabla_1 Q_n(\boldsymbol{\beta}^*;\boldsymbol{\beta}^*)\big]_{\boldsymbol{\gamma}}&=\sqrt{n}\cdot\big[1,-(\mathbf{w}^*)^\top\big]\cdot\nabla_1 Q_n(\boldsymbol{\beta}^*;\boldsymbol{\beta}^*)\\&=\sqrt{n}\cdot\big[1,-(\mathbf{w}^*)^\top\big]\cdot\nabla\ell_n(\boldsymbol{\beta}^*)/n.\end{aligned} \tag{C.41}$$

Recall that $\ell_n(\boldsymbol{\beta}^*)$ is the log-likelihood function defined in (2.2). Hence, $\big[1,-(\mathbf{w}^*)^\top\big]\cdot\nabla\ell_n(\boldsymbol{\beta}^*)/n$ is the average of $n$ independent random variables. Meanwhile, the score function has mean zero at $\boldsymbol{\beta}^*$, i.e., $\mathbb{E}\big[\nabla\ell_n(\boldsymbol{\beta}^*)\big]=\mathbf{0}$. For the variance of the rescaled average in (C.41), we have

$$\begin{aligned}\mathrm{Var}\Big\{\sqrt{n}\cdot\big[1,-(\mathbf{w}^*)^\top\big]\cdot\nabla\ell_n(\boldsymbol{\beta}^*)/n\Big\}&=\big[1,-(\mathbf{w}^*)^\top\big]\cdot\mathrm{Cov}\big[\nabla\ell_n(\boldsymbol{\beta}^*)/\sqrt{n}\big]\cdot\big[1,-(\mathbf{w}^*)^\top\big]^\top\\&=\big[1,-(\mathbf{w}^*)^\top\big]\cdot I(\boldsymbol{\beta}^*)\cdot\big[1,-(\mathbf{w}^*)^\top\big]^\top.\end{aligned}$$

Here the second equality is from the fact that the covariance of the score function equals the Fisher information (up to renormalization). Hence, the variance of each item in the average in (C.41) is

$$\begin{aligned}\big[1,-(\mathbf{w}^*)^\top\big]\cdot I(\boldsymbol{\beta}^*)\cdot\big[1,-(\mathbf{w}^*)^\top\big]^\top&=\big[I(\boldsymbol{\beta}^*)\big]_{\alpha,\alpha}-2\cdot(\mathbf{w}^*)^\top\cdot\big[I(\boldsymbol{\beta}^*)\big]_{\boldsymbol{\gamma},\alpha}+(\mathbf{w}^*)^\top\cdot\big[I(\boldsymbol{\beta}^*)\big]_{\boldsymbol{\gamma},\boldsymbol{\gamma}}\cdot\mathbf{w}^*\\&=\big[I(\boldsymbol{\beta}^*)\big]_{\alpha,\alpha}-\big[I(\boldsymbol{\beta}^*)\big]_{\boldsymbol{\gamma},\alpha}\cdot\big[I(\boldsymbol{\beta}^*)\big]^{-1}_{\boldsymbol{\gamma},\boldsymbol{\gamma}}\cdot\big[I(\boldsymbol{\beta}^*)\big]_{\boldsymbol{\gamma},\alpha}\\&=\big[I(\boldsymbol{\beta}^*)\big]_{\alpha|\boldsymbol{\gamma}},\end{aligned}$$

where the second and third equalities are from (4.1) and (4.2). Hence, by the central limit theorem we obtain (C.30). Finally, combining (C.29) and (C.30) by invoking Slutsky's theorem, we obtain

$$\sqrt{n} \cdot S_n\big(\widehat{\boldsymbol{\beta}}_0, \lambda\big) \xrightarrow{D} N\big(0, \big[I(\boldsymbol{\beta}^*)\big]_{\alpha|\boldsymbol{\gamma}}\big),$$

which concludes the proof of Lemma 5.3. □

## C.4 Proof of Lemma 5.4

*Proof.* Throughout the proof, we abbreviate $w\big(\widehat{\boldsymbol{\beta}}_0, \lambda\big)$ as $\widehat{\mathbf{w}}_0$. Our proof is under the null hypothesis where $\boldsymbol{\beta}^* = \big[\alpha^*, (\boldsymbol{\gamma}^*)^\top\big]^\top$ with $\alpha^* = 0$. Recall that $\mathbf{w}^*$ is defined in (4.1). Then by the definitions of $\big[T_n\big(\widehat{\boldsymbol{\beta}}_0\big)\big]_{\alpha|\boldsymbol{\gamma}}$ and $\big[I(\boldsymbol{\beta}^*)\big]_{\alpha|\boldsymbol{\gamma}}$ in (2.11) and (4.2), we have

$$\begin{aligned}
\big[T_n\big(\widehat{\boldsymbol{\beta}}_0\big)\big]_{\alpha|\boldsymbol{\gamma}} &= \big(1, -\widehat{\mathbf{w}}_0^\top\big) \cdot T_n\big(\widehat{\boldsymbol{\beta}}_0\big) \cdot \big(1, -\widehat{\mathbf{w}}_0^\top\big)^\top \\
&= \big[T_n\big(\widehat{\boldsymbol{\beta}}_0\big)\big]_{\alpha,\alpha} - 2 \cdot \widehat{\mathbf{w}}_0^\top \cdot \big[T_n\big(\widehat{\boldsymbol{\beta}}_0\big)\big]_{\boldsymbol{\gamma},\alpha} + \widehat{\mathbf{w}}_0^\top \cdot \big[T_n\big(\widehat{\boldsymbol{\beta}}_0\big)\big]_{\boldsymbol{\gamma},\boldsymbol{\gamma}} \cdot \widehat{\mathbf{w}}_0, \\
\big[I(\boldsymbol{\beta}^*)\big]_{\alpha|\boldsymbol{\gamma}} &= \big[I(\boldsymbol{\beta}^*)\big]_{\alpha,\alpha} - \big[I(\boldsymbol{\beta}^*)\big]_{\boldsymbol{\gamma},\alpha}^\top \cdot \big[I(\boldsymbol{\beta}^*)\big]_{\boldsymbol{\gamma},\boldsymbol{\gamma}}^{-1} \cdot \big[I(\boldsymbol{\beta}^*)\big]_{\boldsymbol{\gamma},\alpha} \\
&= \big[I(\boldsymbol{\beta}^*)\big]_{\alpha,\alpha} - 2 \cdot (\mathbf{w}^*)^\top \cdot \big[I(\boldsymbol{\beta}^*)\big]_{\boldsymbol{\gamma},\alpha} + (\mathbf{w}^*)^\top \cdot \big[I(\boldsymbol{\beta}^*)\big]_{\boldsymbol{\gamma},\boldsymbol{\gamma}} \cdot \mathbf{w}^*.
\end{aligned}$$

By triangle inequality, we have

$$\begin{aligned}
\Big|\big[T_n\big(\widehat{\boldsymbol{\beta}}_0\big)\big]_{\alpha|\boldsymbol{\gamma}} + \big[I(\boldsymbol{\beta}^*)\big]_{\alpha|\boldsymbol{\gamma}}\Big| \leq &\underbrace{\Big|\big[T_n\big(\widehat{\boldsymbol{\beta}}_0\big)\big]_{\alpha,\alpha} + \big[I(\boldsymbol{\beta}^*)\big]_{\alpha,\alpha}\Big|}_{\text{(i)}} + 2 \cdot \underbrace{\Big|\widehat{\mathbf{w}}_0^\top \cdot \big[T_n\big(\widehat{\boldsymbol{\beta}}_0\big)\big]_{\boldsymbol{\gamma},\alpha} + (\mathbf{w}^*)^\top \cdot \big[I(\boldsymbol{\beta}^*)\big]_{\boldsymbol{\gamma},\alpha}\Big|}_{\text{(ii)}} \\
&+ \underbrace{\Big|\widehat{\mathbf{w}}_0^\top \cdot \big[T_n\big(\widehat{\boldsymbol{\beta}}_0\big)\big]_{\boldsymbol{\gamma},\boldsymbol{\gamma}} \cdot \widehat{\mathbf{w}}_0 + (\mathbf{w}^*)^\top \cdot \big[I(\boldsymbol{\beta}^*)\big]_{\boldsymbol{\gamma},\boldsymbol{\gamma}} \cdot \mathbf{w}^*\Big|}_{\text{(iii)}}. \qquad \text{(C.42)}
\end{aligned}$$

**Analysis of Term (i):** For term (i) in (C.42), by Lemma 2.1 and triangle inequality we have

$$\Big|\big[T_n\big(\widehat{\boldsymbol{\beta}}_0\big)\big]_{\alpha,\alpha} + \big[I(\boldsymbol{\beta}^*)\big]_{\alpha,\alpha}\Big| \leq \underbrace{\Big|\big[T_n\big(\widehat{\boldsymbol{\beta}}_0\big)\big]_{\alpha,\alpha} - \big[T_n(\boldsymbol{\beta}^*)\big]_{\alpha,\alpha}\Big|}_{\text{(i).a}} + \underbrace{\Big|\big[T_n(\boldsymbol{\beta}^*)\big]_{\alpha,\alpha} - \big\{\mathbb{E}_{\boldsymbol{\beta}^*}\big[T_n(\boldsymbol{\beta}^*)\big]\big\}_{\alpha,\alpha}\Big|}_{\text{(i).b}}. \qquad \text{(C.43)}$$

For term (i).a in (C.43), by Condition 4.4 we have

$$\begin{aligned}
\Big|\big[T_n\big(\widehat{\boldsymbol{\beta}}_0\big)\big]_{\alpha,\alpha} - \big[T_n(\boldsymbol{\beta}^*)\big]_{\alpha,\alpha}\Big| &\leq \Big\|T_n\big(\widehat{\boldsymbol{\beta}}_0\big) - T_n(\boldsymbol{\beta}^*)\Big\|_{\infty,\infty} \\
&= O_{\mathbb{P}}\big(\zeta^{\mathrm{L}}\big) \cdot \big\|\widehat{\boldsymbol{\beta}}_0 - \boldsymbol{\beta}^*\big\|_1. \qquad \text{(C.44)}
\end{aligned}$$

Note that we have $\big(\widehat{\boldsymbol{\beta}}_0\big)_\alpha = (\boldsymbol{\beta}^*)_\alpha = 0$ by definition, which implies $\big\|\widehat{\boldsymbol{\beta}}_0 - \boldsymbol{\beta}^*\big\|_1 \leq \big\|\widehat{\boldsymbol{\beta}} - \boldsymbol{\beta}^*\big\|_1$. Hence, by Condition 4.1 we have

$$\big\|\widehat{\boldsymbol{\beta}}_0 - \boldsymbol{\beta}^*\big\|_1 = O_{\mathbb{P}}\big(\zeta^{\mathrm{EM}}\big). \qquad \text{(C.45)}$$

Moreover, combining (C.44) and (C.45), by (4.6) in Assumption 4.5 we have

$$\zeta^{\mathrm{L}} \cdot \zeta^{\mathrm{EM}} \leq \max\bigl\{\|\mathbf{w}^*\|_1,\ 1\bigr\} \cdot s^*_{\mathbf{w}} \cdot \lambda = o(1)$$

for $\lambda$ specified in (4.5). Hence we obtain

$$\begin{aligned}\Bigl|\bigl[T_n(\widehat{\boldsymbol{\beta}}_0)\bigr]_{\alpha,\alpha} - \bigl[T_n(\boldsymbol{\beta}^*)\bigr]_{\alpha,\alpha}\Bigr| &\leq \Bigl\|T_n(\widehat{\boldsymbol{\beta}}_0) - T_n(\boldsymbol{\beta}^*)\Bigr\|_{\infty,\infty} \\ &= O_{\mathbb{P}}\bigl(\zeta^{\mathrm{L}} \cdot \zeta^{\mathrm{EM}}\bigr) = o_{\mathbb{P}}(1). \end{aligned} \tag{C.46}$$

Meanwhile, for term (i).b in (C.43) we have

$$\Bigl|\bigl[T_n(\boldsymbol{\beta}^*)\bigr]_{\alpha,\alpha} - \bigl\{\mathbb{E}_{\boldsymbol{\beta}^*}\bigl[T_n(\boldsymbol{\beta}^*)\bigr]\bigr\}_{\alpha,\alpha}\Bigr| \leq \bigl\|T_n(\boldsymbol{\beta}^*) - \mathbb{E}_{\boldsymbol{\beta}^*}\bigl[T_n(\boldsymbol{\beta}^*)\bigr]\bigr\|_{\infty,\infty} = O_{\mathbb{P}}\bigl(\zeta^{\mathrm{T}}\bigr) = o_{\mathbb{P}}(1), \tag{C.47}$$

where the second last equality follows from Condition 4.3, while the last equality holds because our assumption in (4.6) of Assumption 4.5 implies

$$\zeta^{\mathrm{T}} \leq \max\bigl\{\|\mathbf{w}^*\|_1,\ 1\bigr\} \cdot s^*_{\mathbf{w}} \cdot \lambda = o(1)$$

for $\lambda$ specified in (4.5).

**Analysis of Term (ii):** For term (ii) in (C.42), by Lemma 2.1 and triangle inequality, we have

$$\begin{aligned}&\Bigl|\widehat{\mathbf{w}}_0^\top \cdot \bigl[T_n(\widehat{\boldsymbol{\beta}}_0)\bigr]_{\boldsymbol{\gamma},\alpha} + (\mathbf{w}^*)^\top \cdot \bigl[I(\boldsymbol{\beta}^*)\bigr]_{\boldsymbol{\gamma},\alpha}\Bigr| \\ &\quad \leq \underbrace{\Bigl|(\widehat{\mathbf{w}}_0 - \mathbf{w}^*)^\top \cdot \Bigl\{T_n(\widehat{\boldsymbol{\beta}}_0) - \mathbb{E}_{\boldsymbol{\beta}^*}\bigl[T_n(\boldsymbol{\beta}^*)\bigr]\Bigr\}_{\boldsymbol{\gamma},\alpha}\Bigr|}_{\text{(ii).a}} + \underbrace{\Bigl|(\widehat{\mathbf{w}}_0 - \mathbf{w}^*)^\top \cdot \bigl\{\mathbb{E}_{\boldsymbol{\beta}^*}\bigl[T_n(\boldsymbol{\beta}^*)\bigr]\bigr\}_{\boldsymbol{\gamma},\alpha}\Bigr|}_{\text{(ii).b}} \\ &\qquad + \underbrace{\Bigl|(\mathbf{w}^*)^\top \cdot \Bigl\{T_n(\widehat{\boldsymbol{\beta}}_0) - \mathbb{E}_{\boldsymbol{\beta}^*}\bigl[T_n(\boldsymbol{\beta}^*)\bigr]\Bigr\}_{\boldsymbol{\gamma},\alpha}\Bigr|}_{\text{(ii).c}}. \end{aligned} \tag{C.48}$$

By Hölder's inequality, term (ii).a in (C.48) is upper bounded by

$$\|\widehat{\mathbf{w}}_0 - \mathbf{w}^*\|_1 \cdot \Bigl\|\Bigl\{T_n(\widehat{\boldsymbol{\beta}}_0) - \mathbb{E}_{\boldsymbol{\beta}^*}\bigl[T_n(\boldsymbol{\beta}^*)\bigr]\Bigr\}_{\boldsymbol{\gamma},\alpha}\Bigr\|_\infty \leq \|\widehat{\mathbf{w}}_0 - \mathbf{w}^*\|_1 \cdot \Bigl\|T_n(\widehat{\boldsymbol{\beta}}_0) - \mathbb{E}_{\boldsymbol{\beta}^*}\bigl[T_n(\boldsymbol{\beta}^*)\bigr]\Bigr\|_{\infty,\infty}.$$

By Lemma C.3, we have $\|\widehat{\mathbf{w}}_0 - \mathbf{w}^*\|_1 = O_{\mathbb{P}}(s^*_{\mathbf{w}} \cdot \lambda)$. Meanwhile, we have

$$\Bigl\|T_n(\widehat{\boldsymbol{\beta}}_0) - \mathbb{E}_{\boldsymbol{\beta}^*}\bigl[T_n(\boldsymbol{\beta}^*)\bigr]\Bigr\|_{\infty,\infty} \leq \Bigl\|T_n(\widehat{\boldsymbol{\beta}}_0) - T_n(\boldsymbol{\beta}^*)\Bigr\|_{\infty,\infty} + \bigl\|T_n(\boldsymbol{\beta}^*) - \mathbb{E}_{\boldsymbol{\beta}^*}\bigl[T_n(\boldsymbol{\beta}^*)\bigr]\bigr\|_{\infty,\infty} = o_{\mathbb{P}}(1).$$

where the second equality follows from (C.46) and (C.47). Therefore, term (ii).a is $o_{\mathbb{P}}(1)$, since (4.6) in Assumption 4.5 implies $s^*_{\mathbf{w}} \cdot \lambda = o(1)$. Meanwhile, by Hölder's inequality, term (ii).b in (C.48) is upper bounded by

$$\|\widehat{\mathbf{w}}_0 - \mathbf{w}^*\|_1 \cdot \bigl\|\bigl\{\mathbb{E}_{\boldsymbol{\beta}^*}\bigl[T_n(\boldsymbol{\beta}^*)\bigr]\bigr\}_{\boldsymbol{\gamma},\alpha}\bigr\|_\infty \leq \|\widehat{\mathbf{w}}_0 - \mathbf{w}^*\|_1 \cdot \bigl\|\mathbb{E}_{\boldsymbol{\beta}^*}\bigl[T_n(\boldsymbol{\beta}^*)\bigr]\bigr\|_{\infty,\infty}. \tag{C.49}$$

By Lemma C.3, we have $\|\widehat{\mathbf{w}}_0 - \mathbf{w}^*\|_1 = O_{\mathbb{P}}(s^*_{\mathbf{w}} \cdot \lambda)$. Meanwhile, we have $\mathbb{E}_{\boldsymbol{\beta}^*}\big[T_n(\boldsymbol{\beta}^*)\big] = -I(\boldsymbol{\beta}^*)$ by Lemma 2.1. Furthermore, (4.4) in Assumption 4.5 implies

$$\big\|I(\boldsymbol{\beta}^*)\big\|_{\infty,\infty} \leq \big\|I(\boldsymbol{\beta}^*)\big\|_2 \leq C, \tag{C.50}$$

where $C > 0$ is some absolute constant. Therefore, from (C.49) we have that term (ii).b in (C.48) is $O_{\mathbb{P}}(s^*_{\mathbf{w}} \cdot \lambda)$. By (4.6) in Assumption 4.5, we have $s^*_{\mathbf{w}} \cdot \lambda = o(1)$. Thus, we conclude that term (ii).b is $o_{\mathbb{P}}(1)$. For term (ii).c, we have

$$\begin{aligned}
&\Big|(\mathbf{w}^*)^\top \cdot \Big\{T_n\big(\widehat{\boldsymbol{\beta}}_0\big) - \mathbb{E}_{\boldsymbol{\beta}^*}\big[T_n(\boldsymbol{\beta}^*)\big]\Big\}_{\gamma,\alpha}\Big| \\
&\quad \leq \|\mathbf{w}^*\|_1 \cdot \Big\|T_n\big(\widehat{\boldsymbol{\beta}}_0\big) - \mathbb{E}_{\boldsymbol{\beta}^*}\big[T_n(\boldsymbol{\beta}^*)\big]\Big\|_{\infty,\infty} \\
&\quad \leq \|\mathbf{w}^*\|_1 \cdot \Big\|T_n\big(\widehat{\boldsymbol{\beta}}_0\big) - T_n(\boldsymbol{\beta}^*)\Big\|_{\infty,\infty} + \|\mathbf{w}^*\|_1 \cdot \big\|T_n(\boldsymbol{\beta}^*) - \mathbb{E}_{\boldsymbol{\beta}^*}\big[T_n(\boldsymbol{\beta}^*)\big]\big\|_{\infty,\infty} \\
&\quad = O_{\mathbb{P}}\big(\|\mathbf{w}^*\|_1 \cdot \zeta^{\mathrm{L}} \cdot \zeta^{\mathrm{EM}}\big) + O_{\mathbb{P}}\big(\|\mathbf{w}^*\|_1 \cdot \zeta^{\mathrm{T}}\big) = o_{\mathbb{P}}(1).
\end{aligned}$$

Here the first and second inequalities are from Hölder's inequality and triangle inequality, the first equality follows from (C.46) and (C.47), and the second equality holds because (4.6) in Assumption 4.5 implies

$$\|\mathbf{w}^*\|_1 \cdot \big(\zeta^{\mathrm{L}} \cdot \zeta^{\mathrm{EM}} + \zeta^{\mathrm{T}}\big) \leq \max\big\{\|\mathbf{w}^*\|_1,\ 1\big\} \cdot s^*_{\mathbf{w}} \cdot \lambda = o(1)$$

for $\lambda$ specified in (4.5).

**Analysis of Term (iii):** For term (iii) in (C.42), by (2.12) in Lemma 2.1 we have

$$\begin{aligned}
&\Big|\widehat{\mathbf{w}}_0^\top \cdot \big[T_n\big(\widehat{\boldsymbol{\beta}}_0\big)\big]_{\gamma,\gamma} \cdot \widehat{\mathbf{w}}_0 + (\mathbf{w}^*)^\top \cdot \big[I(\boldsymbol{\beta}^*)\big]_{\gamma,\gamma} \cdot \mathbf{w}^*\Big| \qquad\qquad (\mathrm{C}.51)\\
&\quad \leq \underbrace{\Big|\widehat{\mathbf{w}}_0^\top \cdot \Big\{T_n\big(\widehat{\boldsymbol{\beta}}_0\big) - \mathbb{E}_{\boldsymbol{\beta}^*}\big[T_n(\boldsymbol{\beta}^*)\big]\Big\}_{\gamma,\gamma} \cdot \widehat{\mathbf{w}}_0\Big|}_{\text{(iii).a}} + \underbrace{\Big|\widehat{\mathbf{w}}_0^\top \cdot \big[I(\boldsymbol{\beta}^*)\big]_{\gamma,\gamma} \cdot \widehat{\mathbf{w}}_0 - (\mathbf{w}^*)^\top \cdot \big[I(\boldsymbol{\beta}^*)\big]_{\gamma,\gamma} \cdot \mathbf{w}^*\Big|}_{\text{(iii).b}}.
\end{aligned}$$

For term (iii).a in (C.51), we have

$$\begin{aligned}
\Big|\widehat{\mathbf{w}}_0^\top \cdot \Big\{T_n\big(\widehat{\boldsymbol{\beta}}_0\big) - \mathbb{E}_{\boldsymbol{\beta}^*}\big[T_n(\boldsymbol{\beta}^*)\big]\Big\}_{\gamma,\gamma} \cdot \widehat{\mathbf{w}}_0\Big| &\leq \|\widehat{\mathbf{w}}_0\|_1^2 \cdot \Big\|\Big\{T_n\big(\widehat{\boldsymbol{\beta}}_0\big) - \mathbb{E}_{\boldsymbol{\beta}^*}\big[T_n(\boldsymbol{\beta}^*)\big]\Big\}_{\gamma,\gamma}\Big\|_{\infty,\infty} \\
&\leq \|\widehat{\mathbf{w}}_0\|_1^2 \cdot \Big\|T_n\big(\widehat{\boldsymbol{\beta}}_0\big) - \mathbb{E}_{\boldsymbol{\beta}^*}\big[T_n(\boldsymbol{\beta}^*)\big]\Big\|_{\infty,\infty}. \qquad (\mathrm{C}.52)
\end{aligned}$$

For $\|\widehat{\mathbf{w}}_0\|_1$ we have $\|\widehat{\mathbf{w}}_0\|_1^2 \leq \big(\|\mathbf{w}^*\|_1 + \|\widehat{\mathbf{w}}_0 - \mathbf{w}^*\|_1\big)^2 = \big[\|\mathbf{w}^*\|_1 + O_{\mathbb{P}}(s^*_{\mathbf{w}} \cdot \lambda)\big]^2$, where the equality holds because by Lemma C.3 we have $\|\widehat{\mathbf{w}}_0 - \mathbf{w}^*\|_1 = O_{\mathbb{P}}(s^*_{\mathbf{w}} \cdot \lambda)$. Meanwhile, on the right-hand side of (C.52) we have

$$\begin{aligned}
\Big\|T_n\big(\widehat{\boldsymbol{\beta}}_0\big) - \mathbb{E}_{\boldsymbol{\beta}^*}\big[T_n(\boldsymbol{\beta}^*)\big]\Big\|_{\infty,\infty} &\leq \Big\|T_n\big(\widehat{\boldsymbol{\beta}}_0\big) - T_n(\boldsymbol{\beta}^*)\Big\|_{\infty,\infty} + \big\|T_n(\boldsymbol{\beta}^*) - \mathbb{E}_{\boldsymbol{\beta}^*}\big[T_n(\boldsymbol{\beta}^*)\big]\big\|_{\infty,\infty} \\
&= O_{\mathbb{P}}\big(\zeta^{\mathrm{L}} \cdot \zeta^{\mathrm{EM}} + \zeta^{\mathrm{T}}\big).
\end{aligned}$$

Here the last equality is from (C.46) and (C.47). Hence, term (iii).a in (C.51) is $O_{\mathbb{P}}\bigl[\bigl(\|\mathbf{w}^*\|_1 + s^*_{\mathbf{w}} \cdot \lambda\bigr)^2 \cdot \bigl(\zeta^{\rm L} \cdot \zeta^{\rm EM} + \zeta^{\rm T}\bigr)\bigr]$. Note that

$$\begin{aligned}&\bigl(\|\mathbf{w}^*\|_1 + s^*_{\mathbf{w}} \cdot \lambda\bigr)^2 \cdot \bigl(\zeta^{\rm L} \cdot \zeta^{\rm EM} + \zeta^{\rm T}\bigr)\\ &\quad = \underbrace{\|\mathbf{w}^*\|_1^2 \cdot \bigl(\zeta^{\rm L} \cdot \zeta^{\rm EM} + \zeta^{\rm T}\bigr)}_{\rm (i)} + 2 \cdot \underbrace{s^*_{\mathbf{w}} \cdot \lambda}_{\rm (ii)} \cdot \underbrace{\|\mathbf{w}^*\|_1 \cdot \bigl(\zeta^{\rm L} \cdot \zeta^{\rm EM} + \zeta^{\rm T}\bigr)}_{\rm (iii)} + (\underbrace{s^*_{\mathbf{w}} \cdot \lambda}_{\rm (ii)})^2 \cdot \underbrace{\bigl(\zeta^{\rm L} \cdot \zeta^{\rm EM} + \zeta^{\rm T}\bigr)}_{\rm (iv)}.\end{aligned}$$

From (4.6) in Assumption 4.5 we have, for $\lambda$ specified in (4.5), terms (i)-(iv) are all upper bounded by $\max\bigl\{\|\mathbf{w}^*\|_1, 1\bigr\} \cdot s^*_{\mathbf{w}} \cdot \lambda = o(1)$. Hence, we conclude term (iii).a in (C.51) is $o_{\mathbb{P}}(1)$. Also, for term (iii).b in (C.51), we have

$$\begin{aligned}&\Bigl|\widehat{\mathbf{w}}_0^\top \cdot \bigl[I(\boldsymbol{\beta}^*)\bigr]_{\boldsymbol{\gamma},\boldsymbol{\gamma}} \cdot \widehat{\mathbf{w}}_0 - (\mathbf{w}^*)^\top \cdot \bigl[I(\boldsymbol{\beta}^*)\bigr]_{\boldsymbol{\gamma},\boldsymbol{\gamma}} \cdot \mathbf{w}^*\Bigr|\\ &\quad \le \Bigl|(\widehat{\mathbf{w}}_0 - \mathbf{w}^*)^\top \cdot \bigl[I(\boldsymbol{\beta}^*)\bigr]_{\boldsymbol{\gamma},\boldsymbol{\gamma}} \cdot (\widehat{\mathbf{w}}_0 - \mathbf{w}^*)\Bigr| + 2 \cdot \Bigl|\mathbf{w}^* \cdot \bigl[I(\boldsymbol{\beta}^*)\bigr]_{\boldsymbol{\gamma},\boldsymbol{\gamma}} \cdot (\widehat{\mathbf{w}}_0 - \mathbf{w}^*)\Bigr|\\ &\quad \le \|\widehat{\mathbf{w}}_0 - \mathbf{w}^*\|_1^2 \cdot \bigl\|I(\boldsymbol{\beta}^*)\bigr\|_{\infty,\infty} + 2 \cdot \|\widehat{\mathbf{w}}_0 - \mathbf{w}^*\|_1 \cdot \|\mathbf{w}^*\|_1 \cdot \bigl\|I(\boldsymbol{\beta}^*)\bigr\|_{\infty,\infty}\\ &\quad = O_{\mathbb{P}}\bigl[(s^*_{\mathbf{w}} \cdot \lambda)^2 + \|\mathbf{w}^*\|_1 \cdot s^*_{\mathbf{w}} \cdot \lambda\bigr],\end{aligned}$$

where the last equality follows from Lemma C.3 and (C.50). Moreover, by (4.6) in Assumption 4.5 we have $\max\bigl\{s^*_{\mathbf{w}} \cdot \lambda,\ \|\mathbf{w}^*\|_1 \cdot s^*_{\mathbf{w}} \cdot \lambda\bigr\} \le \max\bigl\{\|\mathbf{w}^*\|_1,\ 1\bigr\} \cdot s^*_{\mathbf{w}} \cdot \lambda = o(1)$. Therefore, we conclude that term (iii).b in (C.51) is $o_{\mathbb{P}}(1)$. Combining the above analysis for terms (i)-(iii) in (C.42), we obtain

$$\Bigl|\bigl[T_n(\widehat{\boldsymbol{\beta}}_0)\bigr]_{\alpha|\boldsymbol{\gamma}} + \bigl[I(\boldsymbol{\beta}^*)\bigr]_{\alpha|\boldsymbol{\gamma}}\Bigr| = o_{\mathbb{P}}(1).$$

Thus we conclude the proof of Lemma 5.4. □

### C.5 Proof of Lemma 5.5

*Proof.* Our proof strategy is as follows. Recall that $\mathbf{w}^*$ is defined in (4.1). First we prove

$$\sqrt{n} \cdot \bigl[\overline{\alpha}(\widehat{\boldsymbol{\beta}}, \lambda) - \alpha^*\bigr] = -\sqrt{n} \cdot \bigl[I(\boldsymbol{\beta}^*)\bigr]^{-1}_{\alpha|\boldsymbol{\gamma}} \cdot \Bigl\{\bigl[\nabla_1 Q_n(\boldsymbol{\beta}^*; \boldsymbol{\beta}^*)\bigr]_\alpha - (\mathbf{w}^*)^\top \cdot \bigl[\nabla_1 Q_n(\boldsymbol{\beta}^*; \boldsymbol{\beta}^*)\bigr]_{\boldsymbol{\gamma}}\Bigr\} + o_{\mathbb{P}}(1). \tag{C.53}$$

Here note that $\bigl[I(\boldsymbol{\beta}^*)\bigr]_{\alpha|\boldsymbol{\gamma}}$ is defined in (4.2) and $\widehat{\boldsymbol{\beta}} = \bigl[\widehat{\alpha}, \widehat{\boldsymbol{\gamma}}^\top\bigr]^\top$ is the estimator attained by the high dimensional EM algorithm (Algorithm 4). Then we prove

$$\sqrt{n} \cdot \bigl[I(\boldsymbol{\beta}^*)\bigr]^{-1}_{\alpha|\boldsymbol{\gamma}} \cdot \Bigl\{\bigl[\nabla_1 Q_n(\boldsymbol{\beta}^*; \boldsymbol{\beta}^*)\bigr]_\alpha - (\mathbf{w}^*)^\top \cdot \bigl[\nabla_1 Q_n(\boldsymbol{\beta}^*; \boldsymbol{\beta}^*)\bigr]_{\boldsymbol{\gamma}}\Bigr\} \xrightarrow{D} N\bigl(0, \bigl[I(\boldsymbol{\beta}^*)\bigr]^{-1}_{\alpha|\boldsymbol{\gamma}}\bigr). \tag{C.54}$$

**Proof of (C.53):** For notational simplicity, we define

$$\overline{S}_n(\boldsymbol{\beta}, \widehat{\mathbf{w}}) = \bigl[\nabla_1 Q_n(\boldsymbol{\beta}; \boldsymbol{\beta})\bigr]_\alpha - \widehat{\mathbf{w}}^\top \cdot \bigl[\nabla_1 Q_n(\boldsymbol{\beta}; \boldsymbol{\beta})\bigr]_{\boldsymbol{\gamma}}, \quad \text{where } \ \widehat{\mathbf{w}} = w(\widehat{\boldsymbol{\beta}}, \lambda). \tag{C.55}$$

By the definition of $S_n(\cdot,\cdot)$ in (2.9), we have $\overline{S}_n(\widehat{\boldsymbol{\beta}}, \widehat{\mathbf{w}}) = S_n(\widehat{\boldsymbol{\beta}}, \lambda)$. Let $\widetilde{\boldsymbol{\beta}} = \bigl(\alpha^*, \widehat{\boldsymbol{\gamma}}^\top\bigr)^\top$. The Taylor expansion of $\overline{S}_n(\widehat{\boldsymbol{\beta}}, \widehat{\mathbf{w}})$ takes the form

$$\overline{S}_n(\widehat{\boldsymbol{\beta}}, \widehat{\mathbf{w}}) = \overline{S}_n(\widetilde{\boldsymbol{\beta}}, \widehat{\mathbf{w}}) + (\widehat{\alpha} - \alpha^*) \cdot \bigl[\nabla \overline{S}_n(\boldsymbol{\beta}^\sharp, \widehat{\mathbf{w}})\bigr]_\alpha, \tag{C.56}$$

where $\boldsymbol{\beta}^\sharp$ is an intermediate value between $\widehat{\boldsymbol{\beta}}$ and $\widetilde{\boldsymbol{\beta}}$. By (C.55) and the definition of $T_n(\cdot)$ in (2.8), we have

$$\begin{aligned}\big[\nabla \bar{S}_n(\widehat{\boldsymbol{\beta}}, \widehat{\mathbf{w}})\big]_\alpha &= \big[\nabla_{1,1} Q_n(\widehat{\boldsymbol{\beta}}; \widehat{\boldsymbol{\beta}}) + \nabla_{1,2} Q_n(\widehat{\boldsymbol{\beta}}; \widehat{\boldsymbol{\beta}})\big]_{\alpha,\alpha} - \widehat{\mathbf{w}}^\top \cdot \big[\nabla_{1,1} Q_n(\widehat{\boldsymbol{\beta}}; \widehat{\boldsymbol{\beta}}) + \nabla_{1,2} Q_n(\widehat{\boldsymbol{\beta}}; \widehat{\boldsymbol{\beta}})\big]_{\boldsymbol{\gamma},\alpha} \\ &= \big[T(\widehat{\boldsymbol{\beta}})\big]_{\alpha,\alpha} - \widehat{\mathbf{w}}^\top \cdot \big[T(\widehat{\boldsymbol{\beta}})\big]_{\boldsymbol{\gamma},\alpha}. \end{aligned} \tag{C.57}$$

By (C.57) and the definition in (2.18), we have

$$\bar{\alpha}(\widehat{\boldsymbol{\beta}}, \lambda) = \widehat{\alpha} - \Big\{\big[T(\widehat{\boldsymbol{\beta}})\big]_{\alpha,\alpha} - \widehat{\mathbf{w}}^\top \cdot \big[T(\widehat{\boldsymbol{\beta}})\big]_{\boldsymbol{\gamma},\alpha}\Big\}^{-1} \cdot \bar{S}_n(\widehat{\boldsymbol{\beta}}, \widehat{\mathbf{w}}) = \widehat{\alpha} - \big[\nabla \bar{S}_n(\widehat{\boldsymbol{\beta}}, \widehat{\mathbf{w}})\big]_\alpha^{-1} \cdot \bar{S}_n(\widehat{\boldsymbol{\beta}}, \widehat{\mathbf{w}}).$$

Further by (C.56) we obtain

$$\begin{aligned}\sqrt{n} \cdot \big[\bar{\alpha}(\widehat{\boldsymbol{\beta}}, \lambda) - \alpha^*\big] &= \sqrt{n} \cdot (\widehat{\alpha} - \alpha^*) - \sqrt{n} \cdot \big[\nabla \bar{S}_n(\widehat{\boldsymbol{\beta}}, \widehat{\mathbf{w}})\big]_\alpha^{-1} \cdot \bar{S}_n(\widehat{\boldsymbol{\beta}}, \widehat{\mathbf{w}}) \qquad \text{(C.58)} \\ &= \underbrace{-\sqrt{n} \cdot \big[\nabla \bar{S}_n(\widehat{\boldsymbol{\beta}}, \widehat{\mathbf{w}})\big]_\alpha^{-1} \cdot \bar{S}_n(\widetilde{\boldsymbol{\beta}}, \widehat{\mathbf{w}})}_{\text{(i)}} + \underbrace{\sqrt{n} \cdot (\widehat{\alpha} - \alpha^*) \cdot \Big\{1 - \big[\nabla \bar{S}_n(\widehat{\boldsymbol{\beta}}, \widehat{\mathbf{w}})\big]_\alpha^{-1} \cdot \big[\nabla \bar{S}_n(\boldsymbol{\beta}^\sharp, \widehat{\mathbf{w}})\big]_\alpha\Big\}}_{\text{(ii)}}. \end{aligned}$$

**Analysis of Term (i):** For term (i) in (C.58), in the sequel we first prove

$$\big[\nabla \bar{S}_n(\widehat{\boldsymbol{\beta}}, \widehat{\mathbf{w}})\big]_\alpha = -\big[I(\boldsymbol{\beta}^*)\big]_{\alpha|\boldsymbol{\gamma}} + o_{\mathbb{P}}(1). \tag{C.59}$$

By the definition of $\big[I(\boldsymbol{\beta}^*)\big]_{\alpha|\boldsymbol{\gamma}}$ in (4.1) and the definition of $\mathbf{w}^*$ in (C.53), we have

$$\big[I(\boldsymbol{\beta}^*)\big]_{\alpha|\boldsymbol{\gamma}} = \big[I(\boldsymbol{\beta}^*)\big]_{\alpha,\alpha} - \big[I(\boldsymbol{\beta}^*)\big]_{\boldsymbol{\gamma},\alpha}^\top \cdot \big[I(\boldsymbol{\beta}^*)\big]_{\boldsymbol{\gamma},\boldsymbol{\gamma}}^{-1} \cdot \big[I(\boldsymbol{\beta}^*)\big]_{\boldsymbol{\gamma},\alpha} = \big[I(\boldsymbol{\beta}^*)\big]_{\alpha,\alpha} - (\mathbf{w}^*)^\top \cdot \big[I(\boldsymbol{\beta}^*)\big]_{\boldsymbol{\gamma},\alpha}.$$

Together with (C.57), by triangle inequality we obtain

$$\begin{aligned}&\Big|\big[\nabla \bar{S}_n(\widehat{\boldsymbol{\beta}}, \widehat{\mathbf{w}})\big]_\alpha + \big[I(\boldsymbol{\beta}^*)\big]_{\alpha|\boldsymbol{\gamma}}\Big| \\ &\quad \leq \underbrace{\Big|\big[T_n(\widehat{\boldsymbol{\beta}})\big]_{\alpha,\alpha} + \big[I(\boldsymbol{\beta}^*)\big]_{\alpha,\alpha}\Big|}_{\text{(i).a}} + \underbrace{\Big|\widehat{\mathbf{w}}^\top \cdot \big[T_n(\widehat{\boldsymbol{\beta}})\big]_{\boldsymbol{\gamma},\alpha} + (\mathbf{w}^*)^\top \cdot \big[I(\boldsymbol{\beta}^*)\big]_{\boldsymbol{\gamma},\alpha}\Big|}_{\text{(i).b}}. \end{aligned} \tag{C.60}$$

Note that term (i).a in (C.60) is upper bounded by

$$\Big\|T_n(\widehat{\boldsymbol{\beta}}) + I(\boldsymbol{\beta}^*)\Big\|_{\infty,\infty} \leq \Big\|T_n(\widehat{\boldsymbol{\beta}}) - T_n(\boldsymbol{\beta}^*)\Big\|_{\infty,\infty} + \big\|T_n(\boldsymbol{\beta}^*) - \mathbb{E}_{\boldsymbol{\beta}^*}\big[T_n(\boldsymbol{\beta}^*)\big]\big\|_{\infty,\infty}. \tag{C.61}$$

Here the inequality is from Lemma 2.1 and triangle inequality. For the first term on the right-hand side of (C.61), by Conditions 4.1 and 4.4 we have

$$\Big\|T_n(\widehat{\boldsymbol{\beta}}) - T_n(\boldsymbol{\beta}^*)\Big\|_{\infty,\infty} = O_{\mathbb{P}}(\zeta^{\mathrm{L}}) \cdot \big\|\widehat{\boldsymbol{\beta}} - \boldsymbol{\beta}^*\big\|_1 = O_{\mathbb{P}}(\zeta^{\mathrm{L}} \cdot \zeta^{\mathrm{EM}}). \tag{C.62}$$

For the second term on the right-hand side of (C.61), by Condition 4.3 we have

$$\big\|T_n(\boldsymbol{\beta}^*) - \mathbb{E}_{\boldsymbol{\beta}^*}\big[T_n(\boldsymbol{\beta}^*)\big]\big\|_{\infty,\infty} = O_{\mathbb{P}}(\zeta^{\mathrm{T}}). \tag{C.63}$$

Plugging (C.62) and (C.63) into (C.61), we obtain

$$\Big\|T_n\big(\widehat{\boldsymbol{\beta}}\big)+I(\boldsymbol{\beta}^*)\Big\|_{\infty,\infty}=O_{\mathbb{P}}\big(\zeta^{\mathrm{L}}\cdot\zeta^{\mathrm{EM}}+\zeta^{\mathrm{T}}\big). \tag{C.64}$$

By (4.6) in Assumption 4.5, we have

$$\zeta^{\mathrm{L}}\cdot\zeta^{\mathrm{EM}}+\zeta^{\mathrm{T}}\leq\max\big\{\|\mathbf{w}^*\|_1,\ 1\big\}\cdot s^*_{\mathbf{w}}\cdot\lambda=o(1)$$

for $\lambda$ specified in (4.5). Hence, we conclude that term (i).a in (C.60) is $o_{\mathbb{P}}(1)$. Meanwhile, for term (i).b in (C.60), by triangle inequality we have

$$\begin{aligned}&\Big|\widehat{\mathbf{w}}^\top\cdot\big[T_n\big(\widehat{\boldsymbol{\beta}}\big)\big]_{\boldsymbol{\gamma},\alpha}+(\mathbf{w}^*)^\top\cdot\big[I(\boldsymbol{\beta}^*)\big]_{\boldsymbol{\gamma},\alpha}\Big| \\ &\quad\leq\Big|(\widehat{\mathbf{w}}-\mathbf{w}^*)^\top\cdot\big[T_n\big(\widehat{\boldsymbol{\beta}}\big)+I(\boldsymbol{\beta}^*)\big]_{\boldsymbol{\gamma},\alpha}\Big|+\Big|(\mathbf{w}^*-\widehat{\mathbf{w}})^\top\cdot\big[I(\boldsymbol{\beta}^*)\big]_{\boldsymbol{\gamma},\alpha}\Big|+\Big|(\mathbf{w}^*)^\top\cdot\big[T_n\big(\widehat{\boldsymbol{\beta}}\big)+I(\boldsymbol{\beta}^*)\big]_{\boldsymbol{\gamma},\alpha}\Big|.\end{aligned} \tag{C.65}$$

For the first term on the right hand-side of (C.65), we have

$$\begin{aligned}\Big|(\widehat{\mathbf{w}}-\mathbf{w}^*)^\top\cdot\big[T_n\big(\widehat{\boldsymbol{\beta}}\big)+I(\boldsymbol{\beta}^*)\big]_{\boldsymbol{\gamma},\alpha}\Big|&\leq\|\widehat{\mathbf{w}}-\mathbf{w}^*\|_1\cdot\Big\|T_n\big(\widehat{\boldsymbol{\beta}}\big)+I(\boldsymbol{\beta}^*)\Big\|_{\infty,\infty}\\&=O_{\mathbb{P}}\big[s^*_{\mathbf{w}}\cdot\lambda\cdot\big(\zeta^{\mathrm{T}}+\zeta^{\mathrm{L}}\cdot\zeta^{\mathrm{EM}}\big)\big]=o_{\mathbb{P}}(1).\end{aligned}$$

The inequality is from Hölder's inequality, the second last equality is from Lemma C.3 and (C.64), and the last equality holds because (4.6) in Assumption 4.5 implies

$$\max\big\{s^*_{\mathbf{w}}\cdot\lambda,\ \zeta^{\mathrm{T}}+\zeta^{\mathrm{L}}\cdot\zeta^{\mathrm{EM}}\big\}\leq\max\big\{\|\mathbf{w}^*\|_1,\ 1\big\}\cdot s^*_{\mathbf{w}}\cdot\lambda=o(1)$$

for $\lambda$ specified in (4.5). For the second term on the right hand-side of (C.65), we have

$$\Big|(\mathbf{w}^*-\widehat{\mathbf{w}})^\top\cdot\big[I(\boldsymbol{\beta}^*)\big]_{\boldsymbol{\gamma},\alpha}\Big|\leq\|\widehat{\mathbf{w}}-\mathbf{w}^*\|_1\cdot\big\|I(\boldsymbol{\beta}^*)\big\|_{\infty,\infty}\leq O_{\mathbb{P}}(s^*_{\mathbf{w}}\cdot\lambda)\cdot\big\|I(\boldsymbol{\beta}^*)\big\|_2=o_{\mathbb{P}}(1),$$

where the second inequality is from Lemma C.3, while the last equality follows from (4.4) and (4.6) in Assumption 4.5. For the third term on the right hand-side of (C.65), we have

$$\Big|(\mathbf{w}^*)^\top\cdot\big[T_n\big(\widehat{\boldsymbol{\beta}}\big)+I(\boldsymbol{\beta}^*)\big]_{\boldsymbol{\gamma},\alpha}\Big|\leq\|\mathbf{w}^*\|_1\cdot\Big\|T_n\big(\widehat{\boldsymbol{\beta}}\big)+I(\boldsymbol{\beta}^*)\Big\|_{\infty,\infty}=\|\mathbf{w}^*\|_1\cdot O_{\mathbb{P}}\big(\zeta^{\mathrm{L}}\cdot\zeta^{\mathrm{EM}}+\zeta^{\mathrm{T}}\big)=o_{\mathbb{P}}(1).$$

Here the inequality is from Hölder's inequality, the first equality is from (C.64) and the last equality holds because (4.6) in Assumption 4.5 implies

$$\|\mathbf{w}^*\|_1\cdot\big(\zeta^{\mathrm{L}}\cdot\zeta^{\mathrm{EM}}+\zeta^{\mathrm{T}}\big)\leq\max\big\{\|\mathbf{w}^*\|_1,\ 1\big\}\cdot s^*_{\mathbf{w}}\cdot\lambda=o(1)$$

for $\lambda$ specified in (4.5). By (C.65), we conclude that term (i).b in (C.60) is $o_{\mathbb{P}}(1)$. Hence, we obtain (C.59). Furthermore, for term (i) in (C.58), we then replace $\widehat{\boldsymbol{\beta}}_0=\big(0,\widehat{\boldsymbol{\gamma}}^\top\big)^\top$ with $\widetilde{\boldsymbol{\beta}}=\big(\alpha^*,\widehat{\boldsymbol{\gamma}}^\top\big)^\top$ and $\widehat{\mathbf{w}}_0=w\big(\widehat{\boldsymbol{\beta}}_0,\lambda\big)$ with $\widehat{\mathbf{w}}=w\big(\widehat{\boldsymbol{\beta}},\lambda\big)$ in the proof of (C.29) in §C.3. We obtain

$$\sqrt{n}\cdot\overline{S}_n\big(\widetilde{\boldsymbol{\beta}},\widehat{\mathbf{w}}\big)=\sqrt{n}\cdot\big[\nabla_1Q_n(\boldsymbol{\beta}^*;\boldsymbol{\beta}^*)\big]_\alpha-\sqrt{n}\cdot(\mathbf{w}^*)^\top\cdot\big[\nabla_1Q_n(\boldsymbol{\beta}^*;\boldsymbol{\beta}^*)\big]_{\boldsymbol{\gamma}}+o_{\mathbb{P}}(1),$$

which further implies

$$
\begin{aligned}
&-\sqrt{n}\cdot\big[\nabla\overline{S}_n(\widehat{\boldsymbol{\beta}},\widehat{\mathbf{w}})\big]_{\alpha}^{-1}\cdot\overline{S}_n(\widetilde{\boldsymbol{\beta}},\widehat{\mathbf{w}})\\
&\quad=-\sqrt{n}\cdot\big[I(\boldsymbol{\beta}^*)\big]_{\alpha|\boldsymbol{\gamma}}^{-1}\cdot\Big\{\big[\nabla_1 Q_n(\boldsymbol{\beta}^*;\boldsymbol{\beta}^*)\big]_{\alpha}-(\mathbf{w}^*)^\top\cdot\big[\nabla_1 Q_n(\boldsymbol{\beta}^*;\boldsymbol{\beta}^*)\big]_{\boldsymbol{\gamma}}\Big\}+o_{\mathbb{P}}(1)
\end{aligned}
$$

by (C.59) and Slutsky's theorem.

**Analysis of Term (ii):** Now we prove that term (ii) in (C.58) is $o_{\mathbb{P}}(1)$. We have

$$
\begin{aligned}
&\sqrt{n}\cdot(\widehat{\alpha}-\alpha^*)\cdot\Big\{1-\big[\nabla\overline{S}_n(\widehat{\boldsymbol{\beta}},\widehat{\mathbf{w}})\big]_{\alpha}^{-1}\cdot\big[\nabla\overline{S}_n(\boldsymbol{\beta}^\sharp,\widehat{\mathbf{w}})\big]_{\alpha}\Big\}\\
&\quad\le\sqrt{n}\cdot\underbrace{|\widehat{\alpha}-\alpha^*|}_{\text{(ii).a}}\cdot\underbrace{\Big|1-\big[\nabla\overline{S}_n(\widehat{\boldsymbol{\beta}},\widehat{\mathbf{w}})\big]_{\alpha}^{-1}\cdot\big[\nabla\overline{S}_n(\boldsymbol{\beta}^\sharp,\widehat{\mathbf{w}})\big]_{\alpha}\Big|}_{\text{(ii).b}}.
\end{aligned}
\tag{C.66}
$$

For term (ii).a in (C.66), by Condition 4.1 and (4.6) in Assumption 4.5 we have

$$
|\widehat{\alpha}-\alpha^*|\le\big\|\widehat{\boldsymbol{\beta}}-\boldsymbol{\beta}^*\big\|_1=O_{\mathbb{P}}\big(\zeta^{\mathrm{EM}}\big)=o_{\mathbb{P}}(1).
\tag{C.67}
$$

Meanwhile, by replacing $\widehat{\boldsymbol{\beta}}$ with $\widetilde{\boldsymbol{\beta}}=\big(\alpha^*,\widehat{\boldsymbol{\gamma}}^\top\big)^\top$ in the proof of (C.59) we obtain

$$
\big[\nabla\overline{S}_n(\widetilde{\boldsymbol{\beta}},\widehat{\mathbf{w}})\big]_{\alpha}=-\big[I(\boldsymbol{\beta}^*)\big]_{\alpha|\boldsymbol{\gamma}}+o_{\mathbb{P}}(1).
\tag{C.68}
$$

Combining (C.59) and (C.68), for term (ii).b in (C.66) we obtain

$$
\Big|1-\big[\nabla\overline{S}_n(\widehat{\boldsymbol{\beta}},\widehat{\mathbf{w}})\big]_{\alpha}^{-1}\cdot\big[\nabla\overline{S}_n(\boldsymbol{\beta}^\sharp,\widehat{\mathbf{w}})\big]_{\alpha}\Big|=o_{\mathbb{P}}(1),
$$

which together with (C.67) implies that term (ii) in (C.58) is $o_{\mathbb{P}}(1)$. Plugging the above results into terms (i) and (ii) in (C.58), we obtain (C.53).

**Proof of (C.54):** By (C.30) in §C.3, we have

$$
\sqrt{n}\cdot\big[\nabla_1 Q_n(\boldsymbol{\beta}^*;\boldsymbol{\beta}^*)\big]_{\alpha}-\sqrt{n}\cdot(\mathbf{w}^*)^\top\cdot\big[\nabla_1 Q_n(\boldsymbol{\beta}^*;\boldsymbol{\beta}^*)\big]_{\boldsymbol{\gamma}}\xrightarrow{D}N\big(0,\big[I(\boldsymbol{\beta}^*)\big]_{\alpha|\boldsymbol{\gamma}}\big),
$$

which implies (C.54). Finally, combining (C.53) and (C.54) with Slutsky's theorem, we obtain

$$
\sqrt{n}\cdot\big[\overline{\alpha}(\widehat{\boldsymbol{\beta}},\lambda)-\alpha^*\big]\xrightarrow{D}N\big(0,\big[I(\boldsymbol{\beta}^*)\big]_{\alpha|\boldsymbol{\gamma}}^{-1}\big).
$$

Thus we conclude the proof of Lemma 5.5. □

## C.6 Proof of Lemma 4.8

*Proof.* According to Algorithm 4, the final estimator $\widehat{\boldsymbol{\beta}}=\boldsymbol{\beta}^{(T)}$ has $\widehat{s}$ nonzero entries. Meanwhile, we have $\|\boldsymbol{\beta}^*\|_0=s^*\le\widehat{s}$. Hence, we have $\big\|\widehat{\boldsymbol{\beta}}-\boldsymbol{\beta}^*\big\|_1\le 2\cdot\sqrt{\widehat{s}}\cdot\big\|\widehat{\boldsymbol{\beta}}-\boldsymbol{\beta}^*\big\|_2$. Invoking (3.19) in Theorem 3.7, we obtain $\zeta^{\mathrm{EM}}$.

For Gaussian mixture model, the maximization implementation of the M-step takes the form

$$
M_n(\boldsymbol{\beta})=\frac{2}{n}\sum_{i=1}^n\omega_{\boldsymbol{\beta}}(\mathbf{y}_i)\cdot\mathbf{y}_i-\frac{1}{n}\sum_{i=1}^n\mathbf{y}_i,\quad\text{and}\quad M(\boldsymbol{\beta})=\mathbb{E}\big[2\cdot\omega_{\boldsymbol{\beta}}(\boldsymbol{Y})\cdot\boldsymbol{Y}-\boldsymbol{Y}\big],
$$

where $\omega_{\boldsymbol{\beta}}(\cdot)$ is defined in (A.1). Meanwhile, we have

$$\nabla_1 Q_n(\boldsymbol{\beta};\boldsymbol{\beta}) = \frac{1}{n}\sum_{i=1}^{n}\bigl[2\cdot\omega_{\boldsymbol{\beta}}(\mathbf{y}_i)-1\bigr]\cdot\mathbf{y}_i-\boldsymbol{\beta}, \quad \text{and} \quad \nabla_1 Q(\boldsymbol{\beta};\boldsymbol{\beta}) = \mathbb{E}\bigl[2\cdot\omega_{\boldsymbol{\beta}}(\boldsymbol{Y})-\boldsymbol{Y}\bigr]-\boldsymbol{\beta}.$$

Hence, we have $\bigl\|M_n(\boldsymbol{\beta})-M(\boldsymbol{\beta})\bigr\|_\infty = \bigl\|\nabla_1 Q_n(\boldsymbol{\beta};\boldsymbol{\beta})-\nabla_1 Q(\boldsymbol{\beta};\boldsymbol{\beta})\bigr\|_\infty$. By setting $\delta = 2/d$ in Lemma 3.6, we obtain $\zeta^{\mathrm{G}}$. □

## C.7 Proof of Lemma 4.9

*Proof.* Recall that for Gaussian mixture model we have

$$Q_n(\boldsymbol{\beta}';\boldsymbol{\beta}) = -\frac{1}{2n}\sum_{i=1}^{n}\Bigl\{\omega_{\boldsymbol{\beta}}(\mathbf{y}_i)\cdot\|\mathbf{y}_i-\boldsymbol{\beta}'\|_2^2+\bigl[1-\omega_{\boldsymbol{\beta}}(\mathbf{y}_i)\bigr]\cdot\|\mathbf{y}_i+\boldsymbol{\beta}'\|_2^2\Bigr\},$$

where $\omega_{\boldsymbol{\beta}}(\cdot)$ is defined in (A.1). Hence, by calculation we have

$$\nabla_1 Q_n(\boldsymbol{\beta}';\boldsymbol{\beta}) = \frac{1}{n}\sum_{i=1}^{n}\bigl[2\cdot\omega_{\boldsymbol{\beta}}(\mathbf{y}_i)-1\bigr]\cdot\mathbf{y}_i-\boldsymbol{\beta}', \quad \nabla_{1,1}^2 Q_n(\boldsymbol{\beta}';\boldsymbol{\beta}) = -\mathbf{I}_d, \tag{C.69}$$

$$\nabla_{1,2}^2 Q_n(\boldsymbol{\beta}';\boldsymbol{\beta}) = \frac{4}{n}\sum_{i=1}^{n}\frac{\mathbf{y}_i\cdot\mathbf{y}_i^\top}{\sigma^2\cdot\bigl[1+\exp\bigl(-2\cdot\langle\boldsymbol{\beta},\mathbf{y}_i\rangle/\sigma^2\bigr)\bigr]\cdot\bigl[1+\exp\bigl(2\cdot\langle\boldsymbol{\beta},\mathbf{y}_i\rangle/\sigma^2\bigr)\bigr]}. \tag{C.70}$$

For notational simplicity we define

$$\nu_{\boldsymbol{\beta}}(\mathbf{y}) = \frac{4}{\sigma^2\cdot\bigl[1+\exp\bigl(-2\cdot\langle\boldsymbol{\beta},\mathbf{y}\rangle/\sigma^2\bigr)\bigr]\cdot\bigl[1+\exp\bigl(2\cdot\langle\boldsymbol{\beta},\mathbf{y}\rangle/\sigma^2\bigr)\bigr]}. \tag{C.71}$$

Then by the definition of $T_n(\cdot)$ in (2.8), from (C.69) and (C.70) we have

$$\bigl\{T_n(\boldsymbol{\beta}^*)-\mathbb{E}_{\boldsymbol{\beta}^*}\bigl[T_n(\boldsymbol{\beta}^*)\bigr]\bigr\}_{j,k} = \frac{1}{n}\sum_{i=1}^{n}\nu_{\boldsymbol{\beta}^*}(\mathbf{y}_i)\cdot y_{i,j}\cdot y_{i,k}-\mathbb{E}_{\boldsymbol{\beta}^*}\bigl[\nu_{\boldsymbol{\beta}^*}(\boldsymbol{Y})\cdot Y_j\cdot Y_k\bigr].$$

Applying the symmetrization result in Lemma D.4 to the right-hand side, we have that for $u>0$,

$$\mathbb{E}_{\boldsymbol{\beta}^*}\Bigl\{\exp\Bigl[u\cdot\Bigl|\bigl\{T_n(\boldsymbol{\beta}^*)-\mathbb{E}_{\boldsymbol{\beta}^*}\bigl[T_n(\boldsymbol{\beta}^*)\bigr]\bigr\}_{j,k}\Bigr|\Bigr]\Bigr\} \le \mathbb{E}_{\boldsymbol{\beta}^*}\Bigl\{\exp\Bigl[u\cdot\Bigl|\frac{1}{n}\sum_{i=1}^{n}\xi_i\cdot\nu_{\boldsymbol{\beta}^*}(\mathbf{y}_i)\cdot y_{i,j}\cdot y_{i,k}\Bigr|\Bigr]\Bigr\}, \tag{C.72}$$

where $\xi_1,\ldots,\xi_n$ are i.i.d. Rademacher random variables that are independent of $\mathbf{y}_1,\ldots,\mathbf{y}_n$. Then we invoke the contraction result in Lemma D.5 by setting

$$f(y_{i,j}\cdot y_{i,k}) = y_{i,j}\cdot y_{i,k}, \quad \mathcal{F}=\{f\}, \quad \psi_i(v)=\nu_{\boldsymbol{\beta}^*}(\mathbf{y}_i)\cdot v, \quad \text{and} \quad \phi(v)=\exp(u\cdot v),$$

where $u$ is the variable of the moment generating function in (C.72). By the definition in (C.71) we have $\bigl|\nu_{\boldsymbol{\beta}^*}(\mathbf{y}_i)\bigr|\le 4/\sigma^2$, which implies

$$\bigl|\psi_i(v)-\psi_i(v')\bigr| \le \bigl|\nu_{\boldsymbol{\beta}^*}(\mathbf{y}_i)\cdot(v-v')\bigr| \le 4/\sigma^2\cdot|v-v'|, \quad \text{for all } v,v'\in\mathbb{R}.$$

Therefore, applying the contraction result in Lemma D.5 to the right-hand side of (C.72), we obtain

$$\mathbb{E}_{\boldsymbol{\beta}^*}\left\{\exp\left[u\cdot\left|\left\{T_n(\boldsymbol{\beta}^*)-\mathbb{E}_{\boldsymbol{\beta}^*}\left[T_n(\boldsymbol{\beta}^*)\right]\right\}_{j,k}\right|\right]\right\}\leq\mathbb{E}_{\boldsymbol{\beta}^*}\left\{\exp\left[u\cdot 4/\sigma^2\cdot\left|\frac{1}{n}\sum_{i=1}^n\xi_i\cdot y_{i,j}\cdot y_{i,k}\right|\right]\right\}. \quad \text{(C.73)}$$

Note that $\mathbb{E}_{\boldsymbol{\beta}^*}(\xi_i\cdot y_{i,j}\cdot y_{i,k})=0$, since $\xi_i$ is a Rademacher random variable independent of $y_{i,j}\cdot y_{i,k}$. Recall that in Gaussian mixture model we have $y_{i,j}=z_i\cdot\beta_j^*+v_{i,j}$, where $z_i$ is a Rademacher random variable and $v_{i,j}\sim N(0,\sigma^2)$. Hence, by Example 5.8 in Vershynin (2010), we have $\|z_i\cdot\beta_j^*\|_{\psi_2}\leq|\beta_j^*|$ and $\|v_{i,j}\|_{\psi_2}\leq C\cdot\sigma$. Therefore, by Lemma D.1 we have

$$\|y_{i,j}\|_{\psi_2}=\left\|z_i\cdot\beta_j^*+v_{i,j}\right\|_{\psi_2}\leq C'\cdot\sqrt{|\beta_j^*|^2+C''\cdot\sigma^2}\leq C'\cdot\sqrt{\|\boldsymbol{\beta}^*\|_\infty^2+C''\cdot\sigma^2}. \quad \text{(C.74)}$$

Since $|\xi_i\cdot y_{i,j}\cdot y_{i,k}|=|y_{i,j}\cdot y_{i,k}|$, by definition $\xi_i\cdot y_{i,j}\cdot y_{i,k}$ and $y_{i,j}\cdot y_{i,k}$ have the same $\psi_1$-norm. By Lemma D.2 we have

$$\|\xi_i\cdot y_{i,j}\cdot y_{i,k}\|_{\psi_1}\leq C\cdot\max\left\{\|y_{i,j}\|_{\psi_2}^2,\ \|y_{i,k}\|_{\psi_2}^2\right\}\leq C'\cdot\left(\|\boldsymbol{\beta}^*\|_\infty^2+C''\cdot\sigma^2\right).$$

Then by Lemma 5.15 in Vershynin (2010), we obtain

$$\mathbb{E}_{\boldsymbol{\beta}^*}\left[\exp(u'\cdot\xi_i\cdot y_{i,j}\cdot y_{i,k})\right]\leq\exp\left[(u')^2\cdot C\cdot\left(\|\boldsymbol{\beta}^*\|_\infty^2+C'\cdot\sigma^2\right)\right] \quad \text{(C.75)}$$

for all $|u'|\leq C''/\left(\|\boldsymbol{\beta}^*\|_\infty^2+C'\cdot\sigma^2\right)$. Note that on the right-hand side of (C.73), we have

$$\begin{aligned}
&\mathbb{E}_{\boldsymbol{\beta}^*}\left\{\exp\left[u\cdot 4/\sigma^2\cdot\left|\frac{1}{n}\sum_{i=1}^n\xi_i\cdot y_{i,j}\cdot y_{i,k}\right|\right]\right\}\\
&\quad\leq\mathbb{E}_{\boldsymbol{\beta}^*}\left(\max\left\{\exp\left[u\cdot 4/\sigma^2\cdot\frac{1}{n}\sum_{i=1}^n\xi_i\cdot y_{i,j}\cdot y_{i,k}\right],\ \exp\left[-u\cdot 4/\sigma^2\cdot\frac{1}{n}\sum_{i=1}^n\xi_i\cdot y_{i,j}\cdot y_{i,k}\right]\right\}\right)\\
&\quad\leq\mathbb{E}_{\boldsymbol{\beta}^*}\left\{\exp\left[u\cdot 4/\sigma^2\cdot\frac{1}{n}\sum_{i=1}^n\xi_i\cdot y_{i,j}\cdot y_{i,k}\right]\right\}+\mathbb{E}_{\boldsymbol{\beta}^*}\left\{\exp\left[-u\cdot 4/\sigma^2\cdot\frac{1}{n}\sum_{i=1}^n\xi_i\cdot y_{i,j}\cdot y_{i,k}\right]\right\}. \quad \text{(C.76)}
\end{aligned}$$

By plugging (C.75) into the right-hand side of (C.76) with $u'=u\cdot 4/(\sigma^2\cdot n)$ and $u'=-u\cdot 4/(\sigma^2\cdot n)$, from (C.73) we have that for any $j,k\in\{1,\ldots,d\}$,

$$\mathbb{E}_{\boldsymbol{\beta}^*}\left\{\exp\left[u\cdot\left|\left\{T_n(\boldsymbol{\beta}^*)-\mathbb{E}_{\boldsymbol{\beta}^*}\left[T_n(\boldsymbol{\beta}^*)\right]\right\}_{j,k}\right|\right]\right\}\leq 2\cdot\exp\left[C\cdot u^2/n\cdot\left(\|\boldsymbol{\beta}^*\|_\infty^2+C'\cdot\sigma^2\right)^2/\sigma^4\right]. \quad \text{(C.77)}$$

Therefore, by Chernoff bound we have that, for all $v>0$ and $|u|\leq C''/\left(\|\boldsymbol{\beta}^*\|_\infty^2+C'\cdot\sigma^2\right)$,

$$\begin{aligned}
\mathbb{P}\left[\left\|T_n(\boldsymbol{\beta}^*)-\mathbb{E}_{\boldsymbol{\beta}^*}\left[T_n(\boldsymbol{\beta}^*)\right]\right\|_{\infty,\infty}>v\right]&\leq\mathbb{E}_{\boldsymbol{\beta}^*}\left\{\exp\left[u\cdot\left\|T_n(\boldsymbol{\beta}^*)-\mathbb{E}_{\boldsymbol{\beta}^*}\left[T_n(\boldsymbol{\beta}^*)\right]\right\|_{\infty,\infty}\right]\right\}\Big/\exp(u\cdot v)\\
&\leq\sum_{j=1}^d\sum_{k=1}^d\mathbb{E}_{\boldsymbol{\beta}^*}\left\{\exp\left[u\cdot\left|\left\{T_n(\boldsymbol{\beta}^*)-\mathbb{E}_{\boldsymbol{\beta}^*}\left[T_n(\boldsymbol{\beta}^*)\right]\right\}_{j,k}\right|\right]\right\}\Big/\exp(u\cdot v)\\
&\leq 2\cdot\exp\left[C\cdot u^2/n\cdot\left(\|\boldsymbol{\beta}^*\|_\infty^2+C'\cdot\sigma^2\right)^2/\sigma^4-u\cdot v+2\cdot\log d\right],
\end{aligned}$$

where the last inequality is from (C.77). By minimizing its right-hand side over $u$, we conclude that for $0 < v \le C'' \cdot (\|\boldsymbol{\beta}^*\|_\infty^2 + C' \cdot \sigma^2)$,

$$\mathbb{P}\Big[\big\|T_n(\boldsymbol{\beta}^*) - \mathbb{E}_{\boldsymbol{\beta}^*}\big[T_n(\boldsymbol{\beta}^*)\big]\big\|_{\infty,\infty} > v\Big] \le 2 \cdot \exp\Big\{-n \cdot v^2 \Big/ \Big[C \cdot \big(\|\boldsymbol{\beta}^*\|_\infty^2 + C' \cdot \sigma^2\big)^2 / \sigma^4\Big] + 2 \cdot \log d\Big\}.$$

Setting the right-hand side to be $\delta$, we have that

$$\big\|T_n(\boldsymbol{\beta}^*) - \mathbb{E}_{\boldsymbol{\beta}^*}\big[T_n(\boldsymbol{\beta}^*)\big]\big\|_{\infty,\infty} \le v = C \cdot \big(\|\boldsymbol{\beta}^*\|_\infty^2 + C' \cdot \sigma^2\big)/\sigma^2 \cdot \sqrt{\frac{\log(2/\delta) + 2 \cdot \log d}{n}}$$

holds with probability at least $1 - \delta$. By setting $\delta = 2/d$, we conclude the proof of Lemma 4.9. □

## C.8 Proof of Lemma 4.10

*Proof.* For any $j, k \in \{1, \ldots, d\}$, by the mean-value theorem we have

$$\begin{aligned}\big\|T_n(\boldsymbol{\beta}) - T_n(\boldsymbol{\beta}^*)\big\|_{\infty,\infty} &= \max_{j,k \in \{1,\ldots,d\}} \Big|\big[T_n(\boldsymbol{\beta})\big]_{j,k} - \big[T_n(\boldsymbol{\beta}^*)\big]_{j,k}\Big| \\ &= \max_{j,k \in \{1,\ldots,d\}} \Big|(\boldsymbol{\beta} - \boldsymbol{\beta}^*)^\top \cdot \nabla\big[T_n(\boldsymbol{\beta}^\sharp)\big]_{j,k}\Big| \le \|\boldsymbol{\beta} - \boldsymbol{\beta}^*\|_1 \cdot \max_{j,k \in \{1,\ldots,d\}} \Big\|\nabla\big[T_n(\boldsymbol{\beta}^\sharp)\big]_{j,k}\Big\|_\infty,\end{aligned} \tag{C.78}$$

where $\boldsymbol{\beta}^\sharp$ is an intermediate value between $\boldsymbol{\beta}$ and $\boldsymbol{\beta}^*$. According to (C.69), (C.70) and the definition of $T_n(\cdot)$ in (2.8), by calculation we have

$$\nabla\big[T_n(\boldsymbol{\beta}^\sharp)\big]_{j,k} = \frac{1}{n} \sum_{i=1}^n \bar{\nu}_{\boldsymbol{\beta}^\sharp}(\mathbf{y}_i) \cdot y_{i,j} \cdot y_{i,k} \cdot \mathbf{y}_i,$$

where

$$\begin{aligned}\bar{\nu}_{\boldsymbol{\beta}}(\mathbf{y}) = &\frac{8/\sigma^4}{\big[1 + \exp\big(-2 \cdot \langle \boldsymbol{\beta}, \mathbf{y}\rangle/\sigma^2\big)\big] \cdot \big[1 + \exp\big(2 \cdot \langle \boldsymbol{\beta}, \mathbf{y}\rangle/\sigma^2\big)\big]^2} \\ &- \frac{8/\sigma^4}{\big[1 + \exp\big(-2 \cdot \langle \boldsymbol{\beta}, \mathbf{y}\rangle/\sigma^2\big)\big]^2 \cdot \big[1 + \exp\big(2 \cdot \langle \boldsymbol{\beta}, \mathbf{y}\rangle/\sigma^2\big)\big]}.\end{aligned} \tag{C.79}$$

For notational simplicity, we define the following event

$$\mathcal{E} = \big\{\|\mathbf{y}_i\|_\infty \le \tau, \quad \text{for all } i = 1, \ldots, n\big\},$$

where $\tau > 0$ will be specified later. By maximal inequality we have

$$\begin{aligned}\mathbb{P}\Big\{\Big\|\nabla\big[T_n(\boldsymbol{\beta}^\sharp)\big]_{j,k}\Big\|_\infty > v\Big\} &\le d \cdot \mathbb{P}\Big(\Big\{\Big|\nabla\big[T_n(\boldsymbol{\beta}^\sharp)\big]_{j,k}\Big|\Big\}_l > v\Big) \\ &= d \cdot \mathbb{P}\Bigg[\bigg|\frac{1}{n}\sum_{i=1}^n \bar{\nu}_{\boldsymbol{\beta}^\sharp}(\mathbf{y}_i) \cdot y_{i,j} \cdot y_{i,k} \cdot y_{i,l}\bigg| > v\Bigg].\end{aligned} \tag{C.80}$$

Let $\bar{\mathcal{E}}$ be the complement of $\mathcal{E}$. On the right-hand side of (C.80) we have

$$\mathbb{P}\Bigg[\bigg|\frac{1}{n}\sum_{i=1}^n \bar{\nu}_{\boldsymbol{\beta}^\sharp}(\mathbf{y}_i) \cdot y_{i,j} \cdot y_{i,k} \cdot y_{i,l}\bigg| > v\Bigg] = \underbrace{\mathbb{P}\Bigg[\bigg|\frac{1}{n}\sum_{i=1}^n \bar{\nu}_{\boldsymbol{\beta}^\sharp}(\mathbf{y}_i) \cdot y_{i,j} \cdot y_{i,k} \cdot y_{i,l}\bigg| > v, \mathcal{E}\Bigg]}_{\text{(i)}} + \underbrace{\mathbb{P}(\bar{\mathcal{E}})}_{\text{(ii)}}. \tag{C.81}$$

**Analysis of Term (i):** For term (i) in (C.81), we have

$$\begin{aligned}
\mathbb{P}\biggl[\biggl|\frac{1}{n}\sum_{i=1}^{n}\overline{\nu}_{\boldsymbol{\beta}^\sharp}(\mathbf{y}_i)\cdot y_{i,j}\cdot y_{i,k}\cdot y_{i,l}\biggr| > v, \mathcal{E}\biggr] &= \mathbb{P}\biggl[\biggl|\frac{1}{n}\sum_{i=1}^{n}\overline{\nu}_{\boldsymbol{\beta}^\sharp}(\mathbf{y}_i)\cdot y_{i,j}\cdot y_{i,k}\cdot y_{i,l}\cdot \mathbb{1}\bigl\{\|\mathbf{y}_i\|_\infty \le \tau\bigr\}\biggr| > v, \mathcal{E}\biggr] \\
&\le \mathbb{P}\biggl[\biggl|\frac{1}{n}\sum_{i=1}^{n}\overline{\nu}_{\boldsymbol{\beta}^\sharp}(\mathbf{y}_i)\cdot y_{i,j}\cdot y_{i,k}\cdot y_{i,l}\cdot \mathbb{1}\bigl\{\|\mathbf{y}_i\|_\infty \le \tau\bigr\}\biggr| > v\biggr] \\
&\le \sum_{i=1}^{n}\mathbb{P}\Bigl[\bigl|\overline{\nu}_{\boldsymbol{\beta}^\sharp}(\mathbf{y}_i)\cdot y_{i,j}\cdot y_{i,k}\cdot y_{i,l}\cdot \mathbb{1}\bigl\{\|\mathbf{y}_i\|_\infty \le \tau\bigr\}\bigr| > v\Bigr],
\end{aligned}$$

where the last inequality is from union bound. By (C.79) we have $|\overline{\nu}_{\boldsymbol{\beta}^\sharp}(\mathbf{y}_i)|\le 16/\sigma^4$. Thus we obtain

$$\mathbb{P}\Bigl[\bigl|\overline{\nu}_{\boldsymbol{\beta}^\sharp}(\mathbf{y}_i)\cdot y_{i,j}\cdot y_{i,k}\cdot y_{i,l}\cdot \mathbb{1}\bigl\{\|\mathbf{y}_i\|_\infty \le \tau\bigr\}\bigr| > v\Bigr] \le \mathbb{P}\Bigl[\bigl|y_{i,j}\cdot y_{i,k}\cdot y_{i,l}\cdot \mathbb{1}\bigl\{\|\mathbf{y}_i\|_\infty \le \tau\bigr\}\bigr| > \sigma^4/16\cdot v\Bigr].$$

Taking $v = 16\cdot\tau^3/\sigma^4$, we have that the right-hand side is zero and hence term (i) in (C.81) is zero.

**Analysis of Term (ii):** Meanwhile, for term (ii) in (C.81) by maximal inequality we have

$$\mathbb{P}(\overline{\mathcal{E}}) = \mathbb{P}\Bigl(\max_{i\in\{1,\ldots,n\}}\|\mathbf{y}_i\|_\infty > \tau\Bigr) \le n\cdot\mathbb{P}\bigl(\|\mathbf{y}_i\|_\infty > \tau\bigr) \le n\cdot d\cdot\mathbb{P}\bigl(|y_{i,j}| > \tau\bigr).$$

Furthermore, by (C.74) in the proof of Lemma 4.8, we have that $y_{i,j}$ is sub-Gaussian with $\|y_{i,j}\|_{\psi_2} = C\cdot\sqrt{\|\boldsymbol{\beta}^*\|_\infty^2 + C'\cdot\sigma^2}$. Therefore, by Lemma 5.5 in Vershynin (2010) we have

$$\mathbb{P}(\overline{\mathcal{E}}) \le n\cdot d\cdot\mathbb{P}\bigl(|y_{i,j}| > \tau\bigr) \le n\cdot d\cdot 2\cdot\exp\bigl[-C\cdot\tau^2/\bigl(\|\boldsymbol{\beta}^*\|_\infty^2 + C'\cdot\sigma^2\bigr)\bigr].$$

To ensure the right-hand side is upper bounded by $\delta$, we set $\tau$ to be

$$\tau = C\cdot\sqrt{\|\boldsymbol{\beta}^*\|_\infty^2 + C'\cdot\sigma^2}\cdot\sqrt{\log d + \log n + \log(2/\delta)}. \tag{C.82}$$

Finally, by (C.80), (C.81) and maximal inequality we have

$$\mathbb{P}\Bigl\{\max_{j,k\in\{1,\ldots,d\}}\Bigl\|\nabla\bigl[T_n(\boldsymbol{\beta}^\sharp)\bigr]_{j,k}\Bigr\|_\infty > v\Bigr\} \le d^2\cdot d\cdot\mathbb{P}\biggl[\biggl|\frac{1}{n}\sum_{i=1}^{n}\overline{\nu}_{\boldsymbol{\beta}^\sharp}(\mathbf{y}_i)\cdot y_{i,j}\cdot y_{i,k}\cdot y_{i,l}\biggr| > v\biggr] \le d^3\cdot\delta$$

for $v = 16\cdot\tau^3/\sigma^4$ with $\tau$ specified in (C.82). By setting $\delta = 2\cdot d^{-4}$ and plugging (C.82) into (C.78), we conclude the proof of Lemma 4.10. □

## C.9 Proof of Lemma 4.12

By the same proof of Lemma 4.8 in §C.6, we obtain $\zeta^{\mathrm{EM}}$ by invoking Theorem 3.10. To obtain $\zeta^{\mathrm{G}}$, note that for the gradient implementation of the M-step (Algorithm 3), we have

$$M_n(\boldsymbol{\beta}) = \boldsymbol{\beta} + \eta\cdot\nabla_1 Q_n(\boldsymbol{\beta};\boldsymbol{\beta}), \quad \text{and} \quad M(\boldsymbol{\beta}) = \boldsymbol{\beta} + \eta\cdot\nabla_1 Q(\boldsymbol{\beta};\boldsymbol{\beta}).$$

Hence, we obtain $\bigl\|\nabla_1 Q_n(\boldsymbol{\beta}^*;\boldsymbol{\beta}^*) - \nabla_1 Q(\boldsymbol{\beta}^*;\boldsymbol{\beta}^*)\bigr\|_\infty = 1/\eta\cdot\bigl\|M_n(\boldsymbol{\beta}^*) - M(\boldsymbol{\beta}^*)\bigr\|_\infty$. Setting $\delta = 4/d$ and $s = s^*$ in (3.22) of Lemma 3.9, we then obtain $\zeta^{\mathrm{G}}$.

### C.10 Proof of Lemma 4.13

*Proof.* Recall that for mixture of regression model, we have

$$Q_n(\boldsymbol{\beta}';\boldsymbol{\beta}) = -\frac{1}{2n}\sum_{i=1}^{n}\Bigl\{\omega_{\boldsymbol{\beta}}(\mathbf{x}_i,y_i)\cdot\bigl(y_i-\langle\mathbf{x}_i,\boldsymbol{\beta}'\rangle\bigr)^2+\bigl[1-\omega_{\boldsymbol{\beta}}(\mathbf{x}_i,y_i)\bigr]\cdot\bigl(y_i+\langle\mathbf{x}_i,\boldsymbol{\beta}'\rangle\bigr)^2\Bigr\},$$

where $\omega_{\boldsymbol{\beta}}(\cdot)$ is defined in (A.3). Hence, by calculation we have

$$\nabla_1 Q_n(\boldsymbol{\beta}',\boldsymbol{\beta})=\frac{1}{n}\sum_{i=1}^{n}\bigl[2\cdot\omega_{\boldsymbol{\beta}}(\mathbf{x}_i,y_i)\cdot y_i\cdot\mathbf{x}_i-\mathbf{x}_i\cdot\mathbf{x}_i^\top\cdot\boldsymbol{\beta}'\bigr],\quad \nabla_{1,1}^2 Q_n(\boldsymbol{\beta}',\boldsymbol{\beta})=-\frac{1}{n}\sum_{i=1}^{n}\mathbf{x}_i\cdot\mathbf{x}_i^\top,\tag{C.83}$$

$$\nabla_{1,2}^2 Q_n(\boldsymbol{\beta}',\boldsymbol{\beta})=\frac{4}{n}\sum_{i=1}^{n}\frac{y_i^2\cdot\mathbf{x}_i\cdot\mathbf{x}_i^\top}{\sigma^2\cdot\bigl[1+\exp\bigl(-2\cdot y_i\cdot\langle\boldsymbol{\beta},\mathbf{x}_i\rangle/\sigma^2\bigr)\bigr]\cdot\bigl[1+\exp\bigl(2\cdot y_i\cdot\langle\boldsymbol{\beta},\mathbf{x}_i\rangle/\sigma^2\bigr)\bigr]}.\tag{C.84}$$

For notational simplicity we define

$$\nu_{\boldsymbol{\beta}}(\mathbf{x},y)=\frac{4}{\sigma^2\cdot\bigl[1+\exp\bigl(-2\cdot y\cdot\langle\boldsymbol{\beta},\mathbf{x}\rangle/\sigma^2\bigr)\bigr]\cdot\bigl[1+\exp\bigl(2\cdot y\cdot\langle\boldsymbol{\beta},\mathbf{x}\rangle/\sigma^2\bigr)\bigr]}.\tag{C.85}$$

Then by the definition of $T_n(\cdot)$ in (2.8), from (C.83) and (C.84) we have

$$\bigl\{T_n(\boldsymbol{\beta}^*)-\mathbb{E}_{\boldsymbol{\beta}^*}\bigl[T_n(\boldsymbol{\beta}^*)\bigr]\bigr\}_{j,k}=\frac{1}{n}\sum_{i=1}^{n}\nu_{\boldsymbol{\beta}^*}(\mathbf{x}_i,y_i)\cdot x_{i,j}\cdot x_{i,k}\cdot y_i^2-\mathbb{E}_{\boldsymbol{\beta}^*}\bigl[\nu_{\boldsymbol{\beta}^*}(Y,\boldsymbol{X})\cdot X_j\cdot X_k\cdot Y_i^2\bigr].$$

Applying the symmetrization result in Lemma D.4 to the right-hand side, we have that for $u>0$,

$$\mathbb{E}_{\boldsymbol{\beta}^*}\Bigl\{\exp\Bigl[u\cdot\Bigl|\bigl\{T_n(\boldsymbol{\beta}^*)-\mathbb{E}_{\boldsymbol{\beta}^*}\bigl[T_n(\boldsymbol{\beta}^*)\bigr]\bigr\}_{j,k}\Bigr|\Bigr]\Bigr\}\le\mathbb{E}_{\boldsymbol{\beta}^*}\Biggl\{\exp\Biggl[u\cdot\Biggl|\frac{1}{n}\sum_{i=1}^{n}\xi_i\cdot\nu_{\boldsymbol{\beta}^*}(\mathbf{x}_i,y_i)\cdot x_{i,j}\cdot x_{i,k}\cdot y_i^2\Biggr|\Biggr]\Biggr\},\tag{C.86}$$

where $\xi_1,\ldots,\xi_n$ are i.i.d. Rademacher random variables, which are independent of $\mathbf{x}_1,\ldots,\mathbf{x}_n$ and $y_1,\ldots,y_n$. Then we invoke the contraction result in Lemma D.5 by setting

$$f\bigl(x_{i,j}\cdot x_{i,k}\cdot y_i^2\bigr)=x_{i,j}\cdot x_{i,k}\cdot y_i^2,\quad \mathcal{F}=\{f\},\quad \psi_i(v)=\nu_{\boldsymbol{\beta}^*}(\mathbf{x}_i,y_i)\cdot v,\quad\text{and}\quad \phi(v)=\exp(u\cdot v),$$

where $u$ is the variable of the moment generating function in (C.86). By the definition in (C.85) we have $\bigl|\nu_{\boldsymbol{\beta}^*}(\mathbf{x}_i,y_i)\bigr|\le 4/\sigma^2$, which implies

$$\bigl|\psi_i(v)-\psi_i(v')\bigr|\le\bigl|\nu_{\boldsymbol{\beta}^*}(\mathbf{x}_i,y_i)\cdot(v-v')\bigr|\le 4/\sigma^2\cdot|v-v'|,\quad\text{for all } v,v'\in\mathbb{R}.$$

Therefore, applying Lemma D.5 to the right-hand side of (C.86), we obtain

$$\mathbb{E}_{\boldsymbol{\beta}^*}\Bigl\{\exp\Bigl[u\cdot\Bigl|\bigl\{T_n(\boldsymbol{\beta}^*)-\mathbb{E}_{\boldsymbol{\beta}^*}\bigl[T_n(\boldsymbol{\beta}^*)\bigr]\bigr\}_{j,k}\Bigr|\Bigr]\Bigr\}\le\mathbb{E}_{\boldsymbol{\beta}^*}\Biggl\{\exp\Biggl[u\cdot 4/\sigma^2\cdot\Biggl|\frac{1}{n}\sum_{i=1}^{n}\xi_i\cdot x_{i,j}\cdot x_{i,k}\cdot y_i^2\Biggr|\Biggr]\Biggr\}.\tag{C.87}$$

For notational simplicity, we define the following event

$$\mathcal{E} = \big\{\|\mathbf{x}_i\|_\infty \le \tau, \quad \text{for all } i = 1, \ldots, n\big\}.$$

Let $\bar{\mathcal{E}}$ be the complement of $\mathcal{E}$. We consider the following tail probability

$$\mathbb{P}\Bigg[4/\sigma^2 \cdot \bigg|\frac{1}{n}\sum_{i=1}^n \xi_i \cdot x_{i,j} \cdot x_{i,k} \cdot y_i^2\bigg| > v\Bigg] \le \underbrace{\mathbb{P}\Bigg[4/\sigma^2 \cdot \bigg|\frac{1}{n}\sum_{i=1}^n \xi_i \cdot x_{i,j} \cdot x_{i,k} \cdot y_i^2\bigg| > v, \mathcal{E}\Bigg]}_{\text{(i)}} + \underbrace{\mathbb{P}(\bar{\mathcal{E}})}_{\text{(ii)}}. \quad \text{(C.88)}$$

**Analysis of Term (i):** For term (i) in (C.88), we have

$$\begin{aligned}\mathbb{P}\Bigg[4/\sigma^2 \cdot \bigg|\frac{1}{n}\sum_{i=1}^n \xi_i \cdot x_{i,j} \cdot x_{i,k} \cdot y_i^2\bigg| > v, \mathcal{E}\Bigg] &= \mathbb{P}\Bigg[4/\sigma^2 \cdot \bigg|\frac{1}{n}\sum_{i=1}^n \xi_i \cdot x_{i,j} \cdot x_{i,k} \cdot y_i^2 \cdot \mathbb{1}\big\{\|\mathbf{x}_i\|_\infty \le \tau\big\}\bigg| > v, \mathcal{E}\Bigg] \\ &\le \mathbb{P}\Bigg[4/\sigma^2 \cdot \bigg|\frac{1}{n}\sum_{i=1}^n \xi_i \cdot x_{i,j} \cdot x_{i,k} \cdot y_i^2 \cdot \mathbb{1}\big\{\|\mathbf{x}_i\|_\infty \le \tau\big\}\bigg| > v\Bigg].\end{aligned}$$

Here note that $\mathbb{E}_{\boldsymbol{\beta}^*}\big(\xi_i \cdot x_{i,j} \cdot x_{i,k} \cdot y_i^2 \cdot \mathbb{1}\{\|\mathbf{x}_i\|_\infty \le \tau\}\big) = 0$, because $\xi_i$ is a Rademacher random variable independent of $\mathbf{x}_i$ and $y_i$. Recall that for mixture of regression model we have $y_i = z_i \cdot \langle\boldsymbol{\beta}^*, \mathbf{x}_i\rangle + v_i$, where $z_i$ is a Rademacher random variable, $\mathbf{x}_i \sim N(0, \mathbf{I}_d)$ and $v_i \sim N(0, \sigma^2)$. According to Example 5.8 in Vershynin (2010), we have $\big\|z_i \cdot \langle\boldsymbol{\beta}^*, \mathbf{x}_i\rangle \cdot \mathbb{1}\{\|\mathbf{x}_i\|_\infty \le \tau\}\big\|_{\psi_2} = \big\|\langle\boldsymbol{\beta}^*, \mathbf{x}_i\rangle \cdot \mathbb{1}\{\|\mathbf{x}_i\|_\infty \le \tau\}\big\|_{\psi_2} \le \tau \cdot \|\boldsymbol{\beta}^*\|_1$ and $\big\|v_i \cdot \mathbb{1}\{\|\mathbf{x}_i\|_\infty \le \tau\}\big\|_{\psi_2} \le \|v_i\|_{\psi_2} \le C \cdot \sigma$. Hence, by Lemma D.1 we have

$$\begin{aligned}\big\|y_i \cdot \mathbb{1}\{\|\mathbf{x}_i\|_\infty \le \tau\}\big\|_{\psi_2} &= \big\|z_i \cdot \langle\boldsymbol{\beta}^*, \mathbf{x}_i\rangle \cdot \mathbb{1}\{\|\mathbf{x}_i\|_\infty \le \tau\} + v_i \cdot \mathbb{1}\{\|\mathbf{x}_i\|_\infty \le \tau\}\big\|_{\psi_2} \\ &\le C \cdot \sqrt{\tau^2 \cdot \|\boldsymbol{\beta}^*\|_1^2 + C' \cdot \sigma^2}.\end{aligned} \quad \text{(C.89)}$$

By the definition of $\psi_1$-norm, we have $\big\|\xi_i \cdot x_{i,j} \cdot x_{i,k} \cdot y_i^2 \cdot \mathbb{1}\{\|\mathbf{x}_i\|_\infty \le \tau\}\big\|_{\psi_1} \le \tau^2 \cdot \big\|y_i^2 \cdot \mathbb{1}\{\|\mathbf{x}_i\|_\infty \le \tau\}\big\|_{\psi_1}$. Further by applying Lemma D.2 to its right-hand side with $Z_1 = Z_2 = y_i \cdot \mathbb{1}\{\|\mathbf{x}_i\|_\infty \le \tau\}$, we obtain

$$\begin{aligned}\big\|\xi_i \cdot x_{i,j} \cdot x_{i,k} \cdot y_i^2 \cdot \mathbb{1}\{\|\mathbf{x}_i\|_\infty \le \tau\}\big\|_{\psi_1} &\le C \cdot \tau^2 \cdot \big\|y_i \cdot \mathbb{1}\{\|\mathbf{x}_i\|_\infty \le \tau\}\big\|_{\psi_2}^2 \\ &\le C' \cdot \tau^2 \cdot \big(\tau^2 \cdot \|\boldsymbol{\beta}^*\|_1^2 + C'' \cdot \sigma^2\big),\end{aligned}$$

where the last inequality follows from (C.89). Therefore, by Bernstein's inequality (Proposition 5.16 in Vershynin (2010)), we obtain

$$\mathbb{P}\Bigg[4/\sigma^2 \cdot \bigg|\frac{1}{n}\sum_{i=1}^n \xi_i \cdot x_{i,j} \cdot x_{i,k} \cdot y_i^2 \cdot \mathbb{1}\big\{\|\mathbf{x}_i\|_\infty \le \tau\big\}\bigg| > v\Bigg] \le 2 \cdot \exp\Bigg[-\frac{C \cdot n \cdot v^2 \cdot \sigma^4}{\tau^4 \cdot \big(\tau^2 \cdot \|\boldsymbol{\beta}^*\|_1^2 + C' \cdot \sigma^2\big)^2}\Bigg], \quad \text{(C.90)}$$

for all $0 \le v \le C \cdot \tau^2 \cdot \big(\|\boldsymbol{\beta}^*\|_1^2 + C' \cdot \sigma^2\big)$ and a sufficiently large sample size $n$.

**Analysis of Term (ii):** For term (ii) in (C.88), by union bound we have

$$\mathbb{P}(\bar{\mathcal{E}}) = \mathbb{P}\Big(\max_{1 \le i \le n} \|\mathbf{x}_i\|_\infty > \tau\Big) \le n \cdot \mathbb{P}\big(\|\mathbf{x}_i\|_\infty > \tau\big) \le n \cdot d \cdot \mathbb{P}\big(|x_{i,j}| > \tau\big).$$

Moreover, $x_{i,j}$ is sub-Gaussian with $\|x_{i,j}\|_{\psi_2} = C$. Thus, by Lemma 5.5 in Vershynin (2010) we have

$$\mathbb{P}(\bar{\mathcal{E}}) \le n \cdot d \cdot 2 \cdot \exp\bigl(-C' \cdot \tau^2\bigr) = 2 \cdot \exp\bigl(-C' \cdot \tau^2 + \log n + \log d\bigr). \tag{C.91}$$

Plugging (C.90) and (C.91) into (C.88), we obtain

$$\begin{aligned}
&\mathbb{P}\Biggl[4/\sigma^2 \cdot \biggl|\frac{1}{n}\sum_{i=1}^{n} \xi_i \cdot x_{i,j} \cdot x_{i,k} \cdot y_i^2\biggr| > v\Biggr] \qquad\qquad (\text{C.92})\\
&\quad \le 2 \cdot \exp\Biggl[-\frac{C \cdot n \cdot v^2 \cdot \sigma^4}{\tau^4 \cdot \bigl(\tau^2 \cdot \|\boldsymbol{\beta}^*\|_1^2 + C' \cdot \sigma^2\bigr)^2}\Biggr] + 2 \cdot \exp\bigl(-C'' \cdot \tau^2 + \log n + \log d\bigr).
\end{aligned}$$

Note that (C.87) is obtained by applying Lemmas D.4 and D.5 with $\phi(v) = \exp(u \cdot v)$. Since Lemmas D.4 and D.5 allow any increasing convex function $\phi(\cdot)$, similar results hold correspondingly. Hence, applying Panchenko's theorem in Lemma D.6 to (C.87), from (C.92) we have

$$\begin{aligned}
\mathbb{P}\Bigl[\Bigl|\bigl\{T_n(\boldsymbol{\beta}^*) - \mathbb{E}_{\boldsymbol{\beta}^*}\bigl[T_n(\boldsymbol{\beta}^*)\bigr]\bigr\}_{j,k}\Bigr| > v\Bigr] &\le 2 \cdot e \cdot \exp\Biggl[-\frac{C \cdot n \cdot v^2 \cdot \sigma^4}{\tau^4 \cdot \bigl(\tau^2 \cdot \|\boldsymbol{\beta}^*\|_1^2 + C' \cdot \sigma^2\bigr)^2}\Biggr]\\
&\quad + 2 \cdot e \cdot \exp\bigl(-C'' \cdot \tau^2 + \log n + \log d\bigr).
\end{aligned}$$

Furthermore, by union bound we have

$$\begin{aligned}
\mathbb{P}\Bigl[\bigl\|T_n(\boldsymbol{\beta}^*) - \mathbb{E}_{\boldsymbol{\beta}^*}\bigl[T_n(\boldsymbol{\beta}^*)\bigr]\bigr\|_{\infty,\infty} > v\Bigr] &\le \sum_{j=1}^{d}\sum_{k=1}^{d} \mathbb{P}\Bigl[\Bigl|\bigl\{T_n(\boldsymbol{\beta}^*) - \mathbb{E}_{\boldsymbol{\beta}^*}\bigl[T_n(\boldsymbol{\beta}^*)\bigr]\bigr\}_{j,k}\Bigr| > v\Bigr]\\
&\le 2 \cdot e \cdot \exp\Biggl[-\frac{C \cdot n \cdot v^2 \cdot \sigma^4}{\tau^4 \cdot \bigl(\tau^2 \cdot \|\boldsymbol{\beta}^*\|_1^2 + C' \cdot \sigma^2\bigr)^2} + 2 \cdot \log d\Biggr]\\
&\quad + 2 \cdot e \cdot \exp\bigl(-C'' \cdot \tau^2 + \log n + 3 \cdot \log d\bigr). \qquad (\text{C.93})
\end{aligned}$$

To ensure the right-hand side is upper bounded by $\delta$, we set the second term on the right-hand side of (C.93) to be $\delta/2$. Then we obtain

$$\tau = C \cdot \sqrt{\log n + 3 \cdot \log d + \log(4 \cdot e/\delta)}.$$

Let the first term on the right-hand side of (C.93) be upper bounded by $\delta/2$ and plugging in $\tau$, we then obtain

$$\begin{aligned}
v = C \cdot &\bigl[\log n + 3 \cdot \log d + \log(4 \cdot e/\delta)\bigr]\\
&\cdot \bigl\{\bigl[\log n + 3 \cdot \log d + \log(4 \cdot e/\delta)\bigr] \cdot \|\boldsymbol{\beta}^*\|_1^2 + C' \cdot \sigma^2\bigr\}/\sigma^2 \cdot \sqrt{\frac{\log(4 \cdot e/\delta) + 2 \cdot \log d}{n}}.
\end{aligned}$$

Therefore, by setting $\delta = 4 \cdot e/d$ we have that

$$\begin{aligned}
&\bigl\|T_n(\boldsymbol{\beta}^*) - \mathbb{E}_{\boldsymbol{\beta}^*}\bigl[T_n(\boldsymbol{\beta}^*)\bigr]\bigr\|_{\infty,\infty} \le v\\
&\quad = C \cdot \bigl(\log n + 4 \cdot \log d\bigr) \cdot \bigl[\bigl(\log n + 4 \cdot \log d\bigr) \cdot \|\boldsymbol{\beta}^*\|_1^2 + C' \cdot \sigma^2\bigr]/\sigma^2 \cdot \sqrt{\frac{\log d}{n}}
\end{aligned}$$

holds with probability at least $1 - 4 \cdot e/d$, which completes the proof of Lemma 4.13. $\square$

### C.11 Proof of Lemma 4.14

*Proof.* For any $j,k\in\{1,\ldots,d\}$, by the mean-value theorem we have

$$\big\|T_n(\boldsymbol{\beta})-T_n(\boldsymbol{\beta}^*)\big\|_{\infty,\infty}=\max_{j,k\in\{1,\ldots,d\}}\Big|\big[T_n(\boldsymbol{\beta})\big]_{j,k}-\big[T_n(\boldsymbol{\beta}^*)\big]_{j,k}\Big| \tag{C.94}$$
$$=\max_{j,k\in\{1,\ldots,d\}}\Big|(\boldsymbol{\beta}-\boldsymbol{\beta}^*)^\top\cdot\nabla\big[T_n(\boldsymbol{\beta}^\sharp)\big]_{j,k}\Big|\le\|\boldsymbol{\beta}-\boldsymbol{\beta}^*\|_1\cdot\max_{j,k\in\{1,\ldots,d\}}\Big\|\nabla\big[T_n(\boldsymbol{\beta}^\sharp)\big]_{j,k}\Big\|_\infty,$$

where $\boldsymbol{\beta}^\sharp$ is an intermediate value between $\boldsymbol{\beta}$ and $\boldsymbol{\beta}^*$. According to (C.83), (C.84) and the definition of $T_n(\cdot)$ in (2.8), by calculation we have

$$\nabla\big[T_n(\boldsymbol{\beta}^\sharp)\big]_{j,k}=\frac{1}{n}\sum_{i=1}^n\bar{\nu}_{\boldsymbol{\beta}^\sharp}(\mathbf{x}_i,y_i)\cdot y_i^3\cdot x_{i,j}\cdot x_{i,k}\cdot\mathbf{x}_i,$$

where

$$\bar{\nu}_{\boldsymbol{\beta}}(\mathbf{x},y)=\frac{8/\sigma^4}{\big[1+\exp\big(-2\cdot y\cdot\langle\boldsymbol{\beta},\mathbf{x}\rangle/\sigma^2\big)\big]\cdot\big[1+\exp\big(2\cdot y\cdot\langle\boldsymbol{\beta},\mathbf{x}\rangle/\sigma^2\big)\big]^2}-\frac{8/\sigma^4}{\big[1+\exp\big(-2\cdot y\cdot\langle\boldsymbol{\beta},\mathbf{x}\rangle/\sigma^2\big)\big]^2\cdot\big[1+\exp\big(2\cdot y\cdot\langle\boldsymbol{\beta},\mathbf{x}\rangle/\sigma^2\big)\big]}. \tag{C.95}$$

For notational simplicity, we define the following events

$$\mathcal{E}=\big\{\|\mathbf{x}_i\|_\infty\le\tau,\ \text{ for all } i=1,\ldots,n\big\},\quad\text{and}\quad\mathcal{E}'=\big\{|v_i|\le\tau',\ \text{ for all } i=1,\ldots,n\big\},$$

where $\tau>0$ and $\tau'>0$ will be specified later. By union bound we have

$$\mathbb{P}\Big\{\Big\|\nabla\big[T_n(\boldsymbol{\beta}^\sharp)\big]_{j,k}\Big\|_\infty>v\Big\}\le d\cdot\mathbb{P}\Big(\Big\{\Big|\nabla\big[T_n(\boldsymbol{\beta}^\sharp)\big]_{j,k}\Big|\Big\}_l>v\Big)$$
$$=d\cdot\mathbb{P}\bigg[\bigg|\frac{1}{n}\sum_{i=1}^n\bar{\nu}_{\boldsymbol{\beta}^\sharp}(\mathbf{x}_i,y_i)\cdot y_i^3\cdot x_{i,j}\cdot x_{i,k}\cdot x_{i,l}\bigg|>v\bigg]. \tag{C.96}$$

Let $\overline{\mathcal{E}}$ and $\overline{\mathcal{E}'}$ be the complement of $\mathcal{E}$ and $\mathcal{E}'$ respectively. On the right-hand side we have

$$\mathbb{P}\bigg[\bigg|\frac{1}{n}\sum_{i=1}^n\bar{\nu}_{\boldsymbol{\beta}^\sharp}(\mathbf{x}_i,y_i)\cdot y_i^3\cdot x_{i,j}\cdot x_{i,k}\cdot x_{i,l}\bigg|>v\bigg]=\underbrace{\mathbb{P}\bigg[\bigg|\frac{1}{n}\sum_{i=1}^n\bar{\nu}_{\boldsymbol{\beta}^\sharp}(\mathbf{x}_i,y_i)\cdot y_i^3\cdot x_{i,j}\cdot x_{i,k}\cdot x_{i,l}\bigg|>v,\mathcal{E},\mathcal{E}'\bigg]}_{\text{(i)}}$$
$$+\underbrace{\mathbb{P}(\overline{\mathcal{E}})}_{\text{(ii)}}+\underbrace{\mathbb{P}(\overline{\mathcal{E}'})}_{\text{(iii)}}. \tag{C.97}$$

**Analysis of Term (i):** For term (i) in (C.97), we have

$$\mathbb{P}\bigg[\bigg|\frac{1}{n}\sum_{i=1}^n\bar{\nu}_{\boldsymbol{\beta}^\sharp}(\mathbf{x}_i,y_i)\cdot y_i^3\cdot x_{i,j}\cdot x_{i,k}\cdot x_{i,l}\bigg|>v,\mathcal{E},\mathcal{E}'\bigg]$$
$$=\mathbb{P}\bigg[\bigg|\frac{1}{n}\sum_{i=1}^n\bar{\nu}_{\boldsymbol{\beta}^\sharp}(\mathbf{x}_i,y_i)\cdot y_i^3\cdot x_{i,j}\cdot x_{i,k}\cdot x_{i,l}\cdot\mathbb{1}\big\{\|\mathbf{x}_i\|_\infty\le\tau\big\}\cdot\mathbb{1}\big\{|v_i|\le\tau'\big\}\bigg|>v,\mathcal{E},\mathcal{E}'\bigg]$$
$$\le\mathbb{P}\bigg[\bigg|\frac{1}{n}\sum_{i=1}^n\bar{\nu}_{\boldsymbol{\beta}^\sharp}(\mathbf{x}_i,y_i)\cdot y_i^3\cdot x_{i,j}\cdot x_{i,k}\cdot x_{i,l}\cdot\mathbb{1}\big\{\|\mathbf{x}_i\|_\infty\le\tau\big\}\cdot\mathbb{1}\big\{|v_i|\le\tau'\big\}\bigg|>v\bigg].$$

To avoid confusion, note that $v_i$ is the noise in mixture of regression model, while $v$ appears in the tail bound. By applying union bound to the right-hand side of the above inequality, we have

$$\mathbb{P}\biggl[\biggl|\frac{1}{n}\sum_{i=1}^{n}\bar{\nu}_{\boldsymbol{\beta}^\sharp}(\mathbf{x}_i,y_i)\cdot y_i^3\cdot x_{i,j}\cdot x_{i,k}\cdot x_{i,l}\biggr|>v,\mathcal{E},\mathcal{E}'\biggr]$$
$$\leq\sum_{i=1}^{n}\mathbb{P}\Bigl[\bigl|\bar{\nu}_{\boldsymbol{\beta}^\sharp}(\mathbf{x}_i,y_i)\cdot y_i^3\cdot x_{i,j}\cdot x_{i,k}\cdot x_{i,l}\cdot\mathbb{1}\bigl\{\|\mathbf{x}_i\|_\infty\leq\tau\bigr\}\cdot\mathbb{1}\bigl\{|v_i|\leq\tau'\bigr\}\bigr|>v\Bigr].$$

By (C.95) we have $\bigl|\bar{\nu}_{\boldsymbol{\beta}^\sharp}(\mathbf{x}_i,y_i)\bigr|\leq 16/\sigma^4$. Hence, we obtain

$$\mathbb{P}\Bigl[\bigl|\bar{\nu}_{\boldsymbol{\beta}^\sharp}(\mathbf{x}_i,y_i)\cdot y_i^3\cdot x_{i,j}\cdot x_{i,k}\cdot x_{i,l}\cdot\mathbb{1}\bigl\{\|\mathbf{x}_i\|_\infty\leq\tau\bigr\}\cdot\mathbb{1}\bigl\{|v_i|\leq\tau'\bigr\}\bigr|>v\Bigr]$$
$$\leq\mathbb{P}\Bigl[\bigl|y_i^3\cdot x_{i,j}\cdot x_{i,k}\cdot x_{i,l}\cdot\mathbb{1}\bigl\{\|\mathbf{x}_i\|_\infty\leq\tau\bigr\}\cdot\mathbb{1}\bigl\{|v_i|\leq\tau'\bigr\}\bigr|>\sigma^4/16\cdot v\Bigr].\qquad\text{(C.98)}$$

Recall that in mixture of regression model we have $y_i=z_i\cdot\langle\boldsymbol{\beta}^*,\mathbf{x}_i\rangle+v_i$, where $z_i$ is a Rademacher random variable, $\mathbf{x}_i\sim N(0,\mathbf{I}_d)$ and $v_i\sim N(0,\sigma^2)$. Hence, we have

$$\bigl|y_i^3\cdot\mathbb{1}\bigl\{\|\mathbf{x}_i\|_\infty\leq\tau\bigr\}\cdot\mathbb{1}\bigl\{|v_i|\leq\tau'\bigr\}\bigr|\leq\Bigl(\bigl|z_i\cdot\langle\boldsymbol{\beta}^*,\mathbf{x}_i\rangle\cdot\mathbb{1}\bigl\{\|\mathbf{x}_i\|_\infty\leq\tau\bigr\}\bigr|+\bigl|v_i\cdot\mathbb{1}\bigl\{|v_i|\leq\tau'\bigr\}\bigr|\Bigr)^3$$
$$\leq\bigl(\tau\cdot\|\boldsymbol{\beta}^*\|_1+\tau'\bigr)^3,$$
$$\bigl|x_{i,j}\cdot x_{i,k}\cdot x_{i,l}\cdot\mathbb{1}\bigl\{\|\mathbf{x}_i\|_\infty\leq\tau\bigr\}\bigr|\leq\bigl|x_{i,j}\cdot\mathbb{1}\bigl\{\|\mathbf{x}_i\|_\infty\leq\tau\bigr\}\bigr|^3\leq\tau^3.$$

Taking $v=16\cdot\bigl(\tau\cdot\|\boldsymbol{\beta}^*\|_1+\tau'\bigr)^3\cdot\tau^3/\sigma^4$, we have that the right-hand side of (C.98) is zero. Hence term (i) in (C.97) is zero.

**Analysis of Term (ii):** For term (ii) in (C.97), by union bound we have

$$\mathbb{P}(\bar{\mathcal{E}})=\mathbb{P}\Bigl(\max_{i\in\{1,\ldots,n\}}\|\mathbf{x}_i\|_\infty>\tau\Bigr)\leq n\cdot\mathbb{P}\bigl(\|\mathbf{x}_i\|_\infty>\tau\bigr)\leq n\cdot d\cdot\mathbb{P}\bigl(|x_{i,j}|>\tau\bigr).$$

Moreover, we have that $x_{i,j}$ is sub-Gaussian with $\|x_{i,j}\|_{\psi_2}=C$. Therefore, by Lemma 5.5 in Vershynin (2010) we have

$$\mathbb{P}(\bar{\mathcal{E}})\leq n\cdot d\cdot\mathbb{P}\bigl(|x_{i,j}|>\tau\bigr)\leq n\cdot d\cdot 2\cdot\exp\bigl(-C'\cdot\tau^2\bigr).$$

**Analysis of Term (iii):** Since $v_i$ is sub-Gaussian with $\|v_i\|_{\psi_2}=C\cdot\sigma$, by Lemma 5.5 in Vershynin (2010) and union bound, for term (iii) in (C.97) we have

$$\mathbb{P}(\overline{\mathcal{E}'})\leq n\cdot\mathbb{P}\bigl(|v_i|>\tau'\bigr)\leq n\cdot 2\cdot\exp\bigl(-C'\cdot\tau'^2/\sigma^2\bigr).$$

To ensure the right-hand side of (C.97) is upper bounded by $\delta$, we set $\tau$ and $\tau'$ to be

$$\tau=C\cdot\sqrt{\log d+\log n+\log(4/\delta)},\quad\text{and}\quad\tau'=C'\cdot\sigma\cdot\sqrt{\log n+\log(4/\delta)}\qquad\text{(C.99)}$$

to ensure terms (ii) and (iii) are upper bounded by $\delta/2$ correspondingly. Finally, by (C.96), (C.97) and union bound we have

$$\mathbb{P}\Bigl\{\max_{j,k\in\{1,\ldots,d\}}\Bigl\|\nabla\bigl[T_n(\boldsymbol{\beta}^\sharp)\bigr]_{j,k}\Bigr\|_\infty>v\Bigr\}\leq d^2\cdot d\cdot\mathbb{P}\biggl[\biggl|\frac{1}{n}\sum_{i=1}^{n}\bar{\nu}_{\boldsymbol{\beta}^\sharp}(\mathbf{x}_i,y_i)\cdot y_i^3\cdot x_{i,j}\cdot x_{i,k}\cdot x_{i,l}\biggr|>v\biggr]\leq d^3\cdot\delta$$

for $v = 16 \cdot \big(\tau \cdot \|\boldsymbol{\beta}^*\|_1 + \tau'\big)^3 \cdot \tau^3/\sigma^4$ with $\tau$ and $\tau'$ specified in (C.99). Then by setting $\delta = 4 \cdot d^{-4}$ and plugging it into (C.99), we have

$$\begin{aligned} v &= 16 \cdot \Big[C \cdot \sqrt{5 \cdot \log d + \log n} \cdot \|\boldsymbol{\beta}^*\|_1 + C' \cdot \sigma \cdot \sqrt{4 \cdot \log d + \log n}\Big]^3 \cdot \Big[C \cdot \sqrt{5 \cdot \log d + \log n}\Big]^3 \Big/ \sigma^4 \\ &\le C'' \cdot \big(\|\boldsymbol{\beta}^*\|_1 + C''' \cdot \sigma\big)^3 \cdot \big(5 \cdot \log d + \log n\big)^3, \end{aligned}$$

which together with (C.94) concludes the proof of Lemma 4.14. □

# D Auxiliary Results

In this section, we lay out several auxiliary lemmas. Lemmas D.1-D.3 provide useful properties of sub-Gaussian random variables. Lemmas D.4 and D.5 establish the symmetrization and contraction results. Lemma D.6 is Panchenko's theorem. For more details of these results, see Vershynin (2010); Boucheron et al. (2013).

**Lemma D.1.** Let $Z_1, \ldots, Z_k$ be the $k$ independent zero-mean sub-Gaussian random variables, for $Z = \sum_{j=1}^k Z_j$ we have $\|Z\|_{\psi_2}^2 \le C \cdot \sum_{j=1}^k \|Z_j\|_{\psi_2}^2$, where $C > 0$ is an absolute constant.

**Lemma D.2.** For $Z_1$ and $Z_2$ being two sub-Gaussian random variables, $Z_1 \cdot Z_2$ is a sub-exponential random variable with

$$\|Z_1 \cdot Z_2\|_{\psi_1} \le C \cdot \max\big\{\|Z_1\|_{\psi_2}^2,\ \|Z_2\|_{\psi_2}^2\big\},$$

where $C > 0$ is an absolute constant.

**Lemma D.3.** For $Z$ being sub-Gaussian or sub-exponential, it holds that $\|Z - \mathbb{E}Z\|_{\psi_2} \le 2 \cdot \|Z\|_{\psi_2}$ or $\|Z - \mathbb{E}Z\|_{\psi_1} \le 2 \cdot \|Z\|_{\psi_1}$ correspondingly.

**Lemma D.4.** Let $\mathbf{z}_1, \ldots, \mathbf{z}_n$ be the $n$ independent realizations of the random vector $\boldsymbol{Z} \in \mathcal{Z}$ and $\mathcal{F}$ be a function class defined on $\mathcal{Z}$. For any increasing convex function $\phi(\cdot)$ we have

$$\mathbb{E}\bigg\{\phi\bigg[\sup_{f \in \mathcal{F}}\bigg|\sum_{i=1}^n f(\mathbf{z}_i) - \mathbb{E}\boldsymbol{Z}\bigg|\bigg]\bigg\} \le \mathbb{E}\bigg\{\phi\bigg[\sup_{f \in \mathcal{F}}\bigg|\sum_{i=1}^n \xi_i \cdot f(\mathbf{z}_i)\bigg|\bigg]\bigg\},$$

where $\xi_1, \ldots, \xi_n$ are i.i.d. Rademacher random variables that are independent of $\mathbf{z}_1, \ldots, \mathbf{z}_n$.

**Lemma D.5.** Let $\mathbf{z}_1, \ldots, \mathbf{z}_n$ be the $n$ independent realizations of the random vector $\boldsymbol{Z} \in \mathcal{Z}$ and $\mathcal{F}$ be a function class defined on $\mathcal{Z}$. We consider the Lipschitz functions $\psi_i(\cdot)$ $(i = 1, \ldots, n)$ that satisfy

$$\big|\psi_i(v) - \psi_i(v')\big| \le L \cdot |v - v'|, \quad \text{for all } v, v' \in \mathbb{R},$$

and $\psi_i(0) = 0$. For any increasing convex function $\phi(\cdot)$ we have

$$\mathbb{E}\bigg\{\phi\bigg[\bigg|\sup_{f \in \mathcal{F}}\sum_{i=1}^n \xi_i \cdot \psi_i\big[f(\mathbf{z}_i)\big]\bigg|\bigg]\bigg\} \le \mathbb{E}\bigg\{\phi\bigg[2 \cdot \bigg|L \cdot \sup_{f \in \mathcal{F}}\sum_{i=1}^n \xi_i \cdot f(\mathbf{z}_i)\bigg|\bigg]\bigg\},$$

where $\xi_1, \ldots, \xi_n$ are i.i.d. Rademacher random variables that are independent of $\mathbf{z}_1, \ldots, \mathbf{z}_n$.

**Lemma D.6.** Suppose that $Z_1$ and $Z_2$ are two random variables that satisfy $\mathbb{E}\big[\phi(Z_1)\big] \le \mathbb{E}\big[\phi(Z_2)\big]$ for any increasing convex function $\phi(\cdot)$. Assuming that $\mathbb{P}(Z_1 \ge v) \le C \cdot \exp(-C' \cdot v^\alpha)$ $(\alpha \ge 1)$ holds for all $v \ge 0$, we have $\mathbb{P}(Z_2 \ge v) \le C \cdot \exp(1 - C' \cdot v^\alpha)$.